%% file: main.tex
\documentclass[11pt]{article}

\usepackage[utf8]{inputenc}
\usepackage[T1]{fontenc}
\usepackage{mathptmx}            
\usepackage[scaled=0.9]{helvet} 
\usepackage{microtype}
\usepackage[a4paper,margin=1in]{geometry}
\usepackage{amsmath,amssymb}
\usepackage{graphicx}
\usepackage{booktabs}
\usepackage{array}
\usepackage{multirow}
\usepackage{longtable}
\usepackage{colortbl}
\usepackage{adjustbox}
\usepackage{float}
\usepackage[section]{placeins}

\usepackage[font=small,labelfont=bf]{caption}
\usepackage{subcaption}
\usepackage{tikz}
\usetikzlibrary{positioning,arrows.meta}
\usepackage{enumitem}
\usepackage{xcolor}
\usepackage{tcolorbox}
\tcbuselibrary{skins,breakable}
\usepackage{titlesec}
\usepackage{titling}
\usepackage{authblk}            

\usepackage[numbers,sort&compress]{natbib}
\usepackage[colorlinks=true,linkcolor=NavyBlue,citecolor=ForestGreen,urlcolor=NavyBlue]{hyperref}
\hypersetup{
  pdftitle={Toward Sub-1 kB Identity-Preserving Face Compression},
  pdfauthor={Petr Hurtik, Jakub Sochor},
  pdfsubject={Sub-1 kB identity-preserving face-image compression benchmark},
  pdfkeywords={face compression, face recognition, sub-1kB, identity preservation, JPEG-AI, learned codec}
}

\definecolor{NavyBlue}{HTML}{1F4E79}
\definecolor{ForestGreen}{HTML}{2E7D32}
\definecolor{ToDoRed}{HTML}{C62828}
\definecolor{TldrBG}{HTML}{EAF2FB}
\definecolor{TldrBar}{HTML}{1F4E79}
\definecolor{HeadGray}{HTML}{333333}

\graphicspath{{figures/}}

\titleformat{\section}{\sffamily\Large\bfseries\color{HeadGray}}{\thesection}{0.6em}{}
\titleformat{\subsection}{\sffamily\large\bfseries\color{HeadGray}}{\thesubsection}{0.6em}{}
\titleformat{\subsubsection}{\sffamily\normalsize\bfseries\color{HeadGray}}{\thesubsubsection}{0.6em}{}

\newtcolorbox{tldrbox}{
  enhanced, breakable, boxrule=0pt, frame hidden,
  colback=TldrBG, sharp corners,
  borderline west={3pt}{0pt}{TldrBar},
  left=10pt, right=8pt, top=5pt, bottom=5pt,
  before skip=6pt, after skip=8pt}
\newcommand{\tldr}[1]{\begin{tldrbox}\small\textbf{\sffamily TL;DR\;}\,#1\end{tldrbox}}

\newcommand{\shadelegend}{\colorbox{green!25}{\textbf{Best}}/%
\colorbox{red!22}{\underline{worst}} per column, as in the other codec tables.}
\newcommand{\shadelegendrow}{\colorbox{green!25}{\textbf{Best}}/%
\colorbox{red!22}{\underline{worst}} per row, as in the other codec tables.}

\title{\sffamily\bfseries Toward Sub-1\,kB Identity-Preserving Face Compression:\\
{\normalsize A Benchmark of Codecs, a Custom Learned Codec, and Studies of
Resolution, Demographic Fairness, Recompression, and Adversarial Robustness}}

\author[]{Petr Hurtik}
\author[]{Jakub Sochor}
\affil[]{Innovatrics}
\date{August 2026}

\begin{document}
\maketitle

\vspace{-1.9em}
{\centering\sffamily\bfseries\large Main findings\par}
\vspace{0.2em}
\noindent{\footnotesize
\renewcommand{\arraystretch}{1.04}
\begin{tabular}{@{}>{\bfseries}p{0.15\textwidth}p{0.82\textwidth}@{}}
\toprule
\normalfont\textbf{Theme} & \textbf{Finding} \\
\midrule
Feasibility & Sub-1\,kB identity-preserving compression is achievable: modern codecs keep $112$\,px EER below ${\sim}0.5\%$ on Color~FERET/ArcFace at 1024\,B (WebP $0.09\%$, up to $0.61\%$ under EdgeFace-XS); AI-Solutions-KK runs higher (\S\ref{sec:benchmark}). \\
1024\,B ranking & JPEG-AI leads identity (Color~FERET id-cosine $0.958$ at $224$\,px); Ours-ACCURATE $0.947$/FAST $0.942$ sit just behind, ahead of AVIF $0.899$/WebP $0.893$; JPEG $0.610$ and JPEG\,2000 $0.585$ collapse. \\
Operating point & At FMR$=10^{-4}$/$1024$\,B, JPEG-AI, WebP and AVIF give the lowest FNMR; JPEG\,2000 is worst (${\sim}38\%$ on Color~FERET). \\
The 512\,B collapse & At $512$\,B, AVIF/HEIF/JPEG\,XL and legacy JPEG collapse to $28$--$98\%$ FNMR; WebP ($6.0\%$), JPEG-AI ($2.9\%$) and the byte-budgeted learned codecs (Ours-ACCURATE $1.8\%$, Ours-FAST $6.7\%$) hold up. CompressAI baselines can't be placed: at $n{=}3{,}375$ impostor pairs, FMR$=10^{-4}$ isn't estimable. \\
Custom codec & Ours-ACCURATE ($18.7$\,M params) tracks JPEG-AI on identity: runner-up at the $224$\,px headline (id-cos $0.947$ vs $0.958$) but \emph{leading} at $112$\,px ($0.948$ vs $0.930$), where it also beats JPEG-AI at $512$\,B on Color~FERET ($1.83\%$ vs $2.86\%$ FNMR) and is \emph{best} on AI-Solutions-KK at $512$\,B ($6.9\%$ vs WebP $24.3\%$) --- confirmed on the held-out CVLFace-IR101 matcher ($0.92\%$ vs WebP $2.55\%$; \S\ref{subsec:codec-heldout}). \\
Model effect & The codec ranking is backbone-invariant (ViT and CNN/IR; Kendall $W{=}0.85$, twelve codecs across all fourteen matchers at $112$\,px, rising to $0.89$ without the four EdgeFace-family matchers): absolute EER tracks matcher strength, not codec order, and ACCURATE\,$>$\,FAST holds on the held-out CVLFace-IR101 backbone. \\
Budget compliance & Several classical codecs barely reach $512$\,B at $168$/$224$\,px ($0$--$5\%$ fit; WebP up to $31\%$); JPEG\,2000 always fits but is worst on identity --- a fit-vs-accuracy tension. \\
Resolution & Downsampling to $112$\,px loses almost no identity (AC energy $>0.99$, embedding cosine $>0.99$ for most matchers, LVFace-L the exception at $0.90$) while freeing bits --- the operating sweet spot. \\
Quality $\neq$ utility & Image quality is an unreliable utility proxy: at $1024$\,B neither PSNR ($\rho{=}{-}0.45$) nor LPIPS ($+0.37$) predicts EER (only SSIM/MS-SSIM clear $p{<}0.05$, weakly); dropping the two coverage-artefact neural baselines ($n{=}10$) makes PSNR/MS-SSIM track firmly ($-0.82$/$-0.92$) while LPIPS still fails. At $512$\,B all predict, DISTS tightest ($+0.96$); the learned codec wins LPIPS/DISTS even where it trails PSNR/SSIM. \\
Fairness & Compression mildly widens the between-group EER gap: JPEG\,2000 amplifies uniform skin-tone disparity $11.2$--$20.0\times$ on the three independent anchors ($+3.4$--$5.0$\,pp); every other codec---ours included---adds at most $1.7$\,pp. Ours-ACCURATE's ArcFace/Color~FERET ethnicity disparity ($0.42$\,pp) is mid-band (JPEG-AI $0.26$, JPEG\,2000 $3.62$); the KK per-tone ordering is non-monotone, so we claim no clean ``darker-skin'' gradient (\S\ref{sec:fairness}). \\
Recompression & Same-codec and modern-to-modern re-encoding is benign, but some chains (JPEG\,XL$\to$HEIF/AVIF) spike EER sharply --- to be avoided in document pipelines. \\
Adversarial & Lossy compression sanitizes no-box (HFC, CLIP, Li-AE) perturbations, but sanitization strength runs opposite identity preservation: at $1024$\,B JPEG\,2000 sanitizes best and is worst on clean identity, while Ours-ACCURATE --- strongest at $112$\,px --- is the \emph{weakest} sanitizer. \\
Significance & Most top-modern-codec pairs are statistically \emph{significant} but practically negligible (ROC-AUC gaps at the 4th decimal; only AVIF$\approx$WebP and HEIF$\approx$JPEG\,XL fail significance on Color~FERET) --- operationally interchangeable; all beat JPEG/JPEG\,2000 decisively (McNemar/DeLong, BH-FDR). \\
Hard budget & A binary search over a frozen $64$-entry gain table with a self-describing container guarantees $\le B$ bytes; a gain-CDF cache makes the learned encode $2.4$--$5\times$ faster. \\
\bottomrule
\end{tabular}\par}
\vspace{0.3em}

\noindent{\footnotesize\itshape
Reading the matcher roster: the four \emph{anchor} matchers (ArcFace-antelopev2,
LVFace-L, TopoFR-R100, EdgeFace-XS) carry \emph{every} metric in this report; the
wider fourteen-model roster appears only in the model-effect panels of
Section~\ref{sec:benchmark}, testing whether the codec ranking depends on the
matcher backbone. CVLFace-IR101 is additionally held out as an independent check
on the custom codec (Section~\ref{subsec:codec-heldout}).\par}
\vspace{0.6em}

\input{sections/_deployment_table}

\clearpage
\begin{abstract}
\input{sections/00_abstract}
\end{abstract}

\vspace{0.8em}
\begin{figure}[H]
  \centering
  \includegraphics[width=0.88\linewidth]{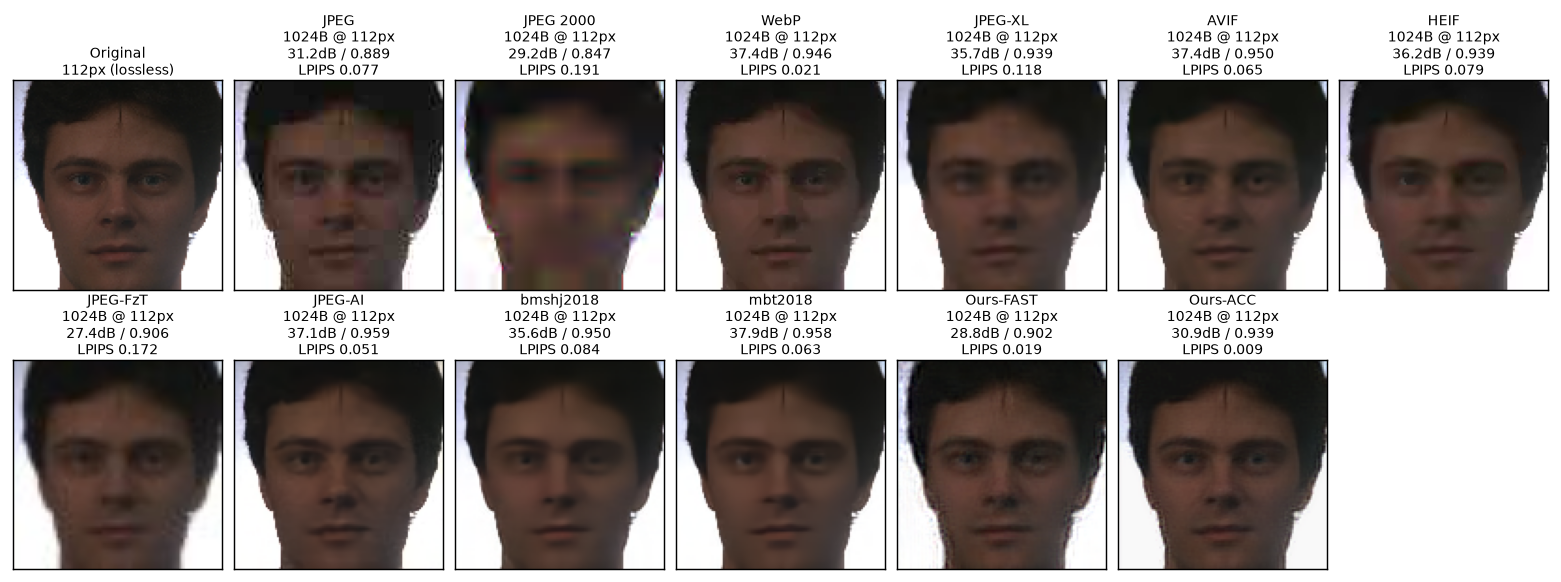}
  \caption{What one kilobyte looks like. A single Color~FERET capture (top left,
  lossless) and its reconstruction by every codec benchmarked in this report, each
  encoded to $\le$$1024$ bytes at the $112$\,px verification working resolution. Tile
  captions give PSNR\,/\,SSIM and LPIPS \emph{on this crop}, not cell medians. JPEG and
  JPEG\,2000 visibly break down; the transform codecs hold structure but soften texture;
  the learned codecs keep the detail a matcher reads --- Ours-ACCURATE reaches LPIPS
  $0.009$, best in the row, while scoring \emph{lower} on SSIM ($0.939$ vs $0.959$): the
  perceptual-versus-distortion decoupling of Section~\ref{sec:quality}. The crop is
  \emph{selected} --- the best-reconstructed of the $300$-image subset the CompressAI
  baselines ran on, chosen so every column is populated. Across the
  full corpus Ours-ACCURATE trails JPEG-AI on SSIM ($0.834$ vs $0.920$) and leads it on
  LPIPS ($0.012$ vs $0.060$); the population figures are in Table~\ref{tab:quality-matrix}.}
\end{figure}
{\small\tableofcontents}
\bigskip

\input{sections/01_introduction}
\input{sections/02_related_work}
\input{sections/03_datasets}
\input{sections/04_protocol}
\input{sections/05_benchmark}
\input{sections/06_quality}
\input{sections/07_custom_codec}
\input{sections/07b_difficulty}
\input{sections/08_ablations}
\input{sections/08c_annex_study}
\input{sections/09_fairness}
\input{sections/10_recompression}
\input{sections/11_adversarial}
\input{sections/12_significance}
\input{sections/13_discussion}
\input{sections/14_conclusion}
\input{sections/15_artifacts}

\bibliographystyle{plainnat}
\bibliography{references}

\end{document}

%% file: sections/_deployment_table.tex
\begin{table}[H]
  \centering
  \footnotesize
  \caption{Deployment recommendations distilled from the study. EER figures are
  Color~FERET at 112\,px under ArcFace unless stated; FNMR@$10^{-4}$ figures are the
  mean across the ArcFace and LVFace-L anchors. Section references give the
  supporting evidence.}
  \label{tab:deployment-guidance}
  \setlength{\tabcolsep}{7pt}
  \adjustbox{max width=\textwidth}{%
  \begin{tabular}{p{0.29\textwidth}p{0.19\textwidth}p{0.42\textwidth}}
    \toprule
    Constraint & Recommendation & Supporting evidence \\
    \midrule
    Best identity at 1024\,B; software or GPU decode acceptable & JPEG-AI \emph{or}
    Ours-ACCURATE &
    These two lead the field and are \textbf{effectively tied}: mean FNMR@$10^{-4}$
    $2.92\%$ vs $2.93\%$ on the in-the-wild AI-Solutions-KK set and $1.00\%$ vs $1.26\%$
    on Color~FERET, with overlapping $95\%$ subject-bootstrap CIs on \emph{all four}
    anchors of \emph{both} datasets (\S\ref{sec:frr-far}). Which one leads depends on the
    anchor, not on the codec: Ours-ACCURATE takes three of the four KK anchors, JPEG-AI
    three of the four Color~FERET ones, and on KK at 112\,px under ArcFace McNemar cannot
    separate them at all ($\chi^2{=}3.2$, $p{=}0.074$; \S\ref{sec:significance}).
    Choose on engineering grounds: JPEG-AI leads identity cosine at 224\,px ($0.958$ vs
    $0.947$, $n{=}64$ crops/cell), Ours-ACCURATE decodes $15\times$ faster on GPU
    ($63$ vs $959$\,ms) from an $18.7$\,M-parameter self-contained model
    (\S\ref{subsec:speed}). Neither has decoder silicon
    (\S\ref{subsec:codec-properties}) \\
    1024\,B with broad software deployability and royalty-free licensing & WebP or
    AVIF & Both reach EER $0.09\%$ at $1024$\,B, level with JPEG-AI and ahead of
    Ours-ACCURATE ($0.14\%$), and they are statistically tied with each other on
    Color~FERET (\S\ref{sec:significance}). This row is won on \emph{licensing and
    ubiquity}, not on accuracy: WebP ($3.1$) and AVIF ($4.1$) sit behind JPEG-AI
    ($1.9$) on mean EER rank across the fourteen matchers, with Ours-ACCURATE level
    with WebP at $3.1$ (Figure~\ref{fig:codec-mean-rank}). Pick these when a
    royalty-free, universally decodable bitstream matters more than the last
    $0.05$\,pp \\
    Hard 512\,B or sub-800\,B operation \emph{at the 112\,px working resolution} &
    byte-budgeted learned codec (JPEG-AI or Ours-ACCURATE) & classical codecs collapse
    below ${\sim}800$\,B (\S\ref{sec:budget-floor}); Ours-ACCURATE is the best codec on
    the in-the-wild set at 512\,B (FNMR@$10^{-4}$ $6.9\%$ vs WebP $24.3\%$; held-out
    confirmation in \S\ref{subsec:codec-heldout}). At 224\,px the 512\,B ranking
    inverts: on Color~FERET the leaders are separated by less than $0.4$\,pp
    (JPEG-AI $3.19\%$, WebP $3.20\%$, Ours-FAST $3.24\%$, JPEG~XL $3.53\%$) with
    Ours-ACCURATE 5th at $4.23\%$ (\S\ref{sec:frr-far}) \\
    Hardware decoder required on legacy/mobile silicon & HEIF (patent-encumbered) or
    JPEG at 1024\,B only & HEVC decode HW is near-universal since ${\sim}2015$
    (\S\ref{subsec:codec-properties}); JPEG holds $0.25\%$ EER at 1024\,B but
    $33.4\%$ at 512\,B (\S\ref{sec:benchmark}) \\
    Encode on a bare CPU core in real time & JPEG, JPEG-FzT, or WebP & encode
    $0.26$/$3.7$/$4.2$\,ms per crop; HEIF costs $172$\,ms and the learned codecs need
    a GPU for interactive encode (\S\ref{subsec:speed}) \\
    Recompression expected downstream (verified for pre-compressed sources only) & fix
    one modern codec end-to-end; safe targets WebP/AVIF/HEIF & modern-to-modern chains
    cost $<{\sim}1$\,pp; forbid JPEG~XL$\rightarrow$HEIF ($+16$--$21$\,pp) and
    JPEG~XL$\rightarrow$AVIF ($+7$--$13$\,pp), and any JPEG~2000 chain
    (\S\ref{sec:recompression}) \\
    Fairness-sensitive deployment & any modern codec except JPEG~2000; Ours-ACCURATE at
    512\,B & JPEG~2000 amplifies the skin-tone disparity $11.2$--$20.0\times$
    ($+3.4$--$5.0$\,pp) at 1024\,B on the three side-stream-independent anchors; all
    others add at most $1.7$\,pp. At the harder 512\,B point the classical gaps widen
    sharply (Color~FERET ethnicity: JPEG $8.3$--$12.3$\,pp, HEIF $7.5$--$8.4$\,pp) but
    Ours-ACCURATE stays lowest ($0.4$--$0.5$\,pp on the independent anchors;
    \S\ref{subsec:fairness-512}) \\
    Avoid in all configurations & JPEG~2000; JPEG at $\le$512\,B & JPEG~2000 has the
    worst identity at every \emph{1024}\,B operating point and is the worst codec on
    average, despite always fitting the budget (\S\ref{sec:budget-compliance}); at the
    tighter $512$\,B/$112$\,px point legacy JPEG is worse still ($33.4\%$ vs $19.5\%$
    EER); catastrophic 512\,B chains (\S\ref{sec:recompression}) \\
    \bottomrule
  \end{tabular}}
\end{table}

%% file: sections/00_abstract.tex
Storing or transmitting face images under a hard sub-kilobyte budget---for
identity documents, smart-card biometrics, and bandwidth-constrained
verification---forces a codec to discard most of the signal while keeping the
part a face matcher actually reads: identity. Generic image codecs optimize for
pixel fidelity, not for the embedding distances that drive verification, so it is
unclear which codec, resolution, and setting best preserve identity at
$\le$1024 bytes, and how this degrades toward an even harder 512-byte point.

This report addresses that gap in three parts. First, we benchmark ten
off-the-shelf general and face-specific codecs across resolutions, byte budgets, two
datasets (controlled Color~FERET and in-the-wild AI-Solutions-KK), and four
face-recognition anchor models, with a fourteen-model roster spanning ViT and CNN
backbones used to check that the codec ranking is backbone-invariant. Second, we train a custom identity-preserving
codec in two variants---a tiny model with no side stream and a larger model with
an identity side-stream and a refine head---that enforces the byte budget exactly
via a binary search over a frozen gain table and packs a self-describing
container. Third, we run four studies: resolution, demographic fairness,
compressed-on-compressed recompression, and no-box adversarial robustness.

Sub-kilobyte identity preservation is feasible, but which codec to deploy depends
entirely on the budget. At 1024 bytes and the 112\,px verification working
resolution the problem is close to solved: the modern codecs hold Color~FERET
equal-error rate under $0.35\%$ on the ArcFace anchor (up to $0.85\%$ on the
weakest, EdgeFace-XS), and the choice between them is nearly immaterial. \emph{At 512 bytes the field re-sorts.} AVIF, HEIF, JPEG~XL and legacy
JPEG collapse to $28$--$98\%$ false-non-match rate at
FMR${=}10^{-4}$, while WebP, JPEG-AI and the byte-budgeted learned codecs stay out of that band ---
single digits on Color~FERET, and, in the wild, $24.3\%$ for WebP against $6.9\%$ for our
accurate variant. That re-sort, not the 1024-byte ranking, is the operational result:
a codec chosen at 1\,kB is not the codec to deploy at half that.

Our accurate variant is the strongest codec at the harder point. On the in-the-wild
AI-Solutions-KK set at 512 bytes it reaches $6.9\%$ FNMR@$10^{-4}$ against WebP's
$24.3\%$, and on Color~FERET $1.83\%$ against JPEG-AI's $2.86\%$ --- confirmed on a
held-out matcher independent of the identity side-stream it is trained against. We
also report the framing that is \emph{least} favourable to our codec: measured by
median identity-cosine at the 224\,px source resolution and 1024 bytes, JPEG-AI
leads at 0.958 and our variant is runner-up at 0.947 (AVIF 0.899, classical JPEG
0.610). The ranking inverts at the 112\,px working point (0.948 vs 0.930), and
image fidelity tells a third story again --- our codec trades broadband pixel
accuracy (PSNR 28.6\,dB against JPEG-AI's 33.3\,dB) for identity and perceptual
quality. That the three metrics disagree is itself a finding: codec choice for
identity has to be validated on verification, not on fidelity.
Section~\ref{sec:discussion} records which panels of the study are complete and
which extensions remain.

%% file: sections/01_introduction.tex
\section{Introduction}
\label{sec:intro}
\tldr{We study how to compress an aligned face crop into at most a few hundred bytes so that it still verifies against its subject, benchmark ten off-the-shelf codecs uniformly, train a custom byte-budgeted identity-preserving codec, and probe resolution, fairness, recompression, and adversarial robustness.}

Face images are increasingly stored not as photographs but as a handful of bytes inside a credential: a two-dimensional barcode on an identity document, a boarding pass, or a smartcard chip. In these settings the storage budget is fixed and tiny, yet the stored face must remain usable by an automated matcher long after capture. This report asks how far a learned, byte-budgeted compressor can be pushed while preserving the identity signal a modern face-recognition (FR) model relies on, and how it compares against the full spread of general-purpose and face-specific codecs under a single, strict per-image budget.

\subsection{Motivation and use cases}
\tldr{Identity documents and machine-readable credentials can spare only a few hundred bytes for a face, so a recognizable face must survive extreme compression; this directly extends prior Innovatrics and NIST work on faces in 2D barcodes.}

The driving use case is the machine-readable travel and identity credential. Standards bodies are moving toward Digital Travel Credentials in which a person's identity is carried in a cryptographically signed, tamper-resistant token, and the natural carrier of that token is a 2D barcode such as a QR code printed on a boarding pass or document \cite{ijcb2026anon}. A barcode has very limited physical area, and once the digital signature, metadata, and error-correction overhead are accounted for, only a fraction of a kilobyte remains for the face photograph itself. The NIST Face Recognition Technology Evaluation report on preparing compact face images for 2D barcodes frames exactly this regime, studying target sizes on the order of a few hundred to roughly twelve hundred bytes across mainstream codecs and several recognition algorithms \cite{nistsp500343}. Our prior conference work in this line established that, at the one-kilobyte target, codec choice dominates downstream recognition accuracy and that some classical codecs largely fail to even reach the required size at full resolution \cite{ijcb2026anon}. This report extends that line: rather than only ranking off-the-shelf codecs, we train a compressor whose objective is identity preservation under a hard byte budget, and we broaden the study to fairness, recompression, and adversarial effects.

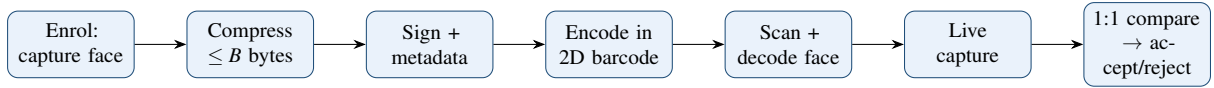
\begin{figure}[t]
  \centering
  \resizebox{\linewidth}{!}{%
  \begin{tikzpicture}[
      >=Stealth, node distance=5mm and 7mm, font=\scriptsize,
      box/.style={draw=TldrBar, rounded corners, fill=TldrBG, align=center,
        minimum height=9mm, inner sep=3pt, text width=15mm}]
    \node[box] (a) {Enrol:\\capture face};
    \node[box, right=of a] (b) {Compress\\$\le B$ bytes};
    \node[box, right=of b] (c) {Sign +\\metadata};
    \node[box, right=of c] (d) {Encode in\\2D barcode};
    \node[box, right=of d] (e) {Scan +\\decode face};
    \node[box, right=of e] (f) {Live\\capture};
    \node[box, right=of f] (g) {1:1 compare\\$\to$ accept/reject};
    \draw[->] (a)--(b); \draw[->] (b)--(c); \draw[->] (c)--(d);
    \draw[->] (d)--(e); \draw[->] (e)--(f); \draw[->] (f)--(g);
  \end{tikzpicture}}
  \caption{The face-on-document workflow this study targets. At enrolment a face is
  compressed to a hard $\le B$-byte budget, signed with metadata, and encoded into a 2D
  barcode; at verification the barcode is scanned, the reference face decoded, and matched
  $1{:}1$ against a live capture. The binding constraint is that the decoded face must
  still verify, not that it look perceptually perfect.}
  \label{fig:workflow}
\end{figure}

A second motivation is bandwidth- and storage-constrained verification more generally: enrolling or transmitting a face template-sized payload over a low-bandwidth channel, or storing many millions of reference faces where per-record size is the binding constraint. In all of these cases the relevant quality metric is not perceptual fidelity but whether the decoded face still verifies against the subject under a deployed FR model. The capture--compress--sign and then scan--match workflow that motivates this work is illustrated in Figure~\ref{fig:workflow}.

\subsection{Problem statement}
\tldr{Given an aligned face crop and a byte budget $B$, produce a bitstream of at most $B$ bytes whose decoded image still verifies against the same subject; we evaluate $B=1024$ and the harder $B=512$.}

We assume the input is a pixel-aligned face crop, so detection and alignment are out of scope and held fixed across all methods. The task is then: given such a crop and a hard byte budget $B$, produce a self-contained bitstream of at most $B$ bytes such that, when decoded, the resulting image yields an FR embedding that verifies against an embedding of the same subject. ``Verifies'' is measured by the FR model's score: the primary metric is the equal-error rate (EER), supported by the false non-match rate (FNMR) at fixed operating points and by statistical significance testing across methods.

Three properties make this regime distinctive. First, the budget is a hard ceiling, not an average rate: every single image must fit, which several classical codecs cannot guarantee at higher resolutions \cite{nistsp500343}. Second, the objective is identity, not appearance, so distortion that is perceptually visible may be harmless and, conversely, perceptually subtle changes may destroy the embedding. Third, the budget is small enough (hundreds of bytes) that resolution, color, and codec overhead all compete for the same scarce bits, forcing explicit trade-offs. We instantiate the problem at $B=1024$ bytes and at a harder $B=512$ bytes, on aligned crops rendered at $64$, $96$, $112$, $168$, and $224$ pixels, with $112$\,px taken as the verification working resolution.

\subsection{Goals and scope}
\tldr{We benchmark off-the-shelf codecs uniformly, build and study a custom byte-budgeted codec, and run controlled studies of resolution, fairness, recompression, and adversarial robustness; detection and alignment are fixed and out of scope.}

Following the task definition for this project, the work has four goals. (1)~Benchmark as many relevant compression algorithms and settings as practical under one uniform protocol, including the choice of resolution, quality setting, and color, all at a fixed byte budget. (2)~Evaluate a representative set of strong, open FR models so that conclusions are not tied to a single matcher. (3)~Train a custom compression architecture aimed specifically at the sub-1\,kB regime and at preserving identity rather than pixels. (4)~Study second-order effects that matter for deployment: demographic and skin-tone bias, facial position and resolution, ``compressed-on-compressed'' recompression (because source images are typically already stored in some compressed form), and no-box adversarial perturbations injected before compression, including whether the codec can sanitize them.

The compared methods span classical and learned codecs: JPEG \cite{wallace1992jpeg}, JPEG\,2000 \cite{taubman2002jpeg2000}, WebP \cite{google2010webp}, JPEG\,XL \cite{alakuijala2019jpegxl}, AVIF \cite{aomavif}, HEIF \cite{sullivan2012hevc}, the F-transform-based JPEG-FzT \cite{perfilieva2021ftransform}, and the learned JPEG-AI standard \cite{ascenso2023jpegai}, together with learned baselines from the literature (bmshj2018 \cite{balle2018hyperprior} and mbt2018 \cite{minnen2018joint}). Against these we place our custom codec in two variants: \emph{Ours-FAST}, a compact model of roughly $1.35$\,M parameters with no side stream, and \emph{Ours-ACCURATE}, a larger model of roughly $18.7$\,M parameters with an identity side-stream and a refinement head. Both variants enforce the hard byte budget exactly via a binary search over a frozen $64$-entry gain table and emit a self-describing container. Two datasets anchor the evaluation: Color~FERET \cite{phillips2000feret}, a controlled set used as the clean, frontal benchmark, and AI-Solutions-KK, an in-the-wild set carrying Monk skin-tone, age, and gender attributes for fairness analysis. Four FR models serve as the primary identity anchors: arcface\_antelopev2 \cite{deng2019arcface}, lvface\_l, topofr\_r100, and edgeface\_xs \cite{george2024edgeface}; a wider fourteen-model roster spanning ViT and CNN/IR backbones (the LVFace, EdgeFace, TopoFR, and CVLFace families) is used for the model-effect analysis that tests whether the codec ranking depends on matcher backbone (Section~\ref{sec:benchmark}). Out of scope are face detection, landmarking, and alignment (held fixed), template-level protection or encryption, and any change to the FR models themselves, which we treat as frozen black boxes.

\subsection{Contributions}
\tldr{A uniform sub-1\,kB/512\,B benchmark of ten off-the-shelf codecs on two datasets and four FR models with statistical testing; a custom byte-budgeted identity-preserving codec in FAST and ACCURATE variants; and controlled studies of resolution, fairness, recompression, and adversarial sanitization.}

This report makes three contributions.

\textbf{(1)~A uniform extreme-compression benchmark.} We evaluate ten off-the-shelf codecs under a single protocol at fixed budgets of $1024$ and $512$ bytes, on two datasets (Color~FERET \cite{phillips2000feret} and AI-Solutions-KK) and four FR models (arcface\_antelopev2 \cite{deng2019arcface}, lvface\_l, topofr\_r100, edgeface\_xs \cite{george2024edgeface}), reporting EER and FNMR with confidence intervals and pairwise statistical significance tests \cite{mcnemar1947,delong1988,benjamini1995fdr}. Coverage is not uniform across the ten: seven run on both datasets at all five resolutions, JPEG-AI runs on both but at $112$ and $224$\,px only on AI-Solutions-KK, and the two CompressAI baselines were run on Color~FERET alone, on a $300$-crop subset. The protocol holds detection, alignment, pairing, and the budget identical across methods so that codec rankings are directly comparable. We further score the benchmark with a fourteen-model roster spanning ViT and CNN/IR backbones, and find that the matchers agree strongly on the codec ranking (Kendall's $W=0.85$ over all fourteen, rising to $0.89$ once the four EdgeFace-family matchers are set aside) even though absolute EER tracks matcher capacity.

\textbf{(2)~A custom byte-budgeted identity-preserving learned codec.} We design and train a codec whose loss targets identity preservation under a hard per-image byte ceiling, in two variants (Ours-FAST $\approx 1.35$\,M parameters, no side stream; Ours-ACCURATE $\approx 18.7$\,M parameters, with an identity side-stream and refine head). The budget is enforced exactly by binary search over a frozen $64$-entry gain table and packed into a self-describing container, so the model guarantees the size constraint rather than approximating an average rate. We are candid about what the side-stream buys: an ablation (Section~\ref{sec:codec}) shows it is \emph{not} what carries the codec's identity advantage --- on matchers independent of its EdgeFace anchor it adds no measurable identity, and reallocating its $90$--$175$\,B to the spatial latent instead leaves independent-matcher identity unchanged ($\Delta$id-cos within $\pm0.005$) while \emph{improving} reconstruction PSNR by ${\sim}0.8$\,dB. Ours-ACCURATE's edge over Ours-FAST is therefore carried by the wider transform, not the headline side-stream; we report this as a negative result rather than overstate the component.

\textbf{(3)~Deployment-oriented studies.} We quantify the resolution--information trade-off across $64$--$224$\,px, audit demographic fairness using Monk skin-tone, age, and gender labels \cite{monk2023mst}, measure the effect of recompressing already-compressed source images, and assess no-box adversarial perturbations applied before compression \cite{goodfellow2015fgsm,kurakin2017physical}, including whether compression sanitizes the attack.

\subsection{Report roadmap}
\tldr{The report proceeds from related work and datasets, through the evaluation protocol and codec benchmark, to our custom codec and a series of controlled studies, closing with discussion and conclusions.}

The remainder of the report is organized as follows. Section~\ref{sec:related} reviews related work on image compression and face recognition relevant to the sub-1\,kB regime. Section~\ref{sec:datasets} describes the two datasets and their attributes, and Section~\ref{sec:protocol} fixes the evaluation protocol, metrics, and FR anchor models. Section~\ref{sec:benchmark} presents the uniform codec benchmark at the $1024$- and $512$-byte budgets, and Section~\ref{sec:quality} reports image-quality and visual comparisons. Section~\ref{sec:codec} introduces our custom byte-budgeted identity-preserving codec and its FAST and ACCURATE variants, and Section~\ref{sec:difficulty} examines which crops are hardest to compress and what makes them so. Section~\ref{sec:ablations} then ablates the codec design and the preprocessing/resolution choices. Section~\ref{sec:fairness} audits demographic fairness, Section~\ref{sec:recompression} studies compressed-on-compressed recompression, and Section~\ref{sec:adversarial} examines no-box adversarial robustness and sanitization. Section~\ref{sec:significance} consolidates the statistical-significance analysis. Section~\ref{sec:discussion} discusses implications and limitations, Section~\ref{sec:conclusion} concludes, and Section~\ref{sec:artifacts} indexes every machine-readable file the tables and figures are built from.

%% file: sections/02_related_work.tex
\section{Background and Related Work}
\label{sec:related}
\tldr{We situate our work against six bodies of literature: general-purpose codecs, learned (neural) compression, face-specific codecs, face-recognition embedders, image-similarity/quality metrics, and no-box adversarial attacks. The recurring lesson is that at extreme rates perceptual quality and biometric utility diverge, that modern codecs decisively beat legacy JPEG, and that downsampling damages identity less than aggressive lossy compression.}

The target of this report---compressing an aligned face crop to at most 1024 (and ideally 512) bytes while preserving the identity signal a matcher needs---draws on several distinct research areas. We review them in turn, emphasising the findings that motivate our design choices: the codec ladder we benchmark against, the learned-compression backbone our custom codec builds on, the identity-loss idea behind our two variants, the embedders we use as anchors, the metrics we report, and the threat model for our robustness study.

\subsection{General-purpose image codecs}
\tldr{The hand-engineered codec ladder runs JPEG $\to$ JPEG\,2000 $\to$ WebP/HEIF $\to$ AVIF $\to$ JPEG\,XL, each following a transform--quantise--entropy-code pipeline; at extreme rates legacy JPEG is unfit and modern codecs (WebP, AVIF) are roughly 30--50\% smaller at matched quality.}

Classical lossy codecs all share a transform-coding pipeline: a fixed, hand-designed invertible transform decorrelates pixels, the coefficients are scalar-quantised, and an entropy coder removes statistical redundancy~\cite{wallace1992jpeg}. JPEG (DCT + Huffman, ISO/IEC 10918) is the baseline~\cite{wallace1992jpeg}; JPEG\,2000 replaces the DCT with a wavelet transform and natively supports target-size encoding~\cite{taubman2002jpeg2000}. WebP codes intra frames with the VP8 toolset~\cite{google2010webp}, HEIF/HEIC uses HEVC intra coding, AVIF uses royalty-free AV1 intra coding~\cite{chen2018av1,aomavif}, and JPEG\,XL combines a modular mode with VarDCT and supports lossless JPEG transcoding and fast encoding~\cite{alakuijala2019jpegxl}. At matched quality AVIF is roughly half the size of reference JPEG, against WebP's roughly 30\% saving~\cite{aomavif,google2010webp}, while JPEG\,XL excels at high fidelity and HEIF is patent-encumbered, which matters for open identity standards~\cite{alakuijala2019jpegxl}.

Two findings from the compression-for-recognition literature directly shape our protocol. First, in the sub-1\,kB regime legacy JPEG is obsolete: NIST SP~500-343 ranks WebP best, then AVIF, then HEIC, then JPEG\,2000, and explicitly deprecates the JPEG variants for preparing faces for 2D barcodes~\cite{nistsp500343}. Second, at a fixed byte budget, reducing resolution does less damage to the identity signal than aggressive lossy compression---the smallest widths tested showed the least shift in the mated-score distribution~\cite{nistsp500343}, a conclusion already reached by the earlier NIST IR~7830 study showing that pre-downsampling before JPEG often improves accuracy~\cite{quinn2011nistir7830}. This is why we sweep resolutions (64--224\,px) rather than fixing one, and why our custom codec downsamples internally. We also include JPEG-FzT, a hybrid that inserts a fuzzy F-transform preprocessing stage that stays backward-compatible with the JPEG bitstream while raising the compression ratio at equal quality~\cite{perfilieva2021ftransform}.

\subsection{Learned image compression}
\tldr{Neural codecs replace the fixed transform with a jointly trained nonlinear autoencoder plus a learned entropy model; the lineage runs factorised prior $\to$ scale hyperprior $\to$ joint autoregressive $\to$ Gaussian-mixture/attention, and the standardised JPEG-AI codec now beats VVC intra. We reuse this machinery, including continuous rate control via learned gain units.}

Learned image compression reframes coding as an end-to-end optimisation problem: an analysis transform $g_a$ maps pixels to a latent, a differentiable surrogate for quantisation (additive uniform noise, the straight-through estimator, or soft-to-hard annealing) makes training possible, and a learned entropy model estimates the rate, all trained jointly to minimise a rate--distortion Lagrangian $L=R+\lambda D$~\cite{balle2017end}. The seminal end-to-end autoencoder already beat JPEG and JPEG\,2000~\cite{balle2017end}. The central architectural lineage is the entropy model: a fully factorised prior~\cite{balle2017end}, then a scale hyperprior that transmits side information predicting per-element Gaussian scales~\cite{balle2018hyperprior}, then a joint autoregressive-and-hierarchical prior that adds a masked-convolution context model and was the first learned method to beat BPG~\cite{minnen2018joint}, then a discretised Gaussian-mixture model with attention reaching parity with VVC intra~\cite{cheng2020learned}. The reference toolbox is CompressAI, which re-implements these baselines (bmshj2018, mbt2018, cheng2020) with pretrained weights~\cite{begaint2020compressai}; we use bmshj2018 and mbt2018 as learned baselines. The third, cheng2020~\cite{cheng2020learned}, is \emph{not} part of this study: its serial autoregressive context model has no stable CUDA path on our hardware and falls back to a CPU decode too slow for a sweep of this size, so it is excluded rather than partially reported, and every codec count in this report is exclusive of it. JPEG-AI (ISO/IEC 6048-1) is the first standardised learned codec, a VAE-based design reporting BD-rate gains over VVC intra ranging from roughly 8\% (base configuration) to roughly 29\% (highest-complexity configuration)~\cite{ascenso2023jpegai}, and we include it among the compared methods.

A practical concern for a single deployable model is variable-rate coding. Rather than train one network per bitrate, learned per-channel ``gain units'' inserted at the encoder output and decoder input let a single model interpolate continuously between operating points~\cite{cui2021asymmetric}. Our custom codec adopts exactly this mechanism: it enforces a hard byte budget with a binary search over a frozen 64-entry gain table, selecting the highest-quality gain vector whose entropy-coded payload still fits the budget. A second relevant distinction is the rate--distortion--perception trade-off: distortion-optimised codecs (PSNR, MS-SSIM) produce blurry low-rate output, while perceptually or generatively optimised codecs trade measurable distortion for realism and can hallucinate detail---a behaviour that is dangerous when identity must be preserved~\cite{blau2018perception}. This tension between realism and identity fidelity is the core risk our identity loss is meant to manage.

Table~\ref{tab:codec-properties} collects the provenance and deployment properties of the
twelve evaluated configurations --- ten off-the-shelf codecs plus our two custom
variants --- introduced above --- ownership, standardisation year, licensing,
hardware-decoder availability, and the hard-budget fit rate --- so that the codec landscape
is visible before the results. Licensing splits the field (JPEG, JPEG\,2000, WebP, JPEG\,XL,
AVIF and the JPEG-AI reference software are royalty-free, whereas HEIF inherits HEVC's patent
pools and the CompressAI baselines ship under a BSD licence with an explicit patent
non-grant); hardware-decoder availability is inversely related to compression strength; and
the fit-rate column previews the budget-compliance result quantified later
(Section~\ref{sec:budget-compliance}). Encode/decode latency is reported separately in
Section~\ref{subsec:speed}.

\begin{table}[t]
  \centering
  \small
  \caption{Codec provenance and deployment properties for the twelve evaluated
  configurations (ten off-the-shelf codecs and our two custom variants).
  ``Fit $\le$1\,kB'' is the fraction of crops that reach the $1024$-byte budget at
  $112$/$224$\,px, measured on Color~FERET (n/a: the CompressAI baselines expose fixed
  quality presets rather than a byte target, so what they achieve is incidental rather
  than enforced and is reported separately in Section~\ref{sec:budget-compliance}: at
  $112$\,px bmshj2018 reaches $59\%$ and mbt2018 $67\%$, at $224$\,px $47\%$ and $66\%$,
  and they are measured at all five resolutions there, on their $300$-crop subset).
  JPEG-AI holds the $1$\,kB budget wherever it was run; both Ours variants
  hold it at $64$/$112$/$224$\,px and emit the over-budget fallback frame on
  $10$--$13\%$ of crops at $96$ and $168$\,px. The budget bites hardest at $512$\,B ---
  all quantified in Section~\ref{sec:budget-compliance}. Encode/decode latency is measured separately
  in Section~\ref{subsec:speed} (Table~\ref{tab:speed}). Source for both Ours
  variants will be released under the MIT licence.}
  \label{tab:codec-properties}
  \adjustbox{max width=\textwidth}{\input{tables/codec_properties.tex}}
\end{table}

\paragraph{Decoder footprint and deployed encode cost.} Two deployment characteristics not
in Table~\ref{tab:codec-properties} matter for the document-scanner scenario. \emph{Decoder
footprint}: the classical codecs decode with library code only (kilobytes of binary, no model
weights), whereas the learned codecs must ship a network --- Ours-FAST is $1.35$\,M parameters
($\approx$5\,MB in FP32), Ours-ACCURATE $18.7$\,M ($\approx$75\,MB), and the JPEG-AI reference
decoder is a substantially larger multi-hundred-MB neural model --- so on constrained
scanner silicon the classical/transform codecs and the tiny Ours-FAST are the deployable
decoders while JPEG-AI and Ours-ACCURATE assume a capable host. \emph{Deployed encode}: the
single-shot latencies in the speed table (Section~\ref{subsec:speed},
Table~\ref{tab:speed}) are per encode call, but the byte-budgeted codecs (the classical
quality search and both Ours variants) run a binary search of $\approx 6$ encodes to hit the
hard budget, so a deployed \emph{budget-compliant} encode costs roughly $6\times$ the
tabulated figure. For Ours, a gain-CDF cache removes the ${\approx}80\%$ of encode cost
that a cold rebuild would cost per gain, bounding the attainable warm-cache speed-up at
up to ${\approx}5\times$ (Section~\ref{sec:codec}); this is a bound derived from that
percentage, not a measured end-to-end figure --- \S\ref{sec:codec} reports none. Decode
is unaffected and remains single-shot.

\subsection{Face-specific compression}
\tldr{Faces are a low-entropy, strongly structured image class, so face-specific codecs (from eigenfaces to generative face video coding) win big on perceptual metrics but risk identity hallucination at extreme rates; the most relevant idea for us is fine-tuning a learned codec with an ArcFace identity loss to preserve recognition accuracy at equal bitrate.}

Because faces carry strong, learnable priors (symmetry, the fixed arrangement of features) and human perception is exquisitely sensitive to identity, faces are the domain where model-based and generative coding beat general codecs by the widest margin---and where the rate--distortion--perception trade-off bites hardest, since at ultra-low rates a decoder must invent plausible detail that can drift from the true identity. The lineage runs from eigenfaces/PCA (a face as a weighted sum of principal components~\cite{turk1991eigenfaces}) and MPEG-4 model-based coding through modern Generative Face Video Coding (GFVC), where a reference frame plus compact per-frame keypoints or semantics drive a generative decoder, achieving 50--75\% BD-rate savings over VVC on perceptual metrics at the cost of pixel fidelity~\cite{chen2023ifvc}. These video methods are now an active JVET standardisation effort but are out of scope for single-image credentials.

Most directly relevant to us is identity-preserving still-image compression: adding an ArcFace\slash CurricularFace identity loss---cosine similarity between the embeddings of the original and the reconstruction---to a learned codec. Joint training of a compression network with a recognition network yields better face-verification accuracy at equal rate than JPEG or JPEG\,2000~\cite{bian2019compnet}, and a recent codebook-quantisation method reports high recognition accuracy at very low bitrate by fine-tuning the decoder with an ArcFace identity loss~\cite{wang2025codebook}. The practical recommendation in this literature---take a general learned-codec backbone and fine-tune it with a face/identity loss rather than design a bespoke architecture---is the recipe behind both of our variants (Ours-FAST and Ours-ACCURATE). The cautionary note is equally important: identity drift is the kill criterion, and it must be validated with embedding cosine similarity on the target data, not with perceptual metrics alone.

\subsection{Face-recognition models}
\tldr{Modern matchers are embedding extractors mapping an aligned crop to a fixed-length vector compared by cosine similarity; we anchor our evaluation on four with public weights---ArcFace (antelopev2), LVFace, TopoFR, and the edge-efficient EdgeFace---spanning the accuracy/compute spectrum.}

Almost every modern face-recognition model is an embedding extractor: it maps an aligned face crop (usually 112$\times$112) to a fixed-length vector, and the same embedding serves verification (cosine similarity versus a threshold) and identification (nearest-neighbour search)~\cite{deng2019arcface}. ArcFace introduced the additive angular-margin loss that produces highly discriminative hypersphere embeddings and remains the de-facto open baseline~\cite{deng2019arcface}; the InsightFace antelopev2 pack ships a ResNet-100 ArcFace trained on Glint360K~\cite{deng2019arcface}. Recent academic leaders include LVFace, a ViT trained with progressive cluster optimisation that took first place on the MFR-Ongoing academic track~\cite{lvface2025}, and TopoFR, which preserves the topological structure of the input space in the embedding via persistent homology~\cite{dan2024topofr}. For edge deployment, EdgeFace adapts an EdgeNeXt CNN--transformer hybrid with a low-rank linear layer, reaching strong accuracy at around 1.77\,M parameters~\cite{george2024edgeface}. We use these four---arcface\_antelopev2, lvface\_l, topofr\_r100, and edgeface\_xs---as the anchor matchers so that conclusions are not tied to a single embedder. Two caveats from this literature shape our protocol: saturated benchmarks (e.g.\ LFW) are uninformative, so we report verification error directly on our data; and recognition accuracy varies across demographic groups, motivating our fairness study~\cite{grother2019frvt}.

\subsection{Image-similarity and quality metrics}
\tldr{We separate fidelity metrics (PSNR, SSIM), perceptual metrics (LPIPS, DISTS), and the identity metric that actually matters here (embedding cosine similarity); the consistent finding is that perceptual quality is only a partial proxy for biometric utility, so identity cosine is our primary criterion.}

Full-reference quality measures fall on a spectrum from signal fidelity to perception. Pixel-error metrics (MSE, PSNR) are trivial, differentiable, and genuine metrics, but correlate poorly with human perception and ignore spatial structure~\cite{wang2004ssim}. SSIM shifted the focus from error visibility to structural fidelity by comparing local luminance, contrast, and structure statistics, and is far better correlated with perception while remaining cheap and differentiable~\cite{wang2004ssim}; MS-SSIM extends it across scales~\cite{wang2004ssim}. Learned perceptual metrics improve further: LPIPS measures distance in deep feature space and predicts human similarity judgments far better than SSIM or PSNR~\cite{zhang2018lpips}, and DISTS unifies structure and texture similarity on deep features, tolerating texture resampling that defeats SSIM and LPIPS~\cite{ding2020dists}. For comparing whole distributions of generated images (rather than pairs), FID/KID and CLIP-feature variants are standard~\cite{radford2021clip}.

Crucially for this work, none of these is a reliable proxy for recognition utility. A codec can look worse to a human yet match better, or vice versa---perceptual metrics correlate only partially with biometric utility, and generative codecs optimised for human preference can produce identity-altering reconstructions~\cite{ijcb2026anon}\cite{mentzer2020hific}. We therefore report PSNR, SSIM, LPIPS, and DISTS for context but treat the cosine similarity between the ArcFace embeddings of the original and the reconstruction (the embed-then-compare paradigm~\cite{deng2019arcface,zhang2018lpips}) as the primary distortion criterion, and downstream verification EER as the ultimate one.

\subsection{No-box adversarial attacks}
\tldr{No-box is the strictest threat model---no model access, no queries, only a handful of self-supplied images---so attacks rely entirely on transferability via tiny surrogates, foundation-model surrogates, or training-free frequency manipulation; we use it to test whether our codec's compression incidentally sanitises such perturbations.}

\paragraph{Why a compression report studies attacks.} A codec in an identity pipeline is
not only a storage decision; it is also the last image-domain operation before the matcher
sees the face, and it is a \emph{lossy, low-pass, requantising} one. That is precisely the
class of transformation the adversarial-defence literature calls input purification. So the
question this report has to answer is not ``can we attack the matcher'' but: \emph{when a
face crop is squeezed to under a kilobyte, what happens to a perturbation that was already
in it?} Two deployment consequences follow. If compression destroys such perturbations, the
byte budget buys a security property for free and a document pipeline gains a defence it did
not pay for. If it does not --- or worse, if the learned codecs preserve perturbations that
the classical ones erase --- then a codec chosen purely on identity-preservation grounds may
quietly weaken the system. No-box is the right threat model for that question because it is
the one an operational attacker actually faces: no access to the deployed matcher and no
queries against it. The rest of this subsection sets up that threat model; the measurement
is in Section~\ref{sec:adversarial}.

The no-box setting assumes the attacker has no access to the victim model's architecture, parameters, or training data, and cannot query it even once; the only resource is a small number of self-collected, in-domain images~\cite{li2020nobox}. Because gradients and query feedback are unavailable, every no-box method relies on transferability---crafting perturbations on a self-supplied proxy and hoping they carry to the unseen victim. The building blocks are the standard white-box attacks reused inside the proxy: FGSM~\cite{goodfellow2015fgsm} and its iterative variant BIM/I-FGSM~\cite{kurakin2017physical}. Three no-box families have emerged: training a tiny surrogate autoencoder from tens of domain images (the seminal prototypical-reconstruction attack, which cut a commercial celebrity-recognition API from 100\% to about 15\% accuracy using ten facial images)~\cite{li2020nobox}; repurposing a generic foundation model such as CLIP~\cite{radford2021clip} as the surrogate after margin-aware fine-tuning~\cite{zhang2023mfclip}; and going fully training-free by suppressing and re-injecting high-frequency image components, which fooled ten ImageNet models at about 98\% success~\cite{zhang2022hfc}.

The defensive angle is what matters for us. Standard input-purification defenses---JPEG compression, bit-depth reduction, randomised resizing---specifically blunt the small-perturbation and frequency-domain routes, because no-box perturbations target shared, often high-frequency, vulnerabilities~\cite{guo2018countering}. Since aggressive lossy compression is itself a low-pass, requantising operation, we ask whether passing an adversarial face through our sub-1\,kB codec incidentally sanitises the attack; the experiment and its sanitisation result are reported later (Section~\ref{sec:adversarial}).

%% file: sections/03_datasets.tex
\section{Datasets}
\label{sec:datasets}
\tldr{Two complementary face sets drive every experiment: Color~FERET, a controlled
frontal-portrait benchmark, and AI-Solutions-KK, an in-the-wild set carrying per-identity
skin-tone, age, and gender labels. Both are aligned identically to five square
resolutions (64--224\,px), with 112\,px as the verification working resolution.}

We evaluate on two datasets chosen to bracket the operating envelope of a face
matcher: one clean and frontal, one uncontrolled and demographically annotated.
Crucially, both are aligned with the \emph{same} canonical five-point template and the
same family of similarity transforms, so the only intended difference between them is
imaging conditions, not framing. This section describes each dataset, the shared
alignment and resolution grid, the verification-pair protocol, and the demographic
attributes used in the fairness study (Section~\ref{sec:fairness} when referenced
later). Table~\ref{tab:dataset-summary} contrasts the two datasets and their
complementary roles.

\begin{table}[t]
  \centering
  \small
  \caption{The two evaluation datasets at a glance. The alignment target (ArcFace
  five-point template at $\{64,96,112,168,224\}$\,px) is identical for both; they
  differ only in imaging conditions and in where the five landmarks come from.}
  \label{tab:dataset-summary}
  \begin{tabular}{lll}
    \toprule
    & Color~FERET & AI-Solutions-KK \\
    \midrule
    Capture conditions & controlled studio portraits & in-the-wild photographs \\
    Aligned images & $11{,}335$ & $17{,}534$ \\
    Identities & $994$ & $105$ \\
    Verification pairs & $\sim$$64$\,M (all $C(N,2)$) & $\sim$$153.7$\,M (all $C(N,2)$) \\
    Landmark source & shipped 23-pt annotation & detector + landmarker \\
    Attribute provenance & dataset-supplied ground truth & model-estimated \\
    Attributes & pose, sex, age, ethnicity & Monk skin tone, sex, age \\
    Role in the study & confirmatory, clean anchor & exploratory, in-the-wild stress \\
    \bottomrule
  \end{tabular}
\end{table}

\subsection{Color~FERET}
\tldr{A controlled, frontal portrait dataset~\cite{phillips2000feret} of general
subjects (not booking photos); after decoding and alignment it yields $11{,}335$
aligned crops used as our clean, frontal benchmark.}

\begin{figure}[t]
  \centering
  \includegraphics[width=0.72\linewidth]{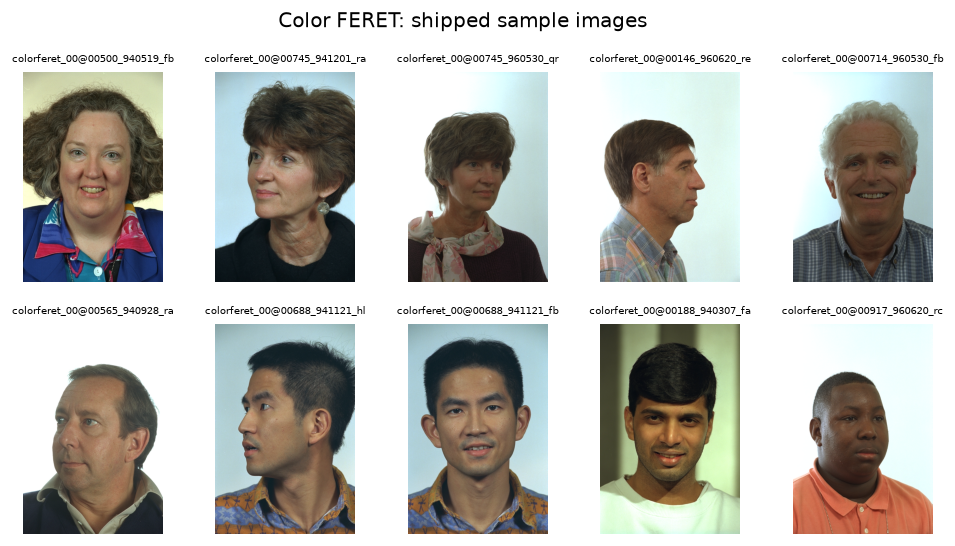}
  \caption{Color~FERET example face crops as shipped in the dataset. The verification
  pipeline operates on aligned-112 crops, as for KK; this panel shows the source
  imagery.}
  \label{fig:ds-cf}
\end{figure}

Color~FERET~\cite{phillips2000feret} is a controlled face-recognition benchmark
collected under studio conditions: uniform lighting, plain backgrounds, and a
standardized set of head poses per subject. We use it as the ``clean'' reference
point of the study, where compression artefacts rather than capture nuisances dominate
identity loss. The canonical set is decoded from an index-aligned image container of
$11{,}338$ source blobs paired with a ground-truth attribute table (gender, race,
glasses, beard/mustache, pose code, and head-pose angles). Color~FERET ships
proprietary 23-point landmark annotations, so no face detector is run during
preparation: we read the five ArcFace landmarks (eye centres, nose tip, mouth corners)
directly from the annotation and similarity-transform each image onto the canonical
template. After dropping images with incomplete landmark annotations, alignment yields
$11{,}335$ aligned crops spanning $994$ subject identities, which form the
Color~FERET evaluation set. Because its attributes are dataset-supplied ground truth,
Color~FERET serves as the confirmatory anchor for the fairness analysis, complementing
the model-estimated attributes of AI-Solutions-KK (below).
Figure~\ref{fig:ds-cf} shows example source captures as shipped and
Table~\ref{tab:cf-attributes} the dataset's ground-truth attribute distribution.

\begin{table}[t]
  \centering
  \small
  \caption{Color~FERET ground-truth attribute distribution over the $11{,}338$ decoded
  captures. Percentages are of the whole corpus. The corpus is majority male
  ($64.6\%$) and majority White ($62.4\%$), which is why Color~FERET is used as the
  \emph{confirmatory} fairness anchor --- its ethnicity subgroups below Hispanic carry
  too few images to support a per-group EER --- while the in-the-wild
  AI-Solutions-KK set carries the skin-tone analysis (Section~\ref{sec:fairness}). The
  pose block is listed in full, so the frontal/non-frontal splits used later can be read
  off directly: the near-frontal conditions (\emph{frontal image}, its repeat, and the
  two $15^{\circ}$ turns) account for $32\%$ of the corpus.}
  \label{tab:cf-attributes}
  \input{tables/cf_attributes.tex}
\end{table}

\subsection{AI-Solutions-KK}
\tldr{An in-the-wild dataset with per-identity Monk skin-tone, age, and gender
attributes~\cite{monk2023mst}; its verification protocol enumerates every
\(C(N,2)\) pair (about 153.7 million) and scores all mated pairs against a fixed
$5$\,M impostor sample (\S\ref{sec:protocol}).}

AI-Solutions-KK is an uncontrolled, ``in-the-wild'' face set: photos vary in pose,
expression, illumination, background, and source resolution, so it stresses a codec on
the conditions a deployed matcher actually faces. Unlike Color~FERET, it ships only raw
per-identity photos with no landmark annotations, so the five ArcFace landmarks are
\emph{predicted} by a face detector and landmarker before the same similarity-transform
alignment is applied (detailed in Section~\ref{sec:datasets-alignment}). Each identity
also carries a Monk skin-tone index~\cite{monk2023mst}, an estimated age, and an
estimated gender, which makes KK the dataset for the demographic-fairness analysis.

\begin{figure}[t]
  \centering
  \includegraphics[width=\linewidth]{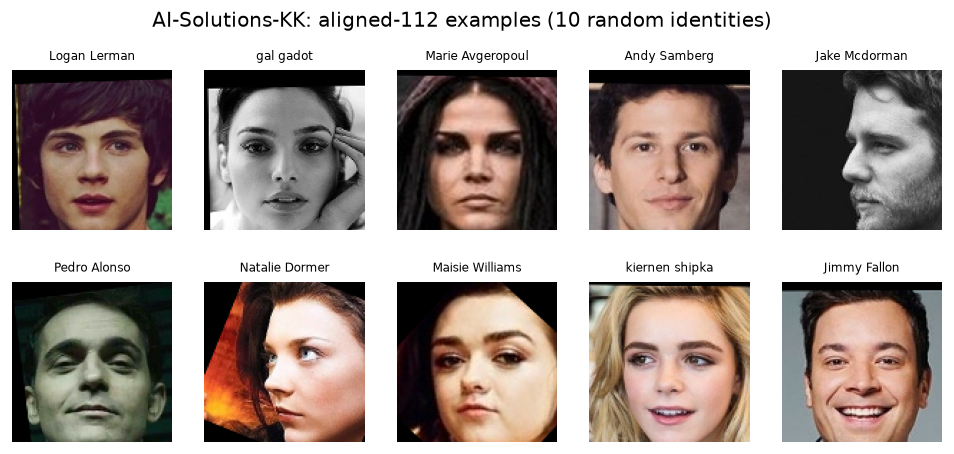}
  \caption{AI-Solutions-KK: aligned-112 example crops, illustrating the in-the-wild
  variation in pose, expression, and illumination.}
  \label{fig:ds-kk-examples}
\end{figure}

The dataset comprises $17{,}534$ aligned crops (the embedded evaluation set) drawn from
$105$ identities. Its
verification protocol enumerates \emph{every} unordered image pair $C(N,2)$,
which at this scale is approximately $153.7$ million pairs (about
$1.52$ million mated and $152.2$ million non-mated); scoring uses all mated pairs and the
fixed $5$\,M impostor sample of \S\ref{sec:protocol}. Example aligned crops are shown in
Figure~\ref{fig:ds-kk-examples}; the original source-image size distributions are
reported under demographic attributes below.

\paragraph{Availability.} Because every result in this report is computed on the
\emph{aligned} crops rather than the raw photos, alignment is part of the protocol and
reproducing the benchmark requires exactly the crops we used. The two headline aligned
resolutions --- $112\times112$ and $224\times224$, produced by the pipeline of
Section~\ref{sec:datasets-alignment} --- are published for AI-Solutions-KK and freely
downloadable at \url{https://huggingface.co/datasets/Cha53c/1kb-face-aligned}
($35{,}068$ images: all $17{,}534$ crops at each of the two resolutions). The
Color~FERET crops are not part of that release. The remaining resolutions used in the
resolution study ($64$, $96$, $168$\,px) are \emph{not} resamplings of the published
crops: each resolution is warped directly from the source photograph with the same
transform scaled by $\mathrm{size}/112$
(Section~\ref{sec:datasets-alignment}), so reproducing them needs the source images and
the same pipeline, not the released crops.

\subsection{Alignment and resolutions}
\label{sec:datasets-alignment}
\tldr{Both datasets are warped to the same canonical five-point template and emitted at
five square resolutions (64, 96, 112, 168, 224\,px), with framing held constant so
resolution is the only variable; 112\,px is the verification working resolution.}

Both datasets are aligned to a single canonical five-point destination template
(eye centres, nose tip, mouth corners) shared by every face-recognition model we
benchmark, using a similarity transform (rotation, uniform scale, translation; no
shear). The two sources differ only in how the five source points are obtained:
Color~FERET reads them from its shipped 23-point annotations, whereas
AI-Solutions-KK predicts them with a face detector and landmarker. This shared target
means the two datasets have identical framing, isolating imaging conditions as the only
between-dataset difference.

For each image we emit one aligned crop at every resolution in
$\{64, 96, 112, 168, 224\}$\,px. The template is defined at $112\times112$ and scaled by
$\mathrm{size}/112$ for the other resolutions, so all crops use the \emph{same} framing
and resolution is the only variable in the resolution study. Unless stated otherwise,
$112$\,px is the working resolution for verification, with the remaining resolutions
used to study the resolution--byte-budget trade-off.

\subsection{Verification pairs}
\tldr{We enumerate the complete pairwise matrix \(C(N,2)\) (same-identity pairs mated,
different-identity non-mated) and score every mated pair; because the impostor set is
$6.4\times10^{7}$ pairs on Color~FERET and $1.5\times10^{8}$ on AI-Solutions-KK, EER/FNMR
use a fixed seeded sample of at most $5\times10^{6}$ impostor
pairs, shared across all conditions (\S\ref{sec:protocol}). Pairs are stored index-based
for fast vectorized cosine scoring over precomputed embeddings.}

Verification accuracy is measured over the \emph{complete} pairwise matrix: for $N$
aligned images, all $C(N,2)$ unordered pairs are enumerated and labelled mated (same
identity/subject) or non-mated (different identity/subject). For
Color~FERET this pair \emph{universe} is roughly $64$ million pairs over the
$\sim\!11.3$\,k aligned images; for AI-Solutions-KK it is the $\sim\!153.7$ million pairs
described above. Scoring uses \emph{all} mated pairs but, since the impostor count is
intractable, a fixed seeded uniform sample of at most $5\times10^{6}$ impostor pairs
(identical across every codec, budget, and matcher; see \S\ref{sec:protocol}); the
$64$/$153.7$\,M figures therefore count the enumerated universe, while the scored impostor
count is the capped $5.0\times10^{6}$. At this scale
a textual \texttt{name1,name2,label} list would be many gigabytes, so pairs are stored
index-based: a compact parquet of integer index pairs plus labels, alongside a small
image-index table mapping each integer index to its relative path, identity/subject, the
per-image alignment residual, and additionally the pose code (Color~FERET) or the
per-image detection score (AI-Solutions-KK). Scoring is then a vectorized
cosine over the precomputed embeddings indexed by the integer pairs, and the index table
lets the accuracy stage slice by pose, identity, or demographic group, and to restrict
to compliant subsets, without re-reading the source labels.

\subsection{Demographic attributes}
\tldr{Each AI-Solutions-KK identity carries a Monk skin-tone
index~\cite{monk2023mst}, an estimated age, and an estimated gender; together with the
images-per-identity and original-size distributions, these drive the fairness study.}

\begin{figure}[t]
  \centering
  \begin{subfigure}{0.49\linewidth}
    \centering
    \includegraphics[width=\linewidth]{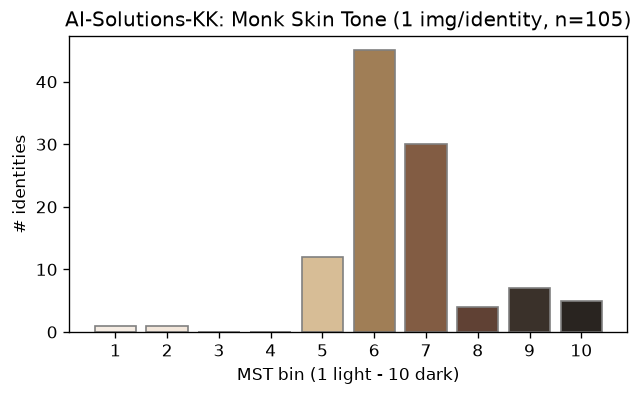}
    \caption{Monk skin-tone index.}
    \label{fig:ds-kk-mst}
  \end{subfigure}
  \hfill
  \begin{subfigure}{0.49\linewidth}
    \centering
    \includegraphics[width=\linewidth]{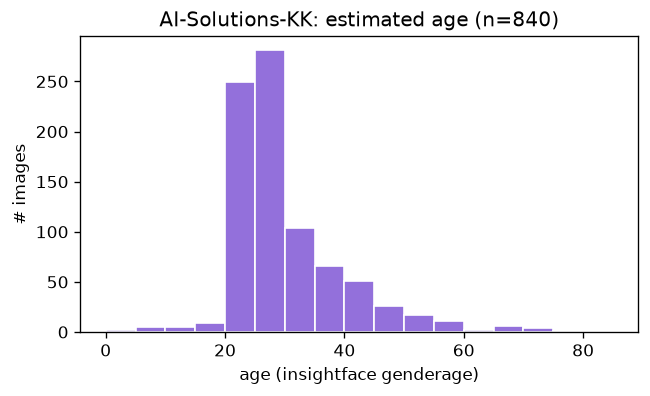}
    \caption{Estimated age.}
    \label{fig:ds-kk-age}
  \end{subfigure}
  \\[0.6em]
  \begin{subfigure}{0.49\linewidth}
    \centering
    \includegraphics[width=\linewidth]{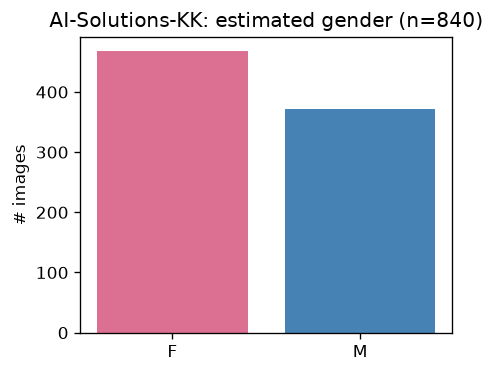}
    \caption{Estimated gender.}
    \label{fig:ds-kk-gender}
  \end{subfigure}
  \hfill
  \begin{subfigure}{0.49\linewidth}
    \centering
    \includegraphics[width=\linewidth]{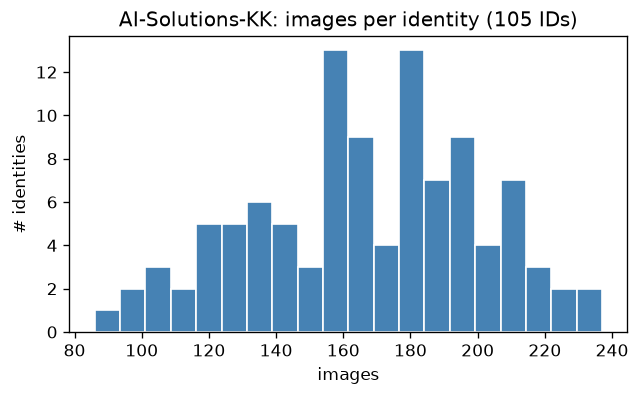}
    \caption{Images per identity.}
    \label{fig:ds-kk-imgs}
  \end{subfigure}
  \caption{AI-Solutions-KK demographic and per-identity statistics:
  (a)~Monk skin-tone index~\cite{monk2023mst}, (b)~estimated age,
  (c)~estimated gender, and (d)~images per identity.}
  \label{fig:ds-kk-demo}
\end{figure}

AI-Solutions-KK provides the demographic attributes used in the fairness analysis. Each
identity carries a Monk skin-tone (MST) index on the 10-tone scale~\cite{monk2023mst}
(light to dark), an estimated age, and an estimated gender. These attributes are
estimated automatically: age and gender from an off-the-shelf gender--age model run on
the aligned 112 crops, and the MST index from a skin-tone estimator on the aligned 224
crops; because skin tone is per-identity stable while age/gender vary little within a
session, the attributes are summarized per identity. Where a per-identity MST label and
the individual image disagree --- typically a light-skinned subject photographed in
shadow --- we fall back on the \emph{Individual Typology Angle} (ITA), a CIELAB-derived
per-image skin-tone measure computed over lit-side facial pixels on which higher angles
mean lighter skin; ITA is what selects the skin-tone exemplars shown in
Section~\ref{sec:quality}. The resulting MST distribution is
strongly concentrated: of the $105$ identities, MST6 carries $45$ and MST7 carries $30$
--- $71\%$ of the set between them --- while MST5, MST9, MST10 and MST8 carry only $12$,
$7$, $5$ and $4$ identities respectively, and MST1 and MST2 a single identity each. A
single-identity group admits no within-group impostor pair, so those two bins cannot
yield a subgroup error rate and the fairness analysis reports the six groups
MST5--MST10, the smallest of them resting on four subjects. Because these labels are
model-estimated rather than curated, and the set is this unbalanced, the KK
fairness results (Section~\ref{sec:fairness}) are exploratory and are read alongside the
ground-truth Color~FERET anchor. Figure~\ref{fig:ds-kk-demo} shows
the MST, age, and gender distributions.

\begin{figure}[t]
  \centering
  \begin{subfigure}{0.62\linewidth}
    \centering
    \includegraphics[width=\linewidth]{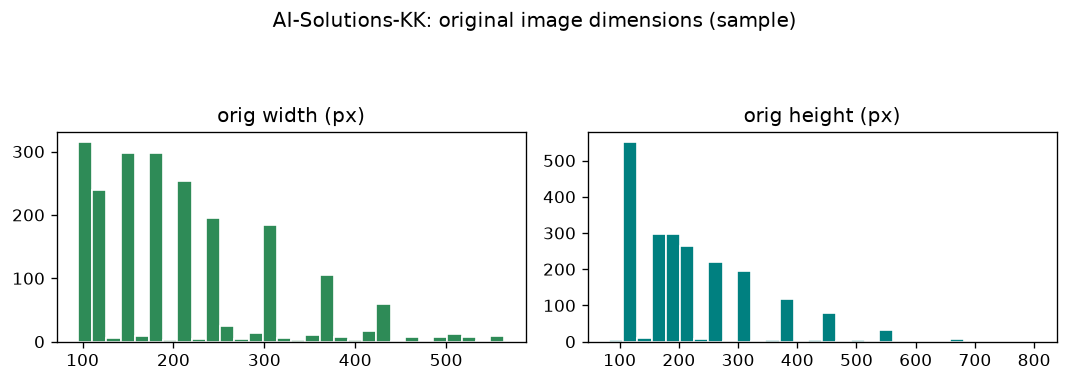}
    \caption{Original image dimensions (sampled).}
    \label{fig:ds-kk-dims}
  \end{subfigure}
  \hfill
  \begin{subfigure}{0.36\linewidth}
    \centering
    \includegraphics[width=\linewidth]{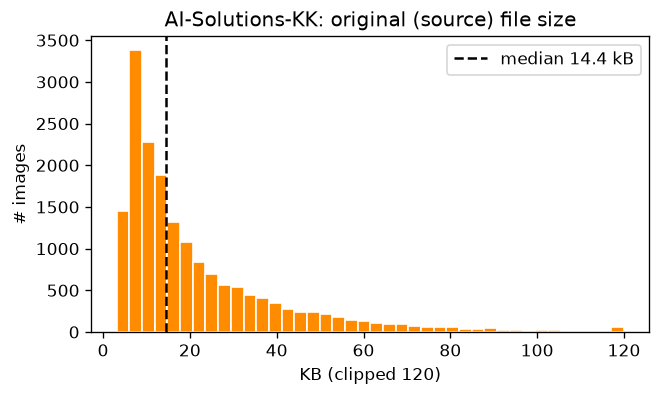}
    \caption{Original file size.}
    \label{fig:ds-kk-filesize}
  \end{subfigure}
  \caption{AI-Solutions-KK source-image statistics before alignment:
  (a)~original width/height distributions and (b)~original file-size distribution.}
  \label{fig:ds-kk-orig}
\end{figure}

Beyond the demographic labels, the source photos vary widely in dimension and file
size, which motivates the controlled alignment step above and contextualizes the
sub-1\,kB target against the original, uncompressed inputs.
Figure~\ref{fig:ds-kk-orig} reports the distributions of original image dimensions and
original (source) file sizes.

%% file: sections/04_protocol.tex
\section{Evaluation Protocol}
\label{sec:protocol}
\tldr{We score every codec at every operating point along two axes that matter for sub-1\,kB face compression: how well a reconstruction preserves the identity signal a matcher needs (verification accuracy) and how faithfully it reproduces the pixels (reconstruction quality), with all comparisons made against a lossless aligned reference and backed by paired significance tests.}

\begin{figure}[t]
  \centering
  \resizebox{\linewidth}{!}{%
  \begin{tikzpicture}[
      >=Stealth, node distance=4mm and 6mm, font=\scriptsize,
      stage/.style={draw=TldrBar, rounded corners, fill=TldrBG, align=center,
        minimum height=11mm, inner sep=3pt, text width=21mm},
      note/.style={align=center, text width=21mm, font=\tiny\itshape}]
    \node[stage] (a) {\textbf{Align}\\ArcFace 5-pt\\template};
    \node[stage, right=of a] (b) {\textbf{Compress}\\hard $\le B$ per\\image};
    \node[stage, right=of b] (c) {\textbf{Embed}\\FR roster};
    \node[stage, right=of c] (d) {\textbf{Score}\\verification\\pairs};
    \node[stage, right=of d] (e) {\textbf{Test}\\paired\\statistics};
    \node[stage, below=of c] (q) {\textbf{Quality}\\full-reference\\metrics};
    \draw[->] (a)--(b); \draw[->] (b)--(c); \draw[->] (c)--(d); \draw[->] (d)--(e);
    \draw[->] (b.south) |- (q.west);
    \node[note, below=1mm of a] {2 datasets\\5 resolutions};
    \node[note, below=1mm of b] {12 configs\\$\{512,1024\}$\,B};
    \node[note, right=2mm of q.east, text width=24mm]
      {PSNR, SSIM, MS-SSIM,\\LPIPS, DISTS};
    \node[note, below=1mm of d] {EER, FNMR@FMR,\\$\Delta$EER};
    \node[note, below=1mm of e] {McNemar, DeLong,\\BH-FDR; Friedman,\\Cliff's $\delta$};
  \end{tikzpicture}}
  \caption{The measurement pipeline. Every codec passes through the identical five
  stages; the grid is 2 datasets $\times$ 12 configurations (10 off-the-shelf codecs
  + Ours in two variants) $\times$ 5 resolutions $\times$ 2 byte budgets, embedded by up to 14 FR
  models. Identity (right branch) and pixel fidelity (lower branch) are always
  reported as separate axes.}
  \label{fig:protocol-pipeline}
\end{figure}
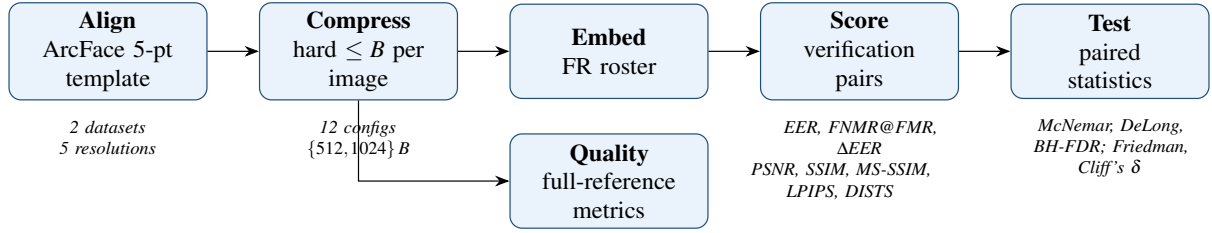

This section fixes the measurement apparatus the rest of the report relies on: the
face-recognition (FR) identity probes, the verification-accuracy metrics computed from
their similarity scores, the reconstruction-quality metrics computed on the pixels, the
resolution/budget grid of operating points, and the statistical machinery for deciding
whether a difference between two conditions is real. Figure~\ref{fig:protocol-pipeline}
summarises the five-stage pipeline and the grid it runs over. All numbers reported here
come from the accuracy and quality pipelines; this section describes only \emph{how} they
are computed, not the values themselves, which appear in the results sections that follow.
Every headline number is also re-derived from those committed outputs into a
machine-readable \emph{findings register}, tracing each quantitative claim to the dataset,
matcher, and operating point that produced it. The artifacts are catalogued in
Section~\ref{sec:artifacts}.

\paragraph{Which table is authoritative for which metric.} Because the report cross-refers
many tables, we state once which is the source of record per metric.
\emph{Identity cosine} (reconstruction vs.\ original): Tables~\ref{tab:codec-cmp-224},
\ref{tab:codec-cmp-112}, \ref{tab:codec-results} and~\ref{tab:codec-results-112}, all measured with the held-out
proprietary \texttt{inno-balanced} embedder at the $112$\,px matcher input (defined below).
Note that the id-cosine \emph{tail} table, Table~\ref{tab:idcos-tail}, instead reports
\texttt{arcface\_antelopev2} percentiles, so its medians differ from the
\texttt{inno-balanced} medians of the codec-comparison tables for the same cell (a
matcher difference, not an inconsistency) --- e.g.\ Ours-ACCURATE at Color~FERET/$112$\,px/$512$\,B
is $0.910$ under \texttt{inno-balanced} (Tables~\ref{tab:codec-cmp-112}
and~\ref{tab:codec-results-112}, the $112$\,px sources; Table~\ref{tab:codec-results}
is the $224$\,px companion and prints $0.797$ for that variant) versus $0.874$ under
ArcFace (Table~\ref{tab:idcos-tail}); each id-cos table
below names its embedder \emph{and its source resolution} in the caption. \emph{Verification accuracy} (EER, FNMR@FMR): the rate--EER
grids (Tables~\ref{tab:rate-eer-cf-arcface}--\ref{tab:rate-eer-kk-topofr}) and the
operating-point FNMR tables (Tables~\ref{tab:frr-summary}--\ref{tab:frr-kk-224}).
\emph{Full-reference image quality} (PSNR/SSIM/MS-SSIM/LPIPS/DISTS): the quality matrix,
Table~\ref{tab:quality-matrix}, is the fullest per-resolution source. The PSNR/SSIM/LPIPS
columns of the codec-comparison tables (Tables~\ref{tab:codec-cmp-224},
\ref{tab:codec-cmp-112}, \ref{tab:codec-results} and~\ref{tab:codec-results-112})
are computed at the codec's native resolution and reconcile with the quality matrix; only the
\emph{identity cosine} in those tables is measured at the $112$\,px matcher input (an earlier
draft computed the codec-comparison fidelity at $112$\,px too, which disagreed with the
quality matrix at $224$\,px; that is fixed).

\subsection{Face-recognition roster}
\tldr{Four embedding models of different families and sizes act as the identity probes, chosen from a wider validated roster of fourteen so that conclusions do not hinge on a single matcher.}

A codec that preserves identity for one matcher may discard exactly the features another
matcher relies on, so we evaluate against a panel of FR models rather than a single
reference. The panel comprises four anchor matchers, selected to span architecture
families, parameter budgets, and training regimes:

\begin{itemize}
  \item \textbf{arcface\_antelopev2} --- a ResNet-100 ArcFace
        model~\cite{deng2019arcface}, our high-capacity reference matcher;
  \item \textbf{lvface\_l} --- a large-variant matcher used as an independent
        high-capacity cross-check \cite{lvface2025};
  \item \textbf{topofr\_r100} --- a ResNet-100 TopoFR matcher~\cite{dan2024topofr}
        used as a second independent high-capacity probe;
  \item \textbf{edgeface\_xs} --- the extra-small EdgeFace
        model~\cite{george2024edgeface}, a deliberately lightweight, edge-deployable
        probe that stresses whether a reconstruction survives a low-capacity matcher.
\end{itemize}

These four anchors are drawn from a wider roster of fourteen models validated on the
aligned reference imagery: \texttt{arcface\_antelopev2}, \texttt{cvlface\_ir101},
\texttt{cvlface\_vit\_b}, \texttt{edgeface\_base}, \texttt{edgeface\_s},
\texttt{edgeface\_xs}, \texttt{edgeface\_xxs}, \texttt{lvface\_b}, \texttt{lvface\_l},
\texttt{lvface\_s}, \texttt{lvface\_t}, \texttt{topofr\_r50}, \texttt{topofr\_r100}, and
\texttt{topofr\_r200}. From this roster we designate four anchors for the headline
results --- \texttt{arcface\_antelopev2}, \texttt{lvface\_l}, \texttt{topofr\_r100}, and
\texttt{edgeface\_xs} --- chosen to span the high-capacity ResNet/ViT families and the
lightweight edge-deployable end of the spectrum; reporting four keeps the result tables
legible while retaining diversity in matcher capacity and inductive bias. Every model consumes the same aligned face crop at the
working resolution, emits a fixed-length embedding, and scores a pair by cosine
similarity. The model identifier is carried through the \texttt{model} column of the
master accuracy grid, so every accuracy figure can be read per matcher.

\paragraph{The held-out identity-cosine metric (\texttt{inno-balanced}).} The
\emph{identity-cosine} figures in the codec-comparison and custom-codec tables
(Tables~\ref{tab:codec-cmp-224}, \ref{tab:codec-cmp-112}, \ref{tab:codec-results}
and~\ref{tab:codec-results-112})
are \emph{not} produced by any of the fourteen roster matchers above. They are computed
with a separate proprietary Innovatrics production embedder, \texttt{inno-balanced} ---
the balanced accuracy/speed operating point of Innovatrics's \{fast, balanced, accurate\}
face-model family, a $512$-dimensional ONNX model. It is \emph{held out of codec
training}: the learned codec is optimised against a differentiable EdgeFace identity loss
(Section~\ref{subsec:codec-side}) and \texttt{inno-balanced} is used only at evaluation,
so its cosine measures generalisation rather than agreement with the training objective.
Being a distinct production model --- not the EdgeFace anchor used in the training loss,
and not one of the four evaluation anchors (ArcFace, LVFace-L, TopoFR-R100, EdgeFace-XS)
--- it is architecturally independent of both. Concretely, ``id-cos'' in those tables is
the median, over a fixed \emph{comparison sample} of $n{=}64$ held-out crops per
cell (the ``held-out set'' referenced in those tables' captions), of the cosine between the
\texttt{inno-balanced} embedding of the reconstruction and of the original $112$\,px crop
(higher is better). It
is distinct from the held-out \emph{verification} matcher \texttt{cvlface\_ir101} of
Table~\ref{tab:heldout-cvlface}: the two are different held-out models, used for the
id-cosine and EER cross-checks respectively.

\subsection{Verification-accuracy metrics}
\tldr{From each matcher's cosine scores over positive and negative pairs we report the equal-error rate, the false non-match rate at three fixed false-match rates, and the change in EER caused by compression relative to the lossless aligned reference.}

Identity preservation is measured in the 1:1 verification setting. For a given dataset,
matcher, resolution, codec, and byte budget, we form a set of genuine (same-identity)
and impostor (different-identity) pairs, score each pair by cosine similarity between the
two embeddings, and sweep a decision threshold to obtain the receiver-operating
characteristic. Three pairing scenarios recur in the report and are named once here:
\emph{compressed-vs-original} (a compressed probe matched against its lossless aligned
reference --- the default asymmetric enrolment case used throughout the benchmark),
\emph{compressed-vs-compressed} (both sides passed through the same codec, used in the
recompression study, Section~\ref{sec:recompression}), and the
\emph{original-vs-original ceiling} (the lossless aligned reference matched against
itself, i.e.\ the \texttt{aligned} baseline that upper-bounds achievable accuracy).
Unless stated otherwise, results are in the compressed-vs-original scenario against a
single global aligned reference per matcher/resolution.

\paragraph{Pair sampling.} We score \emph{all} genuine (mated) pairs, but the full
impostor set is intractable (${\sim}64$\,M pairs on Color~FERET and ${\sim}153.7$\,M on
AI-Solutions-KK; \S\ref{sec:datasets}), so impostor pairs are a single \emph{seeded,
uniform random sample} capped at $5\times10^{6}$ (the \texttt{-{}-sample-nonmated}
parameter, default $5\,000\,000$, drawn once with a fixed seed). The identical impostor
sample is reused across every codec, budget, and matcher, so EER/FNMR differences reflect
the codec and not a change of denominator; when the full impostor count is below the cap
(as on the smaller per-subgroup slices) all impostor pairs are used. This is the sense in
which the protocol is ``exhaustive on mated pairs, fixed-sample on impostor pairs'': the
$\sim$$64$/$153.7$\,M figures in \S\ref{sec:datasets} count the pair \emph{universe}, while
the scored impostor count is the capped $5.0\times10^{6}$. The counts are recorded in the
\texttt{n\_pos} and \texttt{n\_neg} columns of the accuracy grid; for example, the
Color~FERET ArcFace condition uses on the order of $9.6\times10^{4}$ genuine and the capped
$5.0\times10^{6}$ impostor pairs. A handful of cells left from early tiny-pair runs carry a
small \texttt{n\_neg}; the table generators always select the fully-populated row for a
given cell and treat a residual small-$n$ duplicate as superseded.

We summarise each ROC with the following quantities, each a column of the accuracy
grid:

\begin{description}
  \item[EER (\texttt{eer}).] The equal-error rate --- the operating point where the false
        match rate (FMR) equals the false non-match rate (FNMR). A single
        threshold-free scalar; lower is better.
  \item[FNMR at fixed FMR (\texttt{fnmr\_0.01}, \texttt{fnmr\_0.001},
        \texttt{fnmr\_0.0001}).] The false non-match rate at false-match rates of
        $10^{-2}$, $10^{-3}$, and $10^{-4}$. These trace the high-security tail of the
        ROC that the EER alone hides; the $10^{-4}$ point is the most demanding and the
        most sensitive to compression artefacts.
  \item[Reference EER (\texttt{eer\_aligned}) and $\Delta$EER (\texttt{delta\_eer}).] The
        EER of the \emph{lossless aligned} reference for the same matcher/resolution, and
        the signed change in EER induced by compression,
        $\Delta\mathrm{EER}=\mathrm{EER}_{\text{codec}}-\mathrm{EER}_{\text{aligned}}$.
        $\Delta$EER is our headline identity-preservation metric: a value near zero means
        the codec costs essentially no verification accuracy, a positive value quantifies
        identity damage, and a (rare) negative value indicates the compressed image
        scores marginally better than the uncompressed reference for that matcher.
\end{description}

Anchoring every figure to the same lossless aligned reference is what makes the
comparison fair: $\Delta$EER isolates the effect of the codec from the intrinsic
difficulty of the dataset and the matcher. The lossless aligned condition itself appears
as a row in the accuracy grid (with the \texttt{codec} field empty and
\texttt{eer\_aligned} blank), and serves as the zero of the $\Delta$EER axis throughout
the report.

\subsection{Reconstruction-quality metrics}
\tldr{Independently of any matcher, we measure how faithfully each codec reproduces the pixels using a fidelity ladder of five metrics --- PSNR, SSIM, MS-SSIM, LPIPS, and DISTS --- that runs from pixel error to learned perceptual similarity.}

Verification accuracy tells us whether identity survives, but not \emph{how} a codec
fails the pixels, so we also report classical and perceptual reconstruction quality. For
each dataset/codec/resolution/budget cell, the quality pipeline compares each compressed
crop against its lossless aligned source and averages over the cell; the per-cell sample
count is the \texttt{n} column of the quality summary. The five metrics, in
increasing order of perceptual sophistication, are:

\begin{description}
  \item[PSNR (\texttt{psnr}).] Peak signal-to-noise ratio in decibels --- a
        log-scaled mean-squared-error fidelity measure; higher is better.
  \item[SSIM (\texttt{ssim}).] The structural similarity index~\cite{wang2004ssim},
        which scores local luminance, contrast, and structure agreement in $[0,1]$;
        higher is better.
  \item[MS-SSIM (\texttt{ms\_ssim}).] The multi-scale extension of SSIM, aggregating
        structural similarity across a resolution pyramid; higher is better.
  \item[LPIPS (\texttt{lpips}).] The learned perceptual image patch
        similarity~\cite{zhang2018lpips}, a deep-feature perceptual distance; lower is
        better.
  \item[DISTS (\texttt{dists}).] The deep image structure and texture
        similarity~\cite{ding2020dists}, a perceptual distance that balances structure
        and texture and is comparatively tolerant of texture resampling; lower is better.
\end{description}

The two distortion metrics (LPIPS, DISTS) are reported as distances, so for them lower
values indicate better reconstructions, whereas the three fidelity metrics
(PSNR, SSIM, MS-SSIM) are larger-is-better. Crucially, reconstruction quality and
verification accuracy are reported as \emph{separate} axes and never collapsed into one
score: a codec can post strong PSNR while still moving $\Delta$EER, and a perceptually
oriented codec can preserve identity while sacrificing pixel fidelity. Keeping the axes
distinct lets the results sections expose exactly that tension at the sub-1\,kB budget.

\subsection{Operating points}
\tldr{Every codec is evaluated on a grid of two hard byte budgets (1024 and 512 bytes) crossed with five face-crop resolutions (64, 96, 112, 168, and 224 px), with 112\,px as the verification working resolution.}

The central constraint of this study is a \emph{hard} byte budget. We evaluate two
operating points --- $1024$ bytes (the headline sub-1\,kB target) and a more aggressive
$512$ bytes --- recorded in the \texttt{budget} column of both the accuracy grid and the quality summary. Because the
custom codec packs a self-describing container under a strict byte ceiling, the
\emph{achieved} size never exceeds the budget at the resolutions the container is tuned
for ($64$, $112$ and $224$\,px: $100\%$ compliance for both variants on both datasets and
both budgets --- $\ge 99.98\%$ for Ours-ACCURATE at $224$\,px on AI-Solutions-KK --- with
the single exception of Ours-FAST at $224$\,px$/512$\,B); at the two
intermediate resolutions ($96$ and $168$\,px) its rate floor overshoots the ceiling on a
minority of crops, as quantified below. The \texttt{median\_bytes} column of the
accuracy grid records the actual median encoded size per cell, so a reader can verify
that each condition genuinely respects its ceiling rather than averaging over it.

\paragraph{Scoring rule for budget-non-compliant encodes.} JPEG-AI, and the custom codec
at $64$/$112$/$224$\,px,
enforce the ceiling exactly, but a block codec cannot always reach a small budget at a
given resolution (e.g.\ JPEG at $224$\,px/$512$\,B). Our rule is uniform and introduces no
selection bias: each codec is driven to the largest quality/rate setting whose encode is
$\le B$ bytes, and when \emph{no} setting fits we score the smallest available encode
(which then exceeds $B$) rather than dropping the crop --- we never exclude a crop for
being over budget and never re-encode to force compliance. The per-(resolution, budget)
\emph{fit-rate} --- the fraction of crops that actually met the ceiling --- is reported
separately (Section~\ref{sec:budget-compliance}), and an EER/FNMR cell whose fit-rate is
well below $100\%$ must be read together with that fit-rate. The rate--EER and FNMR
tables carry no per-cell budget marker, so we name the non-compliant custom-codec cells
once, here: both variants fit $100\%$ of crops at $64$ and $112$\,px on both datasets and
both budgets, but at the intermediate resolutions Ours-ACCURATE fits only $80/90\%$ of
crops at $96$\,px and $80/90\%$ at $168$\,px on Color~FERET ($79/90\%$ and $79/90\%$ on
AI-Solutions-KK), and Ours-FAST only $74/87\%$ at $96$\,px, $50/88\%$ at $168$\,px and
$80/100\%$ at $224$\,px on Color~FERET ($74/87\%$, $9/87\%$ and $14/100\%$ on KK), quoted
as $512$\,B$/1024$\,B. Those cells are budget \emph{violations} rather than compliant
operating points. None of them carries a headline claim: every $112$\,px result and the
$224$\,px$/1024$\,B headline cell are fully compliant for both variants. Two
coverage caveats follow directly. \emph{(i)}~For the CompressAI learned baselines
(bmshj2018, mbt2018) only crops with a stored compressed file were embedded, so their
$1024$/$512$\,B budget-sweep cells rest on a small, more-compressible subset
($n_{\mathrm{neg}}\approx3.4$\,k) and are high-variance; the apparently non-monotone
mbt2018 budget behaviour in the sweep (EER $2.50\%$ at $1024$\,B but $1.33\%$ at $512$\,B,
\S\ref{sec:budget-floor}) is a coverage/selection artefact of that reduced subset, not
codec behaviour --- the full-coverage $960$/$768$\,B cells ($n_{\mathrm{neg}}=5$\,M) are
monotone and low ($0.13$--$0.21\%$). \emph{(ii)}~In the JPEG-AI timing/decoder study
(Section~\ref{sec:quality}) the reference encoder is rate-matched to its nearest point,
$1108$--$1113$\,B (${\sim}8.5\%$ over the ceiling), so that study's decoder-profile verdict
is read at a slightly non-compliant operating point, as noted there.

Each budget is crossed with five face-crop resolutions --- $64$, $96$, $112$, $168$, and
$224$ pixels --- carried in the \texttt{res} column. Resolution is a first-class variable
because, under a fixed byte budget, raising the resolution spends the same bits over more
pixels: there is a trade-off between spatial detail and per-pixel fidelity that interacts
with the matcher. Unless noted otherwise, \textbf{112\,px is the verification working
resolution}, matching the native input size of the anchor matchers; the other
resolutions support the dedicated resolution study. The full $2\times5$ budget/resolution
grid yields the ten-column rate--EER tables used throughout the benchmark, which are
populated per dataset and matcher in Section~\ref{sec:benchmark}.
The two headline budgets ($1024$/$512$\,B) are crossed with all five resolutions; to test
the sub-kilobyte ``elevated-error floor'' hypothesis we additionally compress the
intermediate budgets $768$ and $960$\,B at $112$\,px (Section~\ref{sec:budget-floor}). The
higher budgets ($1536$/$2048$\,B) and the extra resolutions ($80$/$160$/$256$\,px) remain
descoped to keep the compress-plus-re-embed cost tractable.

\subsection{Statistical methodology}
\tldr{Differences between conditions are tested with paired statistics --- McNemar on match decisions, DeLong on ROC-AUC --- and the resulting $p$-values are corrected for multiple comparisons with the Benjamini--Hochberg false-discovery-rate procedure.}

The result grid is large --- the product of datasets, four matchers, twelve
configurations (ten off-the-shelf codecs plus two custom variants),
five resolutions, and two budgets --- so a difference in a headline metric can easily be
noise, and running many tests inflates the chance of a spurious ``significant'' result.
We therefore (i) use paired tests that exploit the fact that every codec sees the same
pairs over the same source images, and (ii) apply a multiple-comparison correction across
the whole family of tests.

\begin{description}
  \item[McNemar paired test~\cite{mcnemar1947}.] Compares the \emph{match decisions} of
        two conditions on the identical set of pairs at a fixed operating threshold. By
        conditioning on the discordant pairs --- those one condition gets right and the
        other wrong --- it tests whether two codecs differ in verification accuracy while
        controlling for the shared difficulty of the pairs.
  \item[DeLong test~\cite{delong1988}.] Compares two \emph{correlated} ROC curves by
        testing the difference in their areas under the curve (AUC). Because the two ROCs
        are computed from scores on the same pairs, their AUC estimates are correlated;
        DeLong's covariance-based statistic accounts for that correlation, giving a
        threshold-free significance test that complements the threshold-specific McNemar
        test.
  \item[Benjamini--Hochberg FDR correction~\cite{benjamini1995fdr}.] Controls the
        expected proportion of false discoveries across the full family of pairwise tests.
        We report adjusted $p$-values (or, equivalently, the BH-rejection decision at a
        fixed false-discovery rate) so that a claim of ``significantly better'' survives
        the multiplicity of comparisons rather than being an artefact of running hundreds
        of tests.
  \item[Kendall's $W$ (concordance).] Asks whether the \emph{ranking} of codecs is the
        same across matchers. Each matcher ranks the codecs by EER; $W$ measures how much
        those rankings agree, from $0$ (no agreement beyond chance) to $1$ (identical
        orderings). It is the statistic behind the ``backbone-invariant'' claim: a high
        $W$ means a codec chosen on one matcher is the codec you would have chosen on any
        of them.
  \item[Cliff's $\delta$ (effect size).] A non-parametric effect size for codec-vs-codec
        comparisons \emph{over the matcher roster}: the probability that codec $a$ has the
        lower EER on a randomly chosen matcher, minus the reverse, so $\delta=-1$ means
        $a$ wins on every matcher. It ranges $[-1,1]$. We report it alongside the
        significance tests because with fourteen matchers a $p$-value says only that an
        ordering is consistent, not that it is large: $|\delta|$ reads conventionally as
        ${<}0.147$ negligible, ${<}0.33$ small, ${<}0.474$ medium, above that large.
\end{description}

Throughout the report, a stated difference between two codecs (or between a codec and the
lossless aligned reference) is called significant only when it passes the relevant paired
test \emph{after} Benjamini--Hochberg correction at a false-discovery rate of $q=0.05$
(i.e.\ an adjusted McNemar $p<0.05$). The significance pipeline
that produces these tests is \path{scripts/generate_significance.py}.

\subsection{Environment and per-codec settings}
\label{sec:environment}
\tldr{All results were produced on the hardware and software stack in
Table~\ref{tab:environment}, with each codec driven at the settings in
Table~\ref{tab:codec-settings}; the latency claims (\S\ref{sec:quality}) and the
fine-grained cross-codec gaps must be read against this fixed environment, since
speed/effort presets and chroma handling (\S\ref{sec:fairness}) are known confounds.}

The ms-level encode/decode latencies (Section~\ref{sec:quality}) and every cross-codec
comparison depend on hardware and on the exact encoder settings, so we pin both here.
Classical-codec CPU timings are single-core (pinned with \texttt{taskset}); the learned
codecs and JPEG-AI are timed on one GPU. Two settings are deliberate confounds we do not
hide: WebP runs at maximum effort (\texttt{method=6}) whereas AVIF and JPEG~XL use their
library-default speed presets, and chroma subsampling is each library's default (JPEG and
WebP $4{:}2{:}0$). Both bias the \emph{fine} gaps between the modern codecs; the
qualitative rankings (which codecs survive $512$\,B, which collapse) are robust to them,
and Section~\ref{sec:fairness} flags the chroma confound explicitly.

\begin{table}[t]
  \centering\small
  \caption{Hardware and software environment.}
  \label{tab:environment}
  \begin{tabular}{ll}
    \toprule
    Component & Detail \\
    \midrule
    OS & Ubuntu 24.04.1 LTS (Linux) \\
    CPU & 2$\times$ Intel Xeon E5-2683 v4 @ 2.10\,GHz (16C/32T each; 64 threads) \\
    RAM & 503\,GB \\
    GPU & NVIDIA RTX 2080 Ti (11\,GB) --- latency reference; Quadro RTX 6000 (24\,GB) in fleet \\
    Python / PyTorch & 3.11 / 2.10.0 (CUDA 12.8), torchvision 0.25.0 \\
    Codec libs & Pillow 12.2.0, pillow-heif 1.4.0 (libheif/x265 + libaom), pillow-jxl-plugin 1.3.7 (libjxl) \\
    Learned/quality & compressai 1.2.8, onnxruntime-gpu 1.23.2, timm 1.0.24, lpips 0.1.4, pytorch-msssim 1.0.0 \\
    Numerics & numpy 2.4.6, scipy 1.17.0, pandas 3.0.0, scikit-image 0.26.0, OpenCV, matplotlib 3.11.0 \\
    \bottomrule
  \end{tabular}
\end{table}

\begin{table}[t]
  \centering\small
  \caption{Per-codec encoder configuration. ``Quality search'' means the largest
  setting whose encode is $\le B$ bytes is selected per crop (\S\ref{sec:protocol}); the
  search is capped at quality $94$ (not $100$) to skip the near-lossless plateau, where
  quality increments add bytes without changing the reconstruction at these tiny budgets.}
  \label{tab:codec-settings}
  \adjustbox{max width=\textwidth}{%
  \begin{tabular}{lll}
    \toprule
    Codec & Library / backend & Rate control and key settings \\
    \midrule
    JPEG & Pillow / libjpeg & quality search $2$--$94$, \texttt{optimize=True}, default $4{:}2{:}0$ chroma \\
    WebP & Pillow / libwebp & quality search $2$--$94$, \texttt{method=6} (max effort) \\
    AVIF & pillow-heif / libaom (AV1) & quality search $2$--$94$, default speed preset \\
    HEIF & pillow-heif / x265 (HEVC) & quality search $2$--$94$, default preset \\
    JPEG~XL & pillow-jxl / libjxl & quality search $2$--$94$, default effort \\
    JPEG~2000 & Pillow / OpenJPEG & rate mode, ratio $8$--$400{:}1$ binary-searched to the byte budget \\
    JPEG-AI & ITU-T T.840.1 reference SW & analytic bits-per-pixel (bpp) target searched to budget; profile SOP (BOP/HOP in \S\ref{sec:quality}) \\
    JPEG-FzT & reference C++/Qt implementation & face-tuned transform codec at the target budget \\
    Ours-FAST / ACCURATE & this work (PyTorch) & binary search over a frozen 64-entry gain table to $\le B$; ACCURATE adds an EdgeFace-S identity side-stream + refine head \\
    bmshj2018 / mbt2018 & compressai & pretrained hyperprior / joint models, nearest rate point $\le B$ (partial coverage, \S\ref{sec:protocol}) \\
    \bottomrule
  \end{tabular}}
\end{table}

%% file: sections/05_benchmark.tex
\section{Benchmark of Existing Codecs}
\label{sec:benchmark}
\tldr{At $1024$\,B the split is easy: on Color~FERET/ArcFace, JPEG-AI and the modern block codecs (WebP, AVIF, JPEG~XL, HEIF) preserve identity at 112\,px with equal-error rates well under $0.5\%$, while JPEG~2000 is already an order of magnitude worse and plain JPEG is borderline. \emph{At $512$\,B the field re-sorts} --- AVIF, HEIF and JPEG~XL join the collapse, and only WebP, JPEG-AI and the byte-budgeted learned codecs survive. That re-sort, not the $1024$\,B ranking, is the operational result of this section: a codec chosen at $1$\,kB is not the codec to deploy at half that.}

This section evaluates the off-the-shelf codecs at the two hard operating points of
this study, $1024$ and $512$ bytes per aligned crop, before introducing the trained
codec in its own right. We report two complementary views. First, a rate--identity
trade-off measured directly on the reconstructed pixels: identity cosine between the
embedding of the original and of the decoded crop, alongside the standard distortion
metrics PSNR, SSIM~\cite{wang2004ssim} and LPIPS~\cite{zhang2018lpips}. Second, the
end-to-end verification consequence: equal-error rate (EER) on the verification pair
lists, swept across the resolution/budget grid for all four anchor matchers. The first
view tells us how much identity signal survives a single round-trip; the second tells
us whether that surviving signal is enough to keep a deployed matcher's error rate low.

\subsection{Rate-identity trade-off at the byte budget}
\tldr{At $1024$\,B, JPEG-AI leads on identity cosine (${\sim}0.95$) with our codec next; halving the budget to $512$\,B widens the gap sharply and pushes JPEG and JPEG~2000 to near-random identity.}

We first hold the byte budget fixed and compare what each codec returns. Every codec is
driven to land just under the budget, and we report the median over the held-out crops of
identity cosine ($\uparrow$), PSNR ($\uparrow$, dB), SSIM ($\uparrow$) and
LPIPS ($\downarrow$). Table~\ref{tab:codec-cmp-224} reports the $224$\,px source
resolution and Table~\ref{tab:codec-cmp-112} the $112$\,px verification working
resolution, each for Color~FERET and AI-Solutions-KK at both budgets.

At $1024$\,B, JPEG-AI~\cite{ascenso2023jpegai} is the strongest learned-standard codec on
identity: id-cosine $0.958$ on Color~FERET and $0.948$ on KK at $224$\,px. Our trained
codec (Ours-ACCURATE) is the runner-up on Color~FERET at $0.947$, while on KK the two Ours
variants take second and third (Ours-FAST $0.941$, Ours-ACCURATE $0.934$), placing
Ours-ACCURATE between JPEG-AI and AVIF~\cite{aomavif} on both sets; the modern block codecs
WebP~\cite{google2010webp}
and AVIF cluster just below ($0.89$--$0.90$ on Color~FERET). At the $112$\,px verification working
resolution the ranking inverts: Ours-ACCURATE leads JPEG-AI on identity cosine
($0.948$ vs $0.930$ on Color~FERET; $0.938$ vs $0.921$ on KK). Plain JPEG~\cite{wallace1992jpeg} and
JPEG~2000~\cite{taubman2002jpeg2000} collapse at this budget: at $224$\,px on Color~FERET,
JPEG id-cosine $0.610$ and JPEG~2000 $0.585$, i.e.\ around $40\%$ of the identity signal
already lost in a single compression pass.

Notably, our codec wins on perceptual fidelity (LPIPS) even where it trails on PSNR/SSIM:
at $112$\,px$/512$\,B on Color~FERET it reaches LPIPS $0.018$ versus $0.148$ for JPEG-AI (and
$0.111$ vs $0.281$ at the $224$\,px source resolution), reflecting
its perceptual training objective rather than a pixel-MSE objective. Halving the budget to
$512$\,B widens every gap. JPEG-AI retains $0.865$ id-cosine on Color~FERET while JPEG~2000
drops to $0.228$ and JPEG to $0.357$; on KK the legacy codecs are similarly degraded
(JPEG~2000 $0.221$). Our codec holds $0.797$ (Color~FERET) and $0.776$ (KK) at $512$\,B,
ahead of AVIF on Color~FERET. At the $112$\,px working resolution the ordering is broadly
the same but compressed slightly: on Color~FERET the modern codecs and our codec stay above
$0.85$ id-cosine at $1024$\,B, though on AI-Solutions-KK both HEIF and Ours-FAST dip to
$0.844$ (just below $0.85$); JPEG remains usable ($0.851$ Color~FERET, $0.834$ KK) at
$1024$\,B but collapses at $512$\,B (id-cosine $0.069$ Color~FERET).

Which codec is \emph{weakest} depends on the cell, and JPEG-FzT is the reason. At $1024$\,B
it is a mid-field performer --- $0.870$ id-cosine at $224$\,px and $0.862$ at $112$\,px on
Color~FERET, ahead of Ours-FAST at the working resolution --- but it degrades faster than any
other codec as the budget halves, because its F-transform stage fixes a downsampling ratio
before the inner JPEG ever sees the rate target. At $224$\,px$/512$\,B it is, on Color~FERET,
the worst codec in the table on \emph{every} fidelity column and on identity ($0.104$, below
JPEG~2000's $0.228$; on KK it ties JPEG~2000 on identity at $0.221$ and is worst on PSNR and
SSIM, JPEG~2000 taking the worst LPIPS); at $112$\,px$/512$\,B the ordering
instead runs JPEG ($0.069$) $<$ JPEG~2000 ($0.193$) $<$ HEIF ($0.309$) $<$ JPEG-FzT
($0.466$). So JPEG~2000 is
the most consistently poor codec across the grid, but it is not the floor everywhere: at the
full source resolution and the tight budget, JPEG-FzT is.

\begin{table}[t]
  \centering
  \caption{Single-pass rate--identity trade-off at $224$\,px (medians on the held-out set), at the $1024$\,B and $512$\,B budgets, for the ten-codec roster (seven classical/transform codecs, JPEG-AI, and the two Ours variants). id-cos is identity cosine to the uncompressed crop, measured with the held-out proprietary \texttt{inno-balanced} embedder at the $112$\,px matcher input over a fixed comparison sample of $n{=}64$ held-out crops per cell (\S\ref{sec:protocol}), so small id-cos differences between neighbouring codecs are within sampling noise; PSNR/SSIM/LPIPS are \textbf{native-$224$ full-reference fidelity} from the quality matrix (Table~\ref{tab:quality-matrix}), so this table reconciles with it (its fidelity columns were previously computed at $112$\,px, which disagreed at $224$\,px). Per (dataset, budget) block the \colorbox{green!25}{best} value in each column is shaded green and the \colorbox{red!22}{worst} red (LPIPS is lower-is-better). JPEG-AI's native-$224$ fidelity is now measured on \emph{both} datasets (AI-Solutions-KK: PSNR $31.2$/$28.5$\,dB, SSIM $0.910$/$0.856$ at $1024$/$512$\,B); the CompressAI neural baselines \emph{were} compressed and scored at $224$\,px (Color~FERET only, on their $300$-crop subset) but carry no stored held-out id-cos at this resolution, so they are omitted from this table. Source: \texttt{codec\_comparison/res224} (id-cos) + \texttt{quality\_summary} (fidelity).}
  \label{tab:codec-cmp-224}
  {\scriptsize\input{tables/codec_comparison_224.tex}}
\end{table}

\begin{table}[t]
  \centering
  \caption{Single-pass rate--identity trade-off at the $112$\,px verification working resolution (medians over $n{=}64$ held-out crops per cell; id-cosine from the held-out proprietary \texttt{inno-balanced} embedder, \S\ref{sec:protocol}). Same columns and green/red best/worst shading as Table~\ref{tab:codec-cmp-224}; codecs without a stored held-out id-cos at this resolution are omitted. Source: \texttt{codec\_comparison/res112}.}
  \label{tab:codec-cmp-112}
  {\scriptsize\input{tables/codec_comparison_112.tex}}
\end{table}

\begin{figure}[t]
  \centering
  \includegraphics[width=\linewidth]{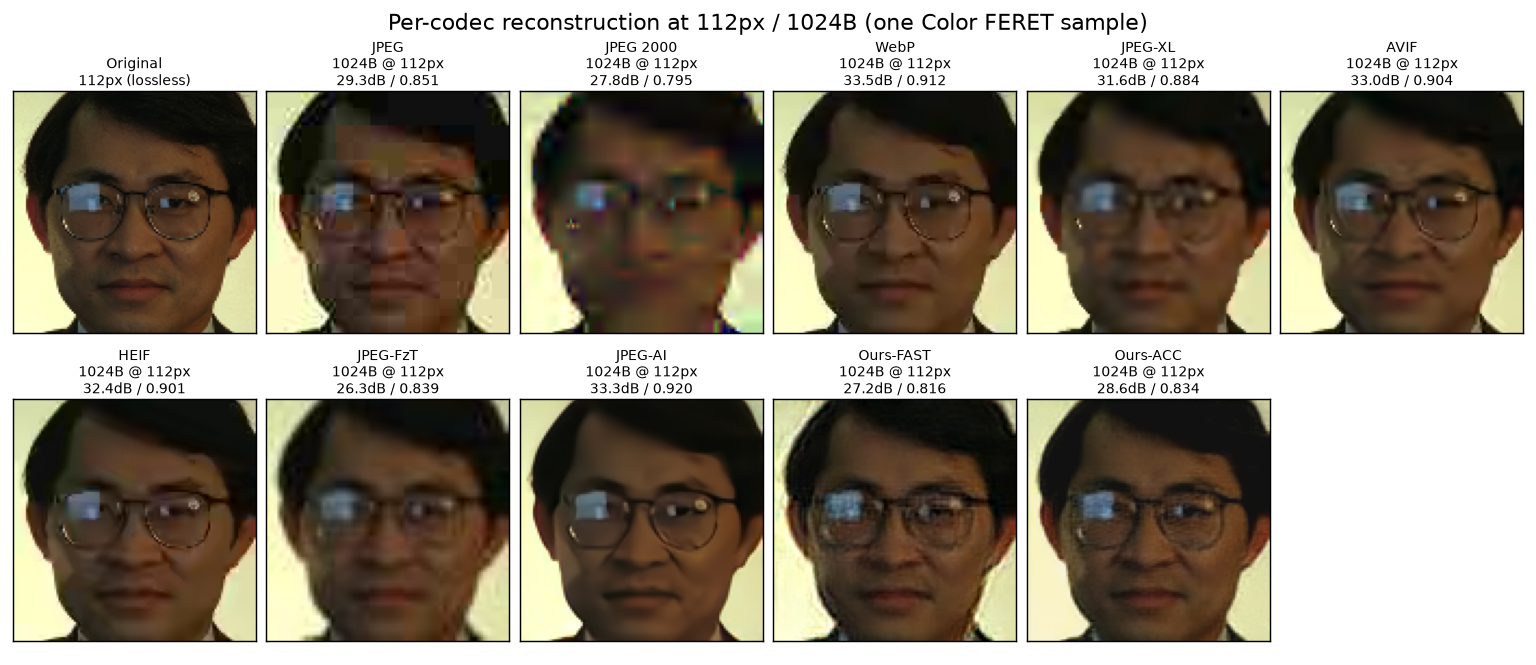}
  \caption{Per-codec reconstruction at the $1$\,kB operating point ($112$\,px), one
  Color~FERET sample, including the learned codecs. JPEG and JPEG\,2000 block and ring;
  the modern transform codecs and JPEG-AI stay clean; Ours-FAST and Ours-ACCURATE
  reconstruct recognisable, colour-correct faces. Tile captions give the dataset-median
  PSNR/SSIM of each codec's cell.}
  \label{fig:cmp-1kb}
\end{figure}

Figure~\ref{fig:cmp-1kb} isolates the $1$\,kB operating point across codecs, showing what
each one returns for the same byte budget.

\subsection{Equal-error rate across the grid}
\tldr{Sweeping EER over five resolutions and two budgets, the modern block codecs and JPEG-AI keep $112$\,px EER well under $0.5\%$ at $1024$\,B on Color~FERET/ArcFace, whereas JPEG inflates it a few-fold and JPEG~2000 by one to two orders of magnitude.}

The pixel-level trade-off above predicts, but does not establish, verification behaviour;
a deployed matcher is judged by its error rate. Tables~\ref{tab:rate-eer-cf-arcface}
through~\ref{tab:rate-eer-cf-topofr} give the EER (\%) grid on Color~FERET for the four
anchor matchers, with columns spanning the five resolutions ($64/96/112/168/224$\,px) at
both budgets and a no-compression \emph{aligned (ref)} baseline row. The corresponding KK
grids are in Tables~\ref{tab:rate-eer-kk-arcface}
through~\ref{tab:rate-eer-kk-topofr}.

The $112$\,px/$1024$\,B column tells the headline story. On Color~FERET/ArcFace
\cite{deng2019arcface}, the aligned reference EER is $0.04\%$; WebP ($0.09\%$),
AVIF ($0.09\%$), JPEG~XL~\cite{alakuijala2019jpegxl} ($0.19\%$), HEIF ($0.18\%$) and
JPEG-AI ($0.09\%$) all stay well under $0.5\%$ at this budget, i.e.\ within striking
distance of the uncompressed matcher. No cell of this anchor grid is exactly $0.00\%$; the
only exactly-zero EER cells anywhere in the grid are three CompressAI-baseline cells scored
on the $300$-crop Color~FERET subset (bmshj2018$\,\times\,$ArcFace at $64$\,px, and
mbt2018$\,\times\,$CVLFace-IR101 at $96$ and $224$\,px). A $0.00\%$ entry (and a bootstrap CI
of $0.00$--$0.00$) means \emph{no genuine/impostor score overlap in this finite sample}, not
error-free recognition; such cells are sample-limited and should be read as ``below the
resolution of the $105$/$994$-identity test set.'' Plain JPEG ($0.25\%$) is borderline at $1024$\,B but
JPEG~2000 ($3.23\%$) is already an order of magnitude worse, and JPEG-FzT
($0.24\%$) is competitive only at the higher resolutions. The collapse becomes
unambiguous at $512$\,B: JPEG jumps to $33.35\%$ and JPEG~2000 to $19.50\%$ EER at
$112$\,px, while WebP ($0.46\%$) and JPEG-AI ($0.27\%$) remain low. The pattern repeats,
at uniformly higher absolute EER, on the harder in-the-wild KK set.

\begin{table}[t]
  \centering
  \caption{Equal-error rate (\%) across the resolution/budget grid, Color~FERET, ArcFace (antelopev2). \emph{aligned (ref)} is the uncompressed baseline; it is \textbf{budget-independent} (no compression) and is therefore listed once per resolution --- the single value applies to both the $1024$ and $512$\,B columns, so the $512$\,B reference is not missing. ``--'' marks an untested cell. $^\dagger$Ours-FAST is over budget at $224$\,px/$512$\,B and at some $96$/$168$\,px cells (its rate floor overshoots the target on $12$--$50\%$ of Color~FERET crops at those cells), so read those Ours-FAST entries as budget violations. \shadelegend}
  \label{tab:rate-eer-cf-arcface}
  \adjustbox{max width=\textwidth}{\input{tables/rate_eer_colorferet_arcface_antelopev2.tex}}
\end{table}

\begin{table}[t]
  \centering
  \caption{Equal-error rate (\%) across the resolution/budget grid, Color~FERET,
  EdgeFace-XS~\cite{george2024edgeface}. Gaps read as in
  Table~\ref{tab:rate-eer-cf-arcface}: the \emph{aligned (ref)} row is budget-independent,
  so its blank $512$\,B cells repeat the $1024$\,B value rather than marking missing data,
  and ``--'' marks an untested cell. \shadelegend}
  \label{tab:rate-eer-cf-edgeface}
  \adjustbox{max width=\textwidth}{\input{tables/rate_eer_colorferet_edgeface_xs.tex}}
\end{table}

\begin{table}[t]
  \centering
  \caption{Equal-error rate (\%) across the resolution/budget grid, Color~FERET, LVFace-L.
  Gaps read as in Table~\ref{tab:rate-eer-cf-arcface}: the \emph{aligned (ref)} row is
  budget-independent, so its blank $512$\,B cells repeat the $1024$\,B value rather than
  marking missing data, and ``--'' marks an untested cell. \shadelegend}
  \label{tab:rate-eer-cf-lvface}
  \adjustbox{max width=\textwidth}{\input{tables/rate_eer_colorferet_lvface_l.tex}}
\end{table}

\begin{table}[t]
  \centering
  \caption{Equal-error rate (\%) across the resolution/budget grid, Color~FERET, TopoFR-R100.
  Gaps read as in Table~\ref{tab:rate-eer-cf-arcface}: the \emph{aligned (ref)} row is
  budget-independent, so its blank $512$\,B cells repeat the $1024$\,B value rather than
  marking missing data, and ``--'' marks an untested cell. \shadelegend}
  \label{tab:rate-eer-cf-topofr}
  \adjustbox{max width=\textwidth}{\input{tables/rate_eer_colorferet_topofr_r100.tex}}
\end{table}

\begin{table}[t]
  \centering
  \caption{Equal-error rate (\%) across the resolution/budget grid, AI-Solutions-KK, ArcFace (antelopev2). As in the Color~FERET tables, \emph{aligned (ref)} is budget-independent and listed once per resolution. The only blank cells in the KK tables are JPEG-AI at $64$, $96$ and $168$\,px: JPEG-AI was compressed in-the-wild at the two headline resolutions ($112$ and $224$\,px) only, because its reference encoder costs ${\approx}3$\,s per crop (Table~\ref{tab:speed}) and the KK set is $17{,}534$ crops per cell --- the intermediate resolutions would have cost tens of GPU-hours for a codec whose resolution trend is already established on Color~FERET, where it \emph{is} measured at all five. Those cells are absent by design, not pending. \shadelegend}
  \label{tab:rate-eer-kk-arcface}
  \adjustbox{max width=\textwidth}{\input{tables/rate_eer_kk_arcface_antelopev2.tex}}
\end{table}

\begin{table}[t]
  \centering
  \caption{Equal-error rate (\%) across the resolution/budget grid, AI-Solutions-KK, EdgeFace-XS~\cite{george2024edgeface}. Blank JPEG-AI cells as in Table~\ref{tab:rate-eer-kk-arcface}. \shadelegend}
  \label{tab:rate-eer-kk-edgeface}
  \adjustbox{max width=\textwidth}{\input{tables/rate_eer_kk_edgeface_xs.tex}}
\end{table}

\begin{table}[t]
  \centering
  \caption{Equal-error rate (\%) across the resolution/budget grid, AI-Solutions-KK, LVFace-L. Blank JPEG-AI cells as in Table~\ref{tab:rate-eer-kk-arcface}. \shadelegend}
  \label{tab:rate-eer-kk-lvface}
  \adjustbox{max width=\textwidth}{\input{tables/rate_eer_kk_lvface_l.tex}}
\end{table}

\begin{table}[t]
  \centering
  \caption{Equal-error rate (\%) across the resolution/budget grid, AI-Solutions-KK, TopoFR-R100. Blank JPEG-AI cells as in Table~\ref{tab:rate-eer-kk-arcface}. \shadelegend}
  \label{tab:rate-eer-kk-topofr}
  \adjustbox{max width=\textwidth}{\input{tables/rate_eer_kk_topofr_r100.tex}}
\end{table}

The grids include the off-the-shelf codecs and the two CompressAI learned
baselines (bmshj2018~\cite{balle2018hyperprior}, mbt2018~\cite{minnen2018joint}) where
those were run; the latter are now scored at \emph{all five} resolutions and both budgets,
but on Color~FERET only (on their $300$-crop subset), so they are absent from the four KK
grids entirely. The trained codec
appears as \emph{Ours-FAST} and \emph{Ours-ACC} in all eight rate--EER grids across the
full $64$--$224$\,px sweep at both budgets. Its EER falls monotonically with resolution, as
expected: on Color~FERET/ArcFace at $1024$\,B, Ours-ACC drops from $1.36\%$ at $64$\,px to
$0.56\%$ ($96$\,px), $0.14\%$ ($112$\,px), $0.12\%$ ($168$\,px) and $0.07\%$ at $224$\,px --
within striking distance of the $0.03\%$ aligned reference at that resolution -- while
Ours-FAST tracks just behind ($2.72\%\!\to\!0.09\%$ over the same sweep, $0.34\%$ at
$112$\,px). Halving the budget to $512$\,B costs little at the working resolutions
(Ours-ACC $0.22\%$ and Ours-FAST $0.55\%$ at $112$\,px). By $224$\,px both variants are on
par with the low-EER block-codec group and JPEG-AI. On LVFace-L the same pattern holds at a
higher floor: Ours-ACC runs from $2.18\%$ ($64$\,px) down to $0.19\%$ ($224$\,px), reaching
$0.44\%$ at $112$\,px, and Ours-FAST from $4.02\%$ to $0.24\%$. The clean Ours embeddings
are complete for every anchor, including Ours-ACCURATE\,$\times$\,TopoFR-R100 (Color~FERET
$112$\,px EER $0.14\%$ at $1024$\,B, $0.23\%$ at $512$\,B): both Ours variants are covered
under all four matchers.
We return to this operating-point comparison -- the deployment-facing one -- in
Section~\ref{sec:frr-far}.

\subsection{Operating-point accuracy: FNMR at fixed FMR}
\label{sec:frr-far}
\tldr{This is the deployment-facing comparison and the headline result for choosing a
codec: fix the false-match rate (FMR --- the older literature calls it FAR) and read the
false-non-match rate (FNMR, older name FRR) there. At the $1024$\,B budget JPEG-AI leads on both datasets, with WebP, AVIF and Ours-ACCURATE clustered behind it (Ours-ACCURATE second in-the-wild); at the aggressive
$512$\,B budget most classical codecs collapse, and the byte-budgeted learned codecs
(JPEG-AI and both Ours variants) degrade most gracefully -- at the $112$\,px working resolution Ours-ACCURATE is in
fact the best codec at $512$\,B on both datasets (at the $224$\,px source resolution the
ordering inverts and WebP leads, \S\ref{sec:frr-far-224}).}

\begin{figure}[t]
  \centering
  \includegraphics[width=\linewidth]{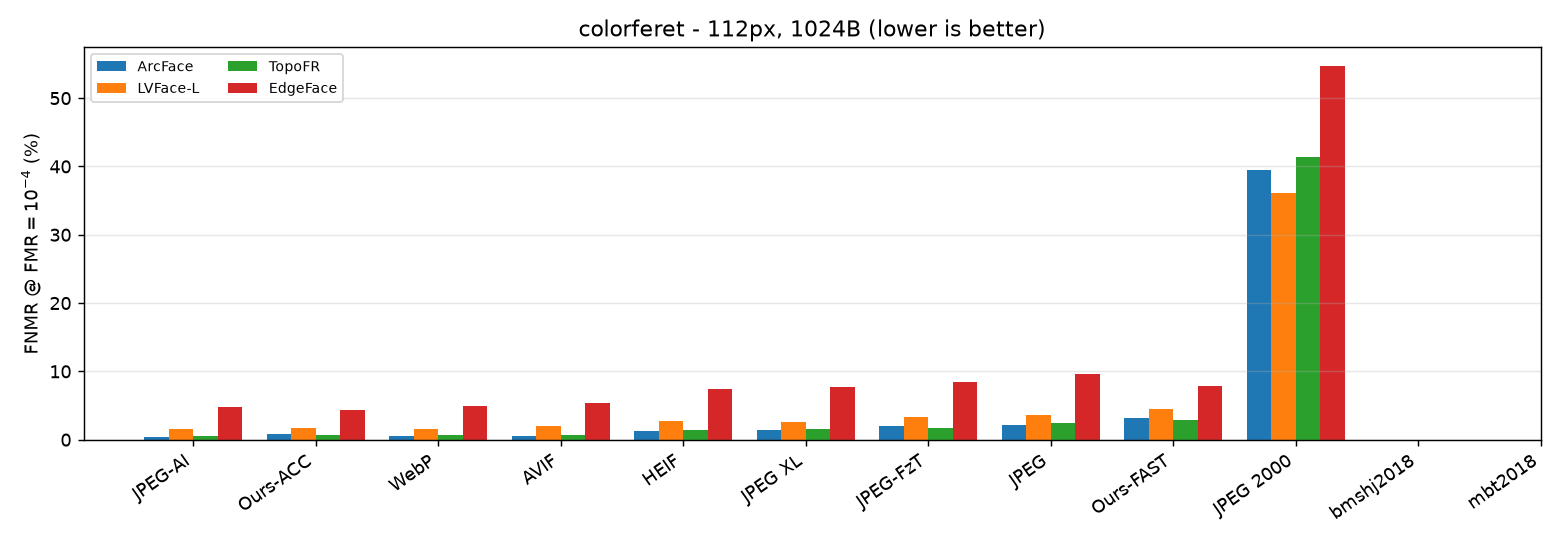}
  \caption{Color~FERET: FNMR at FMR$=10^{-4}$ per codec and matcher, $112$\,px/$1024$\,B
  (lower is better). All four anchor matchers are shown, with the learned codec's clean
  cells populated throughout (tabulated in Table~\ref{tab:frr-cf-1024}).}
  \label{fig:frr-cf}
\end{figure}

\begin{figure}[t]
  \centering
  \includegraphics[width=\linewidth]{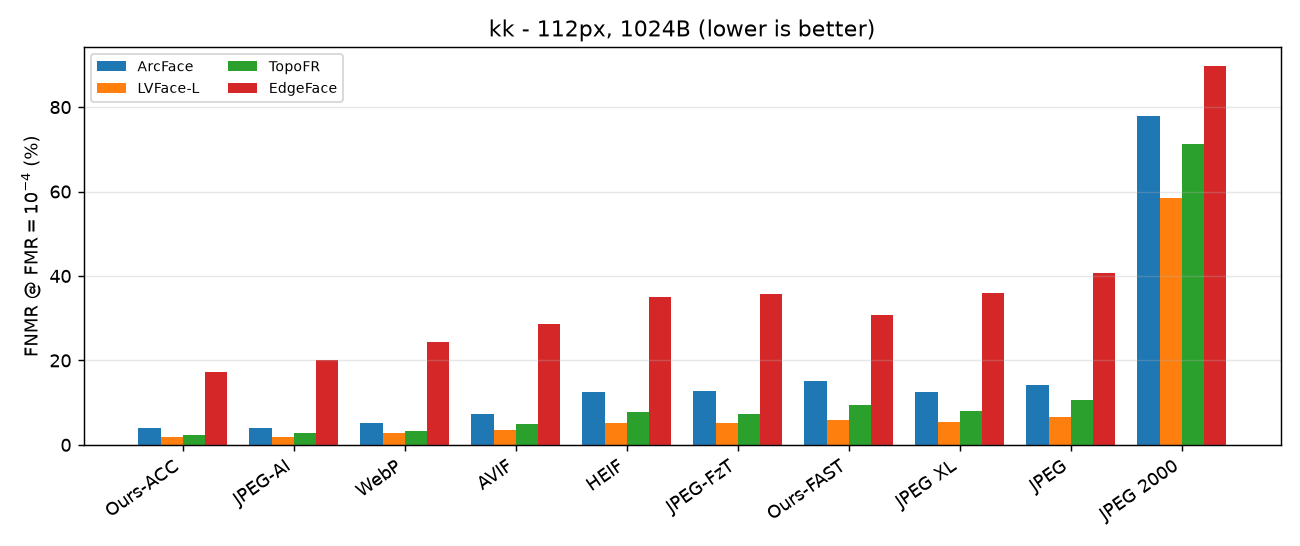}
  \caption{AI-Solutions-KK: FNMR at FMR$=10^{-4}$ per codec and matcher, $112$\,px/
  $1024$\,B.}
  \label{fig:frr-kk}
\end{figure}

\begin{figure}[t]
  \centering
  \includegraphics[width=\linewidth]{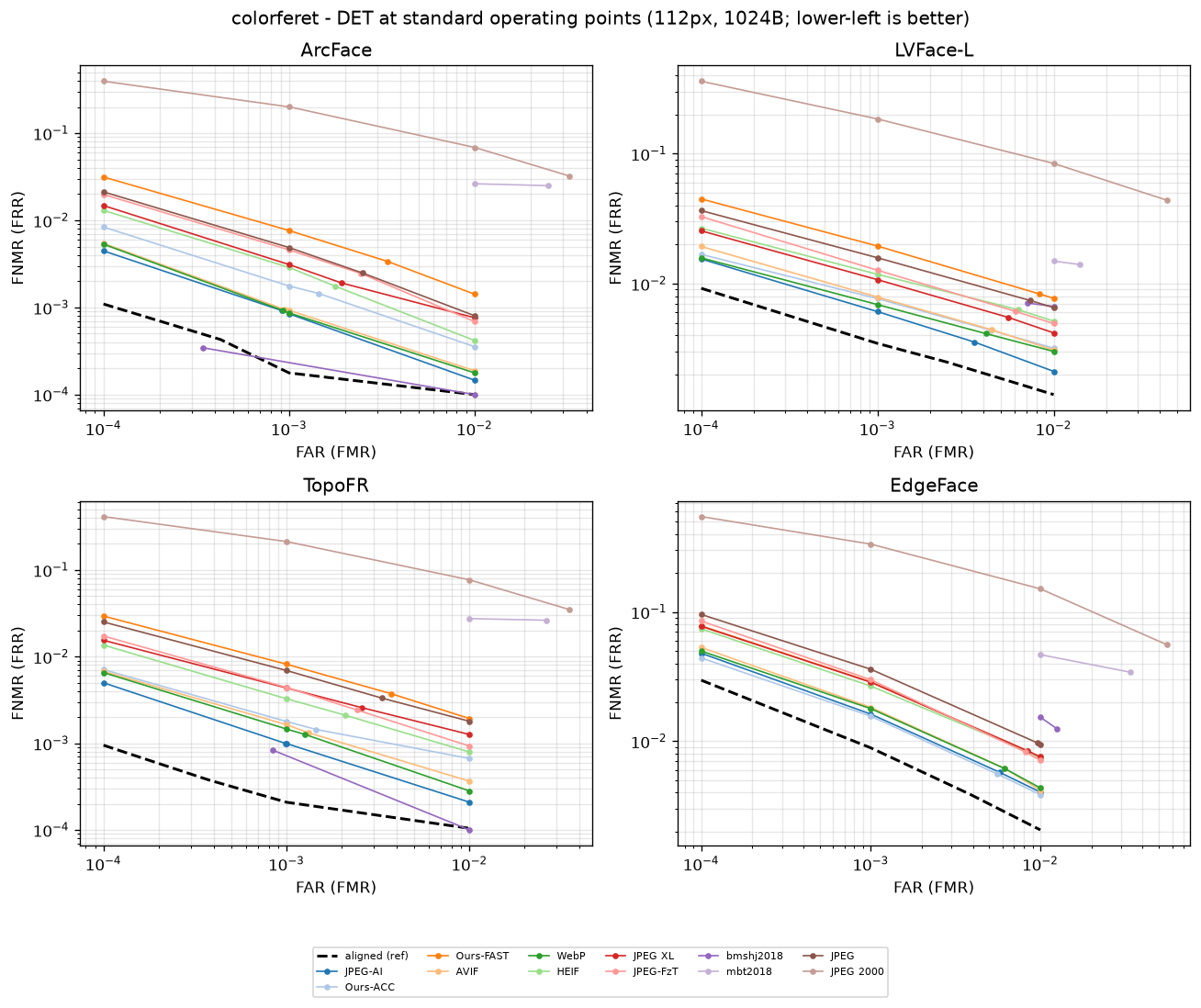}
  \caption{Color~FERET DET curves (FNMR vs FMR, log--log; lower-left is better) at
  $112$\,px/$1024$\,B, one panel per matcher. Each curve passes through the EER point and
  the FNMR at FMR$=10^{-2}/10^{-3}/10^{-4}$ (a continuous curve would require re-scoring the
  raw pair cosines); the dashed black line is the lossless reference. JPEG~2000 is the
  upper (worst) curve under every matcher; the learned codec is shown under all four
  anchors, its clean cells populated throughout.}
  \label{fig:det-cf}
\end{figure}

\begin{figure}[t]
  \centering
  \includegraphics[width=\linewidth]{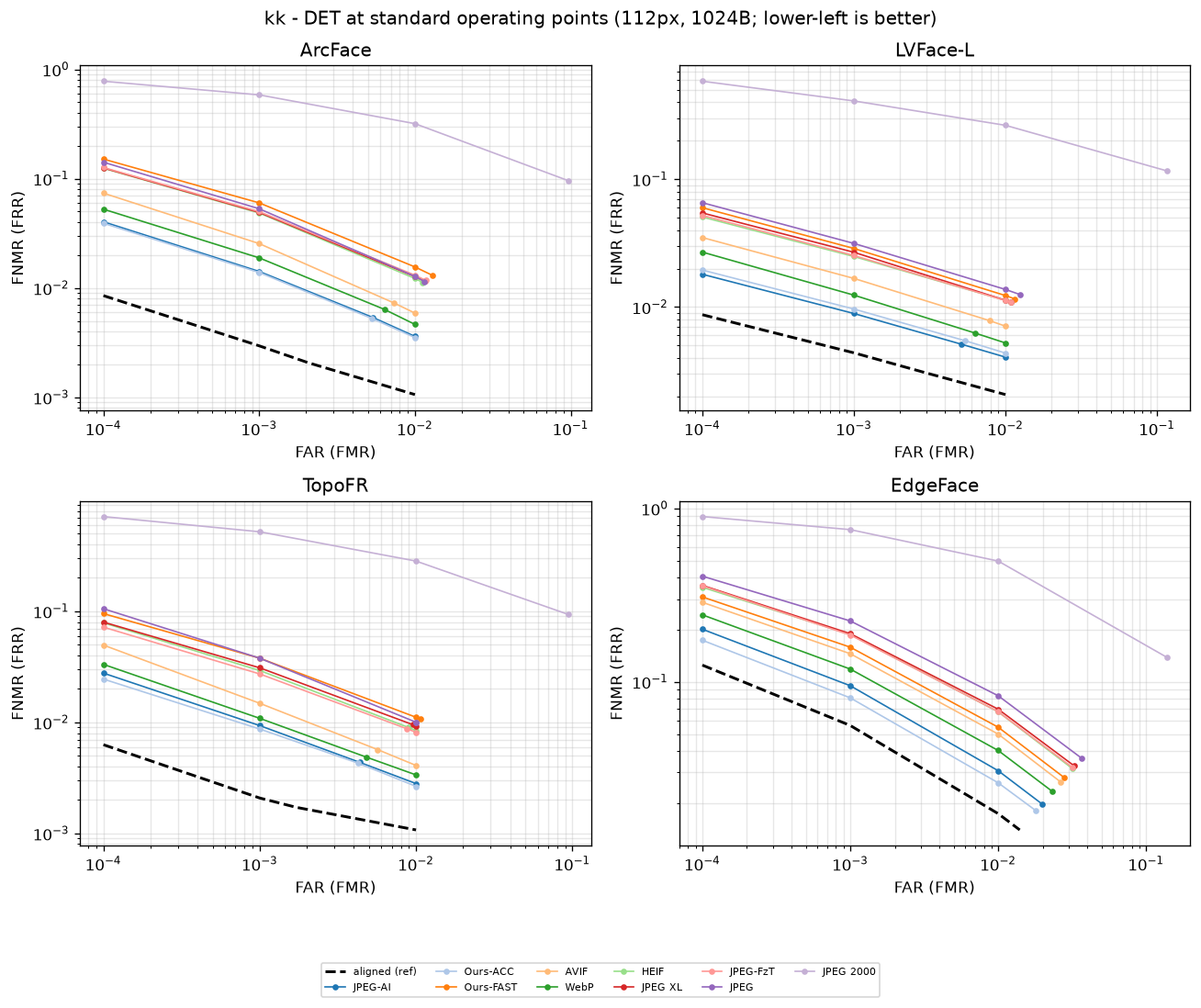}
  \caption{AI-Solutions-KK DET curves at $112$\,px/$1024$\,B, one panel per matcher.}
  \label{fig:det-kk}
\end{figure}

The rate--EER grid fixes the break-even threshold; an operational system instead fixes a
false-match rate (e.g.\ FMR$=10^{-3}$ or $10^{-4}$) and is judged by the false rejections
it incurs there. This is the metric that decides which codec to deploy.
Table~\ref{tab:frr-summary} ranks the codecs by the cross-matcher mean FNMR at
FMR$=10^{-4}$, averaged over two anchors of differing backbone family, ArcFace and
LVFace-L. (All four anchors now carry a complete set of operating points for every codec
that has an estimable FNMR$@10^{-4}$; the two-anchor average is retained for comparability
with the earlier revisions of this table.) Tables~\ref{tab:frr-cf-1024}--\ref{tab:frr-kk-512} give the full
per-matcher FNMR at FMR$=10^{-3}$ and $10^{-4}$ for both datasets and budgets.
Figures~\ref{fig:frr-cf}--\ref{fig:frr-kk} plot that operating point as grouped bars, and
Figures~\ref{fig:det-cf}--\ref{fig:det-kk} give the corresponding DET curves (FNMR vs
FMR across the standard operating points), the full graphical form of the tables.

\begin{table}[t]
  \centering
  \caption{\textbf{Which codec preserves identity best.} Cross-matcher mean FNMR at
  FMR$=10^{-4}$ (\%, lower is better), $112$\,px, averaged over ArcFace and LVFace-L (two
  anchors of differing backbone family; all four anchors now have complete clean coverage
  for every codec with an estimable FNMR$@10^{-4}$).
  CF~=~Color~FERET, KK~=~AI-Solutions-KK. The CompressAI neural baselines
  bmshj2018/mbt2018 are absent from this table for two distinct reasons: on
  AI-Solutions-KK they were never run, while on Color~FERET they \emph{are} scored (EER)
  but only on a $300$-crop subset, whose ${\approx}3.4$\,k impostor pairs cannot resolve
  FMR$=10^{-4}$ at all --- so their FNMR at this operating point is non-estimable rather
  than merely unmeasured. \shadelegend}
  \label{tab:frr-summary}
  \adjustbox{max width=\textwidth}{\input{tables/frr_far_summary.tex}}
\end{table}

\begin{table}[t]
  \centering \footnotesize
  \caption{Color~FERET, $112$\,px/$1024$\,B: FNMR (\%) at FMR$=10^{-3}$ and $10^{-4}$ per
  matcher. ``aligned'' is the lossless reference; every codec\,$\times$\,matcher cell is
  populated for the ten full-corpus codecs, including the learned ones under all four
  anchors. The two CompressAI baselines carry an EER at this cell but no estimable FNMR at
  $10^{-3}$ or $10^{-4}$ (their $300$-crop subset yields only ${\approx}3.4$\,k impostor
  pairs), so they do not appear here. Bracketed values on the
  $10^{-4}$ column are $95\%$ subject-level bootstrap CIs (here and in
  Tables~\ref{tab:frr-kk-1024},~\ref{tab:frr-cf-512},~\ref{tab:frr-kk-512}). \shadelegend}
  \label{tab:frr-cf-1024}
  \adjustbox{max width=\textwidth}{\input{tables/frr_far_colorferet_1024.tex}}
\end{table}

\begin{table}[t]
  \centering \footnotesize
  \caption{AI-Solutions-KK, $112$\,px/$1024$\,B: FNMR (\%) at FMR$=10^{-3}$ and $10^{-4}$
  per matcher. The roster is two rows shorter than its Color~FERET counterpart
  (Table~\ref{tab:frr-cf-1024}): the CompressAI baselines bmshj2018/mbt2018 were run on
  Color~FERET only, so they are omitted here rather than dashed. \shadelegend}
  \label{tab:frr-kk-1024}
  \adjustbox{max width=\textwidth}{\input{tables/frr_far_kk_1024.tex}}
\end{table}

\FloatBarrier

\begin{figure}[t]
  \centering
  \includegraphics[width=\linewidth]{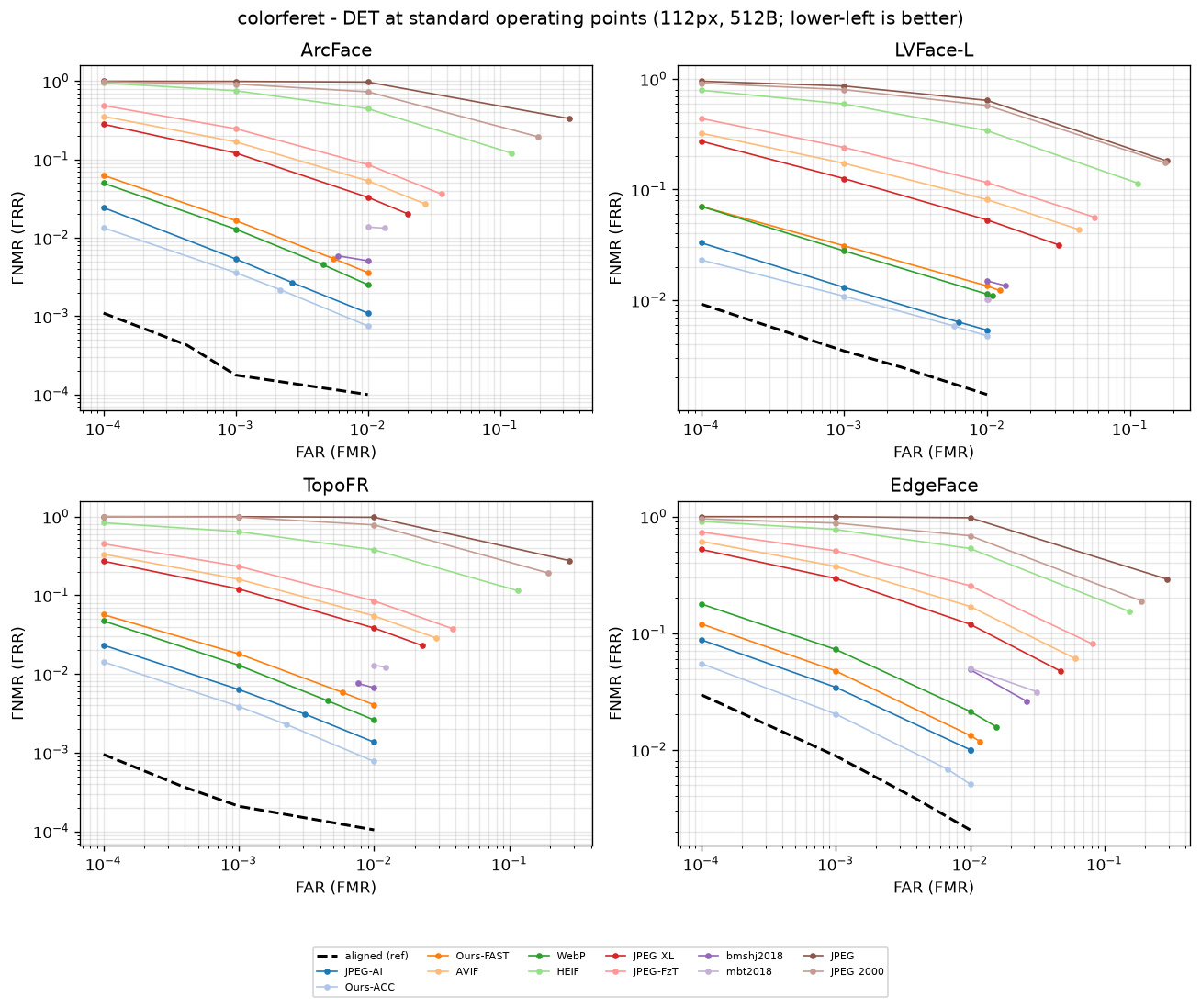}
  \caption{Color~FERET DET curves at the harder $112$\,px/$512$\,B operating point, one
  panel per matcher (same construction as Figure~\ref{fig:det-cf}). The $512$\,B separation
  is where the codecs diverge: JPEG, JPEG~2000, HEIF, AVIF and JPEG~XL climb into the
  upper-right collapse region while JPEG-AI, WebP and Ours-ACCURATE stay in the lower-left
  usable band.}
  \label{fig:det-cf-512}
\end{figure}

\begin{figure}[t]
  \centering
  \includegraphics[width=\linewidth]{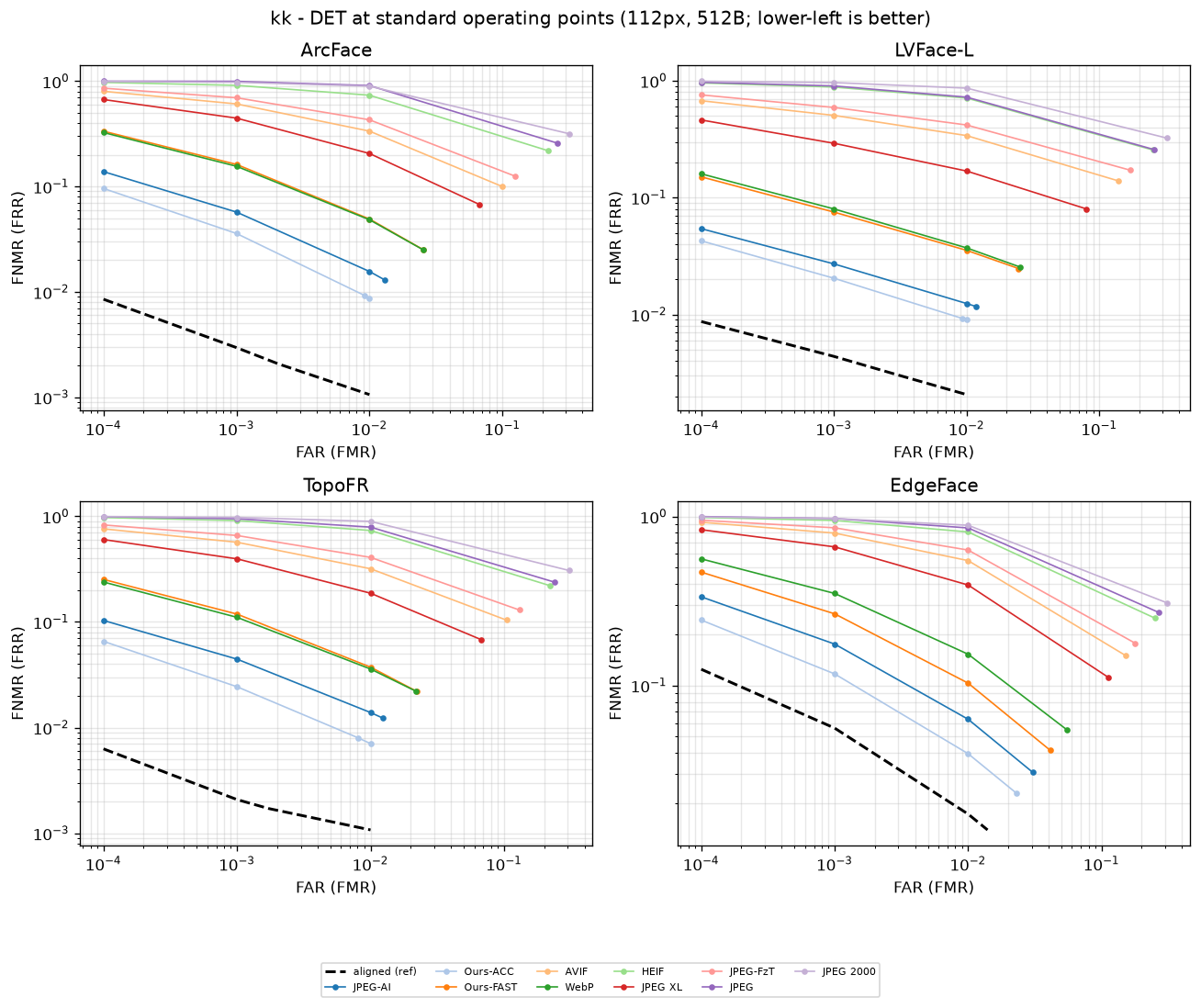}
  \caption{AI-Solutions-KK DET curves at $112$\,px/$512$\,B, one panel per matcher. The
  in-the-wild set spreads the curves further, but the same ordering holds --- only the
  byte-budgeted learned codecs (and WebP, partially) remain in the usable band.}
  \label{fig:det-kk-512}
\end{figure}

At $1024$\,B the field is closely clustered. On Color~FERET, JPEG-AI (mean FNMR$@10^{-4}$ $1.00\%$),
WebP ($1.06\%$) and AVIF ($1.24\%$) lead; Ours-ACCURATE is 4th of the field at $1.26\%$ (behind JPEG-AI and WebP, and effectively tied with AVIF), ahead of
JPEG-FzT and Ours-FAST, while legacy JPEG sits at $2.89\%$ and JPEG~2000 is catastrophic
($37.8\%$). On AI-Solutions-KK --- with JPEG-AI scored on the in-the-wild grid ---
JPEG-AI \emph{leads} (mean FNMR$@10^{-4}$ $2.92\%$) with Ours-ACCURATE ($2.93\%$)
effectively tied for the lead, WebP ($3.98\%$) and AVIF ($5.44\%$) next, so the
$1024$\,B identity ordering holds in-the-wild. The decisive separation appears at $512$\,B
(Tables~\ref{tab:frr-cf-512}--\ref{tab:frr-kk-512}): AVIF, HEIF, JPEG~XL, JPEG and
JPEG~2000 mostly collapse (FNMR$@10^{-4}$ from $\sim$$28\%$ to $\sim$$99\%$); WebP degrades
but stays out of the collapse band ($6.0\%$ on CF, $24.3\%$ on KK), while
JPEG-AI ($2.86\%$ on CF) and Ours-ACCURATE degrade the most gracefully. \emph{At the
$112$\,px working resolution} Ours-ACCURATE is the \emph{best} codec at $512$\,B on both
datasets --- $6.93\%$ vs WebP's $24.28\%$ on AI-Solutions-KK (a $17.35$\,pp gap, a factor of
$3.5$), and $1.83\%$ vs JPEG-AI's
$2.86\%$ on Color~FERET --- with JPEG-AI the graceful runner-up in-the-wild too ($9.6\%$ on
KK/$512$, next after Ours-ACCURATE). The KK/$512$ gap is well outside sampling noise despite
the $105$-identity set: the subject-level bootstrap $95\%$ CIs do not overlap on either
anchor (Ours-ACCURATE $[7.5,12.6]$ vs WebP $[26.9,38.1]$ on ArcFace; $[3.1,5.5]$ vs
$[12.0,19.6]$ on LVFace-L); these $95\%$ subject-bootstrap CIs are recorded in
the accuracy grid and printed in brackets on the FNMR@$10^{-4}$ column of the
operating-point tables (Tables~\ref{tab:frr-cf-1024}--\ref{tab:frr-kk-512}). This
ordering is resolution-conditional, however: at
$224$\,px$/512$\,B it inverts (\S\ref{sec:frr-far-224}), where WebP leads and Ours-ACCURATE
falls to $5$th on CF and to the worst of the budget-compliant \emph{modern} codecs on KK
(Table~\ref{tab:frr-kk-224}). This is the core operating-point evidence: a
kilobyte is comfortable for several codecs, but at half a kilobyte---at $112$\,px---only the
learned codecs (JPEG-AI and ours) keep verification usable; the $512$\,B DET panels
(Figures~\ref{fig:det-cf-512}--\ref{fig:det-kk-512}) show the same separation graphically.

\begin{table}[t]
  \centering \footnotesize
  \caption{Color~FERET, $112$\,px/$512$\,B: FNMR (\%) at FMR$=10^{-3}$ and $10^{-4}$ per
  matcher. The harder budget separates graceful (JPEG-AI, Ours-ACCURATE, WebP) from
  collapsing codecs. \shadelegend}
  \label{tab:frr-cf-512}
  \adjustbox{max width=\textwidth}{\input{tables/frr_far_colorferet_512.tex}}
\end{table}

\begin{table}[t]
  \centering \footnotesize
  \caption{AI-Solutions-KK, $112$\,px/$512$\,B: FNMR (\%) at FMR$=10^{-3}$ and $10^{-4}$
  per matcher. Ours-ACCURATE is the lowest-FNMR codec at this operating point. As in
  Table~\ref{tab:frr-kk-1024}, the CompressAI baselines bmshj2018/mbt2018 were run on
  Color~FERET only and are omitted. \shadelegend}
  \label{tab:frr-kk-512}
  \adjustbox{max width=\textwidth}{\input{tables/frr_far_kk_512.tex}}
\end{table}

\paragraph{Coverage caveat.} The learned codec's \emph{clean} embeddings are complete
for every codec\,$\times$\,matcher cell, including Ours-ACCURATE\,$\times$\,TopoFR-R100. An
earlier decode/matcher path returned a non-finite reconstruction on this one pairing;
running the decode on the GPU path resolved it, and the cell is populated throughout. All
four anchors now carry every codec at this operating point; the cross-matcher ranking
(Table~\ref{tab:frr-summary}) is computed over the ArcFace and LVFace-L pair, so it stays
apples-to-apples; Ours-ACCURATE's
behaviour on a backbone independent of its EdgeFace side-stream is further confirmed on the
held-out CVLFace-IR101 matcher (Section~\ref{subsec:codec-heldout}).

\FloatBarrier

\subsection{Operating points at the 224\,px source resolution}
\label{sec:frr-far-224}
\tldr{The operating-point comparison above is at the $112$\,px working resolution; here we
repeat it at the $224$\,px source resolution. At $1024$\,B the ordering is unchanged---modern
transform/learned codecs low, JPEG and JPEG~2000 an order of magnitude worse, Ours-ACCURATE
level with JPEG-AI. \emph{The $512$\,B story does not carry over}, though: at
$224$\,px$/512$\,B Ours-ACCURATE's advantage disappears---WebP now leads and Ours-ACCURATE is
$5$th on Color~FERET and the worst of the budget-compliant \emph{modern} codecs on
AI-Solutions-KK
(FNMR$@10^{-4}$ $30.4\%$ vs WebP $6.6\%$, ArcFace)---so the ``best at $512$\,B'' result is
specific to the $112$\,px working resolution. Ours-FAST additionally cannot reliably reach
$512$\,B at $224$\,px (its rate floor over-fills the budget: only $80\%$ of Color~FERET and
$14\%$ of AI-Solutions-KK crops fit).}

Because a deployer may store the crop at its native $224$\,px rather than downsample to the
$112$\,px working point, Tables~\ref{tab:frr-cf-224}--\ref{tab:frr-kk-224} give the same
FNMR-at-fixed-FMR comparison at $224$\,px for both budgets. The qualitative picture is
identical to $112$\,px: at $1024$\,B Ours-ACCURATE is level with JPEG-AI on Color~FERET (FNMR
$@10^{-4}$ $0.30\%$ vs JPEG-AI $0.29\%$ on ArcFace) and ahead of WebP/AVIF, while on
AI-Solutions-KK JPEG-AI leads the field (FNMR$@10^{-4}$ $2.45\%$ vs Ours-FAST $2.94\%$),
mirroring its $112$\,px in-the-wild lead; JPEG and
JPEG~2000 sit two orders of magnitude worse ($\sim$$28$--$31\%$ on Color~FERET,
$\sim$$58$--$69\%$ on AI-Solutions-KK). At $512$\,B, however, the
$112$\,px ordering does \emph{not} reproduce: the block codecs still collapse, but so does
Ours-ACCURATE---it falls to $5$th on Color~FERET ($4.17\%$ ArcFace, behind WebP $2.60$,
JPEG-AI $2.73$, JPEG~XL $3.03$ and even Ours-FAST $3.23$) and to the worst of the
budget-compliant \emph{modern} codecs on
AI-Solutions-KK ($30.43\%$ ArcFace, versus WebP $6.64$, JPEG~XL $11.45$, JPEG-AI $14.24$;
only the budget-compliant legacy codecs JPEG~2000 $99.12$ and JPEG-FzT $99.94$ are worse).
So the ``Ours-ACCURATE is best at $512$\,B'' result of \S\ref{sec:frr-far} is specific to the
$112$\,px working resolution and does not hold at the $224$\,px source resolution---WebP is
the most budget-robust codec at $224$\,px$/512$\,B. The other
$224$\,px-specific effect is that Ours-FAST, lacking the identity side-stream that absorbs
the high-resolution rate floor, cannot reliably hold $512$\,B at $224$\,px: it emits a clean
frame that overshoots the budget on $20\%$ of Color~FERET and $86\%$ of AI-Solutions-KK
crops (Section~\ref{sec:codec}). Its $224$\,px/$512$\,B cells
\emph{are} populated in the tables, but must be read as budget violations rather than
compliant results. Ours-ACCURATE, whose
trained low-$\lambda$ levels and identity-only fallback keep it within budget, is unaffected.

\begin{table}[t]
  \centering \footnotesize
  \caption{Color~FERET, $224$\,px: FNMR (\%) at FMR$=10^{-3}$/$10^{-4}$ per matcher, at
  $1024$\,B (top) and $512$\,B. Same layout as the $112$\,px tables; all four anchors have
  complete $224$\,px clean coverage, and JPEG-AI is present at \emph{both} budgets. The
  CompressAI neural baselines are omitted although they \emph{were} compressed and scored
  at $224$\,px: on their $300$-crop subset FMR$=10^{-3}$/$10^{-4}$ is not estimable, so they
  have no value to print in these columns. $^\dagger$Ours-FAST is over
  budget at $224$\,px/$512$\,B on $20\%$ of Color~FERET crops, so those
  Ours-FAST results are budget violations rather than compliant operating points. \shadelegend}
  \label{tab:frr-cf-224}
  \adjustbox{max width=\textwidth}{\input{tables/frr_far_colorferet_1024_224.tex}}\\[4pt]
  \adjustbox{max width=\textwidth}{\input{tables/frr_far_colorferet_512_224.tex}}
\end{table}

\begin{table}[t]
  \centering \footnotesize
  \caption{AI-Solutions-KK, $224$\,px: FNMR (\%) at FMR$=10^{-3}$/$10^{-4}$ per matcher, at
  $1024$\,B (top) and $512$\,B. JPEG-AI is included (compressed on AI-Solutions-KK at
  $224$\,px); the CompressAI neural baselines were run on Color~FERET only --- they are
  absent from every AI-Solutions-KK cell, not just this resolution --- and are omitted.
  $^\dagger$Ours-FAST is over
  budget at $224$\,px/$512$\,B on $86\%$ of AI-Solutions-KK crops (median $595$\,B); read those
  Ours-FAST cells as budget violations. \shadelegend}
  \label{tab:frr-kk-224}
  \adjustbox{max width=\textwidth}{\input{tables/frr_far_kk_1024_224.tex}}\\[4pt]
  \adjustbox{max width=\textwidth}{\input{tables/frr_far_kk_512_224.tex}}
\end{table}

\paragraph{Best configuration per codec.} Table~\ref{tab:frr-summary} fixes $112$\,px and
\S\ref{sec:frr-far-224} fixes $224$\,px, but each codec has its own resolution sweet spot, and
Table~\ref{tab:deployment-guidance} implicitly mixes them. Table~\ref{tab:best-config}
resolves this by reporting, for each codec and budget, the resolution that \emph{minimises}
its cross-matcher mean FNMR$@10^{-4}$ (Color~FERET, ArcFace$+$LVFace-L), so codecs are
compared at their respective best operating points rather than at one imposed resolution.
Two things follow. First, at $1024$\,B the byte-budgeted learned codecs are best at the
\emph{full} $224$\,px source (Ours-ACCURATE $0.57\%$@224, Ours-FAST $0.69\%$@224, JPEG-AI
$0.71\%$@224), whereas the classical block codecs peak at coarser resolutions
(WebP $0.93\%$@96, AVIF $1.05\%$@168) --- the learned codecs use the extra pixels the
classical codecs cannot afford under the hard budget. Second, at $512$\,B the sweet spots
diverge sharply: JPEG-AI is best at $168$\,px ($2.17\%$) and Ours-ACCURATE at the $112$\,px
working point ($1.83\%$), while WebP/AVIF/JPEG~XL only become competitive by dropping to a
coarser trade-off at $224$\,px ($3.2$--$5.8\%$). This defuses the ``JPEG-AI was compared at
its worse resolution'' objection: even each codec at its \emph{own} optimum, the
byte-budgeted learned codecs lead both budgets, and JPEG/JPEG~2000 remain well behind
($2.5$/$27.9\%$ at $1024$\,B, rising to $53.8$/$90.3\%$ at $512$\,B).

\begin{table}[t]
  \centering \small
  \caption{Best-configuration summary: each codec at its \emph{own} optimal resolution per
  budget. Cell is the minimum cross-matcher mean FNMR$@10^{-4}$ (\%, over ArcFace and
  LVFace-L on Color~FERET) across the five resolutions, with the minimising resolution
  (px) after the ``@''. \colorbox{green!25}{Best}/\colorbox{red!22}{worst} codec per
  column, as elsewhere. Codecs are compared at
  their respective sweet spots, so this table is resolution-agnostic (unlike
  Tables~\ref{tab:frr-summary} at $112$\,px and \S\ref{sec:frr-far-224} at $224$\,px).
  The CompressAI baselines bmshj2018/mbt2018 are absent: they were size-measured at all
  five resolutions but \emph{scored} only at $112$\,px, so they have no five-resolution
  sweep to minimise over. Table~\ref{tab:frr-summary} reports them at that single
  resolution.}
  \label{tab:best-config}
  \adjustbox{max width=\textwidth}{\input{tables/best_config.tex}}
\end{table}

\FloatBarrier

\subsection{The sub-1\,kB error floor}
\label{sec:budget-floor}
\tldr{Adding $768$ and $960$\,B rows between the $512$ and $1024$\,B budgets exposes a
sharp error floor: the classical codecs are flat and accurate at $\ge 960$\,B but their
EER spikes below $\sim$$800$\,B (AVIF $24\times$, JPEG $120\times$ from $960$ to $512$\,B),
matching the NIST rationale for a sub-1\,kB target. The byte-budgeted learned codecs break
this floor --- Ours-ACCURATE stays within $0.14$--$0.22\%$ EER across the whole
$512$--$1024$\,B range --- which is precisely the operating regime a hard-byte-budget
codec is designed for.}

\begin{figure}[t]
  \centering
  \includegraphics[width=\linewidth]{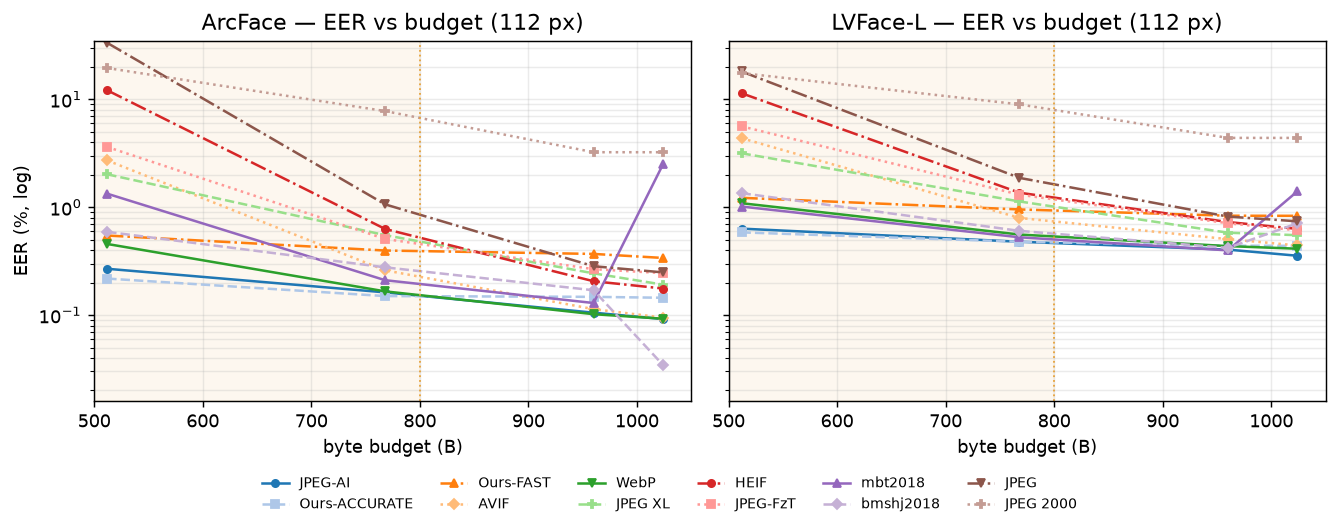}
  \caption{EER vs.\ byte budget at $112$\,px (log scale), ArcFace and LVFace-L, over the
  same full roster as Table~\ref{tab:h4-budget}. The shaded
  band marks the sub-$800$\,B floor zone: classical codecs (AVIF, JPEG) rise steeply into
  it, whereas the byte-budgeted Ours-ACCURATE/Ours-FAST stay nearly flat across the whole
  range. This is the elevated-error-floor test. A codec is drawn only where it has a
  stored EER at two or more budgets; all twelve codecs clear that bar here, each with an
  EER at all four of $512$/$768$/$960$/$1024$\,B. No cell in this sweep is exactly $0\%$,
  so none has to be displaced to the axis floor of the log scale.}
  \label{fig:h4-curve}
\end{figure}

The NIST 2D-barcode work motivates the $\sim$1\,kB target by an \emph{elevated-error
floor}: below a few hundred bytes verification degrades sharply. Our headline grid uses
only $\{512,1024\}$\,B, which cannot resolve where that floor sits. We therefore compress
two intermediate budgets, $768$ and $960$\,B, at the $112$\,px working resolution and
re-score the anchors. Table~\ref{tab:h4-budget} and Figure~\ref{fig:h4-curve} give the
resulting EER-vs-budget curve on Color~FERET (ArcFace shown; the other anchors agree).

Two things stand out. First, \textbf{the floor is real for the general-purpose codecs}: AVIF
and JPEG are flat and near-perfect at $\ge 960$\,B (EER $\le 0.3\%$) but collapse below
$\sim$$800$\,B --- AVIF rises from $0.11\%$ at $960$\,B to $2.71\%$ at $512$\,B
($\sim$$24\times$), and JPEG from $0.28\%$ to $33.4\%$ ($\sim$$120\times$). This is precisely
the sub-1\,kB elevated-error floor the target is chosen to stay above, measured
directly here (supported). WebP is the most floor-resistant classical codec ($0.10\% \to 0.46\%$), while
JPEG~2000 is poor at every budget. Second, and more useful for this project,
\textbf{the byte-budgeted learned codecs flatten the floor}: Ours-ACCURATE holds
$0.14\%$/$0.15\%$/$0.15\%$/$0.22\%$ across $1024$/$960$/$768$/$512$\,B and Ours-FAST
$0.34$--$0.55\%$, because a codec optimised to hit a hard ceiling degrades gracefully
rather than falling off a rate cliff. The intermediate budgets thus both confirm the NIST
floor and quantify the custom codec's advantage precisely where it matters --- in the
sub-$800$\,B regime the classical codecs cannot serve.

\begin{table}[t]
  \centering \small
  \caption{Sub-1\,kB budget sweep on Color~FERET, ArcFace, $112$\,px: verification EER (\%)
  at $512$/$768$/$960$/$1024$\,B. Classical codecs spike below $\sim$$800$\,B; the
  byte-budgeted learned codecs stay flat. JPEG-AI and the CompressAI neural baselines
  (bmshj2018, mbt2018) are included here and in the companion curve
  (Figure~\ref{fig:h4-curve}), which is drawn from the same roster by the same generator.
  \shadelegend{} Codecs tied at the printed precision are all shaded. $^{\ddagger}$ marks a
  cell scored on the CompressAI baselines' $300$-crop subset ($2{,}548$ mated pairs against
  $95{,}839$ elsewhere in the column); it is a per-cell mark --- those two codecs are
  full-support at $768$/$960$\,B --- and marked cells are excluded from the best/worst
  comparison.}
  \label{tab:h4-budget}
  \adjustbox{max width=\textwidth}{\input{tables/h4_budget.tex}}
\end{table}

\FloatBarrier

\subsection{Per-matcher consistency}
\tldr{The codec ranking is stable across the four anchor matchers; absolute EER shifts with matcher capacity, but the same codecs win and the same two legacy codecs collapse.}

A ranking that depends on the matcher would be of little use to a deployer. Comparing the
ArcFace grid (Table~\ref{tab:rate-eer-cf-arcface}) against
EdgeFace-XS~\cite{george2024edgeface} (Table~\ref{tab:rate-eer-cf-edgeface}), LVFace-L
(Table~\ref{tab:rate-eer-cf-lvface}) and TopoFR-R100 (Table~\ref{tab:rate-eer-cf-topofr})
shows the same ordering under all four anchors: the modern block codecs and JPEG-AI form
the low-EER group, JPEG is budget-sensitive, and JPEG~2000 is consistently the worst.
Absolute EER tracks matcher capacity and dataset difficulty -- higher on KK than on
Color~FERET, and higher for lighter matchers -- but the \emph{relative} ranking does not
invert. Concretely, at $112$\,px/$1024$\,B on Color~FERET: the modern block codecs and
JPEG-AI cluster low (EER $\le 1.0\%$ for WebP, AVIF, JPEG~XL, HEIF and JPEG-AI under each
anchor --- the wider $14$-model roster, Table~\ref{tab:model-effect}, pushes all five above
$1\%$ under the weakest backbone, EdgeFace-XXS (JPEG-AI $1.08$, WebP $1.14$, AVIF $1.15$,
JPEG~XL $1.46$, HEIF $1.48$), and HEIF alone above $1\%$ under LVFace-T); JPEG is
budget-sensitive (climbing from $0.25$--$0.97\%$ at $1024$\,B to $18$--$33\%$ at $512$\,B);
and JPEG~2000 is the worst codec under each anchor ($3.23\%$ ArcFace, $4.39\%$ LVFace-L,
$5.57\%$ EdgeFace-XS, $3.50\%$ TopoFR-R100) --- and worst under thirteen of the full
$14$-model roster, mbt2018 edging past it only under CVLFace-IR101 ($2.56$ vs $2.38\%$).
The KK grids
(Tables~\ref{tab:rate-eer-kk-arcface},~\ref{tab:rate-eer-kk-edgeface},~\ref{tab:rate-eer-kk-lvface}
and~\ref{tab:rate-eer-kk-topofr}) reproduce this ordering at uniformly higher absolute EER.

\begin{figure}[t]
  \centering
  \includegraphics[width=\linewidth]{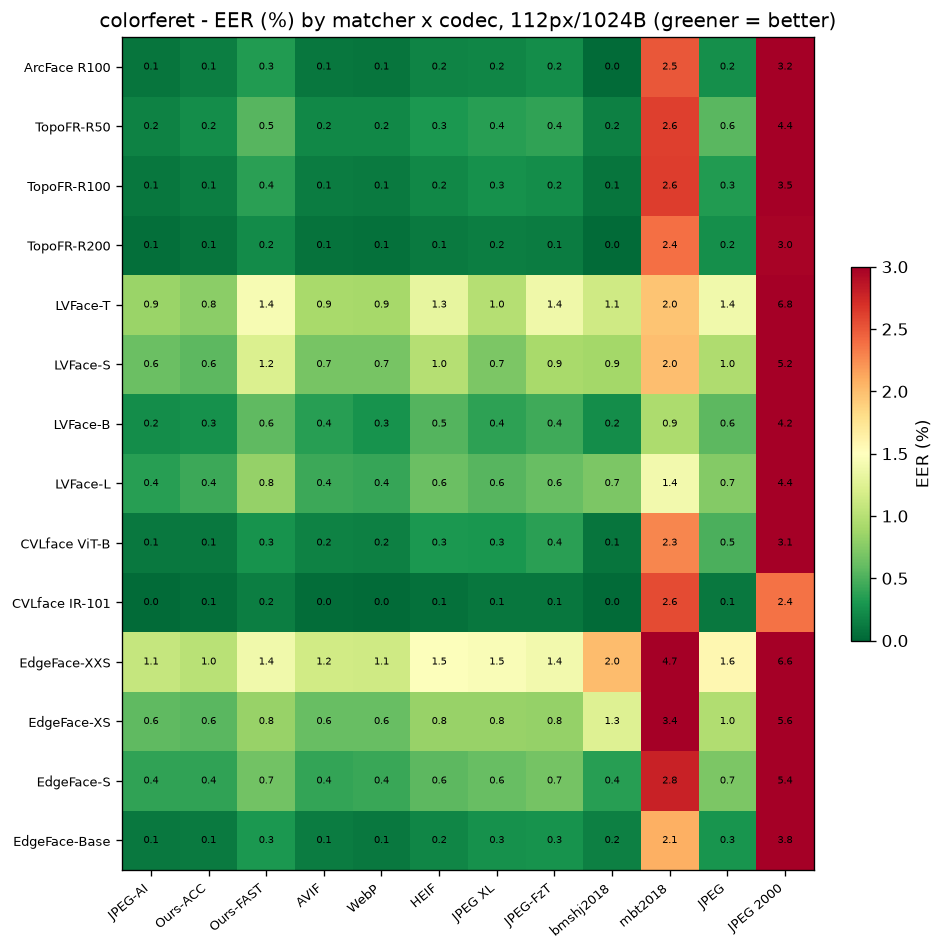}
  \caption{Model effect (Color~FERET, $112$\,px/$1024$\,B): verification EER (\%) for every
  roster matcher (rows, grouped by family/strength) against every codec (columns); greener is
  lower EER. The ranking is consistent down each column; EER tracks matcher strength down the
  rows. The learned-codec columns are populated for every matcher.}
  \label{fig:model-effect}
\end{figure}

\begin{figure}[t]
  \centering
  \includegraphics[width=\linewidth]{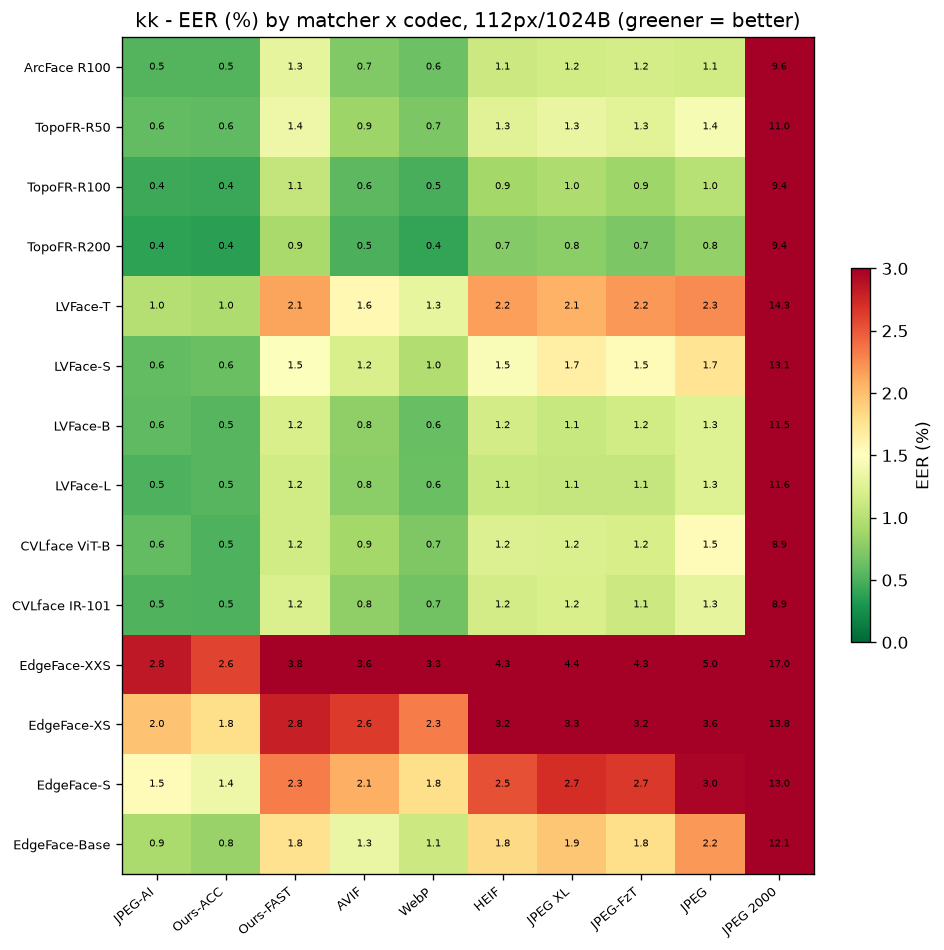}
  \caption{Model effect on AI-Solutions-KK ($112$\,px/$1024$\,B); same axes as
  Figure~\ref{fig:model-effect}. The codec ordering
  reproduces at uniformly higher absolute EER. The in-the-wild panel is now complete: all
  ten codecs run on this dataset --- Ours-ACCURATE, Ours-FAST and JPEG-AI included --- are
  populated across all fourteen matchers. Only the two CompressAI baselines are missing,
  never having been run on AI-Solutions-KK.}
  \label{fig:model-effect-kk}
\end{figure}

\begin{figure}[t]
  \centering
  \includegraphics[width=\linewidth]{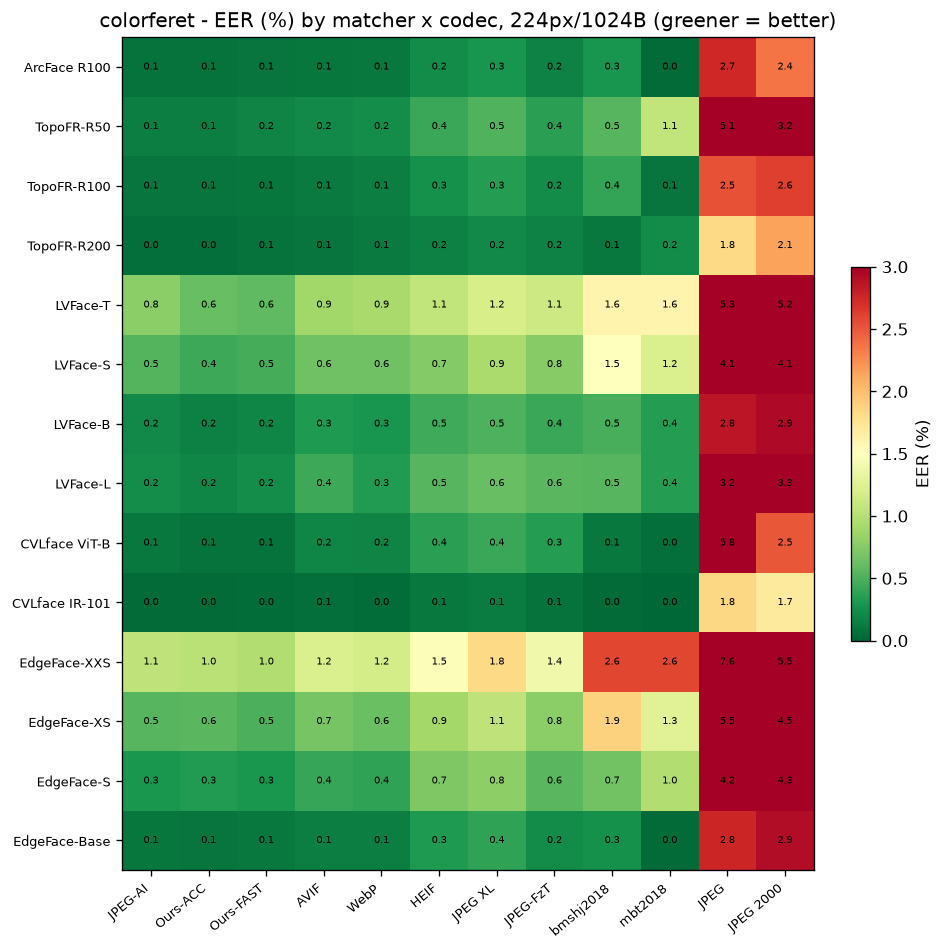}
  \caption{Full-resolution model effect (Color~FERET, $224$\,px/$1024$\,B); same axes as
  Figure~\ref{fig:model-effect}. The seven classical/transform codecs span all fourteen
  matchers, and the codec ordering and the JPEG/JPEG~2000 collapse reproduce at the source
  resolution, at lower absolute EER than $112$\,px. The learned codecs are now equally
  complete: JPEG-AI and \emph{both} Ours variants are scored on all fourteen matchers, as
  are the two CompressAI baselines (on their $300$-crop subset). All three learned codecs
  track each other and lead the classical codecs, exactly as at $112$\,px.}
  \label{fig:model-effect-224-cf}
\end{figure}

\begin{figure}[t]
  \centering
  \includegraphics[width=\linewidth]{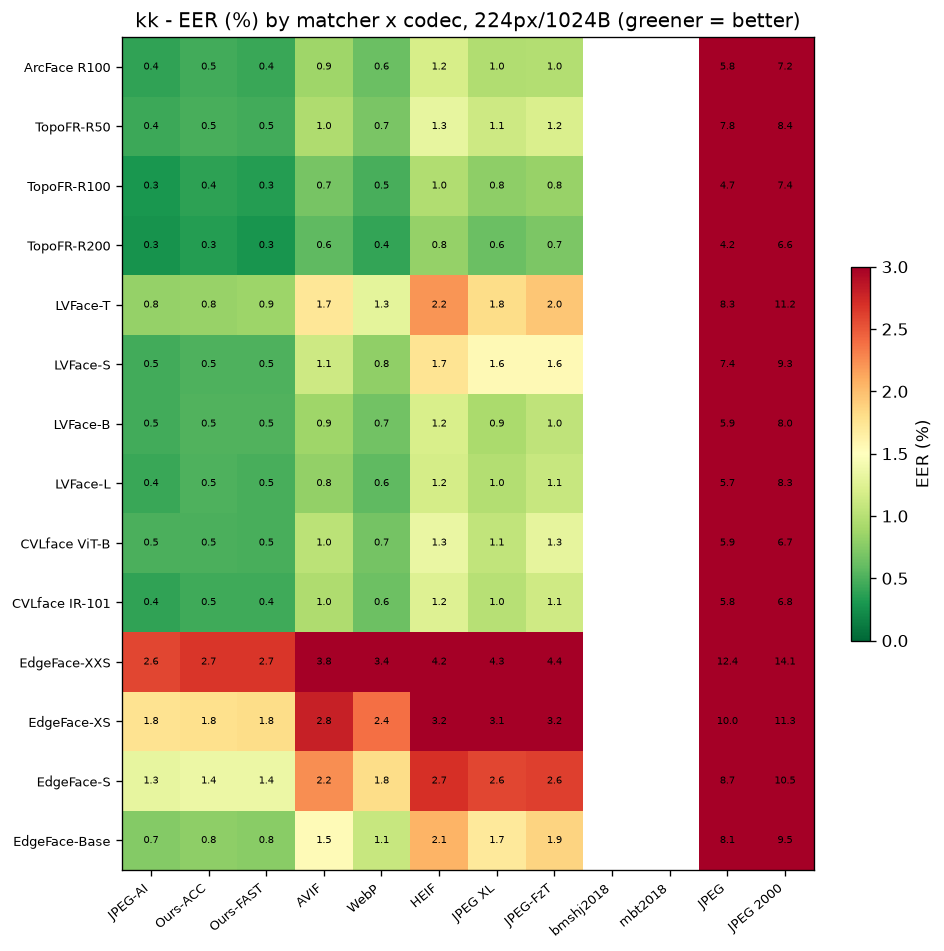}
  \caption{Full-resolution model effect on AI-Solutions-KK ($224$\,px/$1024$\,B); same
  axes as Figure~\ref{fig:model-effect-224-cf}. The classical/transform codecs span all
  fourteen matchers and the codec ordering reproduces at uniformly higher absolute EER,
  exactly as at $112$\,px. The learned codecs are no longer restricted to a handful of
  matchers: the in-the-wild $224$\,px backfill is complete, so JPEG-AI, Ours-ACCURATE and
  Ours-FAST are each scored on all fourteen. Only the two CompressAI baselines are absent,
  never having been run on AI-Solutions-KK.}
  \label{fig:model-effect-224-kk}
\end{figure}

\paragraph{The full 14-model roster (model effect).} The four anchors are one slice of the
project's \textbf{14-model roster} (Section~\ref{sec:protocol}); to test whether the codec
ranking depends on matcher capacity or backbone, we scored \emph{all fourteen} matchers
against the codecs at $112$\,px. Figure~\ref{fig:model-effect} (Color~FERET) and
Figure~\ref{fig:model-effect-kk} (AI-Solutions-KK) give the per-matcher EER heatmaps, with
Tables~\ref{tab:model-effect} and~\ref{tab:model-effect-kk} the exact numbers behind
them. Two results stand out. First, the
\emph{codec} ordering is stable down every matcher column: JPEG-AI, WebP, AVIF and the strong
learned/block codecs stay low, while mbt2018 and JPEG~2000 are the worst under all fourteen
matchers -- the ranking does not invert with matcher choice, and the ViT family (LVFace,
CVLface ViT-B) and the CNN/IR family (TopoFR, ArcFace, CVLface IR-101) agree on which codec to
pick. Second, the absolute EER tracks \emph{matcher strength}: the strongest models
(TopoFR-R200, CVLface IR-101, EdgeFace-Base, LVFace-B) sit near the floor under good codecs
(EER $\le 0.36\%$, and $\le 0.14\%$ for the first three), whereas the tiny/early models (EdgeFace-XXS, LVFace-T) carry a higher
floor and absorb a larger codec penalty -- the gap a lossy codec opens over the aligned
reference widens for weaker matchers.
The heatmaps above are the $112$\,px working point. With the extra-ten matchers
embedded across $64$--$224$\,px, we can confirm the same picture at the $224$\,px source
resolution (Figures~\ref{fig:model-effect-224-cf} and~\ref{fig:model-effect-224-kk}, with
the numbers in Tables~\ref{tab:model-effect-224} and~\ref{tab:model-effect-224-kk}): the
codec ordering is unchanged under all fourteen
backbones and the JPEG/JPEG~2000 collapse persists ($1.7$--$7.6\%$ EER), while absolute
EERs fall with the larger crop as expected (e.g.\ CVLface IR-101 reaches $0.03\%$ and
ArcFace $0.07\%$ under the strong codecs). The model effect is therefore stable across
both matcher backbone \emph{and} resolution. This visual agreement is quantified
statistically in Section~\ref{subsec:rank-based}: the matchers rank the codecs with a
Kendall concordance of $W=0.85$ (Figure~\ref{fig:codec-mean-rank}).

\begin{table}[t]
  \centering \scriptsize
  \caption{Model effect, Color~FERET, $112$\,px/$1024$\,B: EER (\%) per matcher $\times$ codec
  for the full 14-model roster. \shadelegendrow{} Codecs tied at the printed precision are
  all shaded. Column support is \emph{not} equal: the CompressAI baselines
  bmshj2018/mbt2018 were scored on the $300$-crop Color~FERET subset ($2{,}548$ mated pairs
  against $95{,}839$ for every other codec), so their columns are marked $^{\ddagger}$ ---
  the values print, but they take no part in that best/worst comparison. The Color~FERET
  $224$\,px companion (Table~\ref{tab:model-effect-224}) carries the same two ‡-marked
  columns; only the AI-Solutions-KK companion (Table~\ref{tab:model-effect-kk}) drops them,
  since the baselines were never run on that dataset.}
  \label{tab:model-effect}
  \adjustbox{max width=\textwidth}{\input{tables/model_effect_colorferet_1024.tex}}
\end{table}

\begin{table}[t]
  \centering \scriptsize
  \caption{Model effect, \textbf{AI-Solutions-KK}, $112$\,px/$1024$\,B: EER (\%) per matcher
  $\times$ codec for the full 14-model roster --- the numeric companion to the in-the-wild
  heatmap (Figure~\ref{fig:model-effect-kk}), which a deployer needs to read exact values.
  \colorbox{green!25}{Best}/\colorbox{red!22}{worst} per row. All ten codecs run on this
  dataset span all fourteen matchers, JPEG-AI and both Ours variants included; the two
  CompressAI baselines were never run in-the-wild and so have no row.
  Absolute EER is uniformly higher than Color~FERET
  (Table~\ref{tab:model-effect}) but the codec ordering is preserved down every matcher row.}
  \label{tab:model-effect-kk}
  \adjustbox{max width=\textwidth}{\input{tables/model_effect_kk_1024.tex}}
\end{table}

\begin{table}[t]
  \centering \scriptsize
  \caption{Model effect, Color~FERET, $224$\,px/$1024$\,B: EER (\%) per matcher $\times$
  codec for the full 14-model roster. The roster is complete at this cell: every codec
  scored at $224$\,px on Color~FERET --- the seven classical/transform codecs, JPEG-AI,
  both Ours variants and the two CompressAI baselines --- carries an EER under all
  fourteen matchers (see Figure~\ref{fig:model-effect-224-cf}), so no cell is left
  unembedded. \shadelegendrow{} As in Table~\ref{tab:model-effect}, the two CompressAI
  columns are marked $^{\ddagger}$ (scored on the $300$-crop subset) and are excluded from
  that comparison; ties at the printed precision are all shaded.}
  \label{tab:model-effect-224}
  \adjustbox{max width=\textwidth}{\input{tables/model_effect_colorferet_1024_224.tex}}
\end{table}

\begin{table}[t]
  \centering \scriptsize
  \caption{Model effect, AI-Solutions-KK, $224$\,px/$1024$\,B: the numeric companion to
  Figure~\ref{fig:model-effect-224-kk}, and the counterpart of
  Table~\ref{tab:model-effect-224} on the in-the-wild set. Same reading: EER (\%) per
  matcher $\times$ codec over the 14-model roster. All ten codecs run on AI-Solutions-KK
  are populated under all fourteen matchers here; the two CompressAI baseline columns print
  ``--'' in every cell, since they were never run on this dataset --- unlike their
  $112$\,px companion (Table~\ref{tab:model-effect-kk}), which omits the columns outright.
  \shadelegendrow}
  \label{tab:model-effect-224-kk}
  \adjustbox{max width=\textwidth}{\input{tables/model_effect_kk_1024_224.tex}}
\end{table}

\FloatBarrier

\subsection{Budget compliance: can the codec even hit the target?}
\label{sec:budget-compliance}
\tldr{A codec is only usable if it can actually emit a $\le B$-byte file. At $1024$\,B most
codecs comply, but at $512$\,B the block codecs (JPEG, WebP, AVIF, HEIF, JPEG~XL) barely
reach the budget at $168$/$224$\,px ($0$--$5\%$ of crops fit, and only WebP on Color~FERET
climbs higher, to ${\sim}31\%$ at $168$\,px); only
JPEG~2000 (native rate control), JPEG-FzT and the learned codecs (JPEG-AI, ours) hold it --
and JPEG~2000 holds it only by destroying identity.}

\begin{figure}[t]
  \centering
  \includegraphics[width=\textwidth]{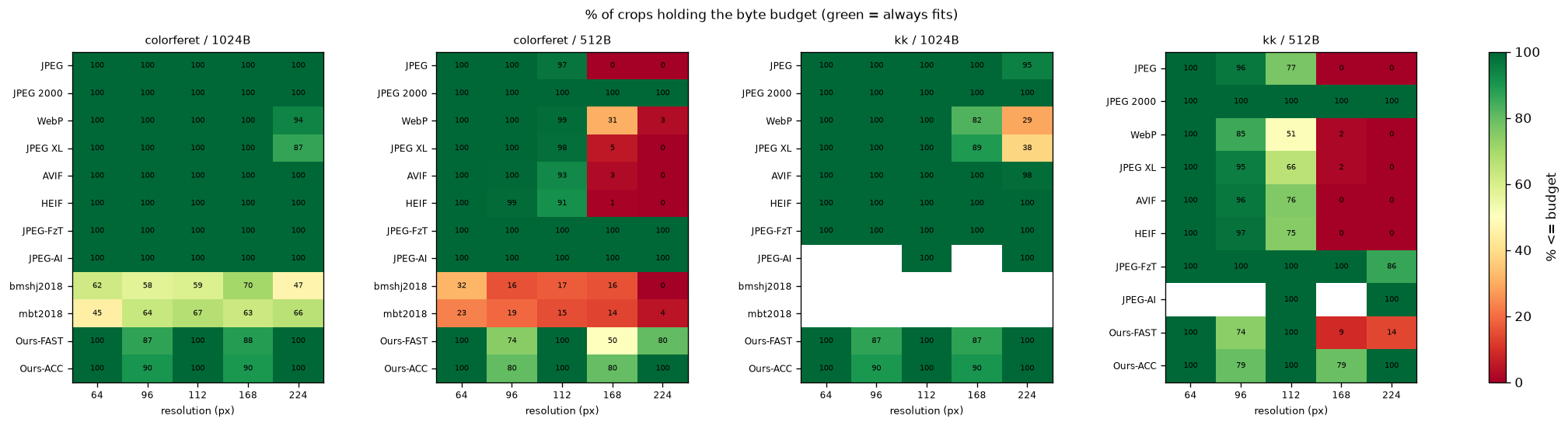}
  \caption{Budget compliance: percentage of crops whose compressed file holds the byte
  budget ($\le B$), per codec $\times$ resolution, for each dataset/budget. Green~$=$~always
  fits, red~$=$~never fits. The red block at $168$/$224$\,px under $512$\,B shows the
  block codecs cannot reach the budget there; JPEG~2000 and the learned codecs
  stay green. A white cell was not encoded at all --- JPEG-AI on AI-Solutions-KK outside
  $112$/$224$\,px, and the CompressAI baselines on that dataset entirely (see
  Table~\ref{tab:fit-kk}) --- and is deliberately blank rather than shaded: a cell measured
  over fewer than $30$ crops is dropped, so no percentage here rests on a handful of
  spot-check files.}
  \label{fig:size-compliance}
\end{figure}

\begin{figure}[t]
  \centering
  \includegraphics[width=\textwidth]{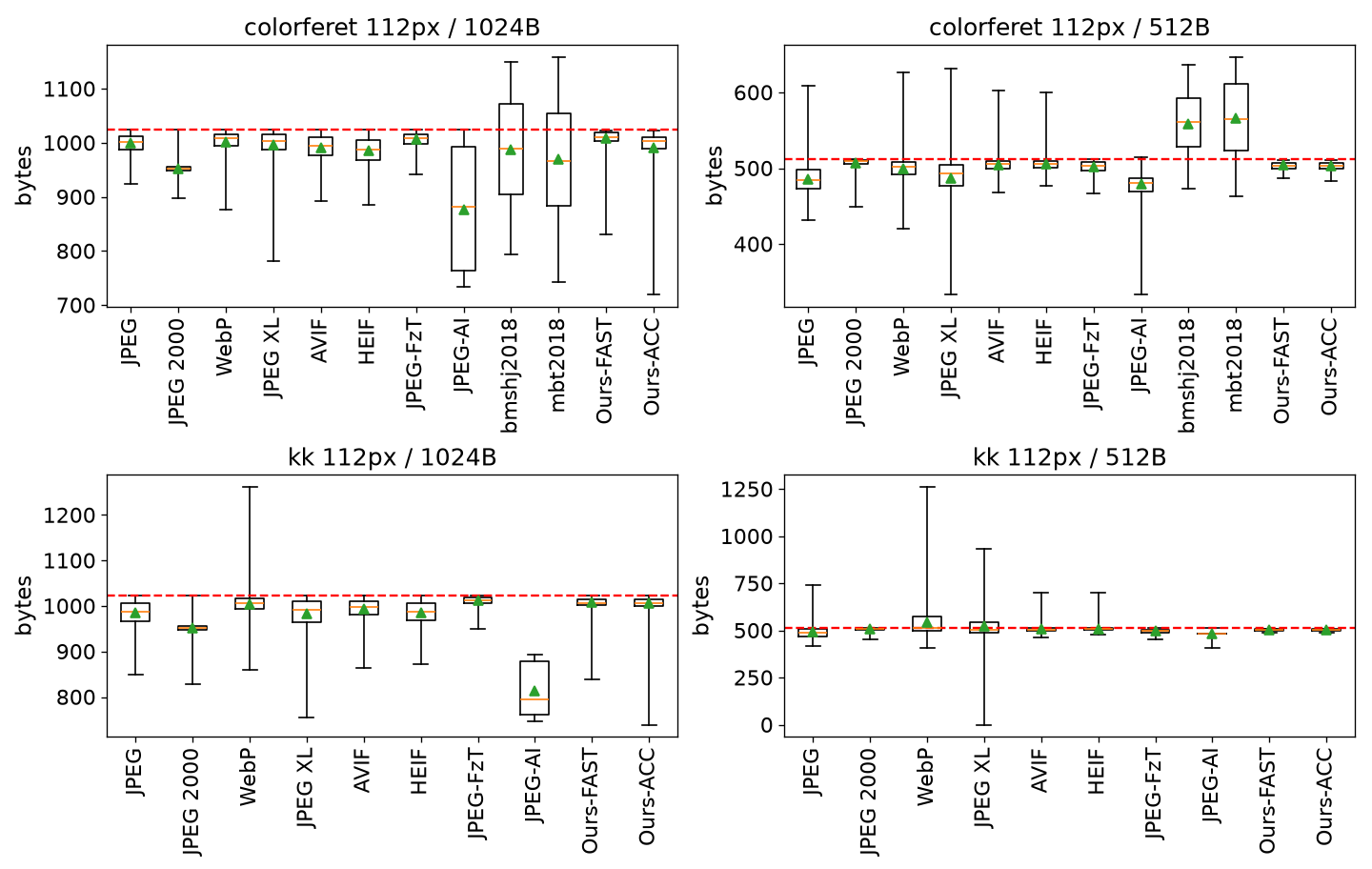}
  \caption{Achieved compressed-file-size distribution per codec at the $112$\,px
  verification working resolution, for both datasets and both budgets. Box =
  inter-quartile range, line = median, triangle = mean, whiskers span the full
  min--max; the dashed red line is the byte budget. Codecs whose box sits above the
  line overshoot the budget for most crops. The remaining resolutions are given
  numerically in the fit-rate tables (Tables~\ref{tab:fit-kk}
  and~\ref{tab:fit-cf}); drawing all twenty cells here would shrink the tick labels
  below legibility.}
  \label{fig:size-box}
\end{figure}

The accuracy tables above implicitly assume the encoder produced a file within the budget;
under a hard $\le B$ constraint that assumption fails for several codecs, resolutions and
image types. We sweep the achieved compressed size of every crop and report, per
(codec, resolution, budget), the fraction \emph{holding} the budget ($\le B$ bytes).
Figure~\ref{fig:size-compliance} is the compliance heatmap; Tables~\ref{tab:fit-kk}
and~\ref{tab:fit-cf} give the numbers at the hardest budget ($512$\,B), with
Table~\ref{tab:fit-1024} the same measure at $1024$\,B for comparison; and
Figure~\ref{fig:size-box} shows the full achieved-size distribution (median line, mean
marker, min--max whiskers) per codec, with the budget drawn as a reference line.

At $1024$\,B nearly every codec complies -- the per-image quality search reaches the target
-- the exceptions being the fixed-rate CompressAI baselines bmshj2018/mbt2018 (only
$59$/$67\%$ of $112$\,px crops fit, since they have no byte target) and a genuine shortfall
for WebP/JPEG~XL at $224$\,px: on Color~FERET they still fit $94$/$87\%$ of crops, but on
AI-Solutions-KK only $29$/$38\%$ (Table~\ref{tab:fit-1024}). At $512$\,B the picture changes sharply: on AI-Solutions-KK the
standard block codecs hold the budget for essentially no crops ($0$--$2\%$) at $168$ and
$224$\,px and only $51$--$77\%$ at $112$\,px -- they simply cannot encode a higher-resolution
face to half a kilobyte. Only JPEG~2000 (a true target-size mode\footnote{This reconciles an apparent
discrepancy with our prior conference work~\cite{ijcb2026anon}, which reported JPEG~2000
as \emph{$0\%$-compliant} at $1$\,kB and excluded it from the budget-matched comparison.
That pipeline drove JPEG~2000 with a narrow quality-layer sweep (Pillow
\texttt{quality\_layers} $\in[80,98]$ under the default rate interpretation), which
bottoms out near ${\sim}1.5$\,kB for a $224^2$ crop and so never reaches the budget. Here
we instead use JPEG~2000's genuine rate control (\texttt{quality\_mode="rates"} with
compression ratios up to $400{:}1$, binary-searched to the exact target), which reaches
any byte budget --- at severe quality cost. The difference is in \emph{how the codec is
driven}, not the codec's capability: JPEG~2000's EBCOT rate allocation supports arbitrary
target rates, whereas the block codecs (JPEG/WebP/AVIF/HEIF) have a genuine minimum size
at a given resolution and therefore drop out of the budget above.}), JPEG-FzT and the
learned codecs (JPEG-AI and ours, budget-targeted by construction) hold the budget
reliably.

This is a
\emph{fit-vs-accuracy} tension: JPEG~2000 always fits but is the worst codec on identity
(Section~\ref{sec:frr-far}), whereas the codecs that best preserve identity at $512$\,B are
exactly the learned ones that are also budget-compliant. The learned codec's hard
byte-budget search (Section~\ref{sec:codec}) guarantees a parseable $\le B$ stream by
construction, and the Ours fit-rate is measured across all five resolutions. It holds
$100\%$ at the $64$, $112$ and (for Ours-ACC) $224$\,px cells, with the over-budget
fallback frame (Section~\ref{sec:codec}) visible only at some $96$ and $168$\,px cells --
e.g.\ on Color~FERET at $1024$\,B Ours-ACC fits $90\%$ ($96$\,px) and $90\%$ ($168$\,px)
(Figure~\ref{fig:size-compliance}), and
the pattern persists at $512$\,B, where Ours-ACC still holds $80\%$ at those two
resolutions while most block codecs have already collapsed to $0$--$5\%$ at
$168$\,px (WebP excepted, at $31\%$). Ours-FAST emits the fallback slightly more often (down to $50\%$ at $168$\,px/
$512$\,B on Color~FERET, and $9$/$14\%$ at $168$/$224$\,px/$512$\,B on AI-Solutions-KK).

\begin{table}[t]
  \centering \small
  \caption{AI-Solutions-KK, $512$\,B: \% of crops holding the budget, codec $\times$
  resolution. JPEG-AI drives an arbitrary target rate and so fits the budget by
  construction ($100\%$ in both measured columns); our codecs do the same at $64$ and
  $112$\,px but fall back to an over-budget frame elsewhere (Ours-ACC $79\%$ at $96$ and
  $168$\,px, Ours-FAST down to $9\%$ at $168$\,px), so the compliance question is really
  about them and about the block codecs, which have a true minimum size.
  ``--'' marks a cell we did not encode, and the reason is cost rather than choice:
  JPEG-AI encodes at ${\approx}2.9$\,s per crop, so covering KK's $17.5$\,k crops at the
  remaining three resolutions would be ${\approx}87$ GPU-hours for a column whose value is
  known in advance; the CompressAI baselines were run on Color~FERET only
  (Table~\ref{tab:fit-cf}), where -- unlike JPEG-AI and our codecs -- they miss the budget
  badly ($0$--$32\%$), because they expose fixed quality presets rather than a target rate. \colorbox{green!25}{Highest}/\colorbox{red!22}{lowest}
  compliance per resolution column, ranked on the printed (rounded) value; a column where
  every codec ties is left unshaded, and green is reserved for a \emph{uniquely} highest
  cell, so a column in which most codecs reach $100\%$ marks only the failures. ``--'' is a
  cell with no run, or one measured over fewer than $30$ crops --- too few for a percentage
  to mean anything next to the $11$--$17.5$\,k-crop cells beside it.}
  \label{tab:fit-kk}
  \adjustbox{max width=\textwidth}{\input{tables/fit_rate_kk_512.tex}}
\end{table}

\begin{table}[t]
  \centering \small
  \caption{Color~FERET, $512$\,B: \% of crops holding the budget, codec $\times$ resolution.
  Shading as in Table~\ref{tab:fit-kk}. The CompressAI baselines are now measured across
  all five resolutions (they were previously $112$\,px only), but on the $300$-crop subset
  they were compressed on --- $2.6\%$ of the corpus behind every other row here --- so
  their percentages carry a sampling error the $11$k-crop rows do not, and their shading
  should be read as indicative. They are also the only rows that genuinely \emph{miss} the
  budget rather than hitting it by construction: CompressAI exposes fixed quality presets,
  not a target rate.}
  \label{tab:fit-cf}
  \adjustbox{max width=\textwidth}{\input{tables/fit_rate_colorferet_512.tex}}
\end{table}

\begin{table}[t]
  \centering \small
  \caption{The same compliance measure at the easier $1024$\,B budget, both datasets
  (AI-Solutions-KK above, Color~FERET below). Shading and the ``--'' convention as in
  Table~\ref{tab:fit-kk}. These are the rows the $512$\,B tables should be read against:
  a codec that fails at $512$\,B but holds here is rate-limited, not broken.}
  \label{tab:fit-1024}
  \adjustbox{max width=\textwidth}{\input{tables/fit_rate_kk_1024.tex}}

  \vspace{4pt}
  \adjustbox{max width=\textwidth}{\input{tables/fit_rate_colorferet_1024.tex}}
\end{table}

\FloatBarrier

\subsection{Headline ranking and discussion}
\tldr{At $1024$\,B, JPEG-AI is the identity leader, WebP/AVIF/JPEG~XL/HEIF are safe, JPEG and JPEG~2000 are not; our trained codec already sits between JPEG-AI and AVIF on identity cosine, and the $512$\,B point is where standard codecs stop being viable.}

Three conclusions follow for the sub-1\,kB target. First, the budget is the dominant
variable: at $1024$\,B several mature codecs already keep $112$\,px verification EER under
$0.5\%$, but at $512$\,B all standard codecs except JPEG-AI degrade sharply, and the JPEG
family becomes unusable (id-cosine $0.069$--$0.357$ on Color~FERET). Second, a codec's
distortion score is a poor proxy for its identity score: JPEG~2000 attains respectable
PSNR/SSIM yet destroys identity (id-cosine $0.228$ at $512$\,B), so identity cosine and
verification EER -- not PSNR -- are the metrics that matter for this task; this
decoupling is quantified in Section~\ref{sec:quality}
(Figure~\ref{fig:quality-vs-identity}). Third, the
ceiling among off-the-shelf options is set by the learned standard JPEG-AI, with the
modern block codecs WebP, AVIF, JPEG~XL and HEIF as the practical, widely-deployable
tier. Our trained codec is already competitive on identity -- between JPEG-AI and AVIF at
$1024$\,B and ahead of AVIF at $512$\,B on Color~FERET -- and its perceptual-loss training
gives it the best LPIPS at $512$\,B; closing the remaining identity-cosine gap to JPEG-AI
and confirming it on the verification EER grid is the subject of the following sections.

%% file: sections/06_quality.tex
\section{Reconstruction Quality}
\label{sec:quality}
\tldr{Full-reference fidelity ranks the codecs JPEG-AI~$>$ WebP/AVIF~$>$ JPEG~XL~$>$ JPEG-FzT/JPEG~2000~$>$ plain JPEG at the 1\,kB operating point --- but that ranking does not transfer to identity: image-quality metrics are an unreliable proxy for verification EER (\S\ref{subsec:quality-vs-identity}), and our learned codec wins the perceptual LPIPS/DISTS metrics while losing PSNR/SSIM. Read this section as a warning against selecting a codec on fidelity, not as a second ranking; classical codecs hold skin tone but soften texture, JPEG blocks worst, and our learned codec degrades gently toward the 512\,B point.}

This section measures how faithfully each codec reconstructs the face crop at the
sub-1\,kB target. The harder question --- whether identity survives --- is answered by
the verification benchmark of Section~\ref{sec:benchmark}, which precedes this section;
here we ask how well pixel fidelity tracks that identity result. Quality is a proxy: a codec can
score well on pixel fidelity yet still discard the cues a matcher relies on, so the
metric matrix and the montages below should be read together with the rate--EER
results rather than in isolation. All numbers are dataset medians over the full
quality run; the underlying per-cell five-metric tables (PSNR, SSIM, MS-SSIM, LPIPS,
DISTS for every $(\text{codec},\text{res},\text{budget})$ cell) live in the project
quality CSVs. Our own learned codecs (Ours-FAST, Ours-ACCURATE) are scored on the same
full population as the classical codecs ($n{=}11{,}335$ on Color~FERET,
$n{=}17{,}534$ on AI-Solutions-KK).

\subsection{Full-reference metric matrix}
\tldr{Across PSNR, SSIM, MS-SSIM, LPIPS and DISTS at 1024\,B, JPEG-AI and WebP lead on distortion metrics, our learned codec trails on PSNR/SSIM but leads every codec on the perceptual LPIPS/DISTS metrics, and plain JPEG is near-last at both resolutions. Every row is now measured on the full population except the two CompressAI baselines.}

We report five complementary full-reference metrics. PSNR and SSIM~\cite{wang2004ssim}
capture pixel- and structure-level fidelity; MS-SSIM adds a multi-scale structural
view; LPIPS~\cite{zhang2018lpips} and DISTS~\cite{ding2020dists} are learned
perceptual distances (lower is better) that correlate better with human judgement of
texture and identity-bearing detail. Table~\ref{tab:quality-matrix} gives a compact
view at the verification working resolution (112\,px) and at full-size crops
(224\,px), both at the 1024\,B budget, on Color~FERET. The ranking is consistent
across metrics: WebP~\cite{google2010webp} and AVIF~\cite{aomavif} lead the broadly
available classical codecs, JPEG~XL~\cite{alakuijala2019jpegxl} follows, JPEG-FzT and
JPEG~2000~\cite{taubman2002jpeg2000} sit mid-pack, and plain
JPEG~\cite{wallace1992jpeg} is last on every perceptual metric at $224$\,px (LPIPS $0.436$,
DISTS $0.354$) and, at $112$\,px, tenth of twelve on DISTS ($0.206$) though only mid-field
on LPIPS ($0.103$). JPEG-AI~\cite{ascenso2023jpegai} is the
strongest codec on the \emph{structural} metrics at
112\,px (SSIM $0.920$, MS-SSIM $0.984$ at 1024\,B; its PSNR $33.32$\,dB is second behind
WebP's $33.45$ among the full-population rows --- mbt2018's $33.85$ is higher but rests on
$300$ crops, which is why it is marked $^{\ddagger}$ and does not take the crown), though our learned
codec still leads it on the perceptual LPIPS/DISTS metrics. Its edge narrows as the crop
grows: at 224\,px its perceptual quality falls off sharply (LPIPS
$0.060\rightarrow0.205$), consistent
with the 112\,px working-resolution choice. The learned baselines
bmshj2018~\cite{balle2018hyperprior} and mbt2018~\cite{minnen2018joint} (from
CompressAI~\cite{begaint2020compressai}) run on a 300-image subset at every resolution,
and are scored for full-reference quality at 112\,px only; mbt2018 reaches PSNR
$33.85$\,dB / SSIM $0.897$ at 1024\,B, competitive on PSNR but weaker than WebP/AVIF on
LPIPS. That $300$-crop support is $2.6\%$ of Color~FERET, so every bmshj2018/mbt2018
number in this report --- here, in the budget-compliance tables and in the per-matcher
rate--EER tables --- is a lower-precision estimate sitting beside full-population
columns. The accuracy grids (model-effect, FNMR, budget sweep) print those cells with no
in-table support mark, and several of their green column-best crowns are bmshj2018 cells,
so they must be read against this note rather than as full-population wins.

The two rightmost blocks add our own learned codecs, measured on every crop.
The headline finding is a clean perceptual-vs-distortion split: on the pixel- and
structure-fidelity metrics Ours-ACCURATE trails the classical field badly (PSNR
$28.63$\,dB, SSIM $0.834$ at 112\,px / 1024\,B, roughly $5$\,dB below WebP and behind
every neural baseline), yet on the learned perceptual metrics it is the best codec in
the table --- LPIPS $0.012$ and DISTS $0.095$ at that cell beat WebP ($0.045$ / $0.142$),
AVIF ($0.088$ / $0.142$), JPEG~XL ($0.145$ / $0.170$) and both of
bmshj2018/mbt2018 (LPIPS $0.118$--$0.129$); this LPIPS lead is an expected consequence of LPIPS
being part of Ours' training objective, not an independent advantage. Ours-FAST sits
just below Ours-ACCURATE on both PSNR/SSIM ($27.17$\,dB / $0.816$) and the perceptual
metrics (LPIPS $0.029$, DISTS $0.146$), while still beating the classical field
on LPIPS (its DISTS $0.146$ only ties WebP/AVIF's $0.142$). This is the perceptual/distortion trade-off characteristic of learned
codecs: they optimise a perceptual objective and so discard the high-frequency energy
PSNR rewards while preserving the texture cues LPIPS/DISTS --- and, we show later, a
face matcher --- actually rely on. The trend on AI-Solutions-KK is the same ordering,
with absolute PSNR roughly $1$--$3$\,dB lower at matched cells (${\sim}3$\,dB for most classical codecs at $112$\,px, though only $0.4$\,dB for JPEG-FzT, and $1.4$\,dB for Ours-ACCURATE) owing to the in-the-wild
content (Ours-ACCURATE 112\,px / 1024\,B: PSNR $27.26$\,dB, LPIPS $0.014$).

\begin{table}[t]
  \centering
  \caption{Full-reference quality (dataset medians) at the 1024\,B budget on
  Color~FERET, for the verification resolution (112\,px) and full crops (224\,px). Higher
  PSNR/SSIM/MS-SSIM is better; lower LPIPS/DISTS is
  better. The \colorbox{green!25}{best} value in each column per resolution block is
  shaded green and the \colorbox{red!22}{worst} red. Every row is now measured on the full
  population ($n{=}11{,}331$--$11{,}335$ on Color~FERET) except the CompressAI baselines
  bmshj2018/mbt2018, which were run on a $300$-crop subset at $112$\,px only and are
  omitted from the $224$\,px block. Those two rows carry $^{\ddagger}$: their values
  print, but they take no part in the best/worst comparison, as in
  Table~\ref{tab:model-effect}. That rule changes one crown --- the $112$\,px PSNR best
  is WebP's $33.45$ on $11{,}335$ crops, not mbt2018's $33.85$ on $300$. Earlier
  versions of this table scored JPEG-AI on $300$ crops and plain JPEG's $112$\,px cell on
  $20$, both of which flattered them --- JPEG's PSNR there fell $31.37\rightarrow29.34$\,dB
  when the full population was measured.}
  \label{tab:quality-matrix}
  \adjustbox{max width=\textwidth}{\input{tables/quality_matrix.tex}}
\end{table}

The plain-JPEG $112$\,px row was previously computed over only $20$ crops, on the stated
grounds that the encoder rarely lands a valid file at that budget. That reason was wrong:
all $11{,}335$ files exist and all are under budget, and re-measuring them drops the cell
from $31.37$ to $29.34$\,dB --- so the small sample had been flattering JPEG by $2$\,dB.
Full-reference scoring of our own
learned codecs (Ours-FAST, Ours-ACCURATE), by contrast, uses the full population
($n{=}11{,}335$ on Color~FERET, $n{=}17{,}534$ on AI-Solutions-KK), and their rows
appear in the matrix above.

\subsection{Distortion does not predict identity}
\label{subsec:quality-vs-identity}
\tldr{Plotting each codec's image-quality score against its verification EER at matched
budget shows the two decouple. Across all twelve codecs PSNR does \emph{not} significantly
predict EER at 1024\,B ($\rho{=}{-}0.45$, $p{=}0.14$), and neither does the perceptual LPIPS
($+0.37$, $p{=}0.24$); only the structural metrics clear $p{<}0.05$ there, and weakly
(SSIM $-0.66$, MS-SSIM $-0.62$) --- but that
reading is dominated by the two partial-coverage neural baselines (\S\ref{sec:protocol});
with them removed ($n{=}10$) the \emph{fidelity} metrics do rank-track EER at 1024\,B (PSNR
$\rho{=}{-}0.82$, MS-SSIM $-0.92$, both significant) while the \emph{perceptual} LPIPS still
does not ($\rho{=}{+}0.50$, $p{=}0.14$). At the tight 512\,B budget every metric becomes
predictive and the perceptual ones track EER most tightly (DISTS $\rho{=}{+}0.96$, LPIPS
$+0.94$, vs PSNR $-0.74$). Either way the deployment message holds: the ``best'' metric flips
across budgets and equal-quality codecs (JPEG~2000 vs Ours-ACCURATE) differ by more than an
order of magnitude in EER, so codec selection for identity must be validated on verification
metrics.}

\begin{figure}[t]
  \centering
  \includegraphics[width=\linewidth]{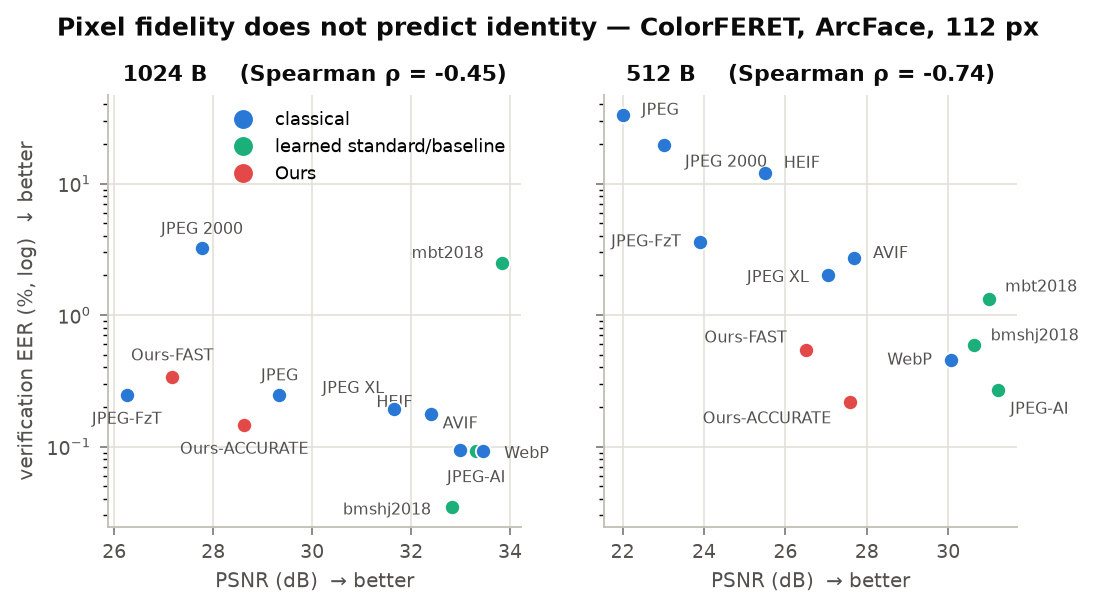}
  \caption{Pixel fidelity versus identity at matched budget (Color~FERET, ArcFace,
  112\,px; EER on a log scale, so vertical gaps of one grid line are order-of-magnitude
  differences). Points are codecs, coloured by family. PSNR does not determine EER:
  at 1024\,B the twelve-codec rank correlation is non-significant ($\rho=-0.45$, $p=0.14$,
  $n=12$), and codecs at equal PSNR differ by more than an order of magnitude in EER (WebP
  $0.09\%$ at $33.5$\,dB vs mbt2018 $2.50\%$ at $33.8$\,dB); removing the two
  partial-coverage neural baselines ($n=10$) makes PSNR predictive
  ($\rho=-0.82$, $p=0.004$), so the null at $n=12$ is largely those two artefact points
  (\S\ref{sec:protocol}).}
  \label{fig:quality-vs-identity}
\end{figure}

The preceding matrix and the benchmark's EER grids measure two different things ---
pixel fidelity and identity survival --- and the sub-1\,kB regime is exactly where
they part ways. Figure~\ref{fig:quality-vs-identity} makes the decoupling explicit by
plotting, for every codec with both measurements at the 112\,px working resolution on
Color~FERET (ArcFace), the median PSNR against the verification EER. If distortion
were a reliable proxy for identity, the points would fall on a monotone
high-PSNR/low-EER curve. They do not. At the 1024\,B budget the Spearman rank
correlation across all twelve codecs is $\rho=-0.45$ and does not reach significance
($p=0.14$); at 512\,B it strengthens to $\rho=-0.74$ ($p=0.006$) but
remains far from deterministic. PSNR is not alone at the comfortable budget: LPIPS is
equally flat there ($\rho=+0.37$, $p=0.24$) and only the structural metrics reach
$p<0.05$, and then weakly (SSIM $-0.66$, $p=0.018$; MS-SSIM $-0.62$, $p=0.031$).
This twelve-codec reading is, however, dominated by the
\textbf{two partial-coverage neural baselines} (bmshj2018/mbt2018), whose EERs rest on the
reduced ${\sim}3.4$\,k-impostor subset and are flagged as a scoring artefact rather than codec
behaviour (\S\ref{sec:protocol}); it is inconsistent to disown mbt2018's $2.50\%$ EER as an
artefact yet let it drive the correlation. Recomputing with those two removed
($n=10$) sharpens the picture: at 1024\,B the \emph{fidelity} metrics now
\emph{do} rank-track EER (PSNR $\rho=-0.82$, $p=0.004$; SSIM $-0.87$; MS-SSIM $-0.92$),
so the weak fidelity signal at the comfortable budget was largely masked by the two
artefact points. The \emph{perceptual} metrics run opposite to the naive perceptual-quality expectation of the
framing: LPIPS still does \emph{not} predict EER at 1024\,B ($\rho=+0.50$, $p=0.14$),
and DISTS is only borderline ($\rho=+0.78$, $p=0.008$ at $n=10$; $+0.53$, $p=0.08$ ---
a coin-flip --- at $n=12$). At 512\,B, where distortion is severe enough to erode identity,
\emph{all} metrics become predictive and the \emph{perceptual} ones track EER most tightly
(DISTS $\rho=+0.96$, LPIPS $+0.94$; MS-SSIM $-0.91$; PSNR $-0.74$; all $p\le0.006$; the
$n=10$ values are within $0.11$). We report both supports; the durable, coverage-robust
point is carried by two concrete disagreements that survive the $n=10$ filter.
First, JPEG~2000 sits mid-field on PSNR ($27.77$\,dB) but is the
worst-performing standard codec on identity ($3.23\%$ EER). Second, the relation
inverts for our perceptually-trained codec: Ours-ACCURATE gives up ${\sim}5$\,dB of
PSNR relative to WebP yet matches the strong codecs on EER ($0.14\%$), because its
objective spends bits on identity-bearing structure rather than on broadband pixel
error --- and it is exactly this codec that also tops the perceptual LPIPS/DISTS
metrics, so its advantage is visible on perceptual quality even where PSNR hides it.
(At $n=12$ the sharpest disagreement is mbt2018's field-best PSNR $33.85$\,dB against a
$2.50\%$ EER, but that is the very coverage artefact we exclude above.) The
practical consequence for a deployer is that a codec cannot be selected
for a face-credential pipeline from image-quality benchmarks of either family --- the
choice must be validated on verification error directly, which is what the
operating-point analysis of Section~\ref{sec:frr-far} provides.

\subsection{Face image quality (FIQ)}
\label{subsec:fiq}
\tldr{Because this is a \emph{face} codec, we add a no-reference face-image-quality
measure (SER-FIQ-style): the stability of a matcher's embedding under mild input
perturbation, swept over the full $\{112,224\}$\,px\,$\times\,\{512,1024\}$\,B grid on
both datasets. At $1024$\,B every codec but legacy JPEG at the $224$\,px source (already ${\sim}0.03$ below the original there) preserves FIQ at the uncompressed level; the
split appears only at $512$\,B, where legacy JPEG collapses hardest (FIQ
$0.95\!\rightarrow\!0.80$) and JPEG-FzT collapses at the $224$\,px source, while the
modern transform codecs and \emph{both} learned variants (Ours-FAST and Ours-ACCURATE)
stay within ${\sim}0.01$ of the original --- more evidence that the learned codecs
degrade gracefully at the hard budget.}

\begin{figure}[t]
  \centering
  \includegraphics[width=\linewidth]{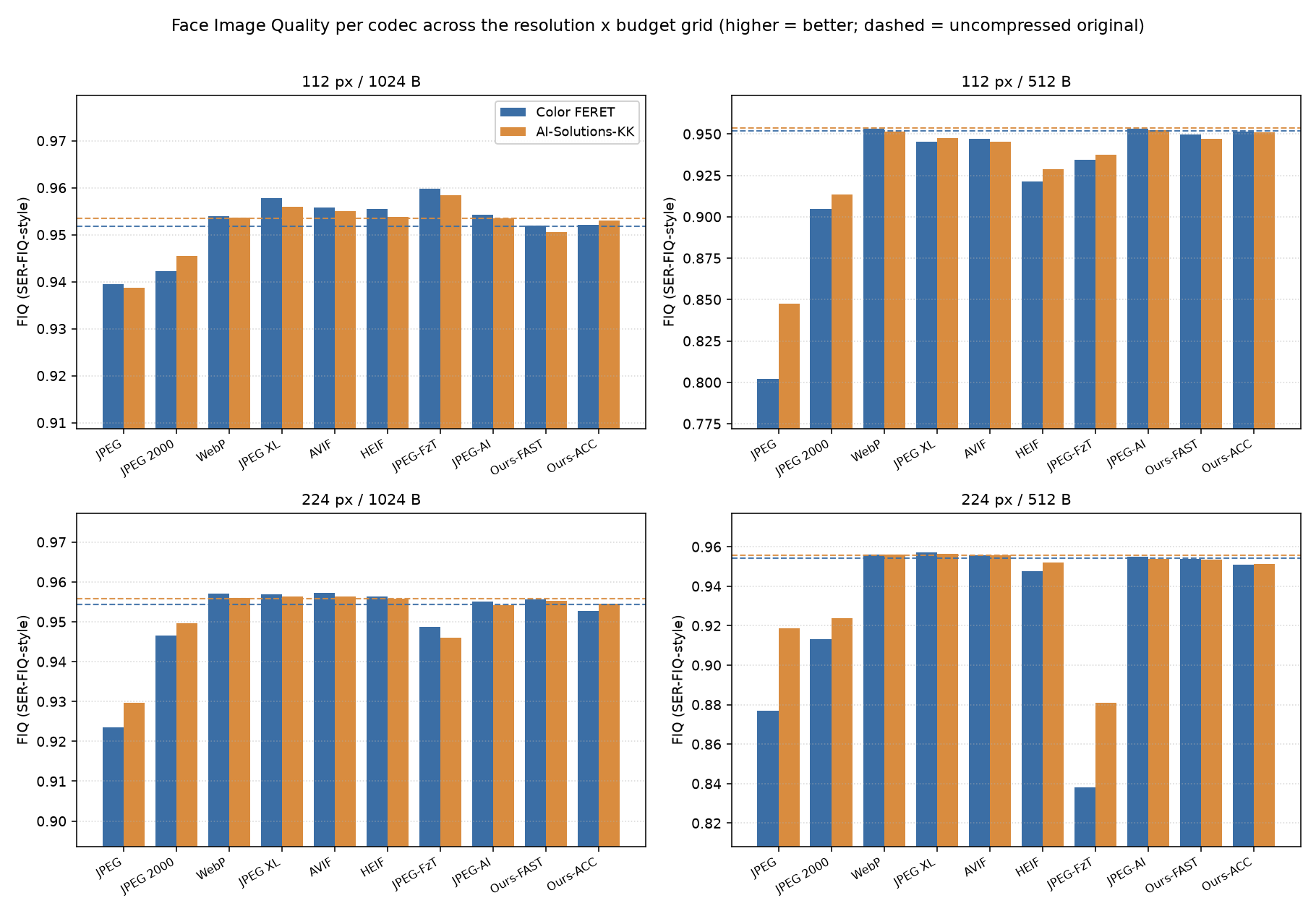}
  \caption{Face image quality per codec across the four resolution$\times$budget cells on
  both datasets (dashed lines: the uncompressed originals). At $1024$\,B (left column of
  panels) every codec tracks the original except legacy JPEG at the $224$\,px source
  (${\sim}0.03$ below it); at $512$\,B (right column) JPEG drops sharply
  and JPEG-FzT drops at $224$\,px, while the modern transform codecs and both Ours
  variants stay at the original level.}
  \label{fig:fiq}
\end{figure}

The full-reference metrics above compare a reconstruction to its own original. A
deployment also cares about a \emph{no-reference} question specific to faces: does the
decoded crop still look like a high-quality face to a matcher? We answer it with a
face image quality assessment (FIQA) score in the SER-FIQ~\cite{terhorst2020serfiq}
family. As the project matchers are deterministic (no dropout), we use the
input-stochasticity variant: for an image $x$ we draw $T{=}10$ mild augmentations
$a_1,\dots,a_T$ (scale-crop $0.90$--$1.00$, horizontal flip, gamma $0.85$--$1.15$,
brightness $\pm 8\%$), embed each with the \texttt{arcface\_antelopev2} oracle, and set
$\mathrm{FIQ}(x)=\operatorname{mean}_{i<j}\cos\!\big(f(a_i(x)),f(a_j(x))\big)$: an
embedding that survives perturbation signals a high-quality, recognisable face. We
report FIQ over the full resolution$\times$budget grid ($\{112,224\}$\,px $\times$
$\{512,1024\}$\,B), $n{=}250$ paired crops per cell per dataset; the learned Ours-* crops
are reconstructed from the paper checkpoints in-process (encode$\to$decode). Every
reconstruction is resized to the matcher's $112$\,px input before augmentation, so all
cells are scored on the operationally-relevant matcher input.
Table~\ref{tab:fiq} and Figure~\ref{fig:fiq} report the result.

\begin{table}[t]
  \centering
  \small
  \caption{No-reference face image quality (SER-FIQ-style, mean pairwise cosine; higher
  is better) across the resolution$\times$budget grid, $n{=}250$ paired crops per cell,
  oracle matcher \texttt{arcface\_antelopev2}. \colorbox{green!25}{Best}/
  \colorbox{red!22}{worst} per column. JPEG-AI is now scored on both datasets (its
  AI-Solutions-KK FIQ sits at the $\sim$$0.95$ original level like the other modern
  codecs). At
  $1024$\,B all codecs except legacy JPEG at $224$\,px ($0.924$ CF / $0.930$ KK) sit at
  the $\sim$$0.95$ original level; the degradation is otherwise a
  $512$\,B effect concentrated in JPEG (and JPEG-FzT at $224$\,px).}
  \label{tab:fiq}
  \adjustbox{max width=\textwidth}{\input{tables/fiq_grid.tex}}
\end{table}

Three readings follow. First, \textbf{at $1024$\,B face image quality is essentially
codec-independent}: every codec --- classical, transform, JPEG-AI and both Ours variants
--- sits within ${\sim}0.015$ of the uncompressed original ($\approx 0.95$) at the
$112$\,px working resolution on both datasets, and within ${\sim}0.01$ at $224$\,px with
the single exception of legacy JPEG ($0.030$ below the original on Color~FERET, $0.026$
on AI-Solutions-KK), so a kilobyte is enough for all of them to preserve a recognisable,
robust face. Second, \textbf{the degradation is a $512$\,B phenomenon and is codec-specific}:
halving the budget barely moves the modern transform codecs (WebP, AVIF, JPEG\,XL) or
JPEG-AI (all still $\ge 0.945$), but legacy JPEG collapses hardest ($0.95\!\rightarrow\!0.80$
at $112$\,px CF, $0.85$ KK), and JPEG-FzT --- fine at $112$\,px --- collapses at the
$224$\,px source ($0.84$ CF, $0.88$ KK), where its F-transform reconstruction of a
half-kilobyte high-resolution crop breaks down; JPEG\,2000 and HEIF lose an intermediate
amount. Third, \textbf{both learned variants degrade gracefully}: Ours-FAST and
Ours-ACCURATE hold FIQ $\ge 0.946$ across \emph{every} cell including the hardest
$224$\,px/$512$\,B point, tracking the strong transform codecs rather than the collapsing
JPEG/JPEG-FzT --- the same graceful-degradation behaviour the verification EER shows,
confirmed on a no-reference face-quality metric. (Small negative $\Delta$FIQ for the
strong codecs at $1024$\,B is a known SER-FIQ artefact: mild compression denoises the
crop, marginally raising augmentation-robustness; read those as ``no degradation''.)

\subsection{Visual comparison}
\tldr{Side-by-side montages confirm the metric story: JPEG-AI keeps the most facial structure, classical codecs preserve skin tone while softening texture, JPEG blocks worst, and the learned codec degrades gently to 512\,B with a mild tone shift on darker skin to watch.}

\begin{figure}[t]
  \centering
  \begin{subfigure}{\linewidth}
    \centering
    \includegraphics[width=\linewidth]{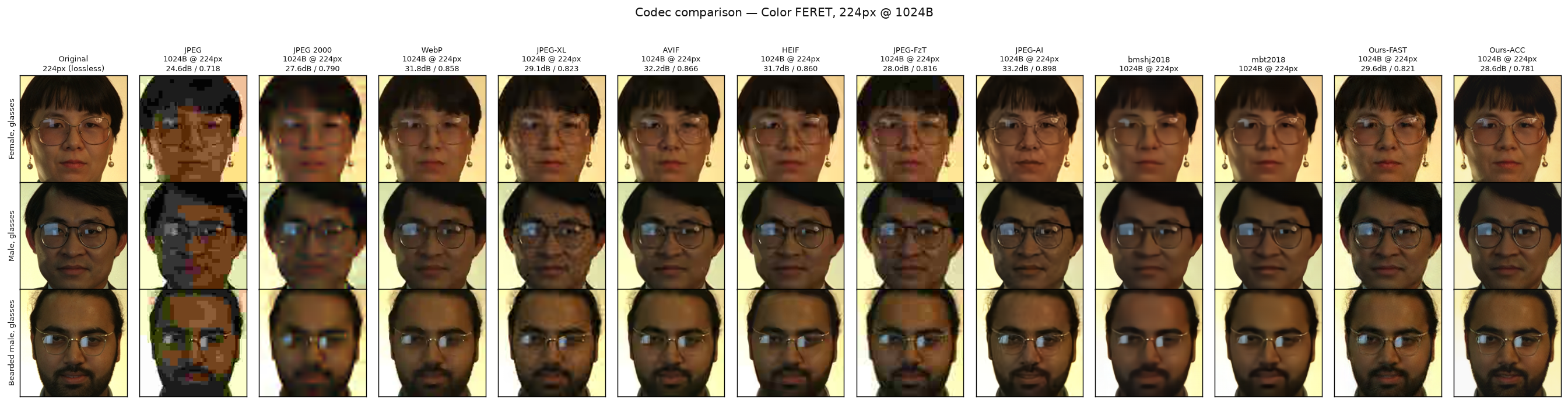}
    \caption{Color~FERET.}
    \label{fig:codec-grid-cf}
  \end{subfigure}\\[0.9em]
  \begin{subfigure}{\linewidth}
    \centering
    \includegraphics[width=\linewidth]{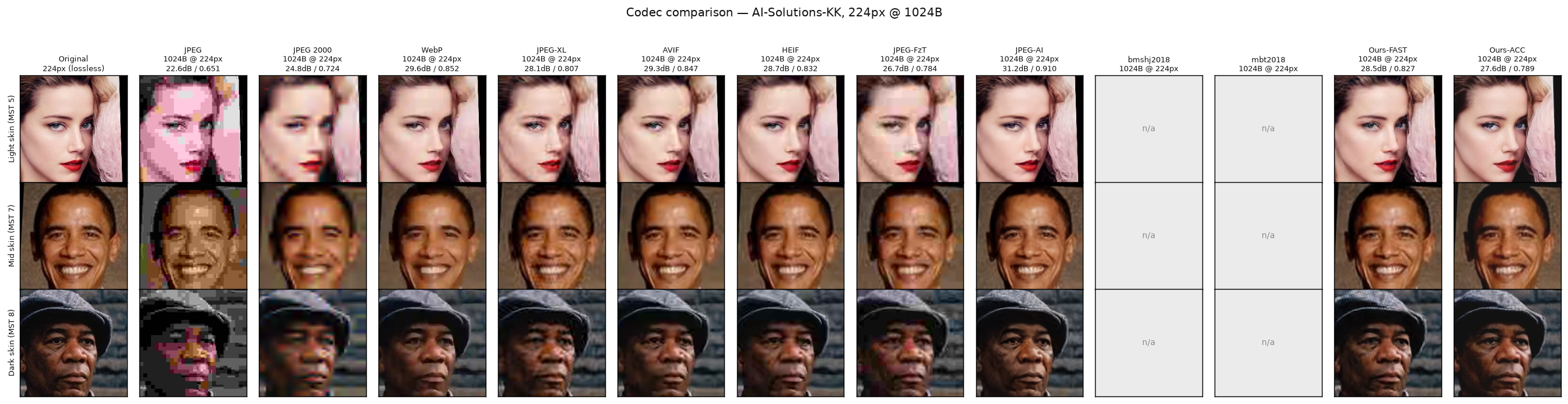}
    \caption{AI-Solutions-KK.}
    \label{fig:codec-grid-kk}
  \end{subfigure}
  \caption{Codec comparison at 224\,px / 1024\,B, decoded crops vs.\ the
  pixel-aligned reference, over the \emph{full} ten-codec$+$Ours roster, so every
  benchmarked method is shown side by side. JPEG-AI keeps the most structure; classical codecs hold skin tone but
  soften texture; plain JPEG blocks worst; Ours-FAST/Ours-ACCURATE reconstruct
  colour-correct faces. JPEG-AI is shown on both datasets; the two CompressAI baselines
  (bmshj2018, mbt2018) appear on Color~FERET only, marked ``n/a'' on AI-Solutions-KK
  because that dataset was never run through them beyond three stray crops apiece
  ($n{=}3$ at $224$\,px$/1024$\,B in the file-size artifact, and no KK quality rows at all)
  --- far below any reportable support. Their role here is as published
  learned-codec reference points, and Color~FERET already fixes that reference. The KK
  rows span the Monk skin-tone scale~\cite{monk2023mst}; the three subjects were picked by
  measured Individual Typology Angle rather than by the dataset's per-image tone label,
  which mislabels light-skinned subjects photographed in shadow. Tile captions report the
  dataset-median PSNR/SSIM for the cell.}
  \label{fig:codec-grids}
\end{figure}

The montages decode one sample per identity with the project's own decoders and crop
to the reference resolution, so the comparison is like-for-like at a fixed byte
budget. Figure~\ref{fig:codec-grids} shows the full codec grid at 224\,px / 1024\,B on
both datasets. On Color~FERET (Figure~\ref{fig:codec-grid-cf}), plain JPEG shows the
worst blocking and ringing around
the eyes and beard, WebP/AVIF/JPEG~XL stay smoother but lose fine texture, and JPEG-AI
retains the most facial structure at this budget --- the eyeglass frames and hairline
are the clearest discriminators. On AI-Solutions-KK (Figure~\ref{fig:codec-grid-kk}) the
same ordering holds; all
classical codecs hold skin tone well at this cell but soften pores and hair, and JPEG
blocks worst on the dark-skin sample where local contrast is low.

\begin{figure}[t]
  \centering
  \begin{subfigure}{\linewidth}
    \centering
    \includegraphics[width=\linewidth]{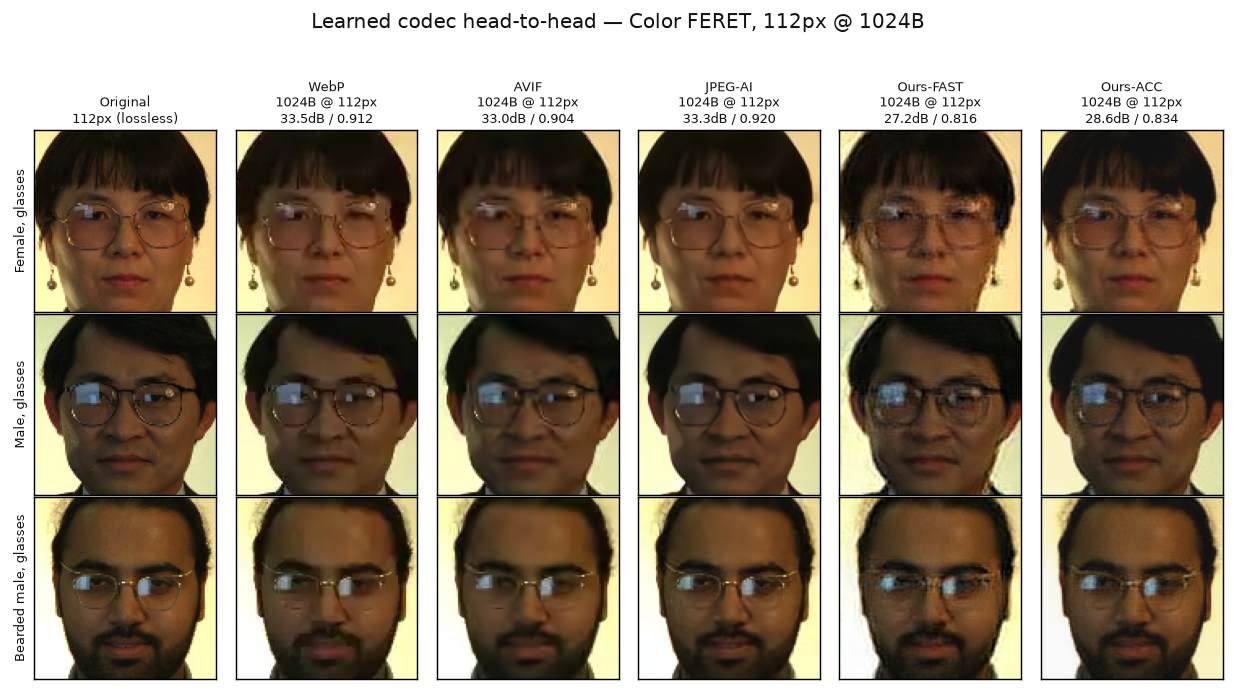}
    \caption{Color~FERET.}
    \label{fig:ours-grid-cf}
  \end{subfigure}\\[0.9em]
  \begin{subfigure}{\linewidth}
    \centering
    \includegraphics[width=\linewidth]{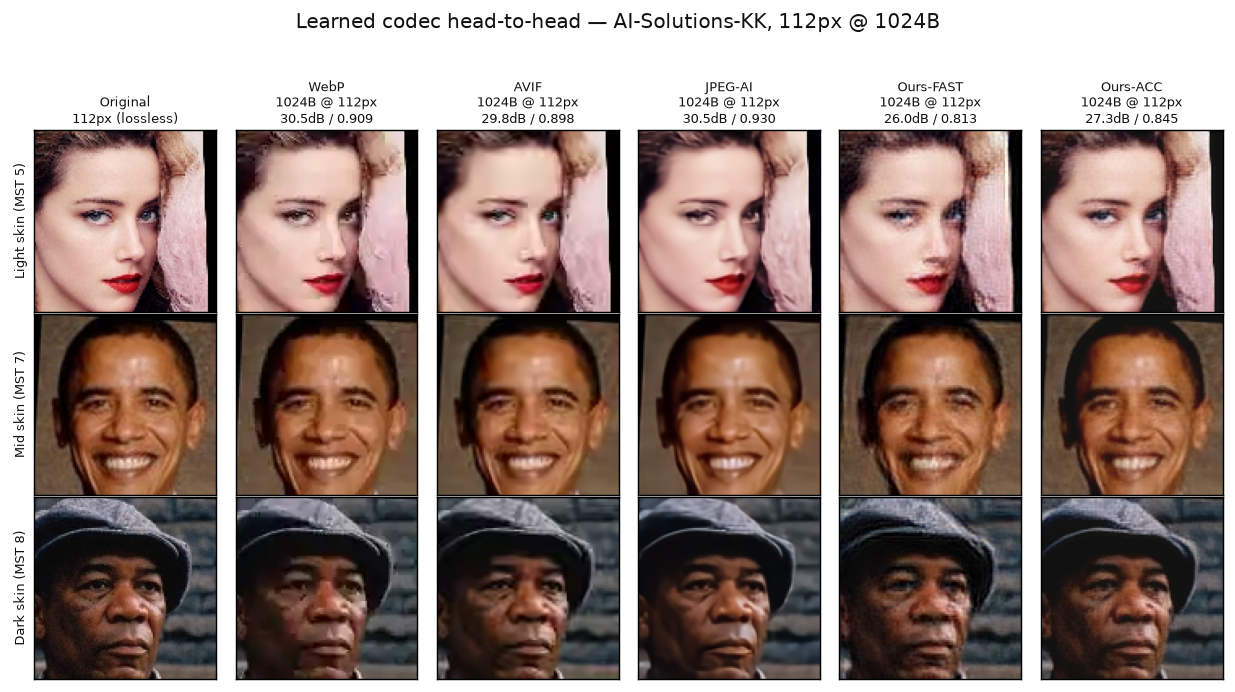}
    \caption{AI-Solutions-KK, three Monk skin tones.}
    \label{fig:ours-grid-kk}
  \end{subfigure}
  \caption{Learned-codec showcase at 112\,px / 1024\,B. Ours-ACCURATE keeps identity
  cues crisper than the classical codecs at matched budget; Ours-FAST is the
  lighter-weight variant. A mild tone shift on darker skin at this checkpoint is
  flagged for the fairness analysis. The three KK rows span Monk tones $5$, $7$ and $8$:
  KK's Monk labels are model-estimated and propagated per identity, so many individual
  crops carry a tone label the image contradicts, and these are the crops whose
  \emph{measured} skin tone (Individual Typology Angle over lit-side facial pixels:
  $76^{\circ}$, $24^{\circ}$, $15^{\circ}$) agrees with the stored label. Monk $9$--$10$
  has no well-exposed crop in KK, so tone $8$ is the darkest exemplar the labelling
  honestly supports.}
  \label{fig:ours-grids}
\end{figure}

Figure~\ref{fig:ours-grids} is the learned-codec showcase at 112\,px / 1024\,B, the
verification working resolution and the operating point our codec is trained for. On
Color~FERET (Figure~\ref{fig:ours-grid-cf}), Ours-ACCURATE keeps identity cues (eye
region, beard edge) crisper than
WebP/AVIF at the same budget, with Ours-FAST giving a lighter-weight, slightly softer
reconstruction; JPEG-AI is the strongest non-learned reference here. On
AI-Solutions-KK (Figure~\ref{fig:ours-grid-kk}), across the three Monk
skin-tone~\cite{monk2023mst} samples both
learned variants preserve identity and avoid the blocking the classical codecs show
--- Ours-FAST producing a smoother result and Ours-ACCURATE keeping more fine texture.
One important caveat carries into the fairness analysis: at this 112\,px checkpoint the
learned codec slightly darkens or shifts colour on the mid- and dark-skin crops, a
tone bias that is visible in the backgrounds and not yet corrected; it is verified and
quantified in the fairness analysis (Section~\ref{sec:fairness}).

Two structural notes apply to the montages. First, JPEG-AI is now compressed for
AI-Solutions-KK at both 112 and 224\,px, so both the 112\,px KK showcase
(Figure~\ref{fig:ours-grids}) and the 224\,px KK codec grid show the \emph{real} JPEG-AI
reconstruction, and every codec column in the grids is populated.
Second, the learned codec is decoded from its
\texttt{.bin} container on GPU (CPU decode produces NaNs for these checkpoints) and
appears at both resolutions: Ours-FAST and Ours-ACCURATE are the last two columns of the
$224$\,px codec grid (Figure~\ref{fig:codec-grids}) and of the $112$\,px showcase, where
they reconstruct clean, colour-correct faces competitive with JPEG-AI and free of the
blocking that plain JPEG shows at this budget.

\subsection{Budget and resolution sweeps}
\tldr{Halving the budget to 512\,B hits classical codecs hard (WebP drops 30.5\,dB$\rightarrow$26.8\,dB with visible blocking) while the learned codec stays smooth; at a fixed 1024\,B, smaller crops look cleaner per pixel but a 224\,px crop carries more facial detail under harder compression.}

\begin{figure}[t]
  \centering
  \includegraphics[width=0.5\linewidth]{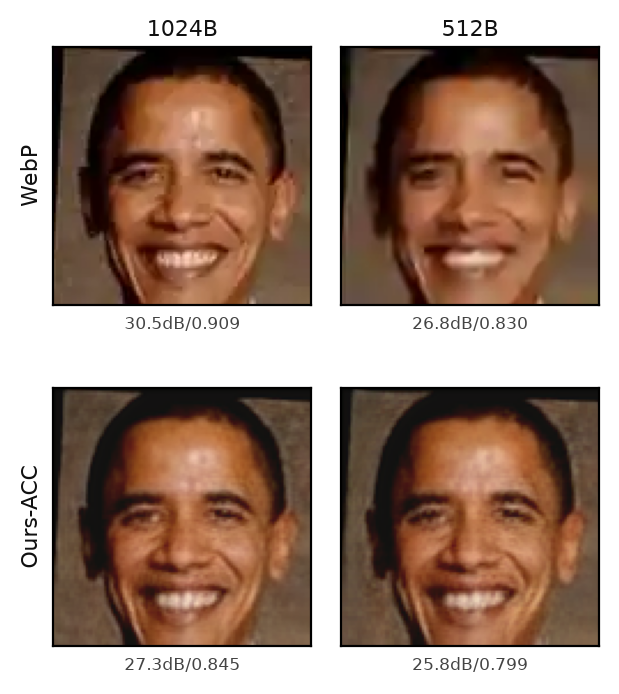}
  \caption{Budget sweep, KK mid-skin sample at 112\,px: 1024\,B (left) vs.\ 512\,B
  (right), for WebP (top) and Ours-ACCURATE (bottom); per-tile PSNR\,dB/SSIM annotated
  below each crop. Halving the budget produces visible blocking and banding in WebP
  (PSNR $30.5\rightarrow 26.8$\,dB) while the learned codec degrades gently. Shown at
  the native crop scale to avoid upscaling.}
  \label{fig:budget-sweep}
\end{figure}

The budget sweep (Table~\ref{tab:budget-sweep}, Figure~\ref{fig:budget-sweep})
isolates what happens when the byte budget is halved from 1024\,B to 512\,B at a fixed
112\,px. On AI-Solutions-KK, WebP's median PSNR drops from $30.46$\,dB to $26.78$\,dB
(SSIM $0.909 \rightarrow 0.830$) with visible blocking, colour banding and eye
artifacts. Ours-ACCURATE degrades far more gently: PSNR falls only $27.26 \rightarrow
25.78$\,dB (SSIM $0.845 \rightarrow 0.799$) and, tellingly, its LPIPS is essentially
flat ($0.014 \rightarrow 0.022$) where WebP's roughly triples ($0.032 \rightarrow
0.092$), so the learned reconstruction stays smooth and recognisable down to 512\,B.
The same pattern holds on Color~FERET (Ours-ACCURATE PSNR $28.63 \rightarrow 27.59$\,dB,
SSIM $0.834 \rightarrow 0.806$; LPIPS $0.012 \rightarrow 0.019$). This graceful
learned-codec behaviour, quantified here, is corroborated later against
verification EER.

\begin{table}[t]
  \centering
  \caption{Budget sweep at 112\,px, 1024\,B vs.\ 512\,B (dataset medians, $n$ = crops
  measured). Halving the budget costs WebP ${\sim}3.4$--$3.7$\,dB of PSNR and roughly
  triples its LPIPS; Ours-ACCURATE loses ${\sim}1$--$1.5$\,dB and its LPIPS stays
  nearly flat. Generated from the quality CSVs, so these medians and the quality
  matrix of Table~\ref{tab:quality-matrix} cannot drift apart.}
  \label{tab:budget-sweep}
  \small
  \input{tables/budget_sweep}
\end{table}

\begin{figure}[t]
  \centering
  \includegraphics[width=\linewidth]{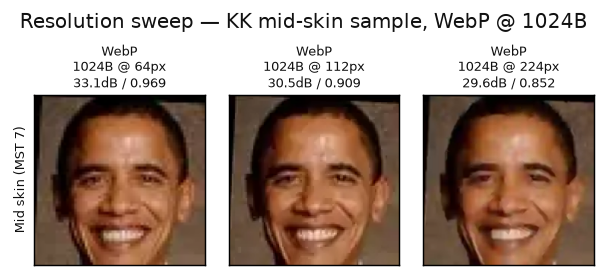}
  \caption{Resolution sweep, WebP @ 1024\,B on the KK mid-skin sample at 64 / 112 /
  224\,px. Smaller crops are cleaner per pixel (PSNR $33.05 \rightarrow 30.46
  \rightarrow 29.65$\,dB) while larger crops carry more detail under harder
  compression; 112\,px is the verification working resolution.}
  \label{fig:resolution-sweep}
\end{figure}

\begin{figure}[t]
  \centering
  \begin{subfigure}{\linewidth}
    \centering
    \includegraphics[width=\linewidth]{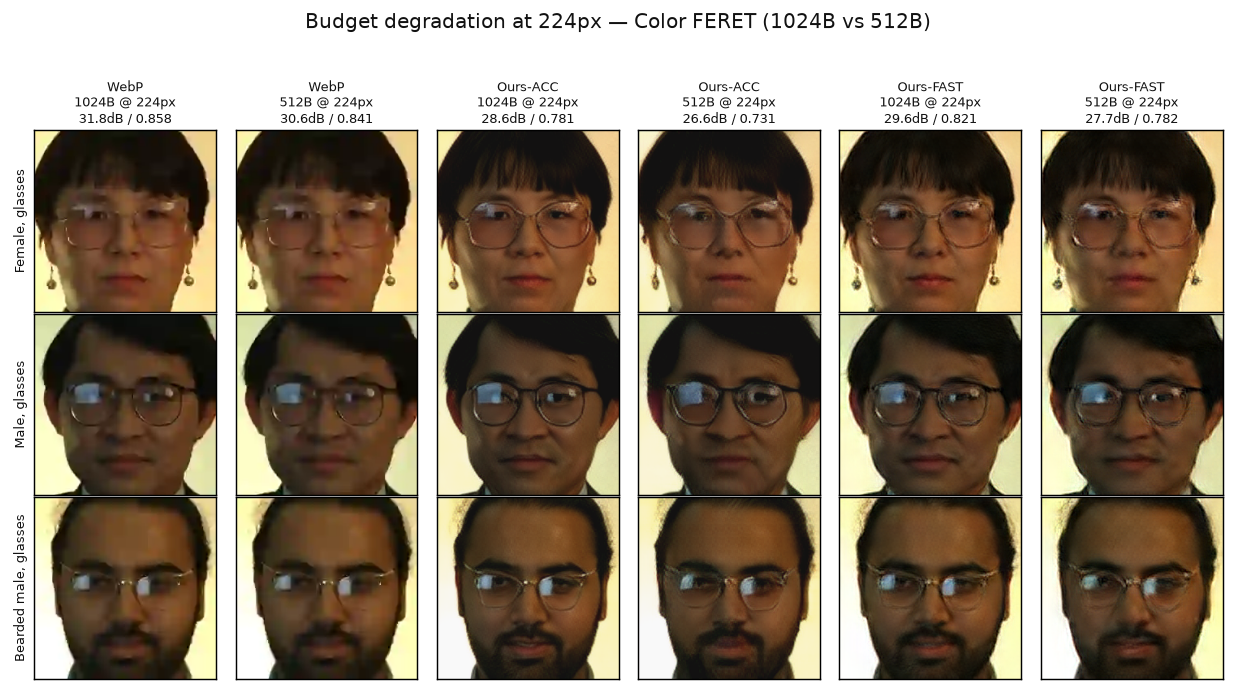}
    \caption{Color~FERET}
    \label{fig:budget224-cf}
  \end{subfigure}\\[4pt]
  \begin{subfigure}{\linewidth}
    \centering
    \includegraphics[width=\linewidth]{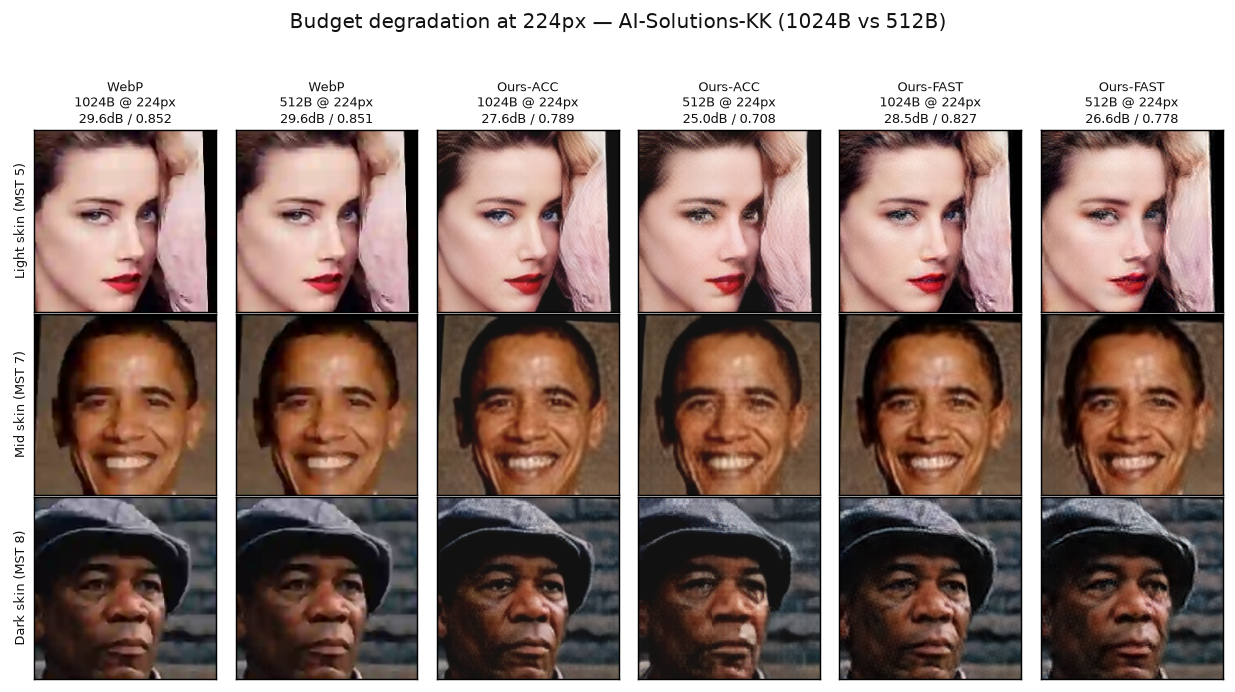}
    \caption{AI-Solutions-KK}
    \label{fig:budget224-kk}
  \end{subfigure}
  \caption{Budget degradation at the $224$\,px source resolution (WebP, Ours-ACCURATE,
  Ours-FAST at $1024$\,B vs $512$\,B), complementing the $112$\,px sweeps above. At
  $224$\,px the $16\times16$ spatial latent and per-image byte ceiling bite harder than at
  $112$\,px: WebP and Ours-ACCURATE degrade gracefully from $1024$\,B to $512$\,B, whereas
  Ours-FAST---which has no identity side-stream to absorb the high-resolution rate
  floor---cannot reach $512$\,B at $224$\,px and emits a clean \emph{over-budget} frame
  (Section~\ref{sec:codec}).}
  \label{fig:budget224}
\end{figure}

The resolution sweep (Figure~\ref{fig:resolution-sweep}) varies the crop size at a
fixed 1024\,B. At a fixed budget a smaller crop spends its bits on fewer pixels and so
looks cleaner per pixel: for WebP on the KK mid-skin sample the median PSNR runs
$33.05$\,dB at 64\,px, $30.46$\,dB at 112\,px and $29.65$\,dB at 224\,px, with LPIPS
rising $0.007 \rightarrow 0.032 \rightarrow 0.150$ over the same range. A 224\,px crop
therefore carries more facial detail in principle but is forced to compress harder, and
Figure~\ref{fig:budget224} shows what that costs once the budget is halved at the
$224$\,px source as well: WebP and Ours-ACCURATE still degrade gracefully on both
Color~FERET (Figure~\ref{fig:budget224-cf}) and AI-Solutions-KK
(Figure~\ref{fig:budget224-kk}), whereas Ours-FAST --- which has no identity side-stream
to absorb the high-resolution rate floor --- mostly cannot reach $512$\,B there. This is
exactly the tension the verification analysis resolves --- 112\,px is the
resolution we adopt as the working point because it balances per-pixel fidelity
against retained facial detail at the target budget.

\subsection{Codec properties and deployment characteristics}
\label{subsec:codec-properties}
\tldr{Beyond identity and pixel fidelity, a deployable document codec must clear
practical gates --- ownership, licensing, hardware-decoder availability, and whether it
can hit the byte budget at each resolution. Table~\ref{tab:codec-properties} collects
these; encode/decode latency is measured separately---and comparably across
codecs---under controlled single-core-CPU / single-GPU conditions in
\S\ref{subsec:speed} (Table~\ref{tab:speed}).}

The rate--identity trade-off is only half of a deployment decision. For a face-on-document
credential the codec must also be licensable, decodable on the target hardware, and able
to guarantee the byte budget at the chosen resolution. Table~\ref{tab:codec-properties}
summarises these provenance and deployment properties across the benchmarked codecs
(the table itself is placed with the codec introductions in
Section~\ref{sec:related}, Table~\ref{tab:codec-properties}).
Three practical patterns stand out. First, \emph{licensing splits the field}: JPEG,
JPEG\,2000, WebP, JPEG\,XL, AVIF, and the JPEG-AI reference software are royalty-free,
whereas HEIF inherits HEVC's patent pools and the CompressAI learned baselines ship under
a BSD licence with an explicit patent non-grant --- a real barrier for a standardised
credential. Second, \emph{hardware-decoder availability} is inversely related to
compression strength: legacy JPEG decodes on essentially every image-signal processor and
HEVC/HEIF on nearly every SoC since $\sim$2015, while the strongest codecs for this task
(JPEG-AI, JPEG\,XL, and the learned codecs, including ours) have no decoder silicon and
run in software or on a GPU/NPU. Third, \emph{the hard-budget fit rate} at $224$\,px
separates the codecs that always reach $1$\,kB (JPEG\,2000, HEIF, JPEG-FzT, JPEG-AI and
both Ours variants, all ${\ge}99.9\%$ on both datasets) from those that cannot on every
crop: on Color~FERET WebP holds $94\%$ and JPEG\,XL $87\%$, and on the in-the-wild
AI-Solutions-KK crops those two collapse to $29\%$ and $38\%$ while AVIF slips to $98\%$
and plain JPEG to $95\%$.
Encode/decode latency is reported separately in Section~\ref{subsec:speed}
(Table~\ref{tab:speed}), measured under controlled single-core-CPU and single-GPU
conditions so the numbers are comparable across codecs.

\subsection{Encode and decode speed}
\label{subsec:speed}
\tldr{Encode and decode wall-clock, measured separately and pinned to a single CPU core
(and, for the learned codecs, on one GPU), over $200$ crops (classical) / $30$ CPU and
$60$ GPU crops (learned) per cell. Classical encode spans three orders of magnitude---JPEG $0.26$\,ms
vs.\ HEIF $172$\,ms per crop at $112$\,px---and is strongly \emph{asymmetric} (decode is
$5$--$140\times$ cheaper than encode for every classical codec except plain JPEG, which is
symmetric), whereas the learned codecs are near-symmetric and
an order of magnitude slower on CPU but recover $4$--$6\times$ on GPU. JPEG-AI is the
outlier at both ends ($2.9$\,s CPU encode, $1.8$\,s even on GPU). Time grows smoothly
with resolution for every codec.}

Deployment on an edge device or a document scanner is gated by latency as much as by
identity. We benchmark encode and decode \emph{separately}, per crop, pinned to a single
CPU core with \texttt{taskset} (and to one GPU for the learned codecs, which support it),
reporting the median over $200$ crops for the classical codecs and, for the learned
ones, $30$ crops on CPU and $60$ on GPU. These are single-shot times at the budget-hitting quality/gain; the \emph{hard}
byte-budget guarantee multiplies the encode by the rate-search count (a $\approx$6-step
binary search for the classical and Ours codecs, $\approx$1--3 for JPEG-AI's analytic fit,
amortised for Ours by its gain-CDF cache), so a deployed encoder pays roughly that
multiple of the single-shot encode below; decode is unaffected.

\begin{figure}[t]
  \centering
  \includegraphics[width=\linewidth]{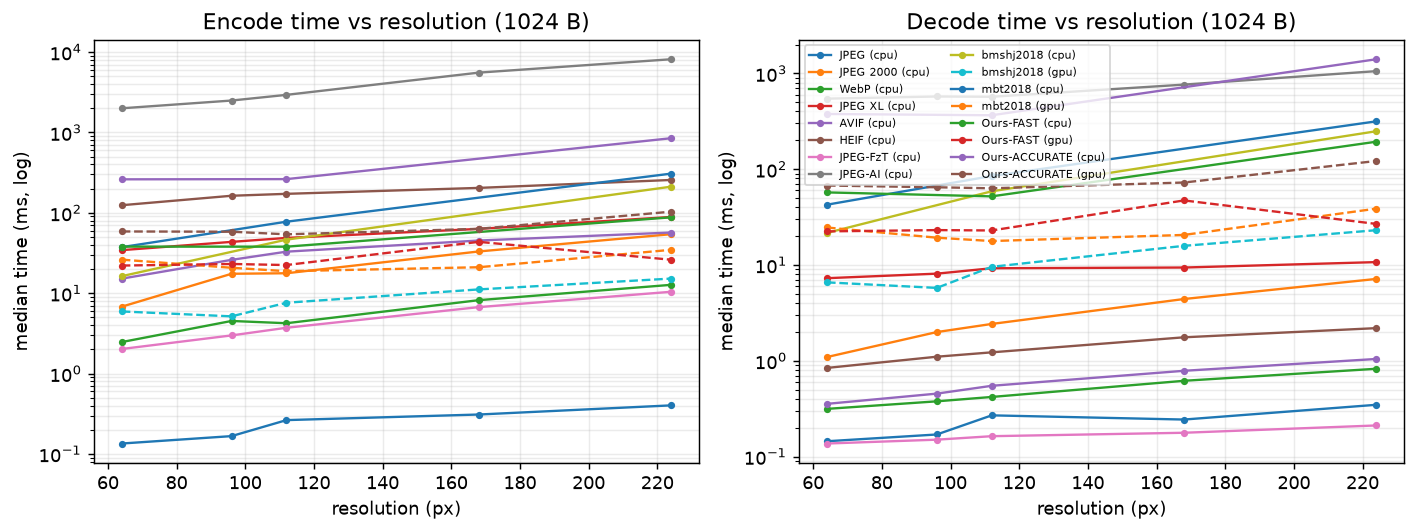}
  \caption{Median encode (left) and decode (right) time vs.\ crop resolution at $1024$\,B
  (log scale). Solid = single CPU core, dashed = GPU. Every codec's cost grows smoothly
  with resolution; classical decode stays far below classical encode, while the learned
  codecs are near-symmetric and converge toward the classical range on GPU. JPEG-AI sits
  at the top of both panels at every resolution --- an order of magnitude above the next
  codec on encode. Only its CPU series is drawn: its GPU cost is measured at the $112$\,px
  working point ($1824$\,ms encode / $959$\,ms decode, Table~\ref{tab:speed}) rather than
  swept, since one point cannot form a trend and a sweep taken while the host is busy
  would not be a property of the codec.}
  \label{fig:speed-trend}
\end{figure}

Table~\ref{tab:speed} reports the $112$\,px$/1024$\,B operating point --- with each row's
\emph{realised} median size, since the two CompressAI baselines cannot reach the budget and
were timed at $\approx$$224$\,B --- and
Figure~\ref{fig:speed-trend} the trend across resolutions. Three findings stand out.
First, \textbf{classical encode cost spans three orders of magnitude}: JPEG encodes a
$112$\,px crop in $0.26$\,ms and JPEG-FzT in $3.7$\,ms, whereas AVIF ($33$\,ms) and
especially HEIF ($172$\,ms, the x265 intra encoder) are hundreds of times slower --- the
same HEVC machinery that made HEIF a strong identity preserver makes it the most expensive
\emph{classical} codec to write. JPEG-AI is in a different regime entirely: $2934$\,ms to
encode and $570$\,ms to decode a $112$\,px crop on one CPU core, roughly $17\times$ the
cost of HEIF and four orders of magnitude above JPEG, and it is the only codec here whose
cost is dominated by a reference-software neural pipeline rather than by a transform. Its
per-crop CPU encode rises from $2.0$\,s at $64$\,px to $8.1$\,s at $224$\,px
(Figure~\ref{fig:speed-trend}), so the identity lead reported in
Section~\ref{subsec:quality-vs-identity} is bought at a latency no CPU-only deployment can
absorb; Section~\ref{subsec:jpegai-speed} takes up what can be done about that. Second, \textbf{classical codecs are strongly asymmetric}: decode is
$5$--$140\times$ cheaper than encode (HEIF $1.2$\,ms against a $172$\,ms encode, AVIF
$0.55$ against $33$\,ms, JPEG~XL $9.2$ against $49$\,ms) --- with plain JPEG the one
exception, symmetric at $0.26$/$0.27$\,ms --- so a document that is written once and read
many times pays the encode cost only at enrolment. The learned codecs, by contrast, are \textbf{near-symmetric and CPU-heavy}:
Ours-FAST is $38$/$52$\,ms (encode/decode) and Ours-ACCURATE $262$/$365$\,ms on one CPU
core, because a neural forward pass dominates both directions. Third, \textbf{the GPU
recovers most of that gap --- for every learned codec except JPEG-AI}: on one GPU
Ours-ACCURATE drops to $54$/$63$\,ms ($\approx$5--6$\times$ faster than CPU), Ours-FAST to
$22$/$23$\,ms, and the CompressAI baselines to $8$--$19$\,ms (at $224$\,B, not $1024$\,B ---
their rate knob does not reach the budget, so that lead is not measured like-for-like; see
the caption), whereas JPEG-AI still costs
$1824$/$959$\,ms, an order of magnitude more than \emph{any} other codec on either device.
The practical reading: JPEG, JPEG-FzT and WebP are real-time on
a bare CPU core; HEIF and our learned codecs are enrolment-time-only on CPU and need a
GPU/NPU for interactive encode; JPEG-AI's reference implementation is not interactive on
either. All costs grow smoothly with resolution
(Figure~\ref{fig:speed-trend}), so the $112$\,px working point is also a favourable
\emph{latency} operating point, not only an accuracy one.

\begin{table}[t]
  \centering \small
  \caption{Median per-crop encode/decode time (ms) at $112$\,px under a $1024$\,B
  \emph{request}: single CPU core
  (\texttt{taskset -c}, $n{=}200$ classical / $30$ CPU and $60$ GPU learned / $10$ CPU and
  $38$ GPU JPEG-AI, whose per-crop
  cost makes a larger sample uninformative) and, where supported, one GPU.
  Single-shot at the budget-hitting setting; the hard-budget search multiplies encode
  (see text). The ``B'' column in each device group is the \emph{realised} median payload
  that row was actually timed at, and it is \textbf{not} uniform: the classical codecs,
  JPEG-AI and both Ours variants land at $831$--$1133$\,B (within $\pm20\%$ of the
  request), but the two CompressAI baselines have no rate knob that reaches it --- their
  lowest quality level already emits only $228$/$238$\,B on CPU and $224$\,B on GPU, under
  a quarter of the budget. So bmshj2018's fastest-on-GPU cell ($7.61$/$9.56$\,ms) is
  \emph{not} measured at the same rate as the rows it beats, and this table does not
  establish it as the fastest codec \emph{at $1024$\,B}. We expect the ordering to hold ---
  these are hyperprior models whose cost is dominated by a fixed-size convolutional pass
  rather than by the bitstream, and at $224$\,px, where bmshj2018 realises $682$\,B
  ($67\%$ of the budget), it is still the fastest learned codec on GPU ($15.2$/$23.1$\,ms
  vs.\ Ours-FAST's $26.1$/$26.8$\,ms at $1007$\,B) --- but we did not measure it at
  $1024$\,B, so the control was not held and we do not claim it was.
  ``--'' = no GPU path (classical codecs) / not applicable. Note the scale:
  JPEG-AI's CPU encode is in seconds, every other codec's is in milliseconds.
  \colorbox{green!25}{Fastest}/\colorbox{red!22}{slowest} per \emph{timing} column; the
  realised-rate columns are reported, not ranked.}
  \label{tab:speed}
  \adjustbox{max width=\textwidth}{\input{tables/speed_benchmark.tex}}
\end{table}

\subsection{Why is JPEG-AI so slow, and can it be sped up?}
\label{subsec:jpegai-speed}
\tldr{JPEG-AI is the identity leader but by far the slowest codec: ${\sim}0.6$--$1.2$\,s to
\emph{decode} and $2.7$--$9.3$\,s to \emph{encode} a single sub-1\,kB $224^2$ crop on one
CPU thread --- three to four orders of magnitude slower than JPEG. The cause is structural
(a VAE transform stack plus a sequential arithmetic coder), so it cannot be tuned away; the
two available levers are a GPU (up to $4.8\times$ on encode, but only ${\sim}1.8\times$ on the
heaviest decode) and choosing a lighter \emph{operation point} --- the standard defines three
decoders (SOP/BOP/HOP) trading quality for cost, and SOP/BOP roughly halve HOP's CPU decode
for ${<}1$\,dB PSNR --- and, on a full $n{=}600$ KK verification run, at \emph{no} EER
cost (all three decoders tie at ${\sim}0.14\%$ ArcFace), so the cheapest SOP decoder is the
right default for a face-verification pipeline.}

JPEG-AI is the strongest codec on identity in this study, yet the codec-properties and speed
tables flag it as GPU-bound; this subsection quantifies \emph{how} slow it is, \emph{why},
and \emph{how far} it can be accelerated, using the ISO/IEC 6048-1 reference
software~\cite{ascenso2023jpegai}. All measurements here are taken at the rate-matcher's
nearest point to the budget, an achieved $1108$--$1113$\,B (${\approx}8.5\%$ over the
$1024$\,B target), so the decoder-profile verdict below is read at a slightly non-compliant
operating point.

\paragraph{The baseline is very slow.} On one CPU thread, JPEG-AI decodes a sub-1\,kB
$224^2$ face crop in $0.57$--$1.19$\,s and \emph{encodes} it in $2.7$--$9.3$\,s
(Table~\ref{tab:jpegai-speed}). Against the classical codecs of Table~\ref{tab:speed}
(JPEG encodes in $0.26$\,ms and decodes in $0.27$\,ms; even the expensive HEIF encodes in
$172$\,ms) this is a factor of $10^3$--$10^4$ on encode and ${\sim}10^3$ on decode --- JPEG-AI
is, in wall-clock terms, in a different regime from every other codec here.

\paragraph{Why it is slow.} The cost is architectural, not incidental. JPEG-AI is a VAE
codec: a learned convolutional \emph{analysis} transform at encode and a \emph{synthesis}
transform at decode. Its latents are coded by a learned entropy model --- a hyperprior plus
an \emph{autoregressive spatial context module} --- feeding a 32-bit arithmetic (range)
coder. The context model + arithmetic coder are inherently \emph{sequential} (each symbol's
probability depends on already-decoded neighbours), which does not parallelise onto a GPU and
dominates decode; and the encoder additionally runs a bitrate-matcher \emph{rate search}
(repeated analysis passes) to hit the byte target, which is why encode is several times
costlier than decode. Because a sub-1\,kB $224^2$ crop is tiny, the neural transforms are too
small to amortise GPU kernel-launch and host$\leftrightarrow$device transfer overhead, so a
GPU barely helps the decode.

\begin{figure}[t]
  \centering
  \includegraphics[width=\linewidth]{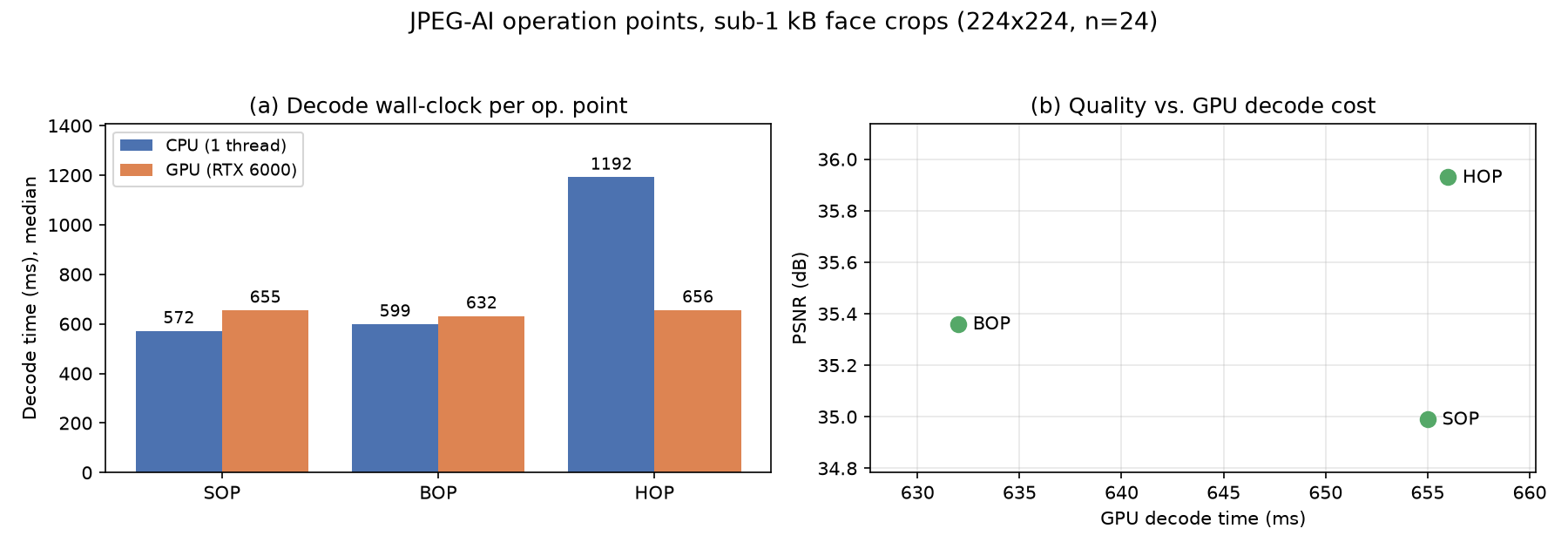}
  \caption{JPEG-AI decoder operation points. \emph{(a)}~Per-crop decode wall-clock, CPU
  (single thread) vs.\ GPU: the GPU helps only the heaviest HOP decoder. \emph{(b)}~Quality
  (PSNR) vs.\ GPU decode cost: on GPU there is essentially no trade-off --- all three
  decoders land within $24$\,ms ($655$/$632$/$656$\,ms) and BOP dominates SOP on both
  axes, being both faster and higher-quality. The ${\sim}2\times$ SOP-vs-HOP decode gap is
  a CPU effect (panel a), so the verification recommendation rests on the CPU column.}
  \label{fig:jpegai-tradeoff}
\end{figure}

\paragraph{Speeding it up: GPU and operation point.} Two levers help, and we measure both
(Table~\ref{tab:jpegai-speed}, Figure~\ref{fig:jpegai-tradeoff}). \emph{(i)~GPU}: moving to
one GPU cuts \emph{encode} by $1.4$--$4.8\times$ (largest at the heavy HOP point) but --- for
exactly the sequential-coder reason above --- gives essentially no decode speed-up at SOP/BOP
and only $1.8\times$ at HOP. \emph{(ii)~Operation point}: JPEG-AI standardises three decoders
of increasing synthesis-transform depth --- SOP (simple, 1 transform), BOP (base, 2), HOP
(high, 3). Dropping from HOP to SOP/BOP roughly \emph{halves} the CPU decode ($1192$\,ms
$\rightarrow$ $572$--$599$\,ms) for a cost of only $0.6$--$0.9$\,dB PSNR and $0.004$ identity
cosine ($0.949\rightarrow0.945$) --- a favourable trade when decode latency matters. Even so,
after both levers JPEG-AI still needs ${\sim}0.63$\,s to decode one crop on a GPU, so the
practical verdict of Section~\ref{subsec:codec-properties} stands: JPEG-AI is an
enrolment-time, server-side codec, not an interactive-decode one.

\begin{table}[t]
  \centering \small
  \caption{JPEG-AI operation points on sub-1\,kB $224^2$ Color~FERET crops ($n{=}24$,
  reference software, CPU single-threaded). Decode/encode are per-crop medians; PSNR/SSIM/
  id-cosine are the reconstruction (device-independent). Quality rises SOP$<$BOP$<$HOP with
  the synthesis-transform depth, at rising decode cost. \colorbox{green!25}{Best}/
  \colorbox{red!22}{worst} per column. Encode CPU (not shown) is $2.7$/$2.8$/$9.3$\,s for
  SOP/BOP/HOP. Achieved size $1108$--$1109$\,B --- the rate-matcher's nearest point,
  ${\approx}8\%$ over the $1024$\,B budget.}
  \label{tab:jpegai-speed}
  \input{tables/jpegai_speed.tex}
\end{table}

\paragraph{Verification-grade measurement on KK.} The timing study above scores
reconstruction quality on $24$ crops; to settle the operation-point choice on downstream
\emph{recognition} we run the full pipeline --- encode$\to$decode$\to$embed$\to$verification
EER --- for all three decoders at $224$\,px$/1024$\,B on a subject-stratified $n{=}600$
AI-Solutions-KK sample (Table~\ref{tab:jpegai-kk}). Two results matter for deployment.
First, \textbf{all three decoders preserve identity}: verification EER stays close to the
uncompressed reference (ArcFace $0.14\%$ vs.\ $0.08\%$ aligned; EdgeFace-XS $\sim$$1.2\%$
vs.\ $0.91\%$), so JPEG-AI at a kilobyte is effectively lossless for verification on
in-the-wild faces. Second, and decisively, \textbf{decoder complexity barely changes
verification}: EER is flat across SOP/BOP/HOP (ArcFace $0.140/0.147/0.140\%$, EdgeFace
$1.25/1.19/1.26\%$) even though reconstruction quality climbs monotonically with
synthesis-transform depth (id-cosine $0.935\!\rightarrow\!0.941$, PSNR
$31.1\!\rightarrow\!32.0$\,dB). The extra fidelity HOP buys is invisible to the matcher, so
the recommendation is unambiguous: for a face-verification pipeline, use the cheapest
\textbf{SOP} decoder --- ${\sim}2\times$ faster to decode than HOP on CPU
(Table~\ref{tab:jpegai-speed}) at no verification-accuracy cost.

\begin{table}[t]
  \centering \small
  \caption{Full verification-grade measurement of the three JPEG-AI decoders at
  $224$\,px$/1024$\,B on AI-Solutions-KK ($n{=}600$ subject-stratified crops):
  encode$\to$decode$\to$embed$\to$EER. EER (\%, lower better) on two anchors, plus the
  reconstruction identity cosine, PSNR and achieved bytes. \colorbox{green!25}{Best}/
  \colorbox{red!22}{worst} per column (ArcFace-EER shading; ``aligned'' is the uncompressed
  reference, excluded from shading, and its synthesis-transform, id-cosine, PSNR and byte
  cells read ``--'' because those quantities are undefined without a codec). Verification EER is statistically flat across the
  three decoders while reconstruction quality rises SOP$<$BOP$<$HOP.}
  \label{tab:jpegai-kk}
  \input{tables/jpegai_kk.tex}
\end{table}

%% file: sections/07_custom_codec.tex
\section{A Custom Identity-Preserving Codec}
\label{sec:codec}
\tldr{We train a learned face codec in two variants --- a tiny pixel-anchored
\emph{Ours-FAST} and a larger \emph{Ours-ACCURATE} with an identity side-stream and a
refine head --- that enforce a hard $\le B$-byte budget at encode time and, at
$224$\,px$/1024$\,B, place second only to JPEG-AI on identity cosine on Color~FERET
(third on AI-Solutions-KK, behind JPEG-AI and Ours-FAST) while keeping the
output verified free of colour cast or black frames.}

The classical and learned baselines benchmarked earlier control rate through a
continuous quality knob, not a byte target, so meeting a strict $\le 1024$\,B (and the
harder $\le 512$\,B) budget per image means re-encoding and searching. We instead build
a codec whose objective is identity preservation under a \emph{guaranteed} byte ceiling:
it optimises a face-recognition (FR) loss during training and enforces the budget
exactly at encode time. Both variants extend a mean-scale
hyperprior~\cite{balle2018hyperprior,minnen2018joint} with gain-unit variable-rate
control~\cite{cui2021asymmetric}, built on the CompressAI
library~\cite{begaint2020compressai}.

\subsection{Architecture}
\tldr{Both variants share a mean-scale hyperprior with a third hyper-downsample (so the
hyperprior latent is $x/128$, shrinking its byte floor) and gain-unit variable-rate
control; FAST is a tiny checkerboard-free synthesis network, ACCURATE adds attention
and an identity side-stream.}

The shared backbone subclasses CompressAI's variable-rate mean-scale
hyperprior~\cite{balle2018hyperprior,minnen2018joint,cui2021asymmetric}: a learned gain
vector sets the quantisation step so a single trained model spans all rate points. To it
we add a \emph{third} stride-2 stage in the hyper-analysis (mirrored by a third upsample
in hyper-synthesis), so the hyperprior latent $z$ is $x/128$ rather than the stock
$x/64$. Quartering the $z$-grid area, together with a small hyperprior channel count,
drives the gain-independent $z$ byte floor down by roughly an order of magnitude
(the backbone notes a $\approx 12$--$16\times$ reduction, from the stock
$\approx 524$\,B at $112$\,px to the $\approx 48$\,B order the budget needs) --- the
change that makes sub-1\,kB latents feasible at all, since at $112$\,px the stock
floor alone would already exhaust most of the budget.
The total geometric downsample is $2^{4+3}=128$, so inputs are edge-padded to a multiple
of $128$ before entropy coding.

Both variants replace the stock synthesis with a conditional synthesis transform whose
every $\times 2$ upsample is a \emph{ResizeConv} (nearest-neighbour upsample followed by
a convolution) rather than a transposed convolution; this removes the checkerboard
artefacts transposed convolutions produce at very low rate. After each inverse-GDN block
a per-stage FiLM layer modulates the features on $(\text{gain}, \text{resolution})$, and
for ACCURATE additionally on the decoded identity code.

\begin{figure}[t]
  \centering
  \resizebox{\linewidth}{!}{%
  \begin{tikzpicture}[
      >=Stealth, node distance=3.5mm and 5mm, font=\scriptsize,
      blk/.style={draw=TldrBar, rounded corners, fill=TldrBG, align=center,
        minimum height=9mm, inner sep=3pt, text width=22mm},
      acc/.style={draw=ToDoRed, rounded corners, fill=white, align=center,
        minimum height=9mm, inner sep=3pt, text width=22mm, dashed},
      io/.style={draw=HeadGray, rounded corners, fill=white, align=center,
        minimum height=9mm, inner sep=3pt, text width=13mm}]
    \node[io] (x) {aligned crop\\$x$};
    \node[blk, right=of x] (ga) {analysis $g_a$\\(gain $\odot$,\\attn.\ in ACC)};
    \node[blk, right=of ga] (search) {binary search over\\frozen 64-gain table\\
      ($\le 6$ rANS encodes,\\real bytes $\le B$)};
    \node[blk, right=of search, text width=25mm] (cont)
      {container $\le B$\\ \texttt{flags$|$res$|$sc$|$y$|$z}\\(7\,B overhead)};
    \node[blk, below=of ga, text width=22mm] (ha)
      {hyperprior $h_a$\\3rd downsample:\\$z$ grid $=x/128$};
    \node[acc, below=of ha] (anchor) {frozen FR anchor\\(EdgeFace-S,\\512-D)};
    \node[acc, right=of anchor, text width=24mm] (proj)
      {learned proj.\ $\to$128-D\\$+$ entropy code\\($\approx$90--175\,B)};
    \node[blk, right=of cont, text width=24mm] (gs)
      {synthesis $g_s$\\ResizeConv $+$ FiLM\\(gain, res., id code)};
    \node[acc, below=of gs, text width=24mm] (refine)
      {gated refine head\\$\hat{x}+0.1\cdot$residual\\(id-conditioned)};
    \node[io, right=of gs] (xhat) {decoded\\$\hat{x}$};
    \draw[->] (x)--(ga);
    \draw[->] (ga)--(search) node[midway, above, font=\tiny] {$y$};
    \draw[->] (search)--(cont);
    \draw[->] (ga)--(ha);
    \draw[->] (ha.east) -| node[pos=0.25, above, font=\tiny] {$z$} (search.south);
    \draw[->, ToDoRed] (x.south) -- ++(0,-32mm) -| (anchor.west);
    \draw[->, ToDoRed] (anchor)--(proj);
    \draw[->, ToDoRed] (proj.east) -| node[pos=0.28, above, font=\tiny]
      {side-channel} (cont.south);
    \draw[->] (cont)--(gs) node[midway, above, font=\tiny] {decode};
    \draw[->] (gs)--(xhat);
    \draw[->, ToDoRed] (gs.south)--(refine.north);
    \draw[->, ToDoRed] (refine.east) -| (xhat.south);
  \end{tikzpicture}}
  \caption{Architecture of the custom byte-budgeted codec. Blue solid blocks are
  shared by both variants: a mean-scale hyperprior with a third hyper-downsample
  (shrinking the gain-independent $z$ byte floor by roughly an order of magnitude),
  gain-unit variable-rate control, and a binary search over a frozen 64-entry gain
  table that measures real rANS bytes so the emitted container is guaranteed
  $\le B$. Red dashed blocks exist only in Ours-ACCURATE: an entropy-coded identity
  side-stream derived from a frozen FR anchor, injected into the decoder by FiLM,
  and a gated refinement head conditioned on the same decoded identity code.}
  \label{fig:codec-arch}
\end{figure}
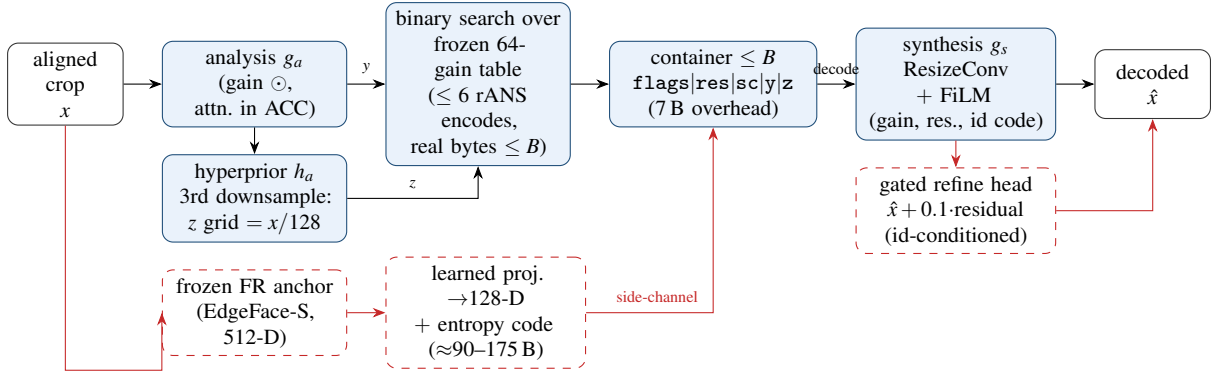

\emph{Ours-FAST} (\texttt{variant\_id}\,$=0$) is a slim, attention-free network with no
side-stream, intended as a real-time pixel-anchored codec. Its on-disk default
constructor instantiates $N{=}64$, $M{=}96$, $N_z{=}48$, totalling
$1{,}352{,}021$ parameters ($\sim$1.35\,M).
\emph{Ours-ACCURATE} (\texttt{variant\_id}\,$=1$) widens the transform
($N{=}192$, $M{=}320$, $N_z{=}64$ in the default constructor), inserts four attention
blocks (two in the analysis transform, two mirrored in synthesis), and adds the
identity side-stream and refine head described below; the network totals
$18{,}712{,}400$ parameters ($\sim$18.7\,M), of which $3{,}652{,}520$ are the frozen
EdgeFace-S anchor, leaving $15{,}059{,}880$ trainable.
Figure~\ref{fig:codec-arch} summarises
how the pieces fit together at encode and decode time; the following subsections
detail each component.

\subsection{Hard byte-budget container}
\tldr{Because the learned rate knob is continuous, the $\le B$ guarantee is enforced at
encode time by a binary search over a frozen 64-entry gain table that measures real rANS
bytes; the result is packed into a self-describing container (7\,B overhead) with an
identity-only fallback that always parses.}

A learned codec cannot be told ``produce $B$ bytes''; it produces whatever the chosen
gain yields. We therefore enforce the budget at \emph{encode} time. Encoding takes a
frozen 64-entry, log-spaced gain table spanning the model's trained gain range (gains
outside that range reconstruct garbage, so the table is derived per model). Because the
emitted size increases monotonically with gain (finer quantisation $\to$ more bytes), a
binary search over the 64 levels finds the largest gain whose \emph{real} rANS byte
count stays $\le B$ in $\approx \log_2 64 = 6$ encodes, measuring actual coder output
each step rather than estimating it. The selected level rides in the header as a 6-bit
rate index, so the decoder reconstructs the exact gain bit-for-bit --- a float-gain
mismatch would silently corrupt the decode, since the reconstruction offsets depend on
the gain.

The result is packed into a self-describing big-endian container
(Table~\ref{tab:codec-container}): a flags byte carrying the variant id (2 bits) and
rate index (6 bits), a 1-byte resolution bucket, a 1-byte side-channel length, the
optional identity side-channel, and the two length-prefixed rANS strings ($y$ and $z$);
rANS streams are not self-delimiting, hence the explicit 2-byte length prefixes. Fixed
overhead excluding payloads is 7\,B, and no CRC is stored because integer rANS is
deterministic. A dedicated identity-only variant code guarantees that even a pathological
crop, or a budget too small for any spatial latent, still emits a parseable $\le B$
stream carrying only the identity side-channel.

\begin{table}[t]
  \centering
  \caption{Byte-budget container layout (big-endian). Overhead excluding payloads is
  7\,B. The side-channel is present only for Ours-ACCURATE; the $H,W$ trailer appears
  only for the raw-geometry resolution code.}
  \label{tab:codec-container}
  \small
  \begin{tabular}{llp{0.45\linewidth}}
    \toprule
    Bytes & Field & Meaning \\
    \midrule
    1 & flags & $(\text{variant\_id} \ll 6)\;|\;\text{rate\_index}$ \\
    1 & res\_bucket & resolution code; $255 =$ raw geometry \\
    1 & sc\_len & identity side-channel length (0 if none) \\
    sc\_len & side-channel & identity payload (ACCURATE only) \\
    2 & len($y$) & length of the $y$ rANS string \\
    len($y$) & $y$ string & spatial latent \\
    2 & len($z$) & length of the $z$ rANS string \\
    len($z$) & $z$ string & hyperprior latent \\
    4 & $H,W$ & only if res\_bucket $=255$ \\
    \bottomrule
  \end{tabular}
\end{table}

\subsection{Variable-rate gain and training}
\tldr{One model spans all budgets: each training step samples a gain level and a
resolution bucket and optimises a rate--distortion objective dominated by an LPIPS and
identity-cosine loss; at encode time the frozen gain table plus binary search turn that
continuous knob into a hard byte target.}

A single model serves every operating point. During training each step samples a gain
level $s$ and weights the distortion by that level's $\lambda$ in the rate--distortion
objective $R + \lambda \cdot D$, so one network learns the whole rate spread; each step
also samples a resolution bucket (default set $\{64,128,192,256\}$, uniform; both the
set and the sampling weights are configurable per run), resizes the $112^2$ crops
to it, and edge-pads to a multiple of $128$. The distortion is dominated by a
learned-perceptual LPIPS term (which the build notes report tracks identity far better
than MSE/SSIM~\cite{zhang2018lpips}) plus MSE and MS-SSIM, with the identity-cosine term
warmed up from zero to avoid the degenerate ``blur fools the matcher'' solution; a
distortion-scaling factor restores the CompressAI $\sim\!255^2$ distortion magnitude so
the model spends bits up to the budget instead of collapsing the latent to near-zero
rate. The trainer is step-based (no epochs): each batch is drawn uniformly at random
with replacement from the full WebFace42M corpus ($\approx 42$\,M images, loaded via
HuggingFace \texttt{datasets}), with no augmentation. Color~FERET and AI-Solutions-KK are
separate evaluation-only corpora, never seen during training. Here ``held-out'' denotes a
disjoint-identity train/val split \emph{within} WebFace42M, assigned by
\texttt{crc32(subject\_id) \% val\_mod}, and the two reported held-out matchers --- the
proprietary \texttt{inno-balanced} embedder for identity-cosine (\S\ref{sec:protocol}) and
\texttt{cvlface\_ir101} for verification EER (\S\ref{subsec:codec-heldout}) --- are both
architecturally independent of the frozen EdgeFace-S side-stream. A train/test overlap check has now been run. For every Color~FERET/KK evaluation
identity we build an ArcFace centroid and record its maximum cosine against a $2$\,M-image
($4.7\%$) random sample of WebFace42M. Color~FERET (non-celebrity studio portraits) shows
\emph{no} overlap; $22$ of the $1099$ evaluation identities ($2.0\%$ --- all
AI-Solutions-KK celebrities, e.g.\ Neil~Patrick~Harris, Anne~Hathaway, Morgan~Freeman)
exceed a $0.8$ cosine, unsurprising as both AI-Solutions-KK and WebFace42M are scraped
celebrity faces. This is \emph{identity} overlap, not \emph{image} overlap: since a
compression model conditions on pixels rather than identity, a \emph{different} photograph
of a known face confers no test-time advantage, and the classical codecs have no training
set at all. Actual leakage --- a shared or near-duplicate image in both training and test,
possible only for the learned codecs (Ours, JPEG-AI, the neural baselines) --- would
register as a cosine near $1.0$; the observed maximum over all evaluation identities is
$0.915$, below the $\approx 0.97$--$1.0$ regime of a reused image. We also ran a direct
image-level check, a perceptual-hash (pHash) near-duplicate scan over the same $2$\,M
sample. Aligned crops confound pHash --- every crop fixes the eyes, nose and mouth in the
same positions, so $34\%$ of evaluation crops fall within a $6$-bit Hamming distance of
\emph{some} training image, making the Hamming count alone uninformative. Visual inspection
of the closest pairs is decisive: the three exact-hash ($0$-bit) matches, and the
low-Hamming Color~FERET matches, are all \emph{different individuals} (e.g.\ a KK actress
hash-colliding with an unrelated bearded man) --- collisions, not reused images. The two
checks agree: evaluation sets share celebrity \emph{identities} with the training corpus
(AI-Solutions-KK only) but no \emph{images}. Both use only a $4.7\%$ sample, so residual
reuse cannot be fully excluded, though neither check shows a positive signal. The
main parameters are optimised with Adam (default LR $10^{-4}$) under a cosine schedule
(\texttt{CosineAnnealingLR}, $\eta_{\min}=10^{-5}$) preceded by a short linear warmup of
$1\%$ of the total steps; a separate Adam aux-optimiser (LR $10^{-3}$) steps the
entropy-bottleneck quantiles. The trainer defaults are $600$k steps and batch $32$. The
distortion weights differ by variant: the MSE ($D$) weight is $2.0$ for ACCURATE versus
$0.3$ for FAST, with the identity-cosine weight $0.5$/$0.4$ and a distortion scale
$\text{DSCALE}=256$. The shipped checkpoints are \texttt{fast\_rbal\_1M} for Ours-FAST
($1{,}000{,}000$ steps) and \texttt{accurate\_3M} for Ours-ACCURATE ($3{,}000{,}000$
steps), both at batch $32$ and each warm-started from a shorter run of the same variant:
FAST from a $300$k run, and ACCURATE from the $990$k-step \texttt{accurate\_1M} (itself
warm-started from a $60$k run), which \texttt{accurate\_3M} resumes and extends to $3$M
steps on a single rebuilt cosine schedule. The FAST retraining replaced the default bucket set
with $\{64,128,192,224,256\}$ under sampling weights $1{:}4{:}3{:}3{:}5$ (i.e.
$\approx\!50\%$ of steps at $224/256$): an earlier $1$M retrain with the default,
low-resolution-heavy buckets improved $112$\,px identity but \emph{regressed}
$224$\,px, and the re-balanced set removes that trade-off; FAST's own $3$M extension
diverged at ${\approx}2.43$M steps and was discarded, so FAST ships at $1$M. ACCURATE kept
the default uniform buckets in both stages --- the first run, \texttt{accurate\_1M}, was
stopped at $990$k steps with the loss converged, and the $3$M extension
(\texttt{accurate\_3M}, the deployed checkpoint) resumed from it under the same bucket
configuration. Both
promotions were validated by a paired A/B against the previous checkpoints at both
$112$ and $224$\,px before any result in this report was regenerated: for
\texttt{accurate\_3M} a $120$-image paired test over both resolutions and both budgets
won $299$ of $480$ image comparisons ($62\%$, $p=8.0\times 10^{-8}$), with no cell
regressing on identity cosine or PSNR. Every Ours-ACCURATE number in this report is
measured on \texttt{accurate\_3M}, and every Ours-FAST number on
\texttt{fast\_rbal\_1M}; the sole exception is the side-stream ablation of
\S\ref{subsec:codec-side}, whose three arms are deliberately held at matched $1$M-step
from-scratch runs --- compared against the $990$k-step \texttt{accurate\_1M} --- so the
comparison is not confounded by training budget.

For ACCURATE the variable-rate range was widened downward so the tight $\le 512$\,B
point is reachable at high resolution: at $224^2/256^2$ the $16\times 16$ spatial latent
plus the identity side-channel impose a high rate floor, so the lower-$\lambda$ levels
are trained to genuinely crush rate. At encode time the frozen, log-spaced 64-entry gain
table and the binary search convert this continuous knob into the hard byte target. We
cache the per-gain entropy-coder CDF buffers: because the 64 table gains recur across
every image and every search step, the first build of each gain is memoised and restored
by reference thereafter. The restored buffers are bit-identical to a rebuild, so the
emitted bytes are unchanged; the implementation notes put the rebuild the cache removes
at ${\approx}80\%$ of encode cost (${\approx}180$\,ms per gain), which is what bounds the
attainable encode speed-up. We report no end-to-end figure for it: the speed benchmark
of \S\ref{subsec:speed} times a single-shot encode at the budget-hitting gain and
deliberately excludes the rate search and its cache.

Two encode/decode defects surfaced and were fixed without retraining. First, an early
\emph{proportional} gain search jumped straight to the first gain that fit coming down
from the maximum; on the non-linear rate curve that overshoots far below budget (emitting
$\approx 295$\,B against a $1024$\,B target). The binary search replaces it and always
uses as much of the budget as the table allows. Second, a colour cast appeared because
the refine head ran in the training \texttt{forward} but not in \texttt{decompress}; the
fix applies the refine at decode using the decoded identity prior so the real decode
reproduces the trained output (Section~\ref{subsec:codec-side}). A NaN guard additionally
sanitises the CPU decode path so a non-finite reconstruction can never propagate into the
downstream FR embeddings.

\subsection{Identity side-stream and refine head}
\label{subsec:codec-side}
\tldr{Ours-ACCURATE spends a few extra bytes on an entropy-coded identity code derived
from a frozen FR anchor; that code FiLM-modulates the decoder and drives a lightly-gated
refine head, giving a ``hard identity floor'' that survives even when the spatial latent
is crushed.}

Ours-ACCURATE carries a small identity side-stream alongside the spatial latent. A frozen
FR anchor embeds the input (512-D, L2-normalised); a learned projection compresses it to
a 128-D code, and a factorised entropy bottleneck codes that code to a small payload (the
code documents $\approx 90$--$175$\,B). The
decoded code drives per-stage FiLM $(\gamma,\beta)$ injected into the decoder synthesis,
with the affine bounded ($\pm 20\%$ scale, $\pm 0.2$ shift) so it modulates rather than
recolours.

\paragraph{The side-stream anchor runs at inference time, not only in training.} To avoid
any ambiguity: for Ours-ACCURATE the EdgeFace-S anchor is part of the deployed
\emph{encoder}. At encode time it embeds the input crop, the projection and entropy
bottleneck turn that embedding into the identity payload, and the payload is written into
the container; at decode time the stored payload is entropy-decoded and consumed by the
FiLM and refine head. It is therefore \emph{not} a training-only teacher whose cost
disappears at deployment --- an Ours-ACCURATE encoder must run one forward pass of
EdgeFace-S per image (its $\sim$90--175\,B are already counted inside the byte budget). The
frozen anchor's own weights are never updated. Ours-FAST has no side-stream and never
invokes any FR model, at training or inference. This side-channel is never dropped: it is the ``hard identity floor'' that
survives spatial-latent collapse, and at the tightest budgets it backs the identity-only
fallback frame.
The shipped anchor is the torch-roster EdgeFace-S model (the ACCURATE constructor's
default \texttt{anchor="edgeface\_s"}); the spec's separately-sourced
ArcFace-iresnet100 weights are not on disk here, so the EdgeFace anchor is used in their
place. Because the side-stream's anchor is an EdgeFace model, the held-out EdgeFace
evaluator is no longer fully architecturally independent of ACCURATE; we therefore treat
ACCURATE's EdgeFace numbers with caution and base its headline ranking on the
architecturally independent anchors (ArcFace, LVFace-L), as in
Section~\ref{sec:frr-far}.

A gated refinement head then predicts a small residual on the decoded image, conditioned
on the same decoded identity code, returning $\mathrm{clamp}(\hat{x} +
0.1\cdot\text{residual}, 0, 1)$. The low gate keeps the residual a \emph{sharpener} of a
faithful reconstruction rather than a generator that could hallucinate a wrong face; the
spatial latent stays dominant. Critically, the refine head must run at decode using the
\emph{decoded} prior --- the bug above was that it ran only in \texttt{forward}, so the
base synthesis' colour bias survived as a cast until the decode path was made to apply
the same refinement.

\paragraph{What the side-stream buys on an independent matcher (ablation).} To test
whether the side-stream contributes identity that survives on a matcher it was never
anchored to, we retrained Ours-ACCURATE from scratch under one protocol in three variants
--- side-stream \emph{on} (the control), side-stream \emph{off}, and side-stream anchored
to a non-EdgeFace backbone (TopoFR-R100) --- and scored median identity-cosine (recon vs.\
original) over a Color~FERET\,+\,AI-Solutions-KK sample ($n{=}200$ each, $112$\,px, both
budgets) under two matchers architecturally independent of every anchor, ArcFace and
CVLFace-IR101. The result is candid: \textbf{on these independent matchers the side-stream
buys essentially nothing} --- the control-minus-no-side identity-cosine gap averages
$-0.0002$ (range $[-0.004,+0.002]$ across matcher/dataset/budget), i.e.\ within noise of
zero. Re-anchoring the side-stream to TopoFR instead of EdgeFace likewise changes little
($-0.003$ on average), confirming the effect is not anchor-specific. This is the honest
counterpart to the strong EdgeFace-family numbers: the side-stream's measurable identity
gain is largely confined to its own backbone family, and Ours-ACCURATE's competitiveness
on the \emph{independent} matchers (Sections~\ref{sec:frr-far},
\ref{subsec:codec-heldout}) is carried by the spatial latent, not the stored identity
code. (All three ablation arms trained from scratch to $1{,}000{,}000$ steps and are compared
against the $990$k-step \texttt{accurate\_1M} on the same protocol, where that
warm-started reference stays marginally strongest. The deployed \texttt{accurate\_3M} is
deliberately \emph{excluded} here: swapping one arm to a $3$M-step run would confound
training budget with the ablated variable, so the ablation is matched at $1$M and speaks
to the side-stream, not to the shipped checkpoint.)

\paragraph{Budget reallocation: spend the side-stream's bytes on the spatial latent.} The
side-stream costs $90$--$175$\,B --- $9$--$17\%$ of a $1024$\,B container and $18$--$34\%$ of
a $512$\,B one. The natural follow-up is: if we remove it, those bytes flow to the spatial
latent (the binary search still fills the same budget), so does reconstruction improve without
costing identity? The side-off variant answers this directly, because it \emph{is} the
reallocated-budget codec. Across the four cells and two independent matchers, removing the
side-stream changes median identity cosine by $+0.0002$ on average (range $[-0.002,+0.004]$)
--- i.e.\ no identity is lost --- while reconstruction PSNR \emph{rises} by
$0.69$--$1.02$\,dB (mean $+0.82$\,dB), because the freed bits go to the broadband latent.
So on independent matchers the $90$--$175$\,B identity code is not merely neutral: the same
bytes buy strictly more when spent on the spatial latent. This is the quantitative case for
cutting the side-stream in a redesign; the deployed \texttt{accurate\_3M} keeps it only
because it inherits it from the warm-started \texttt{accurate\_1M} lineage it resumes, and
because it helps the EdgeFace family it is anchored to
(Section~\ref{subsec:codec-side}). The side-stream also mildly \emph{worsens} adversarial
sanitization (the refine head re-injects high-frequency detail, so Ours-ACCURATE leaves the
larger adversarial residual in $117$ of the $128$ comparable
dataset\,$\times$\,matcher\,$\times$\,attack\,$\times$\,$\varepsilon$\,$\times$\,budget
cells, the eleven exceptions all on AI-Solutions-KK/LVFace-L;
Section~\ref{sec:adversarial}), reinforcing
that its net contribution outside the EdgeFace family is neutral-to-negative.

\paragraph{The stored side-code is a linkable biometric (privacy).} Because the side-stream
stores a projection of a frozen EdgeFace-S embedding, the $90$--$175$\,B identity code is
itself a biometric descriptor, so a face-on-document deployment must ask whether it is
invertible or linkable across documents. We extract the stored $128$-d code for $2{,}500$
Color~FERET crops ($842$ identities) and test re-identification from the \emph{code alone}. A
naive cosine attack is weak (EER $35.8\%$, ROC-AUC $0.63$) because the code carries a large
shared bias direction, but a trivial attacker who mean-centers the codes recovers a
strong linker: \textbf{EER $17.0\%$, ROC-AUC $0.914$}, versus $50.2\%$/$0.498$ for a
random code of the same dimension. Two documents of the same subject are thus matchable from
their stored side-codes alone with ${\sim}83\%$ accuracy: the side-channel is a stored,
linkable biometric template, not benign conditioning metadata, and would need the same
template protection as any enrolled descriptor. Together with the ablation above (no
independent-matcher identity gain, $90$--$175$\,B cost, weaker sanitization), this is a further
reason to remove or protect the side-stream in a deployment. A full
template-reconstruction (invertibility) attack is left to future work; linkability alone
already establishes the descriptor is identity-bearing.

\subsection{Results versus classical and JPEG-AI}
\tldr{At $224$\,px$/1024$\,B Ours-ACCURATE reaches identity cosine $0.947$ on
Color~FERET --- second only to JPEG-AI ($0.958$) and above AVIF and WebP --- and $0.934$
on KK; a quality-regression check confirms the codec is verified clean (no cast/black
frames, $\approx 98\%$ budget fill).}

Table~\ref{tab:codec-results} reports median identity cosine and PSNR on $224$\,px
held-out pairs at both budgets, drawn from the codec comparison (the $224$\,px cell is
the headline operating point quoted throughout this section and the abstract; the
$112$\,px identity cosines are in fact at or above the $224$\,px values ---
$\approx 0.948$/$0.938$ for Color~FERET/KK at $1024$\,B, versus $0.947$/$0.934$ at
$224$\,px, so $112$\,px does not cost identity for Ours --- and are tabulated in
Table~\ref{tab:codec-results-112}). At $1024$\,B
Ours-ACCURATE reaches identity cosine $0.947$ (native-$224$ PSNR $28.6$\,dB) on Color~FERET,
second only to JPEG-AI ($0.958$)~\cite{ascenso2023jpegai} and ahead of AVIF
($0.899$)~\cite{aomavif} and WebP ($0.893$)~\cite{google2010webp}; on the in-the-wild KK
set it holds $0.934$, behind JPEG-AI ($0.948$) and also behind our own FAST variant
($0.941$). At the harder $512$\,B point
identity cosine drops to $0.797$ (Color~FERET) and $0.776$ (KK), where JPEG-AI, WebP,
JPEG~XL and Ours-FAST retain an edge on both datasets, and AVIF on KK. That comparison
is not like-for-like, however: at $224$\,px$/512$\,B only JPEG-AI ($100\%$ of crops
within budget) and Ours-ACCURATE ($100\%$/$99.98\%$) actually meet the budget, whereas
WebP holds it on $2.8\%$ (Color~FERET) and $0.2\%$ (KK) of crops --- median $756$ and
$1150$\,B --- JPEG~XL on $0.2\%$/$0.1\%$ (median $872$/$1090$\,B) and AVIF on
$\le 0.01\%$ (median $696$/$834$\,B); Ours-FAST holds it on $80\%$/$14\%$
(Table~\ref{tab:fit-cf}, Table~\ref{tab:fit-kk}). Notably, Ours-ACCURATE attains these identity scores at markedly lower
PSNR than JPEG-AI ($28.6$ vs $33.2$\,dB at CF$/1024$\,B, native-resolution fidelity from the
quality matrix, Table~\ref{tab:quality-matrix}), consistent with an objective
that spends its limited bits on the identity-relevant signal rather than on broadband
pixel fidelity, and the learned codecs attain the lowest LPIPS of the field at $512$\,B
(native-$224$ Ours-ACCURATE $0.111$ CF / $0.102$ KK and Ours-FAST $0.112$/$0.082$, versus
$0.16$--$0.62$ for the classical codecs).

The $112\leftrightarrow224$\,px inversion is central and easy to misread across the
rate--EER and codec-comparison tables, so Tables~\ref{tab:codec-results}
and~\ref{tab:codec-results-112} report the two resolutions side by side over the same
roster, budgets and held-out metric --- read one against the other and the inversion is
a row-by-row comparison rather than an inference across sections.
For \emph{our} codec, $112$\,px is at least as good as $224$\,px at $1024$\,B and
markedly better at $512$\,B (Color~FERET id-cosine $0.910$ at $112$ vs $0.797$ at $224$),
because a fixed byte budget buys proportionally more of a $112$\,px crop; for JPEG-AI the
trend reverses ($224 > 112$), as its transform exploits the extra resolution. This is why
Ours-ACCURATE leads at the $112$\,px working point but is the runner-up at the $224$\,px
headline cell on Color~FERET, and third there on AI-Solutions-KK.

\begin{table}[t]
  \centering
  \caption{Median identity cosine ($\uparrow$, measured with the held-out proprietary
  \texttt{inno-balanced} embedder at the $112$\,px matcher input; see \S\ref{sec:protocol}) and
  native-$224$ PSNR/dB ($\uparrow$, full-reference from the quality matrix,
  Table~\ref{tab:quality-matrix}) at the $224$\,px source resolution on
  held-out pairs, $\le B$ bytes, over the codec roster.
  \colorbox{green!25}{Best}/\colorbox{red!22}{worst} per column within each dataset;
  the CompressAI neural baselines are omitted. JPEG-AI's native-$224$ PSNR is now measured
  on both datasets (AI-Solutions-KK $31.2$\,dB at $1024$\,B). JPEG-AI is the only
  codec consistently above Ours-ACCURATE on identity. Each id-cosine cell is the median over $n{=}64$ held-out crops. Source:
  \texttt{codec\_comparison}
  (id-cos) + \texttt{quality\_summary} (PSNR).}
  \label{tab:codec-results}
  {\small\input{tables/codec_results.tex}}
\end{table}

\begin{table}[t]
  \centering
  \caption{The $112$\,px companion to Table~\ref{tab:codec-results}: same held-out
  \texttt{inno-balanced} identity cosine and same roster, at the $112$\,px verification
  working resolution (PSNR is native-$112$ full-reference).
  \colorbox{green!25}{Best}/\colorbox{red!22}{worst} per column within each dataset.
  Read against Table~\ref{tab:codec-results} this is the resolution inversion in one
  place: Ours-ACCURATE moves from runner-up at $224$\,px on Color~FERET (third on
  AI-Solutions-KK, behind JPEG-AI and Ours-FAST) to \emph{first} at $112$\,px on both
  datasets, while JPEG-AI moves the other way, because a fixed byte budget buys
  proportionally more of a $112$\,px crop than of a $224$\,px one. Each id-cosine cell is
  the median over $n{=}64$ held-out crops.}
  \label{tab:codec-results-112}
  {\small\input{tables/codec_results_112.tex}}
\end{table}

An independent quality-regression check (a spot check over $n{=}8$ crops per cell)
confirms the codec is verified clean: across both variants, both datasets, both
resolutions ($112$ and $224$\,px) and both budgets, every row passes with an \texttt{ok}
flag, no non-finite pixels, negligible brightness delta ($|\Delta|\le 0.013$) and colour
cast ($\le 0.004$). For Ours-ACCURATE the byte fill is $0.977$--$0.995$ of budget
(median $\approx 0.98$) --- the binary-search fix at work --- and no rows are flagged for
the cast or black-frame artefacts the earlier decode bug had produced. The \texttt{ok}
flag does not test budget compliance, however: Ours-FAST's AI-Solutions-KK
$224$\,px$/512$\,B row passes it at a byte fill of $1.09$, the over-budget behaviour
discussed under known limitations below. Reconstructions of both variants across
budgets and resolutions on the two datasets are shown in the learned-codec showcase of
Section~\ref{sec:quality} (Figure~\ref{fig:ours-grids}), which we do not duplicate here.

\subsection{Held-out independent-matcher evaluation}
\label{subsec:codec-heldout}
\tldr{Because the identity side-stream is anchored to an EdgeFace model, we re-score the
reconstructions on \texttt{cvlface\_ir101} --- an IR-101 trained on WebFace12M, disjoint
from both the EdgeFace side-stream and the ArcFace/LVFace evaluation anchors. On this
truly held-out backbone Ours-ACCURATE tracks the strongest codecs at $1024$\,B and is the
\emph{best} codec on both datasets at $512$\,B, and the
ACCURATE\,$>$\,FAST ordering and the modern-vs-legacy ranking are preserved.}

The headline concern for the custom codec is evaluator circularity: its identity
side-stream is anchored to a frozen EdgeFace-S embedding
(Section~\ref{subsec:codec-side}), and EdgeFace is also an in-grid evaluator. To test
identity preservation on a backbone that is architecturally independent of \emph{both}
the side-stream and the primary anchors, we re-embed every reconstruction with
\texttt{cvlface\_ir101} --- a CVLFace IR-101 model trained on WebFace12M, from a family
used nowhere else in the codec --- and recompute verification EER.
Table~\ref{tab:heldout-cvlface} reports the result at the $112$\,px working point.

Three points hold on this held-out matcher, over the full ten-codec roster (only the
Color-FERET-only CompressAI baselines are absent). First, at $1024$\,B Ours-ACCURATE is
fourth of ten on Color~FERET (EER $0.05\%$: behind the field-best $0.03\%$ shared by WebP
and JPEG-AI and AVIF's $0.04\%$, but ahead of HEIF's $0.07\%$, JPEG~XL's and JPEG-FzT's
$0.10\%$ and JPEG's $0.12\%$); on KK it is the \emph{best} codec at $0.51\%$, a whisker
ahead of JPEG-AI's now-populated $0.52\%$ --- in both cases far above legacy JPEG~2000
($2.38\%$/$8.92\%$), so identity is preserved on a matcher the codec never saw.
Second, at the harder $512$\,B budget Ours-ACCURATE is the \emph{single best} codec on
both datasets ($0.08\%$ on Color~FERET versus JPEG-AI's $0.13\%$; $0.92\%$ on KK versus
JPEG-AI's $1.25\%$, WebP's $2.55\%$ and JPEG~2000's $29.4\%$) --- the
graceful-degradation claim survives the removal of the EdgeFace-related evaluators.
Third, the ACCURATE\,$>$\,FAST ordering and the modern-cluster-beats-JPEG~2000 ranking
reproduce exactly, and so does the $512$\,B collapse of the codecs the anchors already
mark as brittle (JPEG $15.75\%$/$21.24\%$, HEIF $7.90\%$/$19.88\%$, JPEG~XL
$1.26\%$/$6.52\%$ on Color~FERET/KK), so the codec ranking is
not an artefact of the side-stream anchor. On the anchor matchers JPEG-AI keeps the lead
at $1024$\,B on Color~FERET, so the honest overall verdict stands there, but the
held-out result removes the circularity objection from the $512$\,B result and from the
cross-codec ranking.

\begin{table}[t]
  \centering
  \small
  \caption{Verification EER (\%, $\downarrow$) at $112$\,px on the held-out
  \texttt{cvlface\_ir101} matcher (IR-101/WebFace12M), architecturally independent of the
  EdgeFace side-stream and the ArcFace/LVFace anchors, with the deployment-relevant tail
  FNMR@FMR$=10^{-4}$ (\%) in \scriptsize{parentheses}\normalsize. The
  \colorbox{green!25}{best} and \colorbox{red!22}{worst} EER in each column are shaded, as
  in the other codec tables. The whole codec roster is re-embedded on this matcher at
  full support ($95\,839$ mated pairs on Color~FERET, $1\,515\,807$ on AI-Solutions-KK,
  matching the anchor grids); the only omissions are the CompressAI neural baselines
  (bmshj2018, mbt2018), which were run on Color~FERET only and scored on a $2\,548$-pair
  subset, so they have no comparable cell in three of the four columns.
  JPEG-AI is now scored on \emph{both} datasets (its KK \texttt{cvlface\_ir101} cells are
  populated). ``aligned'' is the uncompressed reference; it is \textbf{budget-independent}
  (no compression), so its value spans both budget columns of a dataset rather than being
  missing at $512$\,B. The tail confirms the EER story:
  at KK/$512$\,B Ours-ACCURATE leads on both EER ($0.92\%$) and tail ($8.4\%$ vs WebP
  $29.1\%$, JPEG-AI $12.4\%$).}
  \label{tab:heldout-cvlface}
  \adjustbox{max width=\textwidth}{\input{tables/heldout_cvlface.tex}}
\end{table}

\subsection{Known limitations}
\tldr{Ours-FAST mostly cannot reach $512$\,B at $224$\,px --- only $14\%$ of
AI-Solutions-KK and $80\%$ of Color~FERET crops hold the budget there, its gain floor
over-filling it and the encoder by design emitting a clean over-budget frame --- and both
variants also overflow on part of the $96$/$168$\,px grid (Ours-ACCURATE on ${\approx}20\%$
of crops at $512$\,B, Ours-FAST on up to $91\%$); the full-resolution Ours grid is now
complete.}

The hard budget exposes the codecs' rate floors. Ours-FAST, with no side-stream to absorb
the high-resolution rate floor, mostly cannot reach $512$\,B at $224$\,px --- only $14\%$
of AI-Solutions-KK crops (and $80\%$ of Color~FERET crops) hold the budget in that cell
(Tables~\ref{tab:fit-kk} and~\ref{tab:fit-cf}) --- and it
overflows at some $96$/$168$\,px cells too ($74\%$/$9\%$ of KK crops and $74\%$/$50\%$ of
Color~FERET crops hold at $512$\,B). How far it overshoots is \emph{cell-dependent}, so the
honest statement is the measured per-cell size distribution rather than one global ratio:
in the $224$\,px$/512$\,B cell the median emitted container is $595$\,B on AI-Solutions-KK
($1.16\times$ the budget, with $86\%$ of crops over) but $507$\,B on Color~FERET ---
\emph{under} budget, with $20\%$ of crops over --- and at $168$\,px$/512$\,B the medians are
$607$\,B (KK, $91\%$ over) and $511$\,B (Color~FERET, $50\%$ over). The mechanism is that
the search can only choose from the frozen $64$-entry gain table: when even the floor
gain's real rANS output, plus the fixed $7$\,B container overhead, already exceeds $B$,
there is no lower level to fall back to. By design the encoder then emits the smallest valid spatial
frame, flagged \emph{over-budget} --- a clean reconstruction that exceeds $B$ --- rather
than a black identity-only frame, on the rationale that a real reconstruction is more
useful than a blank one; downstream accounting must treat such frames as budget
violations. ACCURATE does not show this at $224$\,px, where its trained low-$\lambda$ levels and
identity-only fallback keep $100\%$ of Color~FERET crops and $99.98$--$99.99\%$ of
AI-Solutions-KK crops inside the budget at both budgets; it still misses on the
$96$/$168$\,px buckets, but less often than FAST
($79$--$80\%$ of crops hold at $512$\,B, $90\%$ at $1024$\,B). For both variants the
non-compliance is confined to the $96$ and $168$\,px buckets and --- for FAST alone ---
the $224$\,px$/512$\,B cell: at $64$ and $112$\,px both variants hold $100\%$ of crops at
both budgets on both datasets, and in the $224$\,px$/1024$\,B headline cell both hold
$\ge 99.99\%$, so no headline claim in this report rests on an over-budget cell.

One further note. The full-resolution Ours grid (all of $64/96/112/168/224$\,px, both
budgets, both datasets) is now complete: Ours-FAST and Ours-ACCURATE appear in all eight
rate--EER grids (Section~\ref{sec:benchmark}), where EER falls with resolution --- at
$1024$\,B monotonically so (Ours-ACCURATE on Color~FERET/ArcFace drops from $1.36\%$ at
$64$\,px through $0.56$/$0.14$/$0.12\%$ at $96$/$112$/$168$\,px to $0.07\%$ at $224$\,px),
while at $512$\,B the fall is not monotone ($0.22\%$ at $112$\,px against $0.57\%$ at
$168$\,px and $0.37\%$ at $224$\,px), because at the tighter budget the larger crops are
the ones the fixed byte ceiling starves. The refine head's adversarial (HiFiC-style) training remains a
planned phase-3 extension; the shipped head trains only as a refiner under the
reconstruction and identity losses.

\subsection{Worst-case identity tail}
\label{subsec:codec-tail}
\tldr{For a credential system the binding quantity is not the median but the \emph{tail} ---
the worst-served subjects. We report the $5$th-percentile (worst $5\%$) per-image
reconstruction-vs-original identity cosine on the independent ArcFace matcher. At the hard
$512$\,B budget Ours-ACCURATE has the \emph{tightest} tail of the field on both datasets ---
its p5 ($0.802$ CF / $0.763$ KK) sits \emph{above WebP's median} ($0.739$/$0.725$), and WebP's
own p5 collapses to $0.60$/$0.58$ and AVIF's to $0.33$/$0.31$. The codec's advantage is thus
larger in the tail than at the median, which is the relevant regime for a document credential.}

The rate--identity tables report medians, but a face-on-document system fails on its
\emph{worst}-served subjects, so the decision-relevant statistic is the low tail of the
per-subject identity distribution. Table~\ref{tab:idcos-tail} gives the $5$th-percentile
(``worst $5\%$'') and median of the per-image reconstruction-vs-original identity cosine at
$512$\,B, scored on ArcFace (independent of the side-stream) over the full held-out set. Two
points follow. First, at this budget the classical codecs have long lower tails --- WebP drops
from a $0.74$ median to a $0.60$ p5, and AVIF from $0.54$ to $0.33$ --- whereas the
byte-budgeted learned codecs stay compact: Ours-ACCURATE's p5 ($0.802$ CF, $0.763$ KK) is the
highest in the table and \emph{exceeds WebP's median}. Second, the median-based ``best at
$512$\,B'' finding (\S\ref{sec:frr-far}) therefore \emph{understates} Ours-ACCURATE's edge:
because the worst subjects are exactly where classical codecs degrade fastest, the gap widens
in the tail. JPEG-AI is the nearest competitor and is now scored on the full corpus on both
datasets (p5 $0.742$ CF, $0.680$ KK); both sit below Ours-ACCURATE.

\begin{table}[t]
  \centering
  \small
  \caption{Worst-case identity tail at $512$\,B: $5$th-percentile (worst $5\%$) and median of
  the per-image reconstruction-vs-original identity cosine, ArcFace matcher (independent of
  the side-stream), $112$\,px. Higher p5 = a tighter, safer tail. Every cell is computed over
  the full corpus ($11\,331$--$17\,534$ crops), so the columns are directly comparable; an
  earlier version of this table scored JPEG-AI's Color~FERET row on a $300$-crop decode pilot,
  which flattered it (p5 $0.776$ against $0.742$ at full support) --- a p5 over $300$ crops is
  the $15$th-worst crop, where a p5 over $11$k is the $567$th-worst. Source: per-image cosine
  between each aligned crop and its own reconstruction, over the committed embeddings.
  \colorbox{green!25}{Best}/\colorbox{red!22}{worst} per column (higher id-cosine is
  better), as in the other codec tables. Roster: the five codecs the $512$\,B claim turns
  on --- JPEG-AI, WebP and AVIF (the $1024$\,B leaders) plus both learned variants; the
  remaining roster codecs (JPEG, JPEG~2000, JPEG~XL, HEIF, JPEG-FzT) are not scored here.}
  \label{tab:idcos-tail}
  \small\input{tables/idcos_tail.tex}
\end{table}

%% file: sections/07b_difficulty.tex
\section{Trivial and Difficult Samples: What Makes a Face Hard to Compress}
\label{sec:difficulty}
\tldr{Which crops lose the most identity under a sub-1\,kB budget, and why? Difficulty is
\emph{modest and detail-driven on Color~FERET}: at the image level ($n{=}300$) high-amplitude
detail predicts it (Laplacian variance $\rho{=}{+}0.39$, $p{<}0.01$; luma contrast $+0.30$,
$p{<}0.01$) and more colourful crops are \emph{easier} ($\rho{=}{-}0.36$, $p{<}0.01$). On
the in-the-wild KK set none of the eight codec-independent image descriptors reaches
significance ($|\rho|\le0.10$), so the predictors are dataset-specific, not universal. (Age
is the one KK attribute with a middling correlation, $\rho{=}{+}0.35$, but it is computed
over only $12$ age-bucket means, not the $300$-image sample, and does not reach significance
there either, $p{=}0.26$ --- too little data to read as a finding.) Pose does \emph{not}
drive it at all
($|$yaw$|$ $\rho{=}{-}0.02$, $p{=}0.74$) and lighting only weakly (brightness $-0.16$) ---
identity loss is scored against the \emph{same-pose, same-lighting} original, so those
factors largely cancel. An earlier ``glasses $4\times$'' claim was a codec-pooling artefact and does
\emph{not} survive image-level testing ($2$ of the $30$ hardest vs.\ $3$ of the $30$ easiest
crops, Fisher $p{=}1.0$); we withdraw it. Skin tone could not be tested here: this crop set
carries no Monk skin-tone labels. Difficulty is only moderately shared
across codecs (mean pairwise $\rho{=}0.33$ on Color~FERET, $0.38$ on KK): the modern block
codecs agree on which crops are hard --- legacy JPEG does not, it fails almost everywhere
--- while the learned codec stays robust even on crops that break the others.}

The aggregate tables report a \emph{median} crop per codec, but deployments fail on the
\emph{tail}: the worst crops, not the median, drive the false-non-match rate. This section
therefore turns from per-codec medians to per-\emph{image} behaviour and asks which
individual crops are trivial or hard, and whether the hard ones share measurable
properties a system could flag at capture time.

\paragraph{Setup.} We draw a fixed random sample of $300$ Color~FERET and $300$
AI-Solutions-KK aligned crops, compress each at $112$\,px to both byte budgets with six
classical codecs (JPEG, JPEG~2000, WebP, AVIF, HEIF, JPEG~XL), JPEG-AI and both learned
variants, decode, and score
every reconstruction \emph{per image}. Per-image difficulty is the identity loss
$1-\mathrm{id\text{-}cos}$ (cosine between the reconstruction's and the clean crop's held-out
inno-balanced embedding); $0$ means the compressed face is biometrically indistinguishable
from the original. Because both sides of that cosine share the same pose, expression, and
illumination, difficulty isolates \emph{what compression removed}, not the intrinsic
difficulty of the pose. Alongside each crop we record codec-independent image descriptors
computed on the clean crop --- spatial detail (Laplacian variance, gradient magnitude,
edge density, high-frequency DCT share), grey-level entropy, and photometry (brightness,
contrast, colorfulness) --- and the datasets' face attributes (Color~FERET's ground-truth
yaw, glasses, and facial hair; AI-Solutions-KK's Monk skin tone, age, gender). Difficulty
is analysed at the discriminating $512$\,B budget.

\begin{figure}[t]
  \centering
  \includegraphics[width=\linewidth]{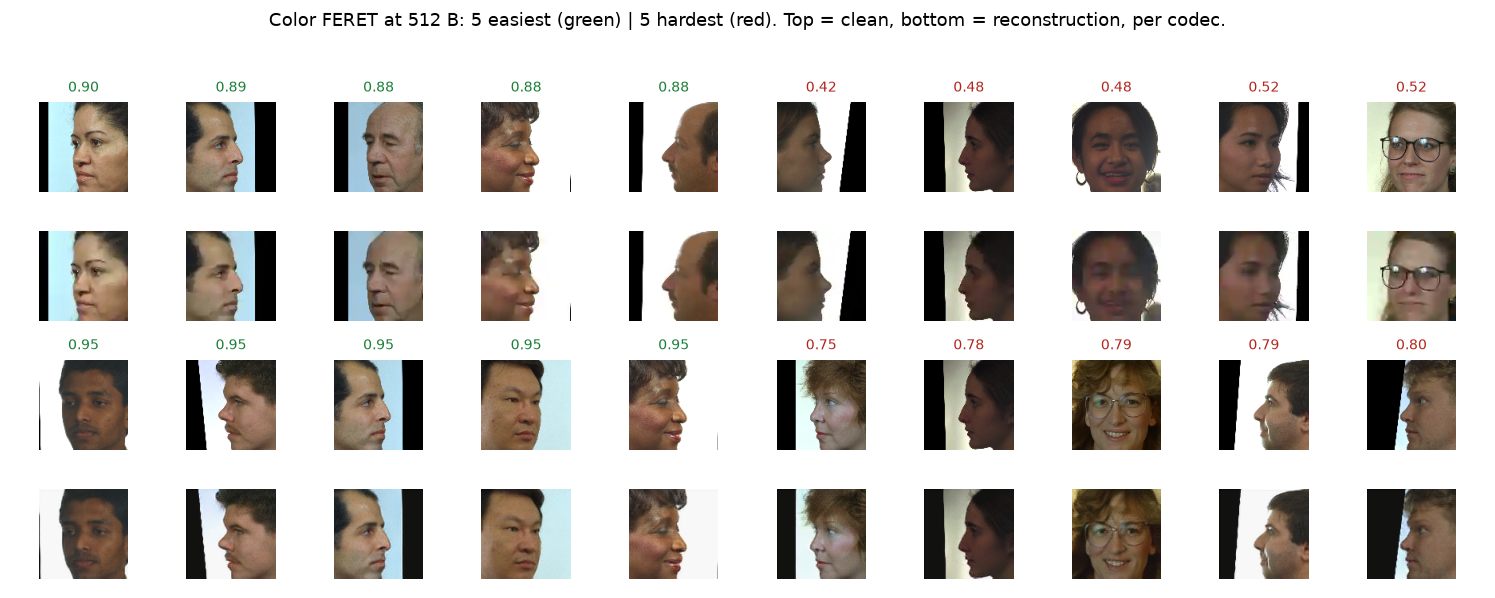}
  \caption{Color~FERET at $512$\,B: the five \emph{easiest} (green id-cosine) and five
  \emph{hardest} (red) crops for the strongest classical codec (WebP, top pair of rows)
  and the robust learned codec (Ours-ACCURATE, bottom pair), clean above and
  reconstruction below. Hard crops carry more fine detail than easy ones (mean Laplacian
  variance $2.4{\times}10^{3}$ vs $1.6{\times}10^{3}$ for WebP); eyewear is not a significant
  driver (Table~\ref{tab:difficulty-contrast}). The learned codec preserves far more identity
  on the hard tail: its hardest five score $0.75$--$0.80$, against $0.42$--$0.52$ for WebP's.}
  \label{fig:difficulty-montage}
\end{figure}

\paragraph{Trivial vs.\ difficult crops.} Figure~\ref{fig:difficulty-montage} shows the five
easiest and five hardest Color~FERET crops at $512$\,B for the strongest classical codec
(WebP) and the robust learned codec (Ours-ACCURATE), with per-image id-cosine annotated.
WebP's hard crops carry markedly more fine detail than its easy ones (mean Laplacian
variance $2.4{\times}10^{3}$ vs.\ $1.6{\times}10^{3}$), which is exactly the image-level
signal quantified below; pose is not what separates them --- the easiest five are the
\emph{more} off-angle set (mean $|$yaw$|$ $45^{\circ}$ vs.\ $31^{\circ}$) --- and just one
crop in each hardest-five wears glasses, which is not a statistically significant driver
(see below). The two codecs agree only partly on which crops are hard (rank $\rho{=}0.32$;
one crop is common to both hardest-five sets), but the learned codec keeps far more identity
on that tail: where WebP's hardest five fall to id-cosine $0.42$--$0.52$, Ours-ACCURATE's own
hardest five never drop below $0.75$ ($0.75$--$0.80$). This is exactly why the learned codec wins the operating-point tables
(Section~\ref{sec:frr-far}) even where its \emph{median} id-cosine only ties the modern
block codecs --- the gain is concentrated on the hard minority.

\begin{figure}[t]
  \centering
  \includegraphics[width=0.82\linewidth]{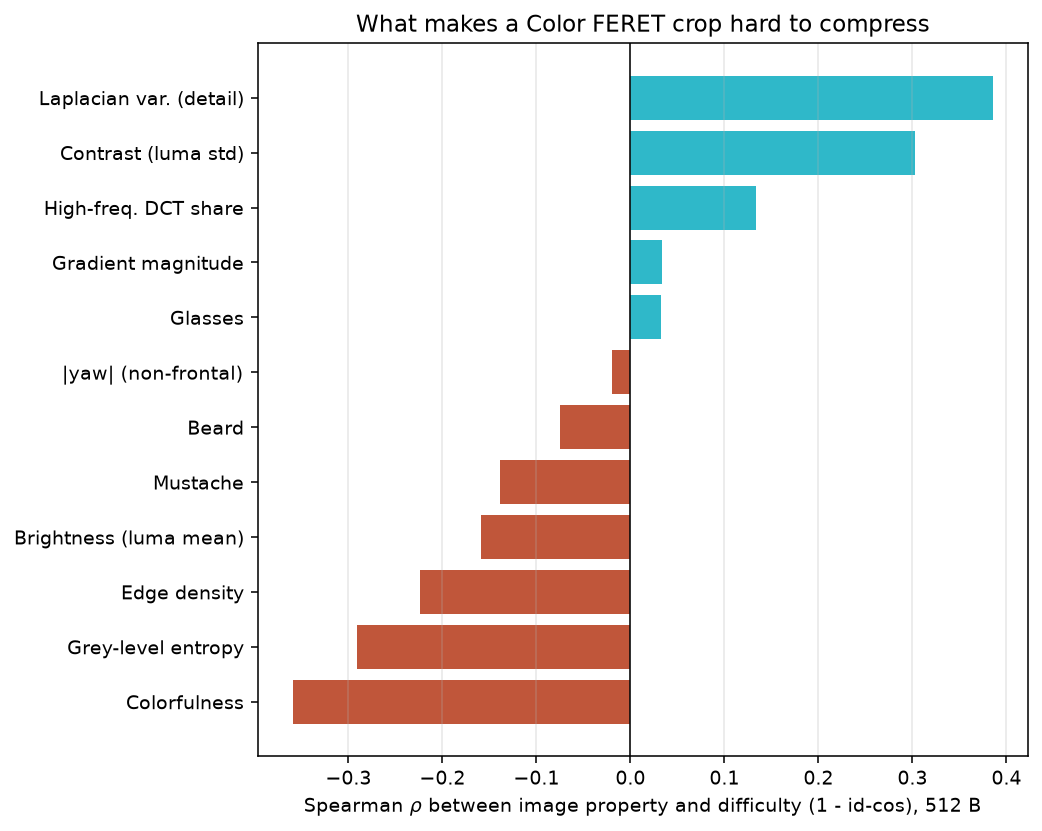}
  \caption{Color~FERET: Spearman correlation of each image property with per-image
  difficulty at $512$\,B (positive $=$ harder). Fine-detail measures lead; pose and
  photometry are near zero.}
  \label{fig:difficulty-predictors}
\end{figure}

\paragraph{What makes a crop hard --- and what does not.}
Table~\ref{tab:difficulty-predictors} correlates per-image difficulty against each
descriptor and attribute at the \emph{image} level: difficulty is averaged over the codecs
to give one point per image, so the reported Spearman $\rho$ carries a proper $n{=}300$
significance test (unlike a codec-pooled correlation, which would count each image nine
times). Three things stand out. First, the effect sizes are \emph{modest}: aligned,
face-centred crops are a homogeneous population, and identity loss under compression is
only weakly predictable from image statistics --- there is no single dominant ``hard-face''
feature. Second, on \textbf{Color~FERET} the strongest predictors are
about spatial detail and colour (Figure~\ref{fig:difficulty-predictors} ranks them):
\textbf{fine detail} makes a crop harder (Laplacian
variance $\rho{=}{+}0.39$, $p{<}0.01$; luma contrast $+0.30$, $p{<}0.01$), while more
\textbf{colourful} crops are \emph{easier} ($\rho{=}{-}0.36$, $p{<}0.01$). This is the
physically expected signal --- under a fixed byte budget the encoder spends its bits on
coarse structure first, so a crop carrying more fine detail loses more of it. The detail
story is not monolithic, though: edge density ($-0.22$, $p{<}0.01$) and grey-level entropy
($-0.29$, $p{<}0.01$) run the \emph{other} way, so it is high-amplitude local detail, not
busyness as such, that costs identity. Crucially,
too, \textbf{none of these predictors reaches significance on the in-the-wild KK set}
(Laplacian variance keeps the same positive sign at $\rho{=}{+}0.10$ but $p{=}0.08$; all
$|\rho|\le0.10$), so the ``detail-driven'' story is specific to the controlled Color~FERET
portraits and should not be read as a universal law. Third, \textbf{pose and
occlusion do not drive difficulty}: $|\text{yaw}|$ correlates $-0.02$ ($p{=}0.74$ --- pose is
simply uninformative here), brightness contributes only weakly ($-0.16$: brighter crops are
marginally \emph{easier}), and glasses ---
despite an earlier codec-pooled ``$4\times$'' impression --- are \emph{not} over-represented
in the hardest crops once counted per image (2 of the 30 hardest and 3 of the 30 easiest
carry glasses; Fisher exact $p{=}1.0$; Table~\ref{tab:difficulty-contrast}). The pose/lighting
null is a direct consequence of the metric --- scoring the reconstruction against the
\emph{same-pose, same-illumination} original cancels those factors. On AI-Solutions-KK this
$300$-crop difficulty sample carries no Monk skin-tone labels, so no correlation could be
computed here; the tone-fairness question itself is answered separately, on the full
labelled set, in Section~\ref{sec:fairness}.

\begin{table}[t]
  \centering\small
  \caption{Image-level Spearman correlation between per-image compression difficulty
  ($1-\mathrm{id\text{-}cos}$ at $512$\,B, averaged over the nine codecs to one point per
  image, $n{=}300$/dataset) and codec-independent image properties. Positive $\rho$ means the
  property makes a crop harder; $^{*}\,p{<}0.05$, $^{**}\,p{<}0.01$. On Color~FERET fine
  detail (Laplacian variance, contrast) and low colourfulness are the strongest predictors;
  \emph{no} property reaches significance on the in-the-wild KK set, so the predictors are
  dataset-specific. Pose is not predictive and brightness only weakly so ($-0.16$); the Monk
  skin-tone index has no labelled crops in this sample, so it yields no estimate. The
  AI-Solutions-KK age row rests on the $12$ crops that carry an age label and is \emph{not}
  significant ($p{=}0.26$).
  (``--'': too few labelled crops for a stable estimate.)}
  \label{tab:difficulty-predictors}
  \adjustbox{max width=\textwidth}{\input{tables/difficulty_predictors.tex}}
\end{table}

\begin{table}[t]
  \centering\small
  \caption{Mean image property in the easiest vs.\ hardest Color~FERET difficulty decile,
  computed at the \emph{image} level ($512$\,B: 30 easiest vs 30 hardest of 300 images, no
  codec pseudoreplication). Laplacian variance ($1.81\times$) and contrast ($1.28\times$)
  are the clearest continuous shifts. Glasses are \emph{not} over-represented once counted
  per image (3/30 easiest vs 2/30 hardest, Fisher exact $p{=}1.00$) --- the earlier
  codec-pooled ``$4\times$'' was a small-sample artefact. The rank correlations in Table~\ref{tab:difficulty-predictors}
  are the more reliable summary.}
  \label{tab:difficulty-contrast}
  \adjustbox{max width=\textwidth}{\input{tables/difficulty_contrast.tex}}
\end{table}

\paragraph{Difficulty is only moderately codec-shared.} Averaged over codec pairs, the
per-image difficulty ranks correlate with mean pairwise Spearman $\rho{=}0.33$ on
Color~FERET (range $0.00$--$0.59$) and $0.38$ on AI-Solutions-KK (range $0.09$--$0.63$). The
agreement is highest among the modern block codecs (AVIF--WebP $0.56$, AVIF--JPEG~XL $0.55$
on Color~FERET) and between the two learned variants ($0.57$); legacy JPEG is the outlier,
correlating at most $0.22$ with any other codec because at $512$\,B it destroys identity
almost everywhere (median id-cosine $0.05$). Ours-ACCURATE shares little with the block
codecs ($\rho{=}0.10$--$0.32$) and does not fail alongside them: on the $30$ crops WebP finds
hardest its median id-cosine is $0.889$, against $0.906$ over all $300$. So ``hardness'' is
partly an image property (the detail signal above, shared by the block codecs) and partly a
codec property (the learned codec simply refuses to fail on the same crops).

\paragraph{Practical implication.} Because the difficulty signal, weak as it is, comes from
cheap codec-independent descriptors computed at capture time, a deployment can still
\emph{triage}: on controlled portraits a high-Laplacian-variance, high-contrast crop is a
candidate for recapture or a larger budget, whereas a smooth, evenly-lit face is safe at the
tightest budget --- though the signal did not replicate on the in-the-wild KK set, so this
remains a Color~FERET-only recommendation. More operationally useful is the tail behaviour itself: the codecs separate not on
the easy majority of crops --- where all of them preserve identity --- but on a hard
minority carrying the most fine detail, and the learned codec's advantage is
concentrated exactly there.

%% file: sections/08_ablations.tex
\section{Ablations and Preprocessing}
\label{sec:ablations}
\tldr{At a fixed byte budget the input resolution sets a sweet spot: smaller crops spend their bits on fewer, cleaner pixels but discard facial detail, while larger crops carry more detail than 1\,kB can encode well; 112\,px is the best compromise and is our working resolution. One-factor-at-a-time preprocessing (chroma/colour/background/high-frequency/ROI), alignment tightness, and the JPEG-FzT decode radius are studied separately and inform the codec design.}

This section isolates the choices that sit \emph{upstream} of the codec: how big the
input crop should be, what preprocessing (if any) helps before encoding, how tightly the
face is cropped, and a codec-internal knob (the JPEG-FzT decode radius). Each is studied
one factor at a time so that the effect is attributable. The resolution study is grounded
in three measurements --- equal-error rate (EER) versus resolution, an embedding
decomposition relative to the working resolution, and a spectral-energy retention curve
--- reported below (Section~\ref{sec:artifacts}).

\subsection{Resolution--information trade-off}
\tldr{Going from 224 to 112\,px loses almost no identity signal (AC spectral energy stays above
0.999 and, for strong matchers such as ArcFace/TopoFR/EdgeFace, embedding cosine stays
above 0.99; LVFace-L is the exception at 0.899), but at
a hard 1\,kB budget the smaller crop lets the codec spend its bits where they matter;
below 112\,px identity retention drops off, so 112\,px is the chosen working resolution.}

\paragraph{Why resolution matters at a fixed budget.}
The byte budget is fixed at $\le 1024$\,B (and a harder 512\,B point) regardless of input
size, so the choice of crop resolution is really a choice of \emph{bits per pixel}. A
small crop (e.g.\ 64\,px) has few pixels, so each one is encoded cleanly, but the crop
itself carries less facial detail to begin with. A large crop (e.g.\ 224\,px) carries
more raw detail, but 1\,kB cannot represent $224^2$ pixels without heavy quantisation, so
the reconstruction is blurred or blocky per pixel. The optimum is in between.

\paragraph{Identity content is nearly saturated by 112\,px.}
Before any compression, we quantify how much identity signal each resolution carries by
(i) the fraction of high-frequency (AC) spectral energy retained relative to the 224\,px
reference, and (ii) the cosine similarity between an embedding at each resolution and the
112\,px embedding for the same image (``identity info retained vs.\ 112''). On the
in-the-wild AI-Solutions-KK set, the AC spectral energy retained relative to 224\,px is
already 0.9997 at 112\,px and 0.9996 at 96\,px, dropping only to 0.9990 at 64\,px
(Table~\ref{tab:res-decomp}); almost all of the spectral content a face matcher uses
survives downsampling to 112\,px. The embedding cosine tells the same story for strong
matchers (ArcFace, TopoFR, EdgeFace all retain $\ge 0.93$ at 64\,px and $\ge 0.98$ at
96\,px relative to their own 112\,px embedding), with LVFace-L the notable exception ---
it is markedly more resolution-sensitive (0.710 at 64\,px, 0.834 at 96\,px), which is why
we do not push below 112\,px.

\begin{figure}[t]
  \centering
  \includegraphics[width=\textwidth]{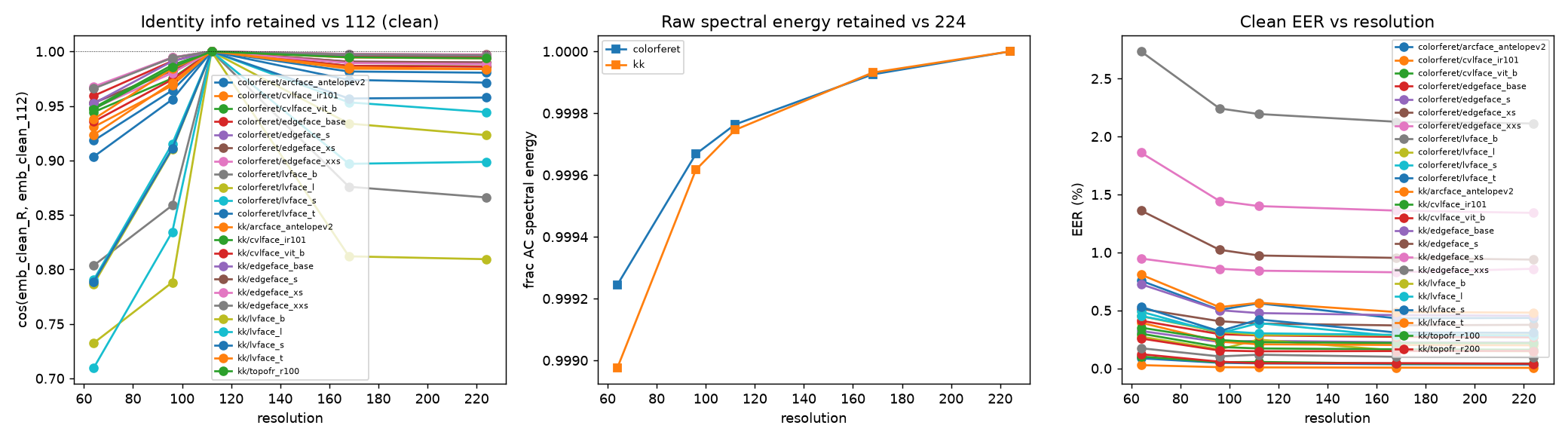}
  \caption{Resolution--information summary across all matchers and both datasets.
  \emph{Left:} identity information retained relative to the 112\,px embedding
  (embedding cosine). \emph{Middle:} fraction of raw AC spectral energy retained relative
  to 224\,px. \emph{Right:} clean EER versus resolution. All three flatten by
  $\sim$112\,px for strong matchers; LVFace variants are the most resolution-sensitive.}
  \label{fig:res-summary}
\end{figure}

\paragraph{Clean EER bottoms out around 112--224\,px.}
On uncompressed crops, EER decreases monotonically from 64\,px up to roughly 112\,px and
then flattens: the gap between 112 and 224\,px is small for every anchor matcher, whereas
dropping from 112 to 64\,px costs noticeably more (e.g.\ ArcFace on KK nearly doubles, from
0.21\,\% at 112\,px to 0.39\,\% at 64\,px; EdgeFace-XS rises a third, from 1.40\,\% to
1.86\,\%). Thus
once compression is in the loop, 112\,px keeps essentially all of the achievable clean
accuracy while giving the codec the fewest pixels to spend its 1\,kB on.
Table~\ref{tab:res-decomp} summarises the three measurements for the four anchor matchers
on KK; Figure~\ref{fig:res-summary} shows the full curves across all matchers and both
datasets.

\begin{table}[t]
  \centering
  \caption{Resolution--information decomposition on AI-Solutions-KK (clean, before
  compression). \emph{AC retained} is the fraction of high-frequency spectral energy kept
  relative to the 224\,px reference (dataset-level). \emph{Id.\ cos vs.\ 112} is the
  cosine between each resolution's embedding and the 112\,px embedding for the same image.
  \emph{EER} is the clean equal-error rate (\%).}
  \label{tab:res-decomp}
  \small
  \begin{tabular}{lccccc}
    \toprule
    & \multicolumn{5}{c}{Resolution (px)} \\
    \cmidrule(lr){2-6}
    Quantity & 64 & 96 & 112 & 168 & 224 \\
    \midrule
    AC spectral energy retained vs.\ 224 & 0.9990 & 0.9996 & 0.9997 & 0.9999 & 1.0000 \\
    \midrule
    \multicolumn{6}{l}{\emph{Identity cosine vs.\ 112 (per matcher)}} \\
    \quad ArcFace (antelopev2) & 0.938 & 0.984 & 1.000 & 0.994 & 0.994 \\
    \quad TopoFR (r100)        & 0.947 & 0.985 & 1.000 & 0.995 & 0.994 \\
    \quad EdgeFace-XS          & 0.968 & 0.995 & 1.000 & 0.998 & 0.997 \\
    \quad LVFace-L             & 0.710 & 0.834 & 1.000 & 0.897 & 0.899 \\
    \midrule
    \multicolumn{6}{l}{\emph{Clean EER (\%, per matcher)}} \\
    \quad ArcFace (antelopev2) & 0.39 & 0.23 & 0.21 & 0.20 & 0.20 \\
    \quad TopoFR (r100)        & 0.30 & 0.18 & 0.17 & 0.16 & 0.16 \\
    \quad EdgeFace-XS          & 1.86 & 1.45 & 1.40 & 1.36 & 1.34 \\
    \quad LVFace-L             & 0.45 & 0.32 & 0.30 & 0.29 & 0.28 \\
    \bottomrule
  \end{tabular}
\end{table}

\paragraph{Decision.}
We adopt 112\,px as the verification working resolution: it sits at the knee of all three
curves, retains essentially all of the clean accuracy of 224\,px, and minimises the
pixel count the codec must reconstruct under a 1\,kB budget.

\subsection{Preprocessing, measured through the codec}
\label{subsec:preproc-through}
\tldr{The operational question is not what a preprocessing operator does to a clean crop,
but what survives \emph{after} the crop is compressed and decoded. Applying each operator
to the aligned $224$\,px input, then round-tripping through the best codecs at $1024$\,B on
the in-the-wild KK set, chroma reduction (A3) is identity-neutral on WebP and
Ours-ACCURATE (it frees bits at no verification cost on \emph{both} anchors) and
near-neutral on AVIF, aggressive spatial denoising
(A1 edge-preserving, A4 non-local-means) smears identity-bearing texture and raises EER,
and the learned Ours-ACCURATE codec is markedly the most robust to any preprocessing.}

\paragraph{Method.}
Because the codec must hit a hard byte budget, any preprocessing that removes
\emph{identity-irrelevant} content frees bits for the face --- but a preprocessing choice
is only useful if its benefit survives compression. We therefore measure each operator
\emph{through the codec}: we apply it to the aligned $224$\,px input, compress and decode
at the $1024$-byte budget with three of the best codecs (WebP, AVIF, and our
Ours-ACCURATE),
embed the reconstruction, and report verification EER together with the identity cosine of
the reconstruction against the \emph{unprocessed} original. As this is an ablation we fix
the source resolution to $224$\,px and the budget to $1024$\,B, and we run it \emph{only on
AI-Solutions-KK}: Color~FERET's near-homogeneous studio background makes background
operators trivially favourable and its clean crops are too easy at this budget to separate
the operators. We evaluate the pure-image operators A1 (edge-preserving mild), A2
(bilateral strong), A3 (chroma reduction), A4 (non-local-means denoise), and B2 (MediaPipe
background flattening); the BiRefNet (B1) and combined (C1/C2) operators are omitted here
as their matting model was not available for the KK $224$\,px run. The sample is $n{=}800$
subject-stratified KK images, scored with the ArcFace and EdgeFace-XS anchors.

\begin{figure}[t]
  \centering
  \includegraphics[width=\textwidth]{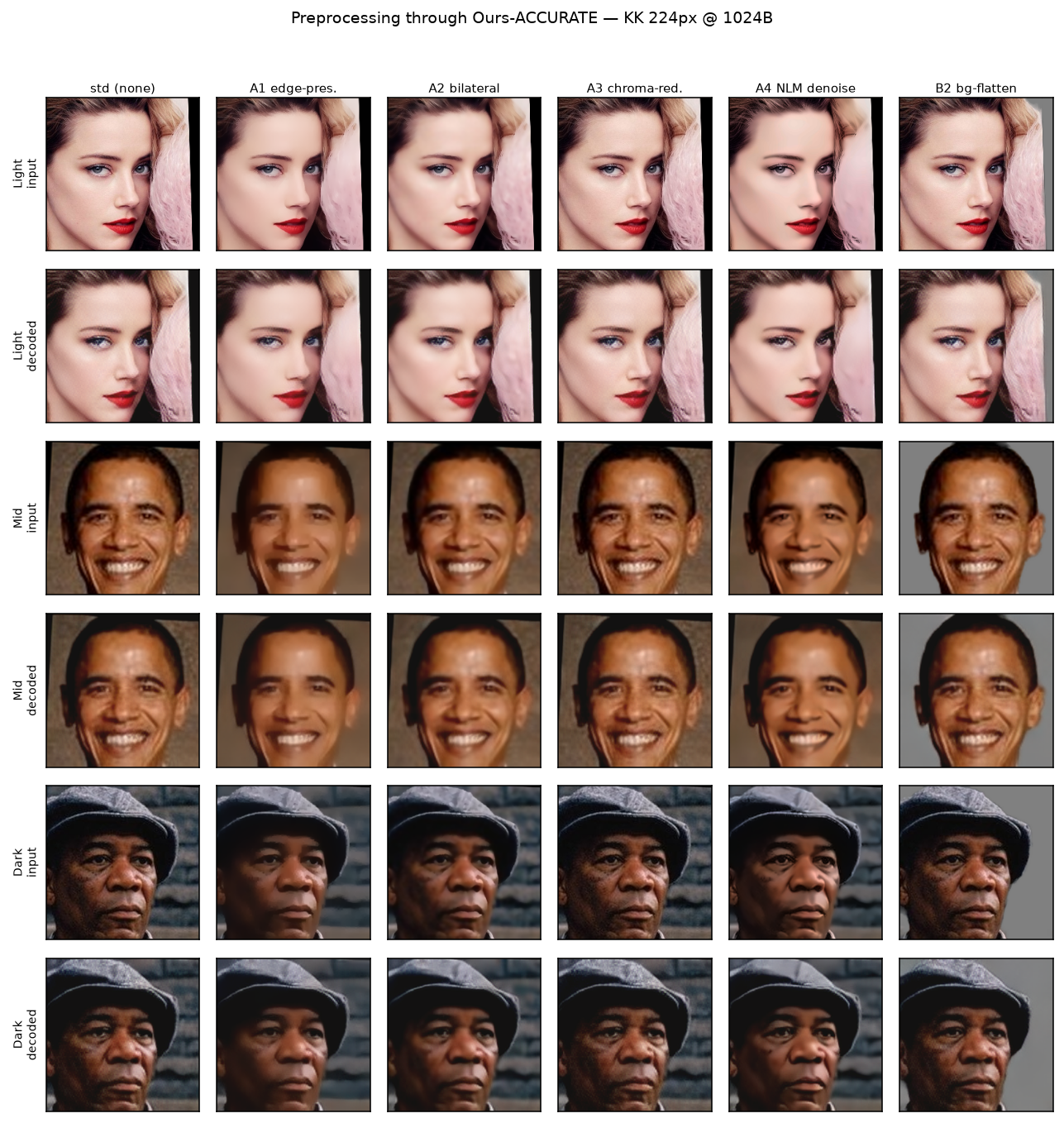}
  \caption{Preprocessing operators through the codec on three AI-Solutions-KK skin tones
  ($224$\,px$/1024$\,B, decoded with Ours-ACCURATE). For each subject the top row is the
  preprocessed \emph{input} and the bottom row the \emph{decoded} reconstruction. B2
  flattens the background to a compressible field without touching the face; A1/A4 visibly
  soften identity-bearing skin and edge texture, whereas A3 (chroma reduction) is
  perceptually and biometrically near-neutral.}
  \label{fig:preproc-through}
\end{figure}

\paragraph{Result: what survives compression.}
Table~\ref{tab:preproc-through} gives the through-codec EER and reconstruction identity
cosine (ArcFace) per operator and codec, and Table~\ref{tab:preproc-through-edge} repeats the
EER under the second, architecturally independent anchor (EdgeFace-XS);
Figure~\ref{fig:preproc-through} shows the
operators applied to three KK skin tones with their Ours-ACCURATE reconstructions. All
three findings hold cleanly across \emph{both} anchors. First, \textbf{chroma reduction (A3) is
identity-neutral through the codec}: its EER matches the unprocessed
\texttt{std} pipeline to within noise on ArcFace (WebP A3 $0.39\%$ vs.\ std $0.42\%$) and on
EdgeFace-XS (WebP A3 $2.05\%$ vs.\ std $2.07\%$), and its
reconstruction id-cosine is unchanged --- it discards chroma the matcher ignores, so the
freed bits are pure gain. Second, \textbf{aggressive
spatial denoising is the most damaging}: edge-preserving filtering (A1) and non-local-means
(A4) smear identity-bearing skin and edge texture, dropping the classical codecs'
reconstruction id-cosine to $0.75$--$0.80$ and raising their EER by up to $+1.0$\,pp (WebP A1
$0.42\!\rightarrow\!1.41\%$, AVIF A1 $0.64\!\rightarrow\!1.60\%$); bilateral (A2) is a
milder version of the same effect. Third, and notably, \textbf{Ours-ACCURATE is the most
robust to preprocessing among the codecs tested}: its worst operator (B2) costs only
$+0.63$\,pp, and A1 --- the operator most damaging to the classical codecs --- costs it just
$+0.38$\,pp (vs.\ up to $+1.0$\,pp for WebP/AVIF); it holds the highest reconstruction
id-cosine in every column, because its identity-preserving objective re-supplies structure
the preprocessing removed.

\emph{Background flattening (B2) --- a caveated, anchor-dependent result.} B2 leaves the
reconstruction id-cosine essentially unchanged but carries a verification cost that is
\emph{strong on ArcFace} (Ours-ACCURATE $0.36\!\rightarrow\!0.99\%$, WebP
$0.42\!\rightarrow\!0.99\%$) yet \emph{mild on EdgeFace-XS} (WebP $2.07\!\rightarrow\!2.58\%$,
Table~\ref{tab:preproc-through-edge}), so unlike A3 and A1/A4 the B2 effect does not
replicate cleanly across anchors. More importantly, the through-codec study used \emph{only}
the MediaPipe matting (B2), which the clean-crop ablation (Table~\ref{tab:ablation-preproc})
already identified as the \emph{weaker} of the two background operators --- BiRefNet (B1)
scored $0.05\%$ ArcFace EER against MediaPipe's $0.63\%$ on clean crops. The B2 cost measured
here is therefore a floor from the known-worse implementation, not a verdict on background
flattening in general; a through-codec rerun with B1 is left to future work, and we do
\emph{not} draw an operational conclusion against background flattening from B2 alone.

The practical rule for a byte-budgeted face pipeline is thus
confirmed on data that includes the codec: free bits with chroma reduction, never with
spatial denoising (and background flattening is promising but must be re-evaluated with the
stronger B1 matting before it is recommended).

\begin{table}[t]
  \centering
  \caption{Preprocessing \emph{through the codec} on AI-Solutions-KK, $224$\,px$/1024$\,B,
  $n{=}800$ subject-stratified crops. Each operator is applied to the aligned input, then
  compressed/decoded by the codec; \emph{EER} (ArcFace, \%, lower better) is verification
  error on the reconstruction and \emph{id-cos} is the reconstruction's identity cosine to
  the \emph{unprocessed} original. \colorbox{green!25}{Best}/\colorbox{red!22}{worst} per
  column. Chroma reduction (A3) is identity-neutral; denoising (A1/A4) is the most damaging;
  Ours-ACCURATE is the most robust. Background flattening here uses only the weaker MediaPipe
  operator (B2); its EER cost is anchor-dependent (see Table~\ref{tab:preproc-through-edge})
  and the stronger BiRefNet (B1) is not yet run through the codec, so no operational verdict
  on background flattening is drawn from B2 alone.}
  \label{tab:preproc-through}
  \small
  \adjustbox{max width=\textwidth}{\input{tables/preproc_through_codec.tex}}
\end{table}

\begin{table}[t]
  \centering
  \caption{Through-codec preprocessing under the \emph{second} anchor
  (EdgeFace-XS, architecturally independent of ArcFace), same $n{=}800$ KK crops at
  $224$\,px$/1024$\,B; verification EER (\%) and the reconstruction's EdgeFace-XS identity
  cosine to the unprocessed original. \colorbox{green!25}{Best}/\colorbox{red!22}{worst}
  per column; both this table and its ArcFace counterpart are generated from the same
  artifact, so they carry the same three codecs. The A3-neutral and A1/A4-most-damaging
  findings replicate; the B2
  background-flattening cost, strong on ArcFace (Table~\ref{tab:preproc-through}), is
  \emph{mild} here (e.g.\ WebP $2.07\!\rightarrow\!2.58\%$), so it does not generalise across
  anchors.}
  \label{tab:preproc-through-edge}
  \small
  \adjustbox{max width=\textwidth}{\input{tables/preproc_through_codec_edgeface.tex}}
\end{table}

\begin{figure}[t]
  \centering
  \includegraphics[width=\textwidth]{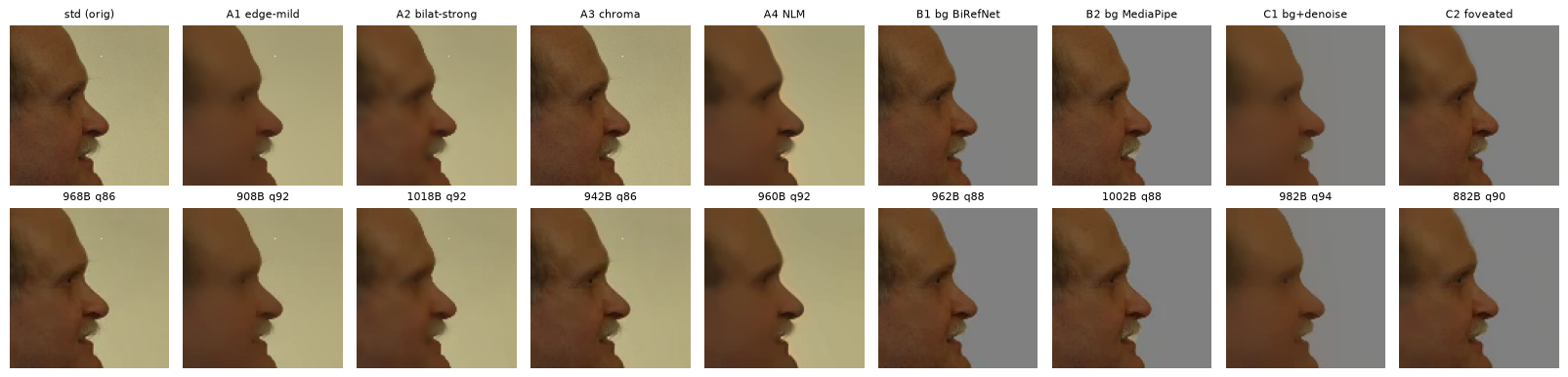}
  \caption{Pre-compression preprocessing, one factor at a time, on a single crop. Top
  row: input variant; bottom row: reconstruction at $\le 1024$\,B with achieved byte size
  and JPEG quality factor annotated. Columns: \texttt{std} (no preprocessing), A1--A4
  filters, B1/B2 background removal, C1/C2 combined.}
  \label{fig:preproc-impact}
\end{figure}

\paragraph{Corroboration on clean crops.}
The same ordering holds \emph{before} compression, isolating the operator effect from the
codec: on clean $112$\,px crops (Table~\ref{tab:ablation-preproc},
Figure~\ref{fig:preproc-impact}, four anchors) chroma de-emphasis (A3) and BiRefNet
background flattening (B1) are identity-free while non-local-means (A4) and edge-preserving
(A1) filtering raise ArcFace EER $7$--$9.5\times$ (from $0.04$ to $0.28$--$0.38\%$) and
roughly triple EdgeFace-XS EER --- the clean measurement that first motivated
the codec design's reliance on chroma reduction and background flattening rather than
spatial denoising.

\begin{table}[t]
  \centering
  \caption{Clean-crop (pre-compression) corroboration: preprocessing one factor at a time
  on $112$\,px crops. \emph{EER} (\%) per anchor matcher; \emph{id-cos vs.\ std} is the
  processed crop's identity cosine to its unprocessed counterpart. Chroma de-emphasis (A3)
  and BiRefNet flattening (B1) are near-neutral; denoising (A1/A4/C1) is the most damaging.
  \colorbox{green!25}{Best}/\colorbox{red!22}{worst} per column; the \texttt{std} row's
  id-cosine is $1.000$ by construction (compared against itself) and is excluded from that
  column's ranking.}
  \label{tab:ablation-preproc}
  \small
  \adjustbox{max width=\textwidth}{\input{tables/ablation_preprocessing.tex}}
\end{table}

\subsection{Alignment tightness}
\tldr{How tightly the face is cropped (inter-ocular distance as a fraction of the crop)
changes how many bits land on the central face; tighter crops enlarge the eyes/nose/mouth
but clip the hairline and jaw, so we verify the matcher's landmark alignment is sane and
compare standard/tight/mid/fill crops.}

\begin{figure}[t]
  \centering
  \includegraphics[width=\textwidth]{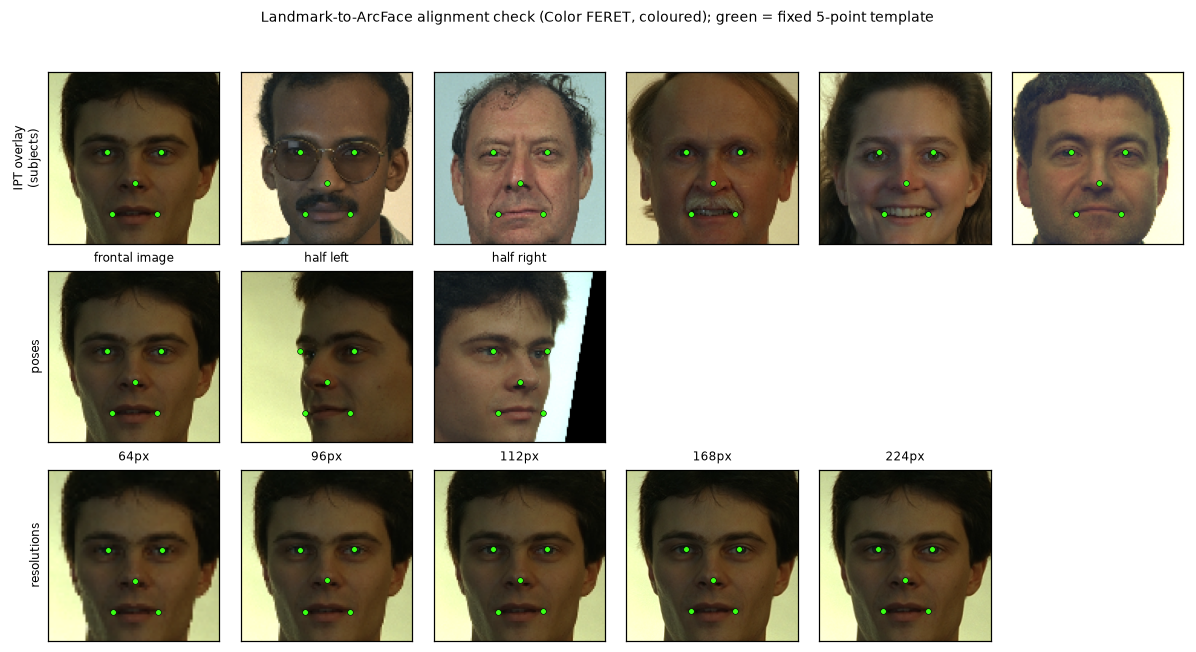}
  \caption{Landmark-to-ArcFace alignment check on Color~FERET. Rows: detected interest
  points overlaid across subjects (top), frontal/left/right poses (middle), and the five
  crop resolutions $64/96/112/168/224$\,px (bottom), confirming the alignment is stable.}
  \label{fig:align-check}
\end{figure}

\begin{figure}[t]
  \centering
  \includegraphics[width=0.85\linewidth]{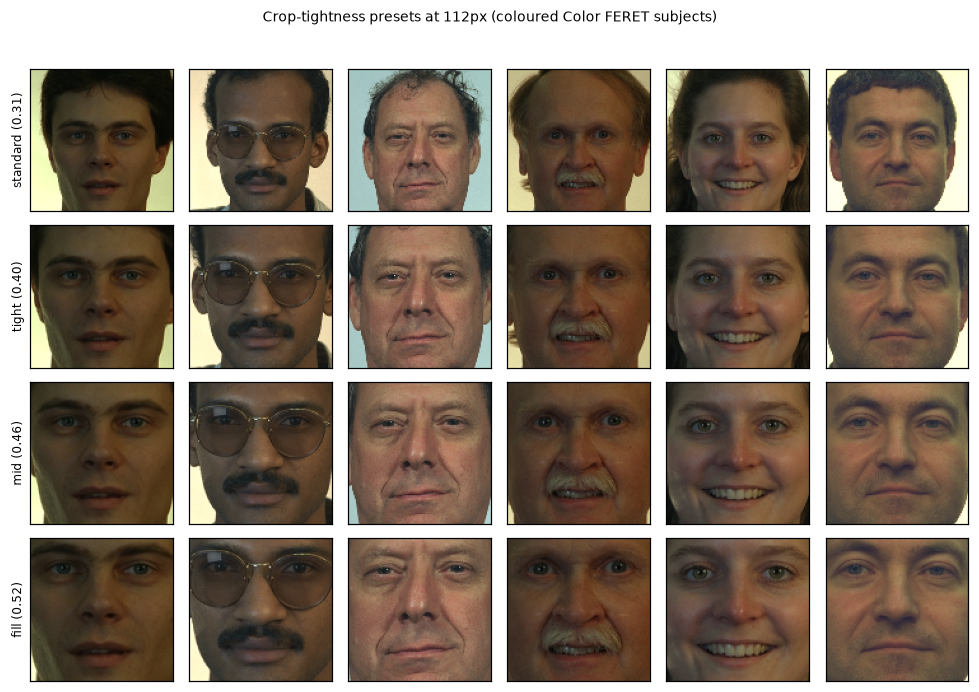}
  \caption{Crop-tightness presets at 112\,px. Rows: \emph{standard} (IOD\,$\approx$\,0.31),
  \emph{tight} (0.40), \emph{mid} (0.46), \emph{fill} (0.52). Tighter crops enlarge the
  inner face but clip peripheral context and are less robust on non-frontal poses.}
  \label{fig:align-variants}
\end{figure}

\paragraph{Method.}
The crop tightness controls the inter-ocular distance (IOD) as a fraction of the crop
width and therefore how the budget is distributed: a tighter crop devotes more pixels to
the inner face but discards peripheral context (hairline, ears, jaw). We first confirm
that the landmark-to-ArcFace alignment is correct --- Figure~\ref{fig:align-check} overlays
the detected interest points and shows the resulting crop across subjects, poses, and the
five resolutions --- and then compare four tightness presets:
\emph{standard} (IOD $\approx 0.31$, mouth row $y=92$), \emph{tight} (IOD $\approx 0.40$),
\emph{mid} (IOD $\approx 0.46$), and \emph{fill} (IOD $\approx 0.52$), shown in
Figure~\ref{fig:align-variants}.

\paragraph{Why this remains a clean (pre-compression) study.}
Unlike the operator preprocessing of Section~\ref{subsec:preproc-through}, which we measure
\emph{through} the codec because a filter's benefit may or may not survive compression, crop
tightness is a \emph{geometric} choice made entirely at the input: a tighter crop that clips
the hairline and jaw discards that context \emph{before} any codec sees the image, and no
decoder can restore pixels that were never encoded. The effect is therefore codec-independent
and is correctly isolated on clean crops; routing it through a codec would only add the
(separately measured) compression penalty on top of the geometric one. We consequently report
crop tightness on clean $112$\,px crops across all four anchors below; a through-codec
crop-tightness sweep on KK at $224$\,px would require regenerating the KK alignment presets
at that resolution and is left to future work.

\paragraph{Observations.}
Tighter crops (mid/fill) enlarge the eyes, nose, and mouth --- the regions matchers weight
most --- but at the extreme (fill) they clip the forehead and jawline and can crop poorly
on non-frontal poses, where the off-centre face no longer fits the assumed geometry. The
standard preset keeps the whole face plus a margin of context and aligns cleanly across
poses and resolutions in the alignment check.

\paragraph{Quantitative impact.}
The clean per-matcher EER on 112\,px crops confirms this: accuracy degrades
\emph{monotonically} as the crop tightens (Table~\ref{tab:ablation-crop}). Going from
\emph{standard} (IOD\,$\approx$\,0.31) to \emph{fill} (0.52) raises ArcFace EER from
0.04\,\% to 5.60\,\%, LVFace-L from 0.25\,\% to 10.44\,\%, and TopoFR-R100 from 0.04\,\%
to 3.88\,\%. The effect is most severe for the compact matcher: EdgeFace-XS climbs from
0.39\,\% at \emph{standard} to 4.73\,\% at \emph{tight}, 12.85\,\% at \emph{mid}, and
26.71\,\% at \emph{fill} --- roughly a 68$\times$ increase, and far larger than for the
heavier backbones. Aggressive crops that clip the hairline and jaw badly hurt identity,
especially for the smallest model, confirming that the \emph{standard} crop is the correct
default.

\begin{table}[t]
  \centering
  \caption{Alignment-tightness ablation: clean equal-error rate (\%) at 112\,px for each
  crop preset (before compression), indexed by inter-ocular distance as a fraction of crop
  width (IOD/W). EER degrades monotonically as the crop tightens; the compact EdgeFace-XS
  matcher is by far the most sensitive. \colorbox{green!25}{Best}/\colorbox{red!22}{worst}
  per column.}
  \label{tab:ablation-crop}
  \small
  \adjustbox{max width=\textwidth}{\input{tables/ablation_croptightness.tex}}
\end{table}

\subsection{JPEG-FzT radius}
\tldr{JPEG-FzT has a decode-time interpolation radius; increasing it from rad=1 to rad=2
cuts perceptual distortion (LPIPS) by nearly $40\%$ at the same byte size and PSNR, so the
larger radius is the better default for this codec at the 1\,kB point.}

\begin{figure}[t]
  \centering
  \includegraphics[width=\textwidth]{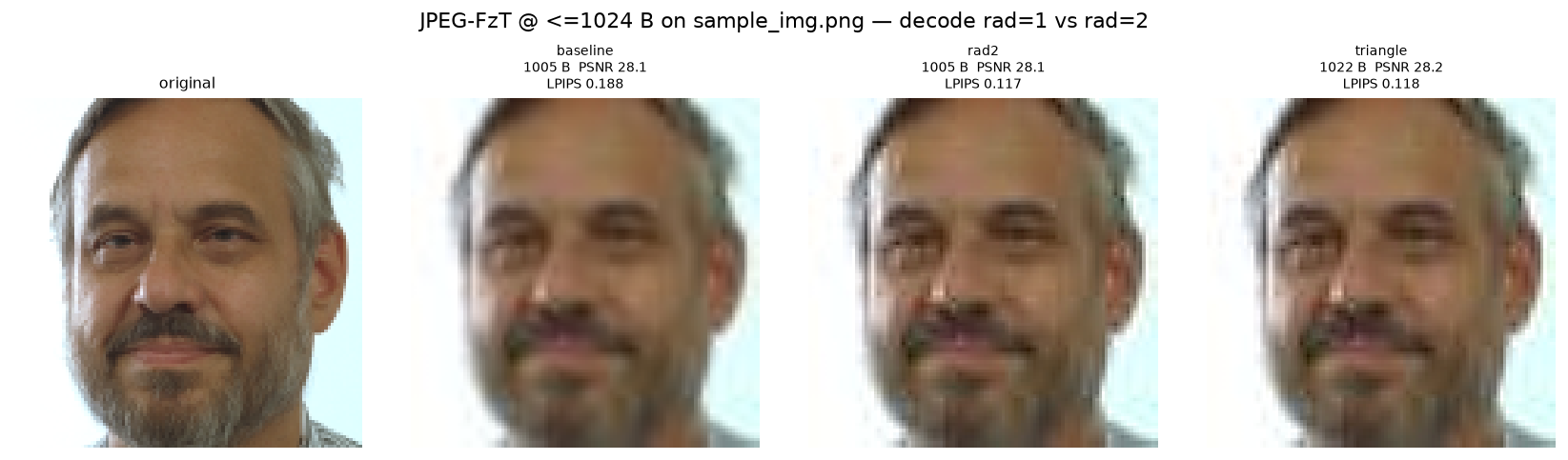}
  \caption{JPEG-FzT decode-radius sweep at $\le 1024$\,B on a sample crop. Left to right:
  original, baseline (rad=1, LPIPS 0.188), rad=2 (LPIPS 0.117), triangle kernel
  (LPIPS 0.118). At matched byte size and PSNR, the larger radius cuts LPIPS by nearly $40\%$.}
  \label{fig:fzt-ablation}
\end{figure}

\paragraph{Method.}
JPEG-FzT exposes a decode-time interpolation radius that controls how aggressively the
sparse transform coefficients are spread when reconstructing the image. At the
$\le 1024$\,B operating point we compare the baseline (rad=1), rad=2, and a triangle
kernel variant on a sample crop, holding the byte budget fixed and reporting PSNR and
LPIPS (Figure~\ref{fig:fzt-ablation}).

\paragraph{Observations.}
At essentially the same byte size and PSNR ($\approx$1005\,B, PSNR\,$\approx$\,28.1), the
larger radius substantially improves the perceptual metric: LPIPS drops from 0.188 at
rad=1 to 0.117 at rad=2, with the triangle kernel giving a near-identical 0.118 at a
slightly larger size (1022\,B, PSNR\,28.2). PSNR is insensitive to the radius (it is
dominated by low-frequency energy), but the perceptual/structural improvement is large and
visible in the reconstruction. We therefore use rad=2 as the JPEG-FzT default in the codec
benchmark.

%% file: sections/08c_annex_study.tex
\section{The ISO/IEC 29794-5 Annex E/F parameter tables}
\label{sec:annex}
\tldr{The Annex~E/F tables were fitted by maximising \emph{self-similarity} --- the cosine
between an image and its own compressed version, under a single matcher --- with a
one-factor-at-a-time search from a fixed $112$\,px baseline, so no interaction between
resolution, colour and masking was ever tested. Reproducing them under a paired
subject-level bootstrap, two thirds of the head-to-head comparisons are statistically
indistinguishable from our own configurations: at $1024$\,B most reasonable parameter sets
tie. The exceptions are real and one-sided --- our JPEG~2000 recommendation is wrong by
$1.4$--$4.7$\,pp, their parameters genuinely win for JPEG and HEIF on the in-the-wild
corpus, and their $56$\,px AVIF entry genuinely loses on Color~FERET. A joint sweep over
resolution $\times$ colour $\times$ masking improves on both only for JPEG~2000 and JPEG,
and \emph{overfits} for WebP. Self-similarity correlates with verification error
($\rho=-0.77$ over $336$ cells) but is not discriminative near the optimum, and is biased
towards blur and low resolution --- the two choices that most distinguish the annex tables
from ours.}

Annexes~E and~F of the emerging ISO/IEC~29794-5 text specify, per codec, a resolution, a
colour mode, an image manipulation and a set of CLI flags for compressing a face to
$1024$\,B~\cite{Andreas2026a,Andreas2026b}. Because those tables are normative-in-spirit
--- implementers will follow them --- it matters whether they are optimal, and whether the
objective they were fitted to is the right one. This section answers both questions on our
data and matchers.

\subsection{How the annex parameters were derived}
\label{subsec:annex-method}

Their published implementation\footnote{\url{https://github.com/dasec/1kB-FaceImage}, MIT.}
makes the derivation explicit, and two properties of it drive everything below.

\paragraph{The objective is self-similarity, not verification.}
Every Optuna case maximises the mean cosine between each original image and its own
compressed version, scored with \emph{one} matcher (CVLface AdaFace ViT-B KP-RPE,
WebFace12M). There are no impostor pairs, no mated pairs across captures, and no EER
anywhere in the fitting loop. A configuration is therefore rewarded for reconstructing
\emph{that image} faithfully, which is not the same as preserving what separates one
identity from another.

\paragraph{The search was one factor at a time.}
Each case fixes resolution at $112$\,px, preprocessing at ``default'' and a colour mode,
then sweeps exactly one flag; the remaining axes are commented out and the study is re-run
under a new name. The annex tables nevertheless present a \emph{combination} per codec.
Interactions --- greyscale frees bits, which shifts the optimal resolution, which changes
what masking can buy --- were never in scope. That is the gap Section~\ref{subsec:annex-joint}
measures.

\paragraph{The manipulations, resolved from the source.}
\texttt{rectangle\_mean} blurs the whole image once with a $3\times3$ box filter and then
restores, through a \emph{hard} (unfeathered) mask, the rectangle spanning the five
landmarks grown by $20\,\%$ of the total resolution on each side; \texttt{ofiq\_landmarks}
restores an OFIQ 98-point face-contour polygon instead; \texttt{mean} keeps the blur
everywhere. The manipulation is applied \emph{after} the resize, and each resolution is
produced by a fresh similarity transform at that size.

\subsection{Protocol}
\label{subsec:annex-protocol}

We reproduce their pipeline inside ours. Crops are aligned to the ArcFace 5-point template
at each target resolution independently, exactly as in Section~\ref{sec:datasets};
because that alignment is canonical, the five landmarks land on the template and the
\texttt{rectangle\_mean} rectangle is exact without a detector. Every cell --- a
(codec, resolution, colour, manipulation, flags) tuple --- is compressed to
$\le 1024$\,B by a binary search on the codec's quality knob, decoded, resized to
$112$\,px and embedded, with nothing written to disk in between.

Scoring uses both objectives: \emph{self-similarity} exactly as they define it (their
matcher, mean cosine to the uncompressed original), and \emph{verification} EER over the
fixed pair set of Section~\ref{sec:protocol}. We report EER under their single matcher
(``their'') and averaged over our four anchors (``anchors''), so that ``their parameters
are worse'' can be separated from ``their parameters are tuned to one ViT''. Both the
symmetric protocol (both sides compressed, the convention of this report) and the
asymmetric one (compressed reference against an uncompressed probe, the document scenario
the annexes describe) are computed; conclusions are identical under both, and we quote the
symmetric numbers.

Experiment~1 runs at the full population. The $336$-cell grid of
Section~\ref{subsec:annex-joint} runs on a $2000$-crop prefix (whole identities, so mated
pairs survive) and its winners are then re-confirmed at the full population, which is why
the two are never compared across tables without being re-scored on the same rows.

\paragraph{Four deviations, stated up front.}
(i) OFIQ's ADNet landmarker is not available in this environment, so the contour polygon
is built from InsightFace's 106-point contour --- same construction, different landmarker.
(ii) libjxl exposes effort~$10$ only behind \texttt{-{}-allow\_expert\_options}; the annexes'
``effort~$=10$'' therefore runs as~$9$. (iii) WebP's \texttt{sns} knob is not exposed by
Pillow, so WebP cells carrying it are encoded with \texttt{cwebp}; an engine-matched
control confirms this does not drive the comparison. (iv) Our JPEG~2000 rate ladder had to
be extended to ratio~$2$: at $56$\,px a $1024$\,B budget is a third of a bit per pixel, and
a ladder stopping at ratio~$12$ caps the file at ${\sim}260$\,B, which would have handed the
annexes' $56$\,px JPEG~2000 configurations a fictitious handicap.

\subsection{Experiment 1 --- are the annex parameters better than ours?}
\label{subsec:annex-exp1}

Table~\ref{tab:annex-exp1} is the head-to-head: the $12$ unique Annex~E/F configurations
against our recommended configuration per codec, on both datasets, under both objectives
and both matcher sets.

\begin{table}[htbp]
\centering
\caption{Experiment~1: the Annex~E/F configurations against ours, at $1024$\,B. ``self-sim''
is their objective under their matcher; EER (\%) is ours, symmetric protocol, under their
matcher and averaged over our four anchors. Full population, both datasets.}
\label{tab:annex-exp1}
\adjustbox{max width=\textwidth}{\input{tables/annex_exp1}}
\end{table}

\paragraph{Point estimates alone would mislead here.}
Read as point estimates, Table~\ref{tab:annex-exp1} says our configurations win on
Color~FERET and theirs win on AI-Solutions-KK. Most of those margins are, however, within
sampling error. Table~\ref{tab:annex-ci} therefore resamples \emph{identities} with
replacement ($200$ draws) and reports the \emph{paired} difference against our
configuration, with the four anchor matchers fused at score level. Eight of the twelve
comparisons have a confidence interval containing zero.

\paragraph{What survives.} Three findings do.
\emph{(i) Our JPEG~2000 configuration is the wrong place to have measured that codec},
decisively and on both corpora: the annexes' $56$\,px greyscale entry is
$-1.38$\,pp [$-1.59$, $-1.19$] on Color~FERET and $-4.65$\,pp [$-5.07$, $-4.29$] on
AI-Solutions-KK. Colour JPEG~2000's rate allocation collapses inside $1024$\,B, and the
annexes found the greyscale escape route that we did not --- which, since this report's
``avoid JPEG~2000'' verdict rests on the colour measurement, matters beyond this section
(Section~\ref{subsec:annex-wg3}).
\emph{(ii) Their JPEG and HEIF entries are genuinely better on the in-the-wild corpus}
($-0.34$\,pp [$-0.41$, $-0.25$] and $-0.09$\,pp [$-0.13$, $-0.03$]), while being
indistinguishable from ours on Color~FERET.
\emph{(iii) Their $56$\,px AVIF entry is genuinely worse} on Color~FERET
($+0.06$\,pp [$+0.02$, $+0.10$]) --- the only case where a specific annex choice is
measurably harmful, and it is a resolution choice.

\paragraph{Everything else is a tie.} JPEG~XL is indistinguishable on both corpora, and so
are AVIF on AI-Solutions-KK and HEIF and JPEG on Color~FERET. At $1024$\,B, with a modern
codec at a sane resolution, the parameter set simply does not matter much --- which is
itself a result worth putting in front of WG~3, because a normative table implies that it
does.

\begin{table}[htbp]
\centering
\caption{Paired subject-level bootstrap ($200$ resamples of identities) of the EER
difference against our configuration, in percentage points; negative favours the
alternative. Four anchor matchers fused at score level, full population.
$^\dagger$ marks an interval excluding zero.}
\label{tab:annex-ci}
\adjustbox{max width=\textwidth}{\input{tables/annex_ci}}
\end{table}

\subsection{Experiment 2 --- is there a better set than either?}
\label{subsec:annex-joint}

Their one-factor-at-a-time search could not see interactions, and ours never varied colour
or masking at all. Stage~A therefore sweeps resolution $\times$ colour $\times$
manipulation jointly at default flags ($7\times2\times4=56$ cells per codec, $336$ in
total); Stage~B then sweeps each codec's own flags around its two best Stage-A cells
($62$ further cells). Table~\ref{tab:annex-joint} is the result.

\begin{table}[htbp]
\centering
\caption{Experiment~2: the joint sweep against both parameter sets, EER (\%), symmetric,
mean of ArcFace-antelopev2 and their ViT-B, on the common $2000$-crop prefix of
Color~FERET --- the set the sweep was selected on. Table~\ref{tab:annex-ci} gives the
held-out verdict.}
\label{tab:annex-joint}
\adjustbox{max width=\textwidth}{\input{tables/annex_joint}}
\end{table}

On the prefix the grid was selected on, the joint sweep beats both parameter sets in $5$ of
$6$ codecs, by up to $10.6\times$ (JPEG~2000). That margin is largely an artefact of
selecting and evaluating on the same $2000$ crops under the same two matchers. Re-measured
at the full population under the four anchors --- three of which played no part in the
selection --- the sweep's advantage over our configuration is significant only for
JPEG~2000 (both corpora) and JPEG (AI-Solutions-KK), is indistinguishable from ours for
AVIF, HEIF and JPEG~XL, and is significantly \emph{worse} for WebP on AI-Solutions-KK
($+0.16$\,pp [$+0.11$, $+0.21$]; Table~\ref{tab:annex-ci}). The WebP case is instructive:
its winning cell added contour masking at $168$\,px, a combination that was ahead by
$0.017$\,pp on the selection set and behind by $0.16$\,pp on a held-out corpus. A grid this
large, scored on a subsample, will manufacture such winners; only the JPEG~2000 result is
big enough to be immune.

\paragraph{Where the gain comes from.}
Table~\ref{tab:annex-axes} gives the marginal effect of each axis. Resolution dominates:
the mean EER at $56$\,px is $0.92\,\%$ against $0.43\,\%$ at $96$--$168$\,px, so the
annexes' $56$--$64$\,px choices sit on the wrong side of the optimum for five of the seven
codecs. Greyscale looks strongly beneficial marginally ($0.49$ vs $0.66\,\%$) --- but that
average is carried almost entirely by JPEG~2000, and at each codec's own best resolution
and colour mode the picture reverses.

\begin{table}[htbp]
\centering
\caption{Stage~A marginal axis effects: mean EER (\%) over all cells at each level.}
\label{tab:annex-axes}
\adjustbox{max width=\textwidth}{\input{tables/annex_axes}}
\end{table}

\paragraph{Masking looks good on the selection set and mostly does not transfer.}
Averaged over the grid, \texttt{rectangle\_mean} is the best manipulation ($0.494$ vs
$0.580\,\%$ for no manipulation), and held at each codec's own best resolution and colour
mode (Table~\ref{tab:annex-manip}) some manipulation still wins for four of six codecs.
That is the evidence on which a masking column in a parameter table would be written. It
does not survive a held-out corpus: of the five per-codec winners that carry a
manipulation, only two reproduce (JPEG's rectangle, and JPEG~2000's full blur --- both
codecs whose rate allocation is starved at $1024$\,B), two are ties, and WebP's contour
mask is significantly \emph{worse} on AI-Solutions-KK. Masking buys back bits where the
codec is short of them, which at this budget means JPEG and JPEG~2000; elsewhere it
removes detail for nothing.

\begin{table}[htbp]
\centering
\caption{Manipulation held at each codec's best resolution and colour mode. EER (\%),
Color~FERET $2000$-crop prefix.}
\label{tab:annex-manip}
\adjustbox{max width=\textwidth}{\input{tables/annex_manip}}
\end{table}

\paragraph{Flags matter less than the axes, with one instructive exception.}
Stage~B moves EER by $\le 0.02$\,pp for AVIF, HEIF, JPEG~XL and JPEG~2000; HEIF's chroma
downsampling makes no measurable difference at all ($0.116\,\%$ for all three settings),
and JPEG~XL's effort~$10$ is \emph{worse} than the default ($0.144$ vs $0.119\,\%$) --- both
are flags the annex tables specify as though they mattered. JPEG is the exception, and it
is the clearest single illustration of the interaction problem: their own
\texttt{smooth}~$=30$ plus arithmetic coding is genuinely worth having, improving
$0.201 \to 0.139\,\%$ --- but only at \emph{colour} $96$\,px. Combined with the greyscale
and masking their table also prescribes, the same flags land at $0.273\,\%$, worse than
doing nothing. Each of their choices helps alone; together they do not.

\paragraph{What greyscale actually buys.}
Because the premise of greyscale and masking is that they free bits, we record bytes at a
fixed quality setting alongside EER at the fixed budget (Table~\ref{tab:annex-bytes}).
Greyscale frees $14$--$23\,\%$ for JPEG and up to $66\,\%$ for JPEG~2000 at $224$\,px, but
only $5$--$17\,\%$ for AVIF, JPEG~XL and HEIF --- and only for JPEG~2000 does the freed
budget convert into lower EER. For the modern codecs the chroma planes were never the
expensive part, so discarding colour costs identity information and buys almost nothing.

\begin{table}[htbp]
\centering
\caption{Bytes freed by greyscale at a fixed quality setting, as a percentage of the
colour file size (no manipulation).}
\label{tab:annex-bytes}
\adjustbox{max width=\textwidth}{\input{tables/annex_bytes}}
\end{table}

\subsection{Does self-similarity predict verification error?}
\label{subsec:annex-selfsim}

This is the question that decides how the finding should be worded to WG~3, and the honest
answer is more nuanced than ``their objective is wrong''.

Across the $336$ classical grid cells, self-similarity and EER are strongly rank-correlated
under their own matcher: Spearman $\rho = -0.77$ ($p < 10^{-60}$), and between $-0.53$ and
$-0.91$ within every individual codec (Table~\ref{tab:annex-selfsim}). Self-similarity is
therefore a \emph{usable} proxy: it separates a bad configuration from a good one.

\begin{table}[htbp]
\centering
\caption{Their self-similarity objective against verification EER over the Stage-A grid,
under their own matcher. $\rho$ is Spearman's rank correlation; the last four columns
compare the configuration self-similarity selects with the EER-optimal one.}
\label{tab:annex-selfsim}
\adjustbox{max width=\textwidth}{\input{tables/annex_selfsim}}
\end{table}

What it is not is \emph{discriminative near the optimum} --- which is exactly the regime a
parameter table is chosen in. Ranking by self-similarity picks a configuration up to
$1.9\times$ worse than the EER optimum (WebP: $0.160$ vs $0.085\,\%$; HEIF $0.186$ vs
$0.108$; JPEG $0.408$ vs $0.241$), and it errs in a systematic direction: it prefers lower
resolution and the full-blur manipulation, both of which flatter a same-image cosine ---
a blurred image is easy to reconstruct consistently --- while removing detail that
cross-capture matching needs. Only for JPEG~XL do the two objectives agree on the winner.

The defensible message is therefore not that the annex tables optimise a quantity unrelated
to verification error, but that they optimise a quantity that cannot resolve the last
factor of two, on a search that could not see interactions, under a single matcher.

\subsection{JPEG-AI}
\label{subsec:annex-jpegai}

JPEG-AI is the cost constraint of this study (${\sim}2$\,s per encode even with the target
bit-rate carried between images), so its grid runs on a $500$-crop subsample at the two
resolutions the annexes use (Table~\ref{tab:annex-jpegai}). The ordering is unambiguous
anyway, because JPEG-AI's margin over the classical codecs is large: the best cell reaches
$0.11\,\%$ EER at $168$\,px colour with no manipulation, against $0.084\,\%$ for the best
classical cell, and its self-similarity ($0.925$) is far above any classical codec's.

\begin{table}[htbp]
\centering
\caption{JPEG-AI arm of the joint grid, $500$-crop subsample of Color~FERET, $1024$\,B.}
\label{tab:annex-jpegai}
\adjustbox{max width=\textwidth}{\input{tables/annex_jpegai}}
\end{table}

Two of the expansion plan's open questions are answered here empirically. The reference VM
does accept greyscale content (in an RGB container), so Annex~F's greyscale JPEG-AI entry
is implementable --- but it is strictly harmful: greyscale is worse than colour at every
one of the four manipulations, roughly doubling EER, and it frees no bytes at all
($979$ vs $980$\,B median), because the learned entropy model was not spending its budget
on chroma in the first place. And manipulation never helps: the annexes' full blur
(Annex~F) is the worst of the four at both resolutions, and OFIQ-style contour masking
(Annex~E) is no better than nothing.

\paragraph{Selection overfitting, quantified.} Of the six per-codec winners the grid
produced, one (JPEG~2000) reproduces at the full population on both corpora, one (JPEG)
reproduces on one corpus, three are ties and one is a regression. This is the same failure
mode as the annexes' own procedure, one level up: a large configuration space, a single
selection metric, and no held-out confirmation. We report it rather than quietly quoting
the selection-set numbers.

\subsection{Recommendation to WG 3}
\label{subsec:annex-wg3}

Three things follow from the numbers above.

\begin{enumerate}
  \item \textbf{The annex resolutions are too small.} Averaged over the grid, mean EER is
  $0.92\,\%$ at $56$\,px against $0.43\,\%$ at $96$--$168$\,px, and the one annex parameter
  choice that is measurably harmful in the head-to-head is a $56$\,px entry (AVIF on
  Color~FERET). JPEG-AI's optimum is $168$\,px. This is the single largest correctable
  error in the tables.
  \item \textbf{Greyscale and masking should be per-codec, not per-table.} They pay for
  JPEG~2000 --- and only there does greyscale free a meaningful fraction of the budget
  ($66\,\%$ at $224$\,px, against $5$--$17\,\%$ for AVIF, JPEG~XL and HEIF). For JPEG-AI
  greyscale frees nothing at all and doubles EER. Prescribing them uniformly, as Annex~F
  largely does, transfers a JPEG~2000 remedy onto codecs that do not need it.
  \item \textbf{The table should not over-specify what the evidence cannot resolve.} Two
  thirds of the head-to-head comparisons are statistically indistinguishable, and several
  specified flags have no measurable effect (HEIF chroma downsampling: $0.116\,\%$ for all
  three settings; JPEG~XL effort~$10$ is worse than the default). Fixing a single
  configuration per codec implies a precision the measurements do not support; a permitted
  range, plus the resolution floor above, would be defensible where a point value is not.
  \item \textbf{If a point value is required, fit it on verification, with held-out
  confirmation.} Self-similarity is a usable coarse filter ($\rho=-0.77$) but cannot
  resolve the last factor of two and is biased towards blur and low resolution. Our own
  joint sweep shows the other half of the lesson: even fitting on EER, a winner selected on
  a subsample under two matchers failed to reproduce for four of six codecs.
\end{enumerate}

The same evidence rebounds on this report. Everywhere else we evaluate JPEG~2000 in
colour, and on that basis Table~\ref{tab:deployment-guidance} tells implementers to avoid
it in all configurations. In greyscale at $224$\,px with a full blur and seven resolution
levels, JPEG~2000's EER falls by $1.4$\,pp on Color~FERET and $4.7$\,pp on
AI-Solutions-KK --- from worst-in-roster to within reach of the other classical codecs.
The ``avoid JPEG~2000'' verdict is therefore a verdict on \emph{colour} JPEG~2000, and
should be re-stated as such rather than treated as settled; a greyscale re-measurement
across the full budget sweep is the obvious follow-up. Our other recommendations stand
unchanged: no other configuration change we tested reproduces on a held-out corpus.

%% file: sections/09_fairness.tex
\section{Demographic Fairness}
\label{sec:fairness}
\tldr{We ask whether squeezing a face to a kilobyte hurts some demographic groups more than others. Across codecs the absolute equal-error rates (EERs) stay low and the rank of the groups barely changes, but compression does inflate the between-group gap, and a few codecs (notably JPEG~2000) amplify it sharply.}

Face-recognition accuracy is known to vary across demographic groups even on uncompressed
imagery~\cite{grother2019frvt}: the NIST FRVT demographic study reported systematically
higher false-match and false-non-match rates for some skin-tone, sex, and age cohorts. The
question this section addresses is narrower and codec-specific: \emph{does pushing a face
crop to the sub-1\,kB budget widen those demographic gaps, and if so, which codecs are
worst?} We answer it by recomputing the verification EER \emph{within} each demographic
subgroup and comparing the spread of subgroup EERs against
the uncompressed (\texttt{aligned}) baseline --- primarily at the 1024-byte operating point,
and, because the learned codec is recommended for the harder 512-byte point, again there
(Section~\ref{subsec:fairness-512}).

Unless stated otherwise, the per-subgroup tables in this section are reported at the 112\,px
working resolution and the 1024-byte operating point with the \texttt{edgeface\_xs}
matcher~\cite{george2024edgeface}, our weakest (and therefore most discriminating) anchor.
The underlying fairness CSVs are, however, the full four-anchor set: they cover all four
recognition anchors -- \texttt{arcface\_antelopev2}~\cite{deng2019arcface},
\texttt{edgeface\_xs}~\cite{george2024edgeface}, \texttt{lvface\_l} and
\texttt{topofr\_r100} -- across the 512- and 1024-byte budgets and resolutions from 64 to
224\,px. All values are taken from the per-subgroup fairness tables and the max$-$min disparity
tables (Section~\ref{sec:artifacts}); the
cross-anchor disparity summary in Table~\ref{tab:fairness-disparity-kk} draws on all four.
Color~FERET supplies controlled pose, sex, age, and a coarse ethnicity label;
AI-Solutions-KK supplies Monk skin-tone~\cite{monk2023mst}, sex, and age.
\emph{The KK attributes are model-estimated}, not curated: Monk skin-tone comes from an
automatic tone estimator and age/gender from a predictor, propagated per identity to
reduce per-image noise, on a set that is not demographically balanced. Crucially, the tone
estimator's own error is skin-tone-correlated, so the extreme (darkest and lightest) cells are
the least reliable, and the per-subgroup identity counts are tiny (MST8 spans $4$
identities, MST10 $5$ and MST9 $7$; only MST6 and MST7, at $45$ and $30$, are
well populated). The KK fairness
results are therefore exploratory, and Color~FERET --- whose pose, sex, age, and
ethnicity labels are dataset-supplied --- is the confirmatory ground-truth anchor. We
read the two datasets together for that reason. The custom learned
codec is scored in these fairness CSVs as well: we report both variants (Ours-ACCURATE
and Ours-FAST) alongside the classical codecs in the subgroup tables
below,\footnote{The Ours-* rows were \emph{refreshed} under the promoted $3$M-step ACCURATE
checkpoint (uniform MST8-excluded basis); the classical JPEG~2000 amplification
($11.2$--$20.0\times$, KK skin-tone) is checkpoint-independent. On KK skin tone,
Ours-ACCURATE is the second-lowest-disparity codec on ArcFace ($0.75$\,pp), behind JPEG-AI
($0.72$\,pp) and ahead of WebP ($0.90$\,pp). On Color~FERET ethnicity, \texttt{accurate\_3M}
measures $0.42$\,pp on ArcFace --- mid-band, with WebP ($0.32$), AVIF ($0.35$) and JPEG-AI
($0.26$) below and JPEG ($0.58$) and JPEG~2000 ($3.62$) above. We make no cross-checkpoint
claim here: the $0.67$\,pp once quoted for the superseded \texttt{accurate\_1M} run used
seven ethnicity cohorts, three with only $8$--$319$ impostor pairs, not comparable to the
four-cohort basis used elsewhere --- these remain max--min spreads over small subgroups at
low error rates.} and revisit its
reconstruction brightness bias as a separate watch item at the end of the section. Because the
subgroup tables in this section are anchored on \texttt{edgeface\_xs}, and the Ours identity
side-stream is trained against a frozen EdgeFace embedding, the Ours rows on any
EdgeFace-family anchor are confounded (not independent); the trustworthy Ours comparisons are
the independent matchers -- \texttt{arcface\_antelopev2} in these tables and the held-out
CVLFace-IR101 evaluation reported elsewhere in this report.

\subsection{Subgroup equal-error rates}
\tldr{Within every subgroup the EER stays well under a few percent for the strong codecs (AVIF, HEIF, WebP, JPEG-AI), but every lossy codec raises subgroup EERs above the uncompressed baseline, and JPEG~2000 raises them by an order of magnitude.}

We define a subgroup's EER as the equal-error rate computed only over pairs in which the
relevant attribute value matches, so each subgroup is scored on its own genuine/impostor
balance. The uncompressed Color~FERET baseline (\texttt{edgeface\_xs}) is already low:
overall skin-tone EER 0.46\%, with the well-populated cohorts spanning White 0.28\% to Black
0.98\% and Asian 0.87\%; the thinly-populated Pacific-Islander cell sits higher at 2.51\%.
On AI-Solutions-KK the uncompressed overall skin-tone EER is 1.49\%, with the worst Monk
subgroup MST8 at 3.79\% and the darkest subgroup MST10 at 0.44\%. The stronger
anchors are uniformly lower: on AI-Solutions-KK the uncompressed overall skin-tone EER is
0.22\% (\texttt{arcface\_antelopev2}), 0.19\% (\texttt{topofr\_r100}) and 0.31\%
(\texttt{lvface\_l}), so the \texttt{edgeface\_xs} figures we tabulate are a conservative
upper bound on the demographic spread.

Compression raises these uniformly. On Color~FERET at 1024 bytes
(\texttt{edgeface\_xs}) the strong codecs land in a tight band -- WebP overall skin-tone EER
0.70\%, AVIF 0.70\%, JPEG-AI~\cite{ascenso2023jpegai} 0.73\%, HEIF 0.96\% -- while plain
JPEG~\cite{wallace1992jpeg} rises to 1.11\% and JPEG~2000~\cite{taubman2002jpeg2000} jumps to
6.08\%, an order-of-magnitude increase over its own uncompressed reference. The same
ordering holds on AI-Solutions-KK, where the overall skin-tone EER at 1024 bytes is 2.47\%
(WebP), 2.76\% (AVIF), 3.22\% (HEIF), 3.69\% (JPEG), 3.24\% (JPEG-FzT), 3.35\%
(JPEG~XL~\cite{alakuijala2019jpegxl}), and 13.27\% (JPEG~2000). The learned codec lands
inside this strong-codec band rather than as an outlier: its overall skin-tone EER is 0.64\%
(Ours-ACCURATE) and 0.96\% (Ours-FAST) on Color~FERET and 1.89\% / 2.91\% on
AI-Solutions-KK, i.e.\ low single-digit percent and inside the classical-codec range ---
Ours-ACCURATE is in fact the lowest overall EER of any codec on both datasets --- with
the more accurate variant consistently the fairer of the two (lower EER in every tabulated
subgroup on both datasets). The same codec ranking --
WebP/AVIF best, JPEG~2000 worst by an order of magnitude -- reproduces on all four anchors:
on AI-Solutions-KK the JPEG~2000 overall skin-tone EER is 8.67\%
(\texttt{arcface\_antelopev2}), 9.08\% (\texttt{topofr\_r100}) and 11.58\%
(\texttt{lvface\_l}), each roughly an order of magnitude above the same anchor's WebP figure
(0.69\%, 0.51\% and 0.62\% respectively).
Table~\ref{tab:fairness-subgroup-cf} and Table~\ref{tab:fairness-subgroup-kk} give the
per-subgroup breakdown for a representative spread of codecs. Because these subgroup tables
are scored on \texttt{edgeface\_xs}, the Ours rows in them are confounded with the codec's
EdgeFace-anchored identity side-stream; the independent-anchor (ArcFace/LVFace/TopoFR)
disparities for the learned codec are the ones to trust and are reported in the disparity
table (Table~\ref{tab:fairness-disparity-cf}).

\begin{table}[t]
  \centering
  \small
  \caption{Color~FERET subgroup EER (\%) at 112\,px, 1024\,B, \texttt{edgeface\_xs}. Pose
  subgroups; ethnicity (skin-tone label) subgroups. ``base'' is uncompressed \texttt{aligned}.}
  \label{tab:fairness-subgroup-cf}
  \adjustbox{max width=\textwidth}{%
  \begin{tabular}{lrrrrrrr}
    \toprule
    Subgroup & base & WebP & AVIF & JPEG & JPEG~2000 & Ours-ACC & Ours-FAST \\
    \midrule
    \multicolumn{8}{l}{\emph{Pose}}\\
    frontal  & 0.00 & 0.02 & 0.02 & 0.03 & 2.35 & 0.00 & 0.02 \\
    quarter  & 0.00 & 0.02 & 0.02 & 0.08 & 2.29 & 0.00 & 0.02 \\
    half     & 0.03 & 0.20 & 0.20 & 0.23 & 4.45 & 0.08 & 0.15 \\
    profile  & 0.67 & 0.92 & 0.94 & 1.41 & 5.18 & 0.96 & 1.19 \\
    \midrule
    \multicolumn{8}{l}{\emph{Ethnicity}}\\
    White            & 0.28 & 0.48 & 0.49 & 0.86 & 5.19 & 0.45 & 0.78 \\
    Hispanic         & 0.31 & 0.53 & 0.64 & 1.40 & 7.43 & 0.45 & 1.01 \\
    Asian            & 0.87 & 1.27 & 1.19 & 1.68 & 7.67 & 1.18 & 1.51 \\
    Black            & 0.98 & 1.42 & 1.67 & 2.30 & 9.85 & 1.33 & 2.00 \\
    Pacific-Islander & 2.51 & 1.93 & 2.51 & 3.43 & 10.63 & 3.43 & 4.11 \\
    \bottomrule
  \end{tabular}}
\end{table}

\begin{table}[t]
  \centering
  \small
  \caption{AI-Solutions-KK subgroup EER (\%) at 112\,px, 1024\,B, \texttt{edgeface\_xs}.
  Monk skin-tone subgroups~\cite{monk2023mst} (MST5 = lightest present, MST10 = darkest).
  ``base'' is uncompressed \texttt{aligned}. Attributes are model-estimated and subgroups are
  small (MST8 spans $4$ identities, MST10 $5$, MST9 $7$), so per-cell EERs are noisy; MST8 in
  particular rests on a pair pool of $n_{\text{pos}}{=}47{,}266$ / $n_{\text{neg}}{=}4{,}349$,
  identical for every codec, and it is the cohort excluded from the uniform disparity basis of
  Table~\ref{tab:fairness-disparity-kk} for that reason.}
  \label{tab:fairness-subgroup-kk}
  \adjustbox{max width=\textwidth}{%
  \begin{tabular}{lrrrrrrr}
    \toprule
    Subgroup & base & WebP & AVIF & JPEG & JPEG~2000 & Ours-ACC & Ours-FAST \\
    \midrule
    MST5  & 0.81 & 1.71 & 2.23 & 3.48 & 15.74 & 1.25 & 2.36 \\
    MST6  & 1.50 & 2.48 & 2.80 & 3.87 & 14.27 & 1.94 & 3.00 \\
    MST7  & 1.16 & 1.96 & 2.21 & 3.13 & 12.85 & 1.51 & 2.39 \\
    MST8  & 3.79 & 4.79 & 5.27 & 6.91 & 17.56 & 4.30 & 6.71 \\
    MST9  & 1.97 & 3.25 & 3.67 & 5.08 & 17.44 & 2.42 & 3.74 \\
    MST10 & 0.44 & 0.95 & 1.09 & 1.95 & 10.95 & 0.71 & 1.42 \\
    \bottomrule
  \end{tabular}}
\end{table}

MST8 is the hardest Monk cohort in the uncompressed baseline (\texttt{edgeface\_xs} EER
3.79\%) and stays hardest under every codec (WebP 4.79\%, AVIF 5.27\%, JPEG 6.91\%,
JPEG~2000 17.56\%), so compression deepens the existing cohort ordering rather than
reordering it. It is also the thinnest cohort in the set --- $4$ identities, and a pair pool
of $n_{\text{pos}}{=}47{,}266$ / $n_{\text{neg}}{=}4{,}349$ that is identical for every codec
--- which is why the disparity summary of Section~\ref{sec:fairness} is reported on a uniform
MST5/6/7/9/10 basis that excludes it: a max$-$min spread anchored on a $4$-identity cell
would be dominated by that cell's noise rather than by the codec. We tabulate MST8 here, but
read the disparity on the uniform basis. On Color~FERET the two thinnest
ethnicity labels -- Native-American ($n_{\text{neg}}{=}8$) and Other ($n_{\text{neg}}{=}22$)
-- are too sparse to be reliable and are omitted from Table~\ref{tab:fairness-subgroup-cf};
the Pacific-Islander cell ($n_{\text{neg}}{=}319$) is small but populated and we report it
with that caveat.

\subsection{Skin-tone effects}
\tldr{Reading the Monk scale as light/mid/dark tertiles, no codec inverts the baseline tone ordering, but that ordering is non-monotone -- the darkest tone MST10 is among the easiest and the mid-dark MST8 the hardest -- so we claim no clean darker-is-worse gradient from the (model-estimated) KK tones; the robust demographic signal is the Color~FERET ground-truth ethnicity anchor, and the between-group gap widens sharply under JPEG~2000.}

To summarise the skin-tone effect on AI-Solutions-KK we group the Monk subgroups into
tertiles: light (MST5--MST6), mid (MST7--MST8) and dark (MST9--MST10), following the Monk
skin-tone scale~\cite{monk2023mst} convention that higher indices are darker. For the
rather than impute a weighted tertile mean from cohorts of very different size we report the
Monk subgroups directly. The sparse Monk tones rest on only
$4$--$7$ identities each (MST8 the smallest at $4$), so the
per-tone EER cells are individually high-variance; we therefore read the coarse tertile
grouping, rather than any single cell, as the reliable tone signal.

The per-tone pattern on AI-Solutions-KK is non-monotone rather than a clean darker-is-worse
gradient, echoing the fact that unconstrained FRVT-style orderings~\cite{grother2019frvt} need
not increase with tone: the darkest subgroup MST10 is among the \emph{easiest} at baseline
(EER 0.44\%), alongside the lightest-present MST5, while the mid-dark MST8 is the single
hardest (3.79\%) and MST9 the next hardest. Compression preserves this ordering
rather than reshuffling it -- MST8 stays the worst Monk subgroup under every codec (WebP
4.79\%, AVIF 5.27\%, JPEG 6.91\%, JPEG~2000 17.56\%) with MST9 next (WebP 3.25\%, JPEG~2000
17.44\%), and MST10 stays among the easiest (WebP 0.95\%). The absolute increases are not
tone-monotone either: for WebP the dark MST9 rises from a 1.97\% baseline to 3.25\%
($+$1.28\,pp) while the light MST5 rises by $+$0.90\,pp (0.81\% to 1.71\%) and the darkest
MST10 by only $+$0.51\,pp; under JPEG~2000 MST9 rises by roughly 15\,pp. So the answer to ``does compression widen the
skin-tone gap?'' is yes for the spread, but the effect is dominated by JPEG~2000 and is mild
for the perceptually strong codecs (WebP/AVIF/HEIF); because the KK ordering is non-monotone
and the KK tones are model-estimated, we do not claim a clean darker-skin penalty from KK
alone and instead treat the Color~FERET ground-truth ethnicity anchor (below) as the robust
signal. The cohort ordering is also anchor-dependent, which is a further reason to read the
tertiles rather than single cells: on \texttt{lvface\_l}, where MST8 is the hardest cell at
baseline (0.67\% vs.\ MST9 0.45\%), it is MST9 that is hardest under compression (WebP 0.75\%
vs.\ MST8 0.61\%; JPEG 1.65\% vs.\ 1.12\%; JPEG~2000 12.19\% vs.\ 10.60\%) --- the opposite of
the \texttt{edgeface\_xs} ordering, and not evidence of a monotone darker-is-worse trend on
either anchor.

The learned codec follows the same tone ordering as the classical codecs rather than
introducing or removing a bias. MST8 is squarely the hardest Monk subgroup for both variants
(Ours-ACCURATE 4.30\%, Ours-FAST 6.71\%) exactly as it is for the classical codecs, while the
darkest subgroup MST10 is the easiest (0.71\% / 1.42\%); Ours-ACCURATE is in fact the lowest
of any codec in that hardest cell (4.30\% against JPEG-AI 4.47\% and WebP 4.79\%), while
Ours-FAST (6.71\%) sits mid-pack among the classical codecs. Ours-ACCURATE is uniformly lower than Ours-FAST across all six
Monk subgroups, and its spread from MST10 to MST8 ($3.59$\,pp) is well inside the
classical-codec range; on the uniform MST8-excluded basis its EdgeFace disparity ($1.71$\,pp)
is just over a quarter of JPEG~2000's ($6.49$\,pp).

On Color~FERET the ethnicity labels behave the same way (Table~\ref{tab:fairness-subgroup-cf}):
Asian and Black are the hardest well-populated cohorts at baseline and remain so after
compression, and their absolute EER increase under JPEG and JPEG~2000 is among the largest.
The learned codec keeps this ordering as well -- White and Hispanic are the easiest
well-populated cohorts (Ours-ACCURATE 0.45\% / 0.45\%, Ours-FAST 0.78\% / 1.01\%) and Asian
and Black the hardest (Ours-ACCURATE 1.18\% / 1.33\%, Ours-FAST 1.51\% / 2.00\%) -- with
Ours-ACCURATE fairer than Ours-FAST in every tabulated cohort and both squarely within the
strong-codec band rather than approaching the JPEG~2000 spread.
The neural baselines (\texttt{neural\_bmshj2018}, \texttt{neural\_mbt2018\_mean}),
however, have only two populated Color~FERET ethnicity subgroups
($n_{\text{subgroups}}{=}2$ in the disparity table, for every anchor), so their skin-tone
disparities are not comparable to the full-coverage codecs and we omit them from the tone
discussion.

The tone distribution these subgroups are cut from is plotted in
Figure~\ref{fig:ds-kk-mst}: the dark tertile (MST9--MST10) is the minority of the set,
which is why its subgroup EERs carry the widest confidence intervals here.

\subsection{Pose, age, and gender}
\tldr{Profile pose is by far the biggest within-attribute swing on Color~FERET; sex shows a small, stable female-over-male EER gap in the same direction on both datasets, much narrower on AI-Solutions-KK; age effects are modest and do not reorder under compression.}

\textbf{Pose (Color~FERET).} Pose is the largest non-codec driver of EER. At baseline the
profile subgroup is already the hardest (\texttt{edgeface\_xs} 0.67\% vs.\ $\sim$0\% frontal),
and compression amplifies exactly this subgroup: profile EER reaches 0.92\% (WebP), 0.94\%
(AVIF), 1.41\% (JPEG) and 5.18\% (JPEG~2000), whereas frontal stays at or below 0.03\% for the
strong codecs. The same profile-is-worst pattern holds on the other anchors (e.g.\ JPEG~2000
profile reaches 3.60\% for \texttt{arcface\_antelopev2}, 4.37\% for \texttt{topofr\_r100} and
4.25\% for \texttt{lvface\_l}). One anchor-specific degeneracy is worth flagging: for
\texttt{lvface\_l} the JPEG-FzT pose disparity collapses to a 0.0--0.0\% range across all four
pose subgroups (\texttt{eer\_std}${=}0.0$), which reflects sparse/degenerate scoring rather
than perfect pose invariance and should not be read as a positive result; the same JPEG-FzT
pose disparity is a normal 1.10\,pp for \texttt{edgeface\_xs} and 0.55\,pp for
\texttt{arcface\_antelopev2}.

\textbf{Gender.} Sex shows a small but consistent female-over-male EER gap that compression
does not invert. It holds in $42$ of the $44$ AI-Solutions-KK codec$\times$anchor cells as
well as on Color~FERET, though it is far smaller in the wild --- under $0.7$\,pp for every
codec but JPEG~2000, and reversed by $0.02$\,pp at the one \texttt{edgeface\_xs} baseline
we tabulate. On Color~FERET (\texttt{edgeface\_xs}) the uncompressed gap is 0.33\% (male)
vs.\ 0.82\% (female); under compression both rise but the ordering holds (e.g.\ JPEG~2000
5.08\% male vs.\ 8.92\% female; WebP 0.54\% vs.\ 1.20\%). On AI-Solutions-KK the baseline
sexes are nearly identical (1.73\% F vs.\ 1.75\% M) and compression keeps them within
0.7\,pp of each other for every codec but JPEG~2000 (e.g.\ WebP 2.90\% F vs.\ 2.84\% M,
JPEG 4.73\% vs.\ 4.04\%; JPEG~2000 is the exception at 17.61\% F vs.\ 13.71\% M), so the KK
gender gap is essentially compression-neutral outside JPEG~2000. The learned codec preserves
the Color~FERET direction rather than altering it:
on Color~FERET (\texttt{edgeface\_xs}) Ours-ACCURATE shows a female-over-male split of 1.10\%
vs.\ 0.49\%, so the learned codec neither introduces nor removes the gender gap. This sex gap
is a separate demographic axis and does not by itself indicate a skin-tone- or
brightness-linked bias.

\textbf{Age.} Age effects are modest on both datasets and the subgroup ordering is stable
under compression. On AI-Solutions-KK (\texttt{edgeface\_xs}, the three populated bins
$\le$25 / 26--35 / 36--50) the 36--50 bin is consistently the easiest (baseline 1.45\%, WebP
2.13\%, JPEG~2000 12.12\%) and the $\le$25 bin the hardest (baseline 1.72\%, WebP 2.74\%,
JPEG~2000 17.20\%); the 26--35 bin sits between. On Color~FERET the spread across age bins is
similar in scale to the skin-tone spread and again does not reorder under compression. The
high-age Color~FERET cells are sparse and should be read with care: the $>$65 bin has only
$n_{\text{neg}}{=}26$ (its EER is reported as 0.0 for every codec except JPEG~2000, at
4.21\%), and the 51--65 bin ($n_{\text{neg}}{=}9120$) is itself an order of magnitude smaller
than the working bins, so neither supports a firm conclusion.

\subsection{Disparity summary}
\tldr{The single fairness number per codec is the max$-$min subgroup EER gap. JPEG~2000 has by far the worst disparity on every attribute and both datasets; WebP, AVIF and HEIF stay closest to the uncompressed level, though all three still widen it.}

We summarise fairness with the disparity $\Delta = \text{EER}_{\max}-\text{EER}_{\min}$ over
the populated subgroups of each attribute, read directly from the disparity CSVs. A codec
that ``widens demographic gaps'' shows a larger $\Delta$ than the uncompressed baseline.
Table~\ref{tab:fairness-disparity-kk} collects the AI-Solutions-KK Monk skin-tone $\Delta$
for all four recognition anchors -- the most populated, most demographically meaningful axis --
and Table~\ref{tab:fairness-disparity-cf} gives the Color~FERET ethnicity and pose
disparities for the \texttt{edgeface\_xs} anchor with the relevant coverage caveats.

\begin{table}[t]
  \centering
  \small
  \caption{AI-Solutions-KK Monk skin-tone disparity
  $\Delta=\text{EER}_{\max}-\text{EER}_{\min}$ (percentage points) at 112\,px, 1024\,B,
  for all four recognition anchors. ``base'' is uncompressed
  \texttt{aligned}. \textbf{Refreshed under the promoted $3$M-step ACCURATE checkpoint}: every row (base,
  classical, JPEG-AI, and both Ours variants) is computed on one uniform 5-subgroup basis
  (MST5/6/7/9/10), so all $\Delta$ are directly comparable and every cell, including
  Ours-ACCURATE\,$\times$\,TopoFR, is populated. MST8 is excluded from \emph{every} row for
  support, not coverage: it spans $4$ identities and $274$ within-group impostor pairs, so its
  per-cell EER is the noisiest in the set. These are the same values as the bracketed
  $1024$\,B column of Table~\ref{tab:fairness-512}, on the same basis.
  $^{\S}$The \texttt{edgeface} column for the Ours rows is confounded --- the Ours identity
  side-stream is EdgeFace-anchored, so an EdgeFace-family score is not independent --- hence
  the \texttt{arcface}/\texttt{topofr}/\texttt{lvface} columns are the ones to trust for the
  Ours rows.}
  \label{tab:fairness-disparity-kk}
  \adjustbox{max width=\textwidth}{%
  \begin{tabular}{lrrrr}
    \toprule
    Codec & arcface & edgeface & lvface & topofr \\
    \midrule
    base (aligned)     & 0.37 & 1.53 & 0.33 & 0.26 \\
    WebP               & 0.90 & 2.30 & 0.52 & 0.47 \\
    AVIF               & 1.17 & 2.58 & 0.81 & 0.68 \\
    HEIF               & 1.53 & 2.95 & 0.97 & 0.90 \\
    JPEG-AI            & 0.72 & 1.90 & 0.63 & 0.31 \\
    JPEG               & 1.67 & 3.13 & 1.09 & 1.12 \\
    JPEG~XL            & 1.61 & 2.94 & 1.04 & 1.08 \\
    JPEG-FzT           & 1.56 & 2.81 & 0.96 & 1.12 \\
    JPEG~2000          & 5.08 & 6.49 & 3.74 & 5.22 \\
    Ours-ACCURATE$^{\S}$ & 0.75 & 1.71 & 0.70 & 0.41 \\
    Ours-FAST$^{\S}$     & 1.58 & 2.32 & 1.10 & 1.26 \\
    \bottomrule
  \end{tabular}}
\end{table}

\begin{table}[t]
  \centering
  \small
  \caption{Color~FERET disparity $\Delta=\text{EER}_{\max}-\text{EER}_{\min}$ (percentage
  points) at 112\,px, 1024\,B, \textbf{reported across all four recognition anchors}. Pose
  spans its full four subgroups; ethnicity is computed over the four well-populated cohorts
  (Asian/Black/Hispanic/White) for every row, baseline included --- the three thin labels
  (Native-American $n_{\text{neg}}{=}8$, Other $22$, Pacific-Islander $319$) are dropped for
  want of impostor support rather than reported as a rate, so every ethnicity row sits on one
  uniform basis and the column is a like-for-like comparison.
  ``base'' is uncompressed. $^{\S}$For the Ours rows the \texttt{edgeface} column is
  confounded (the identity side-stream is EdgeFace-anchored), so the
  \texttt{arcface}/\texttt{lvface}/\texttt{topofr} columns are the ones to trust ---
  these are the independent-anchor ethnicity disparities for the learned codec.
  $^{\ddagger}$ marks a cell computed over fewer cohorts than its block nominal; on the current
  four-cohort ethnicity basis no cell is so restricted.
  \colorbox{green!25}{Lowest}/\colorbox{red!22}{highest} per column, as in the other tables, ranked \emph{within} each block (ethnicity and pose are
  not comparable to each other); ``base (aligned)'' does not compete.}
  \label{tab:fairness-disparity-cf}
  \adjustbox{max width=\textwidth}{%
\input{tables/fairness_disparity_cf.tex}}
\end{table}

Three findings stand out, and all four anchors agree on them. First, JPEG~2000 is the clear
fairness loser: its Monk skin-tone disparity on AI-Solutions-KK is 3.74--6.49\,pp across the
four anchors (5.08 arcface, 6.49 edgeface, 3.74 lvface, 5.22 topofr), an order of
magnitude above the same anchors' uncompressed baselines for the strong matchers, and on Color
FERET its ethnicity disparity reaches 4.66\,pp (\texttt{edgeface\_xs}; 3.31--3.96\,pp on the
independent anchors); it has the worst pose disparity too (2.89\,pp). Second, the perceptually
strong codecs (WebP, AVIF, HEIF) keep the skin-tone disparity closest to the uncompressed
level --- but every one of them still widens it, and we no longer read any codec as improving
on the baseline: on the uniform basis the KK $\Delta$ rises from base 0.37\,pp to
0.90--1.53\,pp on ArcFace (WebP 0.90, AVIF 1.17, HEIF 1.53), from 0.33 to 0.52--0.97 on
LVFace-L and from 0.26 to 0.47--0.90 on TopoFR. Third, plain JPEG and JPEG~XL sit in the
middle (KK skin-tone $\Delta$ 1.04--3.13\,pp across anchors, CF ethnicity $\Delta$ 1.44 and
1.34\,pp for \texttt{edgeface\_xs}). Every ethnicity row is now computed on the same four
well-populated cohorts (Asian/Black/Hispanic/White), so the ethnicity column is a
like-for-like comparison across the whole roster --- no codec's figure is a coverage artefact
of dropped subgroups, and the earlier $^{\ddagger}$ caveat for JPEG-FzT and JPEG-AI no longer
applies.

\paragraph{The learned codec on the robust axis (independent anchors).} Because we designate
the Color~FERET ground-truth ethnicity gap as the most trustworthy demographic signal, it
matters that the learned codec be read on that axis \emph{on an anchor independent of its
EdgeFace side-stream}. The expanded Table~\ref{tab:fairness-disparity-cf} supplies exactly
that: Ours-ACCURATE's CF ethnicity disparity is $0.42$\,pp on ArcFace, $0.49$\,pp on LVFace-L
and $0.38$\,pp on TopoFR --- inside the strong-codec band (WebP $0.32$--$0.49$, AVIF
$0.20$--$0.39$, JPEG-AI $0.26$--$0.42$, JPEG $0.48$--$0.75$) and roughly an order of magnitude
below JPEG~2000 ($3.31$--$3.96$).
JPEG-AI now belongs inside that band rather than beside it: with every ethnicity row computed
on the same four-cohort basis its cells are no longer reduced-coverage. It is the lowest of
the roster on ArcFace ($0.26$\,pp); on the other two independent anchors AVIF is lowest
($0.39$ and $0.20$\,pp), so no single codec owns this axis.
The $0.89$\,pp figure in the \texttt{edgeface\_xs} column is the
\emph{confounded} reading --- EdgeFace is the family the side-stream is anchored to --- and is
not a valid fairness number for our codec. On the independent anchors the honest ranking is
mid-band rather than best: Ours-ACCURATE sits a little \emph{above} WebP ($0.32$\,pp on
ArcFace), AVIF and JPEG-AI, and below JPEG and JPEG~2000. The learned codec therefore adds no
unusual ethnicity disparity, but on this axis it does not lead either. (Ours-FAST is higher
again, $0.41$--$0.76$\,pp on the independent anchors, still inside the band.)

\begin{figure}[t]
  \centering
  \includegraphics[width=\linewidth]{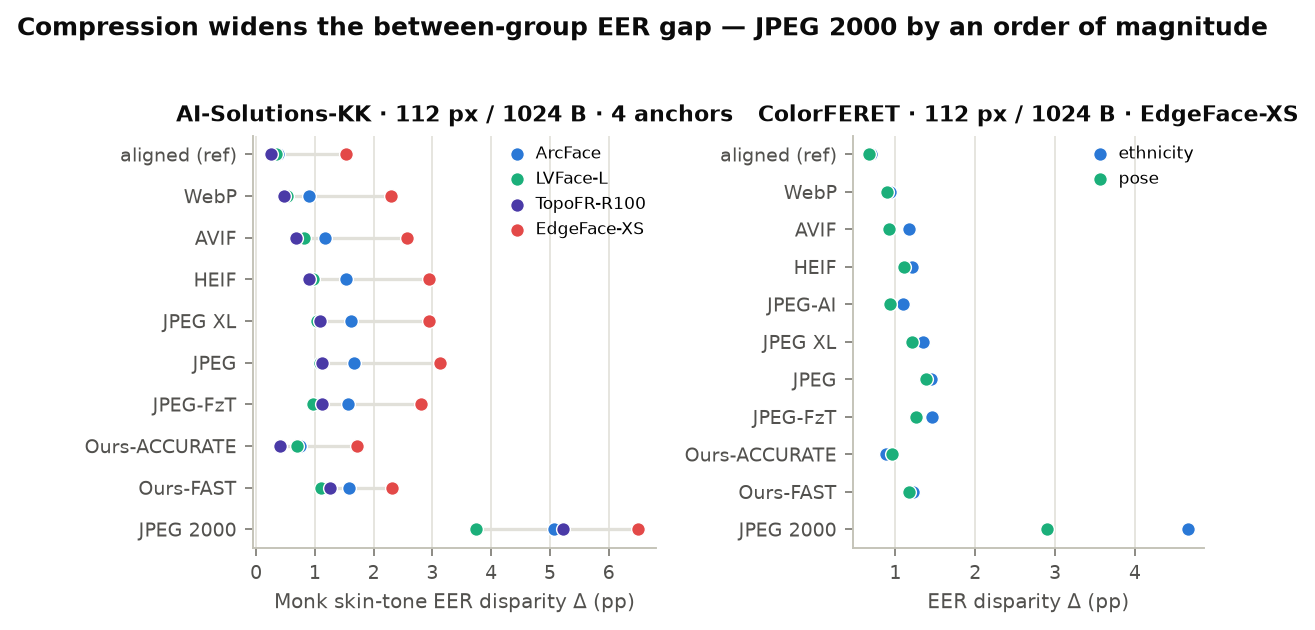}
  \caption{Between-group EER disparity $\Delta=\text{EER}_{\max}-\text{EER}_{\min}$
  per codec at 112\,px / 1024\,B. \emph{Left:} AI-Solutions-KK Monk skin tone, one
  marker per anchor matcher, recomputed on the uniform MST5/6/7/9/10 basis (MST8
  excluded for every codec as the thinnest cohort, $4$ identities; all
  four anchors including Ours-ACCURATE\,$\times$\,TopoFR-R100 are populated).
  \emph{Right:} Color~FERET ethnicity and pose disparities for the
  \texttt{edgeface\_xs} anchor, every row on the same four well-populated ethnicity
  cohorts. Every codec except JPEG~2000 stays within a few
  percentage points of the uncompressed baseline (top row).}
  \label{fig:fairness-disparity}
\end{figure}

\paragraph{Uniform-basis comparison.} Because the MST8 coverage difference confounds a
direct reading of the disparity tables (the classical codecs lose the MST8 cell under
compression while the Ours variants retain it), Figure~\ref{fig:fairness-disparity}
recomputes the KK Monk skin-tone disparity on a \emph{uniform} subgroup basis ---
MST5/6/7/9/10, i.e.\ excluding MST8 for every codec and the baseline alike --- so that
every marker in the panel spans the same subgroups. On this apples-to-apples basis the
qualitative conclusion sharpens into a quantitative one: relative to each anchor's own
uncompressed baseline, JPEG~2000 amplifies the skin-tone disparity by a factor of
$13.7\times$ (ArcFace: $0.37\rightarrow5.08$\,pp), $11.2\times$ (LVFace-L:
$0.33\rightarrow3.74$\,pp), and $20.0\times$ (TopoFR-R100: $0.26\rightarrow5.22$\,pp),
whereas every other codec --- including both Ours variants --- keeps its disparity in a
$0.3$--$3.1$\,pp band ($0.3$--$1.7$ on the three independent anchors; e.g.\ ArcFace: WebP $0.90$\,pp; Ours-ACCURATE is in
fact the \emph{second-lowest}-disparity codec at $0.75$\,pp, behind only JPEG-AI
($0.72$\,pp), and stays in the strong-codec band on every independent anchor). The learned codec therefore neither introduces nor removes a skin-tone
disparity at the kilobyte budget; the fairness risk at this operating point is
concentrated entirely in the choice of codec, with JPEG~2000 the single outlier. The
Color~FERET panel of the same figure shows the ethnicity and pose disparities for the
tabulated \texttt{edgeface\_xs} anchor, where JPEG~2000 is again the extreme point
($4.66$\,pp ethnicity) and all remaining codecs cluster within $0.9$--$1.5$\,pp.

\paragraph{How precise are these amplification factors?} The multiplicative amplification is
a ratio of two small subgroup-EER spreads estimated from only ${\sim}105$ KK identities, so it
comes with real uncertainty. We therefore attach a \emph{subject-level cluster bootstrap}
(resample identities with replacement, $B{=}500$; a pair is kept only if both its subjects are
drawn) to the uniform-basis disparity and its amplification ratio, per anchor
(Table~\ref{tab:disparity-ci}; computed on a reduced $400$\,k-impostor sample for tractability,
so the point ratios run slightly lower than the $5$\,M-impostor Table~\ref{tab:fairness-disparity-kk}
figures but agree in order of magnitude). Two things hold. First, \textbf{the JPEG~2000
amplification is statistically real}: its ratio 95\% CI excludes $1\times$ on \emph{every}
anchor (lower bounds $1.5$--$4.5\times$), so JPEG~2000 genuinely amplifies the skin-tone gap.
Second, \textbf{the exact multiplier is imprecise}: the CIs are wide (e.g.\ ArcFace $12.7\times$
with CI $[4.5,46.6]$, TopoFR-R100 $5.6\times$ $[3.3,38.8]$), so the headline ``$11.2$--$20.0\times$''
should be read as a point estimate of a \emph{large but imprecise} effect, not a sharp
multiplier. By contrast Ours-ACCURATE's amplification is small and tight (ratio $1.1$--$1.8\times$,
all CI upper bounds $\le 3.3\times$) and WebP's CI includes $1\times$ --- i.e.\ neither
significantly amplifies the disparity. This confirms the qualitative reading (JPEG~2000 the
outlier, the learned codec benign) while being honest about the precision of the multiplier.

\begin{table}[t]
  \centering \small
  \caption{Subject-level cluster-bootstrap 95\% CIs on the KK Monk skin-tone amplification
  ratio (uniform MST5/6/7/9/10 basis, ratio $=\Delta_{\text{codec}}/\Delta_{\text{aligned}}$),
  per anchor; $B{=}500$ resamples over the ${\sim}105$ KK identities, $400$\,k-impostor sample.
  JPEG~2000's CI excludes $1\times$ on every anchor (real amplification) but is wide (imprecise
  multiplier); Ours-ACCURATE is small with upper bound $\le 3.3\times$ and WebP's CI includes
  $1\times$. \colorbox{green!25}{Lowest}/\colorbox{red!22}{highest} per column, as in the
  other tables. The first row is \emph{not} a ratio: it gives the uncompressed reference
  disparity $\Delta_{\text{aligned}}$ itself in percentage points --- the denominator of
  every ratio below it --- so it carries no CI and does not compete.}
  \label{tab:disparity-ci}
  \adjustbox{max width=\textwidth}{%
  \input{tables/disparity_ci_kk.tex}}
\end{table}

\subsection{Fairness at the tighter 512-byte budget}
\label{subsec:fairness-512}
\tldr{The deployment claim for the learned codec rests on the 512\,B operating point, so we
repeat the disparity measure there. The gap the classical codecs open up widens sharply at
512\,B --- ethnicity disparities of $7.5$--$12.3$\,pp for JPEG/HEIF, and up to $5.8$\,pp for
the same two on KK skin-tone (JPEG~2000 reaches $7.5$\,pp there) --- while Ours-ACCURATE
stays at the low end on the independent anchors
($0.9$--$1.2$\,pp KK, $0.4$--$0.5$\,pp CF ethnicity) and JPEG~2000 remains the worst codec on
KK skin tone (legacy JPEG overtakes it on both Color~FERET axes, on every anchor). The
learned codec's directional brightness bias does \emph{not} grow at
512\,B (dark/light luma split $+1.24\%/{+}0.22\%$ vs.\ $+1.28\%/{+}0.23\%$ at 1024\,B), so
the watch item stays sub-threshold at the operating point the codec is sold on.}

Every disparity table above is read at the 1024-byte budget. Because Table~\ref{tab:deployment-guidance}
recommends the learned codec specifically for the hard 512-byte point, we recompute the same
$\Delta=\text{EER}_{\max}-\text{EER}_{\min}$ disparity there, on the identical subgroup bases
(the uniform MST8-excluded five-subgroup basis for KK skin tone; for Color~FERET, the full
four-subgroup pose set and the four well-populated ethnicity cohorts, applied to the
baseline row as well as to every codec).
Table~\ref{tab:fairness-512} reports it for all four anchors, with the 1024\,B
value in brackets for direct comparison.

\begin{figure}[t]
  \centering
  \includegraphics[width=\linewidth]{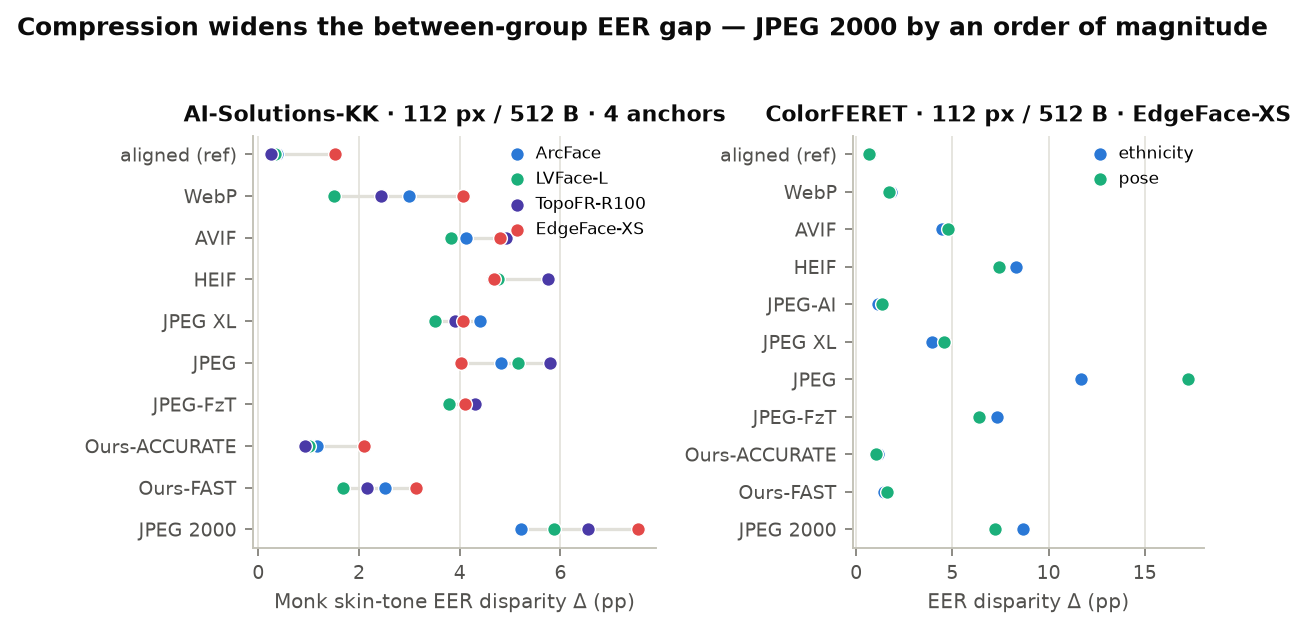}
  \caption{Between-group EER disparity $\Delta=\text{EER}_{\max}-\text{EER}_{\min}$ per codec
  at the \textbf{512-byte} budget, $112$\,px (the 512\,B analogue of
  Figure~\ref{fig:fairness-disparity}). \emph{Left:} AI-Solutions-KK Monk skin tone, one
  marker per anchor, uniform MST5/6/7/9/10 basis. \emph{Right:} Color~FERET ethnicity and pose
  for \texttt{edgeface\_xs}. Halving the budget fans the classical codecs out --- legacy JPEG
  into double digits on both Color~FERET attributes --- while Ours-ACCURATE and the aligned
  baseline stay clustered at the low end. JPEG~2000 remains the extreme outlier on KK skin
  tone, but on Color~FERET legacy JPEG overtakes it on every anchor.}
  \label{fig:fairness-disparity-512}
\end{figure}

Two things stand out. First, \textbf{halving the budget widens the between-group gap, and it
does so overwhelmingly for the codecs that were already weakest}
(Figure~\ref{fig:fairness-disparity-512}): on Color~FERET ethnicity ---
now measured over the four well-populated cohorts (Asian/Black/Hispanic/White), the only ones
carrying enough impostor support to estimate a rate --- the disparity climbs to
$8.3$--$12.3$\,pp for legacy JPEG and $7.5$--$8.4$\,pp for HEIF (from $0.2$--$1.4$\,pp
at 1024\,B), and on KK skin tone HEIF and JPEG reach $4.0$--$5.8$\,pp (from $0.9$--$3.1$\,pp).
JPEG~2000 stays the worst codec at 512\,B on KK skin tone (all four anchors) and is
second-worst on Color~FERET ethnicity on all four, its own disparities spanning
$5.2$--$13.8$\,pp across the three blocks; but on \emph{both} Color~FERET attributes legacy
JPEG overtakes it on every anchor ($30.07$\,pp vs.\ JPEG~2000's $11.42$\,pp on ArcFace
pose). Second,
\textbf{Ours-ACCURATE is the lowest-disparity codec at 512\,B on KK skin tone} on the anchors
independent of its side-stream: $0.93$--$1.17$\,pp (ArcFace/LVFace/TopoFR), and it stays at the
low end on Color~FERET too --- $0.4$--$0.5$\,pp on CF ethnicity and $0.4$--$0.6$\,pp on CF pose,
close to the uncompressed baseline and below WebP ($0.5$--$0.7$\,pp on CF ethnicity): on those
three anchors it is the lowest-disparity codec in the table on both Color~FERET attributes. The
fairness advantage the learned codec shows at 1024\,B therefore \emph{grows} at the budget it is
recommended for, rather than eroding.

This directly addresses the §\ref{subsec:fairness-fmr} watch item that the learned codec's
directional brightness bias ``could become a fairness problem at the tighter 512-byte budget.''
It does not: the Ours-ACCURATE per-tertile luma error on KK at
$112$\,px is $+1.24\%$ (dark) / $+0.22\%$ (light) at $512$\,B, essentially unchanged from
$+1.28\%$/$+0.23\%$ at $1024$\,B --- the tone-dependent shift is stable across budget, and the
subgroup EER disparity above confirms it has not translated into a verification gap even at
$512$\,B. We also recomputed the \emph{differential-FMR} view of §\ref{subsec:fairness-fmr}
at $512$\,B (imposter-acceptance disparity, ArcFace ethnicity): unlike the EER disparity, the
FMR disparity does \emph{not} blow up at the tighter budget --- at FMR$=10^{-2}$ it stays in
a narrow codec-invariant $1.7$--$2.9$\,pp band, comparable to the $2.3$--$3.6$\,pp ArcFace band
at $1024$\,B (Ours-ACCURATE $2.63$\,pp, WebP
$2.56$, AVIF $2.73$; JPEG~2000 $1.81$, again \emph{not} an outlier on FMR), and at
FMR$=10^{-3}$ every codec is under $0.64$\,pp on ArcFace and never above $0.75$\,pp on any of
the four anchors. So neither the EER-disparity picture nor the
security-relevant FMR-disparity picture surfaces a learned-codec fairness problem at $512$\,B.

\begin{table}[t]
  \centering
  \small
  \caption{Between-group EER disparity $\Delta=\text{EER}_{\max}-\text{EER}_{\min}$
  (percentage points) at the \textbf{512-byte} budget, $112$\,px, for the four anchors; the
  bracketed value is the same $\Delta$ at $1024$\,B, recomputed in the same pass and on the
  identical subgroup basis (cf.\ Tables~\ref{tab:fairness-disparity-kk}
  and~\ref{tab:fairness-disparity-cf}), for direct comparison.
  \shadelegend{} applied to the $512$\,B value within each block; the bracketed
  $1024$\,B value is not shaded. The uncompressed
  \emph{base} row is the reference, not a competitor. KK skin tone uses the uniform
  MST8-excluded five-subgroup basis and Color~FERET pose its full four-subgroup set. The
  Color~FERET ethnicity block is computed over the four well-populated cohorts
  (Asian/Black/Hispanic/White) for \emph{every} row, baseline included, as its header
  states and as Table~\ref{tab:fairness-disparity-cf} does at $1024$\,B. The three thin
  labels (Native-American $n_{\text{neg}}{=}8$, Other $22$, Pacific-Islander $319$) are
  dropped for want of impostor support rather than reported as a rate: a max$-$min taken
  over an $8$-impostor cell is not a disparity measurement. Every ethnicity figure here is
  therefore on one uniform basis and comparable across rows and budgets.
  $^{\S}$For the Ours rows the \texttt{edgeface} column is confounded (EdgeFace-anchored
  side-stream); the \texttt{arcface}/\texttt{lvface}/\texttt{topofr} columns are the ones to
  trust.}
  \label{tab:fairness-512}
  \adjustbox{max width=\textwidth}{%
  \input{tables/fairness_disparity_512.tex}}
\end{table}

\subsection{Differential false-match rate (imposter acceptance)}
\label{subsec:fairness-fmr}
\tldr{The disparity tables above are EER/FNMR-based (genuine rejection). The security-relevant
NIST axis~\cite{grother2019frvt} is the \emph{false-match} rate --- imposter acceptance at a
fixed system threshold. On the Color~FERET ethnicity axis (the robust signal) the differential
FMR at a $10^{-2}$ operating point is \textbf{$\sim$$0.7$--$3.6$\,pp for \emph{every} codec,
including the uncompressed baseline}: compression --- ours included --- barely moves it, and,
strikingly, \textbf{JPEG~2000 is \emph{not} an FMR outlier} ($1.4$--$2.6$\,pp), even though it
is the extreme EER-disparity codec. Its fairness problem is therefore false-\emph{non}-match
(genuine rejection), not imposter acceptance.}

We fix a global threshold $\tau$ at a target overall FMR of $10^{-2}$ over each attribute's
within-group impostor pool, then read the per-subgroup FMR$(g)=\Pr[\text{score}\ge\tau\mid
\text{both ends in }g]$ and its max$-$min spread. Table~\ref{tab:fmr-fairness-cf} reports the
Color~FERET ethnicity disparity over the four well-populated cohorts (Asian/Black/Hispanic/White,
each with $\ge 748$ impostor pairs; the rare cohorts are dropped for FMR stability). Two
findings stand out. First, the FMR disparity is essentially \emph{codec-invariant}: every
codec, and the uncompressed baseline, sits in a $0.7$--$3.6$\,pp band across the four anchors,
with Ours-ACCURATE ($1.46$--$2.81$\,pp) indistinguishable from the baseline ($1.13$--$2.87$) and
WebP ($1.74$--$3.17$). The disparity is a property of the matcher$\times$population (the Black
cohort carries the highest FMR under every codec, $\sim$$3.3$--$4.3\%$ vs $\sim$$0.7$--$0.8\%$ for
White on ArcFace), not of the codec. Second, and importantly, \textbf{JPEG~2000 --- the
$3.3$--$4.7$\,pp EER-disparity outlier --- is unremarkable on FMR} ($1.4$--$2.6$\,pp, mid-pack),
which localizes its demographic risk to genuine rejection rather than imposter acceptance. On
AI-Solutions-KK the sparse Monk tones are too small ($4$--$7$ identities each; $274$
within-group impostor pairs for MST8 and $413$ for MST10) to resolve an FMR at the $10^{-2}$
operating point: those two cells quantise in steps of $0.37$ and $0.24$\,pp respectively and
land on exactly zero in $40$ of their $168$ codec$\times$anchor$\times$budget cells, so a KK
disparity would be a difference between two coarsely quantised rates. We therefore do not
report a KK differential FMR and treat
Color~FERET ethnicity as the trustworthy FMR axis, consistent with the rest of this section.

\begin{table}[t]
  \centering
  \small
  \caption{Color~FERET \textbf{differential false-match rate} (imposter acceptance)
  $\Delta\text{FMR}=\text{FMR}_{\max}-\text{FMR}_{\min}$ (percentage points) across the four
  well-populated ethnicity cohorts, at a global threshold set to overall FMR $=10^{-2}$,
  112\,px/1024\,B. ``base'' is uncompressed. $^{\S}$EdgeFace is confounded for the Ours rows
  (side-stream anchor). The roster is shorter than Table~\ref{tab:fairness-disparity-cf}'s: a
  codec whose impostor scores admit no threshold at this target --- every subgroup FMR
  quantises to zero, so the disparity is identically $0.00$\,pp --- is omitted rather than
  reported as perfectly fair, which drops JPEG-AI and the CompressAI baselines here.
  Every codec --- and the baseline --- sits in the same $0.7$--$3.6$\,pp
  band; unlike the EER disparity, JPEG~2000 is not an outlier here. \colorbox{green!25}{Lowest}/\colorbox{red!22}{highest} per column, as in the other tables; ``base''
  does not compete.}
  \label{tab:fmr-fairness-cf}
  \adjustbox{max width=\textwidth}{%
\input{tables/fmr_fairness_cf.tex}}
\end{table}

\paragraph{Scope note: chroma-subsampling bias.} One planned fairness lever is not
exercised here: the effect of chroma subsampling (4:2:0 vs.\ 4:4:4) on the demographic
gap. Every codec is run with its own default chroma handling at the byte budget, so the
chroma axis is confounded with codec choice and we do not isolate it; a controlled
4:2:0-vs-4:4:4 comparison on a fixed codec is left to future work. The disparities
reported here should therefore be read as codec-choice effects, not chroma effects.

\paragraph{Watch item: learned-codec brightness bias.}
The learned codec is scored in the subgroup fairness tables above, but its own
verification check (Section~\ref{sec:codec}) additionally
quantifies a mild, directional skin-tone brightness bias in the reconstruction. For the Ours-ACCURATE variant on
AI-Solutions-KK at 112\,px / 1024\,B, the mean luma error (\texttt{bright\_delta}) is
$+$1.28\% on the dark tertile and $+$0.71\% on the mid tertile but only $+$0.23\% on the light
tertile -- i.e.\ every tertile is reconstructed slightly brighter, and monotonically more so
the darker the tone, a directional tone shift of order $\sim$1\,pp of luma between the dark
and light tertiles. The pattern is stable across budget and resolution (at 224\,px / 512\,B
the dark/light split is $+$1.08\%/$-$0.24\%), and the Ours-FAST variant shows a much smaller
shift ($-$0.04\% dark vs.\ $-$0.21\% light at 112\,px / 1024\,B). Every \texttt{codec\_verify}
row is flagged \texttt{ok} and the shift is below the level that moved verification EER in our
checks, but because it is directional it is exactly the kind of artefact that could become a
fairness problem at the tighter 512-byte budget or with a more tone-sensitive matcher, so we
flag it as a watch item for the learned variants rather than a confirmed disparity. The
subgroup fairness EERs available for both variants (tables above) confirm it stays
sub-threshold at the 1024-byte budget: the learned codec's skin-tone spread sits inside the
strong-codec band and follows the same tone ordering, so the brightness shift has not yet
translated into a subgroup-EER disparity at this budget.

Overall, compression to the 1024-byte budget does not reorder which demographic groups are
hardest -- the pose, tone, sex and age orderings from the uncompressed baseline survive --
but it does inflate the absolute gap, and the size of that inflation is a codec-choice
question: negligible-to-mild for WebP/AVIF/HEIF/JPEG-AI, moderate for JPEG and JPEG~XL, and
severe for JPEG~2000.

%% file: sections/10_recompression.tex
\section{Compressed-on-Compressed Recompression}
\label{sec:recompression}
\tldr{Faces in real document pipelines are often re-encoded after they have already been
compressed once; we measure how a second pass through a different codec moves verification
EER on both Color~FERET and the in-the-wild AI-Solutions-KK set, and find that most chains are
benign but a few (notably JPEG~XL re-encoded as HEIF or AVIF, and anything routed through
JPEG~2000) sharply degrade identity accuracy on both datasets.}

A compressed face crop rarely stays in its original container. Enrolment portals,
document-issuance systems, and messaging gateways routinely decode an already-compressed
image and re-encode it under a different codec or a different byte budget. Each such pass
is lossy, and the second encoder must spend its budget reconstructing the
compression artifacts of the first rather than the original pixels. This section quantifies
the resulting identity cost as a \emph{delta-EER}: the change in equal-error rate of a
two-codec chain relative to compressing once with the second codec alone.

\subsection{Setup}
\label{subsec:recomp-setup}
\tldr{We chain every ordered pair of eight codecs (six standard codecs, the JPEG-FzT
reference codec, and the learned Ours-ACCURATE) at the 112\,px working resolution and the
1024-byte budget on Color~FERET,
then report each chain's EER and its delta against the corresponding single-pass baseline;
JPEG-AI and the neural baselines are excluded (their per-image reference decoders make the
full matrix intractable).}

We take the six classical/standard codecs from the benchmark --- JPEG~\cite{wallace1992jpeg},
JPEG~2000~\cite{taubman2002jpeg2000}, WebP~\cite{google2010webp},
JPEG~XL~\cite{alakuijala2019jpegxl}, AVIF~\cite{aomavif}, and HEIF~\cite{sullivan2012hevc}
--- together with the JPEG-FzT codec~\cite{perfilieva2021ftransform} \emph{and} our
learned Ours-ACCURATE codec, and form all $8\times8$ ordered
source/second pairs, including the diagonal where a codec re-encodes its own output.
Ours-ACCURATE is included as both a source and a second pass (its GPU encode/decode is fast
enough to run the full matrix); the JPEG-AI reference codec and the CompressAI neural
baselines are \emph{excluded} from the recompression matrix because their per-image
reference decoders make a full $8\times8\times$(image-sample) chain prohibitively slow on the
shared hardware --- their single-pass behaviour is already characterised in
Section~\ref{sec:benchmark}. Both passes target the 1024-byte budget at 112\,px, the
verification working resolution. Identity is scored with the lightweight
\texttt{edgeface\_xs} anchor~\cite{george2024edgeface} on
Color~FERET~\cite{phillips2000feret}. For each chain we report the absolute EER and the
delta-EER against the single-pass baseline of the \emph{second} codec, so a positive delta
isolates the damage attributable to having a first encoder in the loop. We restrict the
analysis here to the 112\,px/1024-byte operating point, and we repeat the full $8\times8$ grid
on the in-the-wild AI-Solutions-KK set (Table~\ref{tab:recompression-eer-kk}) to check whether
the same failure pattern generalizes beyond the studio-quality Color~FERET portraits. This
codec-pair chain is a deliberate substitution for the originally-specified recompression
protocol, which fixed a raw source and varied a JPEG source quality factor
(QF~$\in\{95,75,50\}$) before the 1\,kB re-encode. The design here is more
codec-agnostic---it measures the penalty of any first codec, not only a JPEG source---but
it consequently does not include a raw-source or source-QF axis, and it is reported for a
single anchor matcher at 112\,px; the raw-vs-pre-compressed and source-QF questions are
left to future work. Because every chain here is scored with the single
\texttt{edgeface\_xs} anchor, the Ours-ACCURATE rows carry an additional confound:
Ours-ACCURATE's identity side-stream is anchored to a frozen EdgeFace-S embedding, so any
EdgeFace-family matcher---\texttt{edgeface\_xs} included---is not independent of it. We
therefore rescored the matrix on an independent matcher: for the \emph{classical} $7\times7$
grid the ArcFace rescore reproduces the benign/catastrophic pattern
(\S\ref{subsec:recomp-worst}), and the \emph{Ours} rows are now rescored on ArcFace too (at
$1024$\,B), which \emph{confirms} rather than overturns the EdgeFace reading ---
Ours-ACCURATE$\rightarrow$Ours-ACCURATE is $0.44\%$ on ArcFace, inside the benign cluster. So the
modern-to-modern-is-benign conclusion is established across independent matchers for both the
classical and the learned codecs; it remains verified only for pre-compressed (not raw)
sources.

\subsection{Delta-EER results}
\tldr{Same-codec and modern-to-modern chains lose almost nothing (delta-EER under
$\sim$1\,pp), but two JPEG~XL chains and every chain that touches JPEG~2000 inflate EER by
several percentage points or more.}

\begin{figure}[t]
  \centering
  \includegraphics[width=\linewidth]{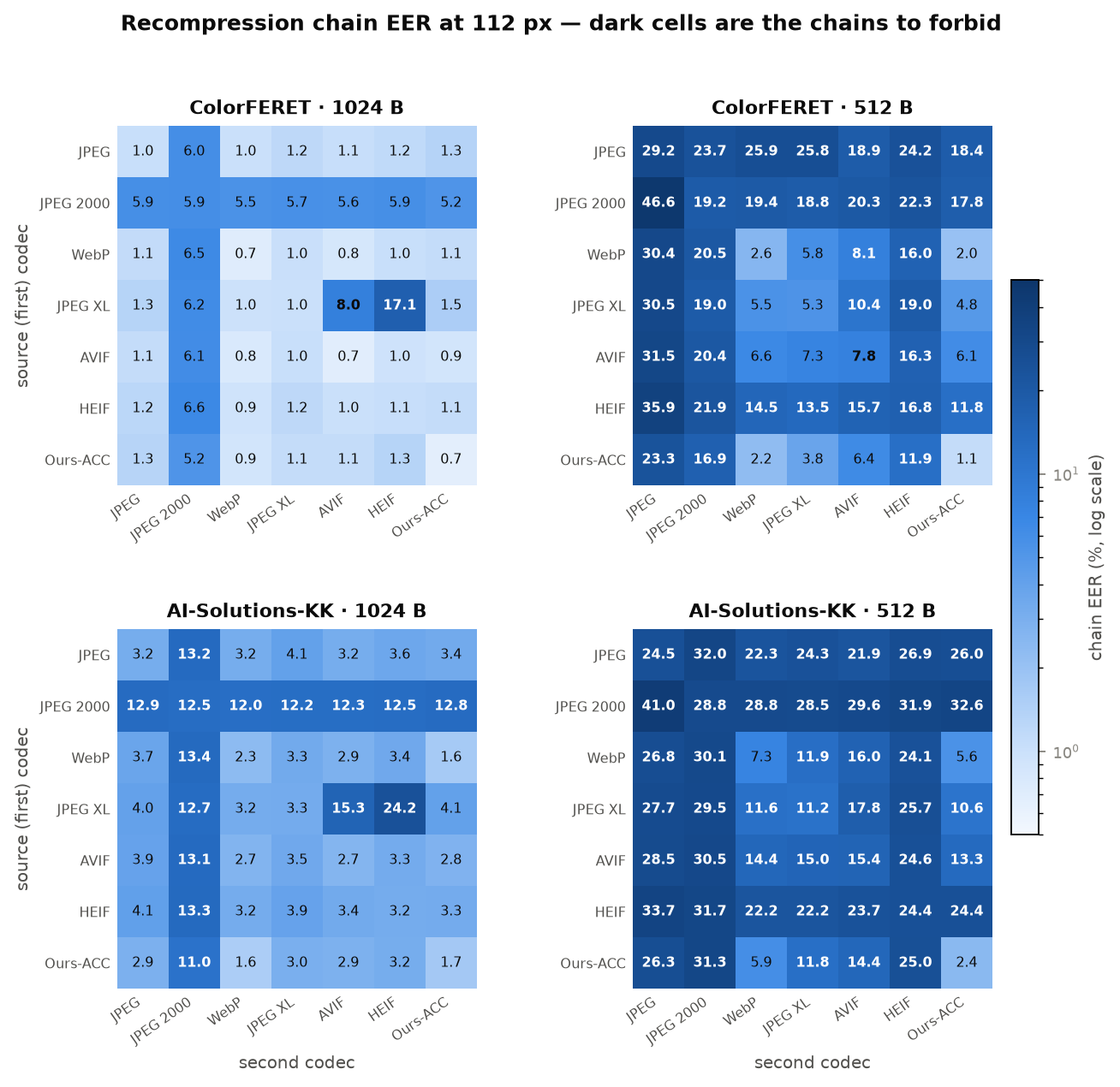}
  \caption{The four recompression matrices (both datasets $\times$ both budgets) as
  annotated heatmaps; cell values are chain EER (\%) on a logarithmic colour scale.
  Three structures recur: the JPEG~2000 \emph{column} (routing any face through a
  second JPEG~2000 pass) and \emph{row} (a JPEG~2000 source) form a uniformly dark
  band; the two isolated dark cells JPEG~XL$\rightarrow$AVIF and
  JPEG~XL$\rightarrow$HEIF are the codec-pair-specific failures; and the remaining
  modern-to-modern region stays light. At 512\,B the whole matrix darkens and the
  JPEG~2000$\rightarrow$JPEG corner becomes the single worst cell on both datasets.}
  \label{fig:recompression-heatmap}
\end{figure}

Table~\ref{tab:recompression-eer-cf} reports the full matrix of chain EER (\%), with rows
indexed by the source (first) codec and columns by the second codec;
Figure~\ref{fig:recompression-heatmap} renders all four matrices (both datasets, both
budgets) as annotated heatmaps so the two failure structures --- the JPEG~2000
row/column band and the two JPEG~XL$\rightarrow$AV1/HEVC cells --- can be read at a
glance. The diagonal and the
``modern-to-modern'' off-diagonal entries (JPEG, WebP, JPEG~XL, AVIF, HEIF re-encoded into
one another) cluster tightly around the single-pass level: AVIF$\rightarrow$AVIF reaches the
lowest \emph{classical} chain EER at 0.68\,\% and WebP$\rightarrow$WebP 0.71\,\%, and the corresponding
delta-EER values stay below roughly 1\,pp. These are the safe chains: re-encoding an
already-WebP/AVIF/HEIF/JPEG face into another modern codec costs little identity signal.

Two failure modes stand out. First, \textbf{JPEG~XL re-encoded as HEIF or AVIF} collapses:
JPEG~XL$\rightarrow$HEIF reaches 17.14\,\% EER, a delta-EER of $+16.31$\,pp over single-pass
HEIF, and JPEG~XL$\rightarrow$AVIF reaches 8.00\,\% EER ($+7.38$\,pp). The JPEG~XL artifact
structure appears to interact pathologically with the HEVC/AV1 intra coders when both are
squeezed to 1\,kB. Second, \textbf{any chain that involves JPEG~2000} stays high regardless
of the partner: every classical entry in the JPEG~2000 \emph{column} sits near 6\,\% (e.g.\
WebP$\rightarrow$JPEG~2000 at 6.51\,\%, HEIF$\rightarrow$JPEG~2000 at 6.60\,\%), and every
entry in the JPEG~2000 \emph{row} sits near 5.2--5.9\,\%, with delta-EER values around
$+4.9$ to $+5.0$\,pp whenever JPEG~2000 is the first codec and a different codec follows
(e.g.\ JPEG~2000$\rightarrow$AVIF at $+5.03$\,pp, JPEG~2000$\rightarrow$WebP at $+4.91$\,pp).
JPEG~2000 is simply a poor fit for a 1\,kB face crop --- its single-pass EER is already the
worst of the group --- and chaining only compounds that.

\begin{table}[t]
  \centering
  \caption{Recompression EER (\%) on Color~FERET at 112\,px and the 1024-byte budget.
  Rows are the source (first) codec; columns are the second codec. The diagonal is
  same-codec re-encoding. Lower is better; the two highlighted JPEG~XL chains and the
  JPEG~2000 row/column are the damaging ones. Column shading marks the
  \colorbox{green!25}{best} / \colorbox{red!22}{worst} \emph{classical} source per second
  codec; the Ours-ACCURATE row and column are left unshaded because the tabulated values are
  EdgeFace-scored (\S\ref{subsec:recomp-worst}).}
  \label{tab:recompression-eer-cf}
  \adjustbox{max width=\textwidth}{\input{tables/recompression_eer_colorferet_1024.tex}}
\end{table}

\paragraph{Interpretation.} The two failure structures have different mechanisms and
therefore different remedies. The JPEG~2000 band is \emph{additive}: JPEG~2000 is
already the weakest single-pass codec at this budget (Section~\ref{sec:benchmark}),
and a second encoder cannot restore what the wavelet quantisation discarded, so every
chain inherits roughly the single-pass penalty regardless of partner ($+4.9$ to
$+5.0$\,pp when JPEG~2000 is the \emph{first} pass on Color~FERET). The remedy is exclusion:
JPEG~2000 should not appear anywhere in a sub-1\,kB face pipeline. The JPEG~XL spikes
are \emph{interactive}: JPEG~XL alone is benign (single-pass EER $0.84\%$ on
Color~FERET), and HEIF/AVIF alone are benign ($0.84\%$/$0.62\%$), yet the composition
reaches $17.14\%$ --- a ${\sim}20\times$ super-additive failure in which the VarDCT artefact
structure of JPEG~XL is apparently expensive for the HEVC/AV1 intra coders to
re-approximate at 1\,kB, leaving too few bits for the face. Because this interaction is
invisible to any single-codec benchmark, recompression matrices of this kind are a
necessary acceptance test for document pipelines that transcode between formats: a
per-codec ranking alone would have declared all three codecs safe.

\paragraph{The learned codec recompresses safely on the anchor matcher.} Adding Ours-ACCURATE to the matrix
shows the learned codec behaves like the modern transform cluster under recompression, not
like the fragile JPEG~XL or JPEG~2000 chains. Same-codec Ours-ACCURATE$\rightarrow$Ours-ACCURATE
costs $0.66\%$ EER on Color~FERET ($1.71\%$ on KK), and every
Ours-ACCURATE$\leftrightarrow$\{WebP, AVIF, HEIF, JPEG, JPEG~XL\} cross-pair stays within
about $1$--$2$\,pp of its single-pass baseline ($0.88$--$1.53\%$ on Color~FERET). The only
expensive Ours-ACCURATE chains are, once again, the JPEG~2000 ones. Two points stand out.
First, Ours-ACCURATE is \emph{among the least-damaging sources before a JPEG~2000 second
pass} ($5.23\%$, behind only JPEG-FzT at $3.93\%$, vs.\ $5.9$--$6.6\%$ for the six classical
sources at $1024$\,B on Color~FERET), because
its identity-preserving reconstruction hands JPEG~2000 a cleaner signal. Second, its
same-codec chain is \emph{budget-robust}: Ours-ACCURATE$\rightarrow$Ours-ACCURATE stays at
$1.09\%$ at $512$\,B against $0.66\%$ at $1024$\,B on Color~FERET ($2.42\%$ on KK), so a
pipeline that carries a face end-to-end in Ours-ACCURATE survives transcoding even at the
half-kilobyte budget. These Ours-ACCURATE chain EERs are measured on a $120$-image subsample ($456$ mated
pairs on Color~FERET, $540$ on KK), so differences below roughly $0.5$\,pp correspond to
single mated pairs and should not be over-read; the stable finding is the \emph{ordering} of
chains, not their exact values.

\emph{Confounded-matcher caveat.} As set out in Section~\ref{subsec:recomp-setup} and
confirmed for the classical matrix in Section~\ref{subsec:recomp-worst}, every
Ours-ACCURATE row and column here is scored with \texttt{edgeface\_xs}, whose backbone
family is shared with the codec's identity side-stream. The tabulated Ours-ACCURATE chain EERs ---
including the low Ours-ACCURATE$\rightarrow$Ours-ACCURATE diagonal, left \emph{un-highlighted}
in the matrices because the table values are EdgeFace-scored --- are now independently
\emph{confirmed} on ArcFace at $1024$\,B (Ours$\to$Ours $0.44\%$, inside the benign band;
Section~\ref{subsec:recomp-worst}), so they are no longer provisional; the \emph{classical}
matrix is likewise rescored on ArcFace and reproduces the pattern, so the
modern-to-modern-is-benign reading is
established independently of EdgeFace.

\subsection{Generalization to AI-Solutions-KK}
\tldr{The same $8\times8$ grid on the in-the-wild AI-Solutions-KK set reproduces the
Color~FERET pattern almost exactly: JPEG~XL$\rightarrow$HEIF and JPEG~XL$\rightarrow$AVIF are
the worst chains, JPEG~2000 stays poor throughout, and modern-to-modern chains remain benign
--- confirming the recompression hazard is not an artifact of studio portraits.}

Table~\ref{tab:recompression-eer-kk} repeats the matrix on AI-Solutions-KK. Absolute EER is
higher across the board than on Color~FERET --- KK is an in-the-wild set with harder capture
conditions, so even single-pass EER sits around 2.5--3\,\% for the modern codecs rather than
$\sim$1\,\% --- but the \emph{ordering} of chains is preserved. The two catastrophic chains
are again the JPEG~XL ones: \textbf{JPEG~XL$\rightarrow$HEIF} reaches 24.17\,\% EER, a
delta-EER of $+21.01$\,pp over single-pass HEIF, and \textbf{JPEG~XL$\rightarrow$AVIF} reaches
15.34\,\% EER ($+12.69$\,pp). Both are markedly larger than on Color~FERET ($+16.31$ and
$+7.38$\,pp), so the pathological interaction between the JPEG~XL artifact structure and the
HEVC/AV1 intra coders is if anything stronger on harder faces.

JPEG~2000 again drags down every chain it touches, with the same row/column structure as on
Color~FERET. JPEG~2000 is so weak on its own at this budget on KK (its whole row sits near
12--13\,\% EER) that adding a first encoder before it barely moves the number: routing a
well-compressed modern face through a second JPEG~2000 pass does inflate absolute EER from
$\sim$3\,\% to $\sim$13\,\%, but because the delta is measured against JPEG~2000's own
single-pass error the JPEG~2000 \emph{column} deltas are small or even slightly negative. The
damage the delta captures sits in the JPEG~2000 \emph{row}: a face that has already been
through JPEG~2000 stays near 12--13\,\% whatever the second codec is, a delta-EER of around
$+8.9$ to $+9.7$\,pp (e.g.\ JPEG~2000 followed by WebP at $+9.67$\,pp, by AVIF at
$+9.62$\,pp, by JPEG at $+9.21$\,pp). Either way the conclusion is unchanged: JPEG~2000 is a poor fit for a
1\,kB face crop and must not appear in a recompression chain. All the same-codec and
modern-to-modern off-diagonal chains stay benign, within about 1\,pp of their single-pass
baselines (e.g.\ WebP$\rightarrow$WebP at 2.31\,\% and AVIF$\rightarrow$AVIF at 2.66\,\% are
the lowest \emph{classical} entries), exactly as on Color~FERET.

\begin{table}[t]
  \centering
  \caption{Recompression EER (\%) on AI-Solutions-KK at 112\,px and the 1024-byte budget.
  Rows are the source (first) codec; columns are the second codec. The diagonal is
  same-codec re-encoding. Lower is better; the two highlighted JPEG~XL chains and the
  JPEG~2000 row and column are the damaging ones, mirroring Color~FERET
  (Table~\ref{tab:recompression-eer-cf}). Column shading marks the \colorbox{green!25}{best} /
  \colorbox{red!22}{worst} \emph{classical} source per second codec; the Ours-ACCURATE row and
  column are left unshaded because the tabulated values are EdgeFace-scored
  (\S\ref{subsec:recomp-worst}).}
  \label{tab:recompression-eer-kk}
  \adjustbox{max width=\textwidth}{\input{tables/recompression_eer_kk_1024.tex}}
\end{table}

\subsection{Worst-case chains}
\label{subsec:recomp-worst}
\tldr{On both Color~FERET and AI-Solutions-KK the chains to forbid in a deployment are
JPEG~XL$\rightarrow$HEIF and JPEG~XL$\rightarrow$AVIF (the sharpest spikes) and anything that
routes through JPEG~2000; the safe default is to keep faces in --- or re-encode them into ---
WebP, AVIF, or HEIF.}

Ranking the matrix by delta-EER makes the operational guidance concrete. The two
catastrophic chains are JPEG~XL$\rightarrow$HEIF and JPEG~XL$\rightarrow$AVIF.
JPEG~XL$\rightarrow$HEIF is the worst cross-codec chain on both datasets --- $17.14\%$ EER on
Color~FERET and $24.17\%$ on AI-Solutions-KK at $1024$\,B, a $+16$--$21$\,pp jump over the
benign JPEG~XL$\rightarrow$JPEG~XL diagonal ($1.01\%$ on Color~FERET, $3.28\%$ on
AI-Solutions-KK) and $+16.31$/$+21.01$\,pp over single-pass HEIF --- while
JPEG~XL$\rightarrow$AVIF reaches $8.00\%$/$15.34\%$ ($+7.38$/$+12.69$\,pp): a system that
stores faces as JPEG~XL and later transcodes them to a HEVC- or AV1-based still format can
multiply its verification error by an order of magnitude at the same byte budget. The next tier is the JPEG~2000 band: with single-pass JPEG~2000
already the weakest codec at this budget, every JPEG~2000-bearing chain remains near 6\,\%
EER and contributes $+4.9$ to $+5.0$\,pp whenever JPEG~2000 is the \emph{first} pass. By contrast,
the benign region is broad: same-codec re-encoding and the WebP/AVIF/HEIF/JPEG cross-pairs
all stay within about 1\,pp of their single-pass baselines.

The practical implication for document pipelines is twofold. First, a face that has already
been compressed should, where possible, be carried forward in its existing modern container
rather than transcoded; if a second pass is unavoidable, WebP, AVIF, and HEIF are safe
targets, and the system should avoid emitting JPEG~XL into an HEVC/AV1 still encoder and
should avoid JPEG~2000 entirely at sub-1\,kB budgets. Second, because these effects are
codec-pair specific rather than uniform, a pipeline that fixes a single canonical face codec
end-to-end will see far smaller recompression loss than one that lets each stage choose its
own format. The AI-Solutions-KK results confirm this guidance holds on in-the-wild faces, where
the JPEG~XL spikes are in fact even larger.

\paragraph{Independent-matcher confirmation (ArcFace).} To remove the single-matcher
objection (\S\ref{sec:recompression} noted the matrix was scored only on
\texttt{edgeface\_xs}), we re-scored the entire $7\times7$ classical recompression matrix on
\texttt{arcface\_antelopev2} --- architecturally independent of EdgeFace. At $1024$\,B the
picture reproduces exactly: the worst chain is again JPEG~XL$\rightarrow$HEIF ($14.2\%$ EER on
ArcFace vs $17.1\%$ on EdgeFace-XS --- same chain, same order of magnitude), same-codec and
modern-to-modern cross-pairs stay benign ($\le 0.39\%$: WebP$\to$WebP $0.12$,
AVIF$\to$AVIF $0.14$, HEIF$\to$HEIF $0.25$, JPEG$\to$JPEG $0.27$), and every
JPEG~2000-bearing chain is the poor band (JPEG~2000$\to$JPEG~2000 $3.49\%$). At $512$\,B the
catastrophic pattern also reproduces on ArcFace: any chain whose \emph{second} pass is legacy
JPEG collapses (worst JPEG~2000$\rightarrow$JPEG $45.7\%$ on ArcFace vs $46.6\%$ on
EdgeFace-XS --- same chain), while modern same-codec re-encoding stays usable (WebP$\to$WebP
$0.84\%$, JPEG~XL$\to$JPEG~XL $2.67\%$, AVIF$\to$AVIF $3.91\%$). So the
``modern-to-modern is benign; JPEG~XL$\to$HEVC/AV1, any-$\to$JPEG at $512$\,B, and JPEG~2000
are not'' conclusion is \emph{not} a single-matcher artefact for the classical matrix.

\emph{The Ours rows are now confirmed on ArcFace too.} We completed the independent-matcher
rescore of the learned-codec chains at \emph{both budgets}, and it removes the confound rather
than overturning the result. At $1024$\,B, Ours-ACCURATE$\rightarrow$Ours-ACCURATE is $0.44\%$ EER on ArcFace ---
\emph{lower} than its EdgeFace-XS value ($0.66\%$) and inside the benign band of the
matrix --- while every modern Ours cross-pair stays benign ($\le 0.44\%$: JPEG$\to$Ours
$0.01$, WebP$\to$Ours $0.01$, AVIF$\to$Ours $0.03$, Ours$\to$JPEG $0.22$) and only the
JPEG~2000 Ours chains are expensive (JPEG~2000$\to$Ours $2.47\%$, Ours$\to$JPEG~2000
$1.12\%$). So the ``Ours chains sit with the benign modern cluster'' reading survives on a matcher
architecturally independent of the codec's identity side-stream, and the recompression matrix
is now independently ArcFace-confirmed for \emph{both} the classical and the learned codecs.
The same holds at $512$\,B: Ours-ACCURATE$\rightarrow$Ours-ACCURATE is $0.64\%$ EER on ArcFace
(vs $1.09\%$ on EdgeFace-XS) and in fact the lowest chain in that matrix, while the costly
chains remain the ones \emph{involving legacy JPEG/JPEG~2000} (e.g.\ JPEG$\rightarrow$Ours
$20.2\%$, JPEG~2000$\rightarrow$Ours $18.0\%$, Ours$\rightarrow$JPEG $28.5\%$ on ArcFace). So
the benign-Ours-diagonal reading holds on
the independent matcher at both budgets, and no recompression conclusion rests on the
EdgeFace confound.

\subsection{Tighter 512-byte budget}
\tldr{At half the byte budget every chain is noisier, but the codec-pair ordering is
preserved and a new worst case appears: JPEG~2000$\rightarrow$JPEG nearly destroys identity
(46.64\,\% EER on Color~FERET, 41.03\,\% on AI-Solutions-KK), while same-codec and
modern-to-modern chains stay comparatively low.}

Tables~\ref{tab:recompression-eer-cf-512} and~\ref{tab:recompression-eer-kk-512} repeat the
full $8\times8$ grid at the harder 512-byte budget on both datasets. Absolute EER rises
sharply everywhere --- halving the budget leaves each encoder far less room to reconstruct
the first pass's artifacts --- but the relative ordering of chains carries over from 1024
bytes: same-codec re-encoding and modern-to-modern cross-pairs remain the safest region
(Ours-ACCURATE$\rightarrow$Ours-ACCURATE is the lowest chain at 1.09\,\% on Color~FERET and
2.42\,\% on AI-Solutions-KK --- EdgeFace-scored here, but the $512$\,B ArcFace rescore
confirms Ours$\to$Ours as the lowest chain independently at $0.64\%$,
\S\ref{subsec:recomp-worst}; and WebP$\rightarrow$WebP is the lowest \emph{classical}
chain at 2.57\,\% and 7.32\,\% respectively), and anything routed through
JPEG~2000 stays poor. The tighter budget adds one stark new failure mode not visible at
1024 bytes: \textbf{JPEG~2000$\rightarrow$JPEG} is catastrophic on both sets, reaching
46.64\,\% EER on Color~FERET and 41.03\,\% on AI-Solutions-KK --- re-encoding an
already-JPEG~2000 face as baseline JPEG under a 512-byte ceiling leaves verification barely
above chance. The overall recompression penalty is therefore larger at 512 bytes than at
1024, but the operational guidance is unchanged: fix a single modern codec end-to-end and
keep both JPEG~2000 and legacy JPEG out of any recompression chain.

\begin{table}[t]
  \centering
  \caption{Recompression EER (\%) on Color~FERET at 112\,px and the tighter 512-byte budget.
  Rows are the source (first) codec; columns are the second codec. The diagonal is
  same-codec re-encoding. Lower is better; the JPEG~2000$\rightarrow$JPEG corner is
  catastrophic. Column shading marks the best (green) / worst (red) \emph{classical} source
  per second codec; the Ours-ACCURATE row and column are left unshaded because the tabulated
  values are EdgeFace-scored (the independent $512$\,B ArcFace rescore confirms
  Ours-ACCURATE$\rightarrow$Ours-ACCURATE as the lowest chain at $0.64\%$,
  \S\ref{subsec:recomp-worst}).}
  \label{tab:recompression-eer-cf-512}
  \adjustbox{max width=\textwidth}{\input{tables/recompression_eer_colorferet_512.tex}}
\end{table}

\begin{table}[t]
  \centering
  \caption{Recompression EER (\%) on AI-Solutions-KK at 112\,px and the tighter 512-byte
  budget. Rows are the source (first) codec; columns are the second codec. The diagonal is
  same-codec re-encoding. Lower is better; the JPEG~2000$\rightarrow$JPEG corner is
  catastrophic. Column shading marks the best (green) / worst (red) \emph{classical} source;
  the EdgeFace-scored Ours-ACCURATE row and column are left unshaded (ArcFace-confirmed at
  $512$\,B on Color~FERET, \S\ref{subsec:recomp-worst}), mirroring Color~FERET
  (Table~\ref{tab:recompression-eer-cf-512}).}
  \label{tab:recompression-eer-kk-512}
  \adjustbox{max width=\textwidth}{\input{tables/recompression_eer_kk_512.tex}}
\end{table}

%% file: sections/11_adversarial.tex
\section{Adversarial Robustness and Sanitization}
\label{sec:adversarial}
\tldr{We ask whether squeezing a face crop to a 512--1024\,B budget also strips
adversarial perturbations. Under three no-box attacks on Color~FERET --- a
high-frequency-content (HFC) attack, a Li-AE prototypical-autoencoder attack, and a
CLIP-surrogate I-FGSM attack --- tighter byte budgets sanitize the perturbation far more
aggressively \emph{for the classical codecs} (the trained codecs are almost
budget-independent), with JPEG~2000 the strongest sanitizer almost everywhere and
Ours-ACCURATE the weakest overall --- it has the highest mean residual of any codec in all
six attack $\times$ dataset tables.
The two surrogate/frequency attacks (Li-AE and HFC) are the strongest and are
statistically interchangeable; the generic CLIP transfer is the weakest. Evaluated as a
defense, our trained codecs are the \emph{weakest} sanitizers of the field at 512\,B and
$\varepsilon \ge 0.06$, while at 1024\,B they are matched or overtaken by WebP and
Ours-FAST falls back to mid-pack; neither approaches JPEG~2000. The same picture reproduces
on the in-the-wild AI-Solutions-KK set.}

A face-compression front end is not only a bitrate component; it is also a fixed,
non-trainable image transform that every uploaded crop must pass through before it ever
reaches a matcher. That makes it a natural \emph{input-purification} defense: if the codec
discards exactly the high-frequency content that carries an adversarial perturbation, the
perturbation is sanitized as a side effect of compression. This section tests that
hypothesis under the strictest practical attacker model and reports what is measured.

\subsection{Threat model}
\tldr{We assume a no-box attacker: no access to the target matcher's architecture, weights,
training data, or query interface --- so the attack must rely on transferable,
model-agnostic image vulnerabilities.}

We adopt the \emph{no-box} threat model, the most restrictive setting in the adversarial
literature: the adversary has zero knowledge of the target face matcher's architecture,
parameters, and training set, \emph{and issues no query against it}
\cite{li2020nobox}. This is
strictly more constrained than any query-based black-box setting, and it is the realistic
model for an attacker facing a verification API or an embedded matcher whose weights and
score feedback are unavailable. Because the attacker cannot adapt to the specific victim,
every no-box attack must instead target \emph{shared} vulnerabilities that many recognition
models hold in common --- chiefly an over-reliance on high-frequency image cues and on
non-robust features. The attacker perturbs the aligned face crop directly under an
$\ell_\infty$ budget $\varepsilon$ on $[0,1]$ images; we sweep
$\varepsilon \in \{0.03, 0.06, 0.10\}$, the standard no-box range. The defender's matcher
is one of our four anchor models (\texttt{arcface\_antelopev2}, \texttt{lvface\_l},
\texttt{topofr\_r100}, \texttt{edgeface\_xs}), none of which the attacker can see.

\subsection{Attacks}
\tldr{We study three no-box attacks spanning the spectrum: a training-free
high-frequency-content (HFC) attack, a Li-AE prototypical-autoencoder attack that trains a
tiny proxy on held-out faces, and a CLIP-surrogate I-FGSM attack that repurposes a
foundation model; all three are run end-to-end on Color~FERET.}

We instantiate three complementary no-box attacks that sit at the three canonical points
of the no-box spectrum~\cite{li2020nobox}: training-free image-statistics manipulation
(HFC), a small surrogate trained only on held-out domain images (Li-AE), and a generic
foundation-model surrogate (CLIP). Running all three lets us complete the attack-strength
ordering we hypothesised at the outset. The measured ordering is
\emph{Li-AE $\approx$ HFC $>$ CLIP-transfer}: the trained proxy-AE and the training-free
frequency attacks are statistically interchangeable at the top and both dominate the
generic transfer attack --- so the strict ``HFC $>$ Li-AE'' separation anticipated in the
plan does not hold (the two are tied), but the broader claim that a
domain-informed surrogate beats a generic one does. The three attack-strength curves are
reported per attack below and summarized in Table~\ref{tab:attack-strength}.

\paragraph{High-frequency-content (HFC) attack.} A \emph{training-free} attack that uses no
surrogate, no training data, and no queries: it suppresses the crop's original
high-frequency components and injects crafted high-frequency noise directly in the frequency
domain, exploiting the fact that recognition models lean heavily on high-frequency cues
\cite{kurakin2017physical, goodfellow2015fgsm}. This is the strictest no-box variant and the
most relevant stress test for a compression defense, because the perturbation lives in
exactly the band a lossy codec is most likely to quantize away. On Color~FERET, the attack
degrades the uncompressed identity cosine (perturbed crop vs.\ clean reference, averaged over
the four matchers) from $1.0$ down to
$0.82$, $0.63$, and $0.41$ at $\varepsilon = 0.03$, $0.06$, and $0.10$ respectively,
confirming it is a strong attack before any defense is applied.

\paragraph{CLIP-surrogate I-FGSM attack.} A \emph{foundation-model-surrogate} attack: the
attacker crafts the perturbation by running iterative FGSM (I-FGSM)
\cite{kurakin2017physical, goodfellow2015fgsm} against a generic pre-trained CLIP image
encoder \cite{radford2021clip} used as a surrogate, then transfers the result to the unseen
face matcher. This probes a different failure mode --- transferable non-robust features
rather than raw high-frequency energy --- and a perturbation that is not concentrated in the
highest frequencies may survive compression more readily. On Color~FERET the CLIP-surrogate
attack is real but milder than HFC: averaged over the four anchor matchers it degrades the
uncompressed identity cosine (perturbed crop vs.\ clean reference) from $1.0$ to $0.86$,
$0.76$, and $0.64$ at $\varepsilon = 0.03$, $0.06$, and $0.10$ --- a transfer attack that
moves identity meaningfully but, at matched $\varepsilon$, less than the frequency-domain
HFC attack ($0.82$/$0.63$/$0.41$), consistent with its energy being spread across the
spectrum rather than packed into the band a codec quantizes hardest.

\paragraph{Li-AE prototypical-autoencoder attack.} A \emph{trained-surrogate} attack, the
seminal no-box method~\cite{li2020nobox}, occupying the middle of the spectrum: unlike HFC
it does train a surrogate, but unlike CLIP that surrogate is a tiny model built from
scratch on the attacker's own held-out faces rather than a foundation model. We train a
small convolutional auto-encoder (a $\sim$1.38\,M-parameter proxy) on $400$ held-out
identities ($3{,}854$ crops) sampled from WebFace42M --- identities disjoint from the
Color~FERET/AI-Solutions-KK evaluation sets, so no victim data or matcher is ever touched
--- using Li 2020's winning \emph{prototypical} objective, in which the decoder
reconstructs the per-identity mean crop rather than the input, forcing the encoder to
learn class-discriminative and thus transferable features from little data. Perturbations
are then crafted by an I-FGSM feature-push on the encoder bottleneck followed by an
Intermediate-Level Attack (ILA) refinement on a mid encoder layer~\cite{huang2019ila},
which amplifies the transferable component of the perturbation. On Color~FERET this attack
degrades the uncompressed identity cosine (averaged over the four anchor matchers) from
$1.0$ to $0.80$, $0.63$, and $0.40$ at $\varepsilon = 0.03$, $0.06$, and $0.10$ --- within
$0.014$ of the training-free HFC attack ($0.82$/$0.63$/$0.41$) at every $\varepsilon$ and
markedly stronger than the CLIP transfer ($0.86$/$0.76$/$0.64$). The trained proxy AE, with
ILA, is therefore as potent as the frequency-domain attack despite crafting against a
$1.38$\,M-parameter model that has never seen the victims.

\paragraph{Attack-strength ordering.} Table~\ref{tab:attack-strength} collects the three
uncompressed attack-strength curves. The measured ordering is
\emph{Li-AE $\approx$ HFC $\gg$ CLIP-transfer}: the two domain-informed attacks (one
training-free, one trained on held-out faces) are indistinguishable within $\le 0.014$
identity-cosine at every $\varepsilon$, and both are far stronger than the generic
foundation-model transfer, whose spread-spectrum perturbation moves identity only about
two-thirds as far at matched budget. This refines the initially hypothesised ``HFC $>$ Li-AE $>$
transfer'' ordering: the surrogate-vs-frequency gap we expected does not exist ---
a properly refined trained proxy matches the frequency attack --- but the surrogate-vs-generic
gap is real and large.

\begin{table}[t]
  \centering
  \caption{Uncompressed no-box attack strength on Color~FERET: identity cosine between the
  perturbed crop and its clean reference (mean over the four anchor matchers
  \texttt{arcface\_antelopev2}, \texttt{lvface\_l}, \texttt{topofr\_r100},
  \texttt{edgeface\_xs}); \emph{lower is a stronger attack}. The clean baseline is $1.0$ by
  construction. Li-AE and HFC are statistically interchangeable at every $\varepsilon$ and
  both dominate the CLIP transfer.}
  \label{tab:attack-strength}
  \small
  \begin{tabular}{lccc}
    \toprule
    Attack & $\varepsilon=0.03$ & $\varepsilon=0.06$ & $\varepsilon=0.10$ \\
    \midrule
    Li-AE (trained proxy AE + ILA) & \textbf{0.803} & \textbf{0.628} & \textbf{0.401} \\
    HFC (training-free, frequency)  & 0.817 & 0.632 & 0.408 \\
    CLIP-surrogate (foundation)     & 0.860 & 0.761 & 0.636 \\
    \bottomrule
  \end{tabular}
\end{table}

\subsection{Compression as a defense}
\tldr{Lossy compression sanitizes all three no-box attacks, and for the classical codecs the
effect is dominated by the byte budget: the 512\,B operating point removes most of the
adversarial identity gap, while 1024\,B leaves much more of it intact (the two trained codecs
are the exception --- their two budgets are essentially tied); JPEG~2000 sanitizes best and
Ours-ACCURATE worst. The three attacks are sanitized in roughly the order they attack
(Li-AE $\approx$ HFC $>$ CLIP survive the codec, Li-AE edging above HFC in most 1024\,B
cells and slightly below it at 512\,B).}

\begin{figure}[t]
  \centering
  \includegraphics[width=\linewidth]{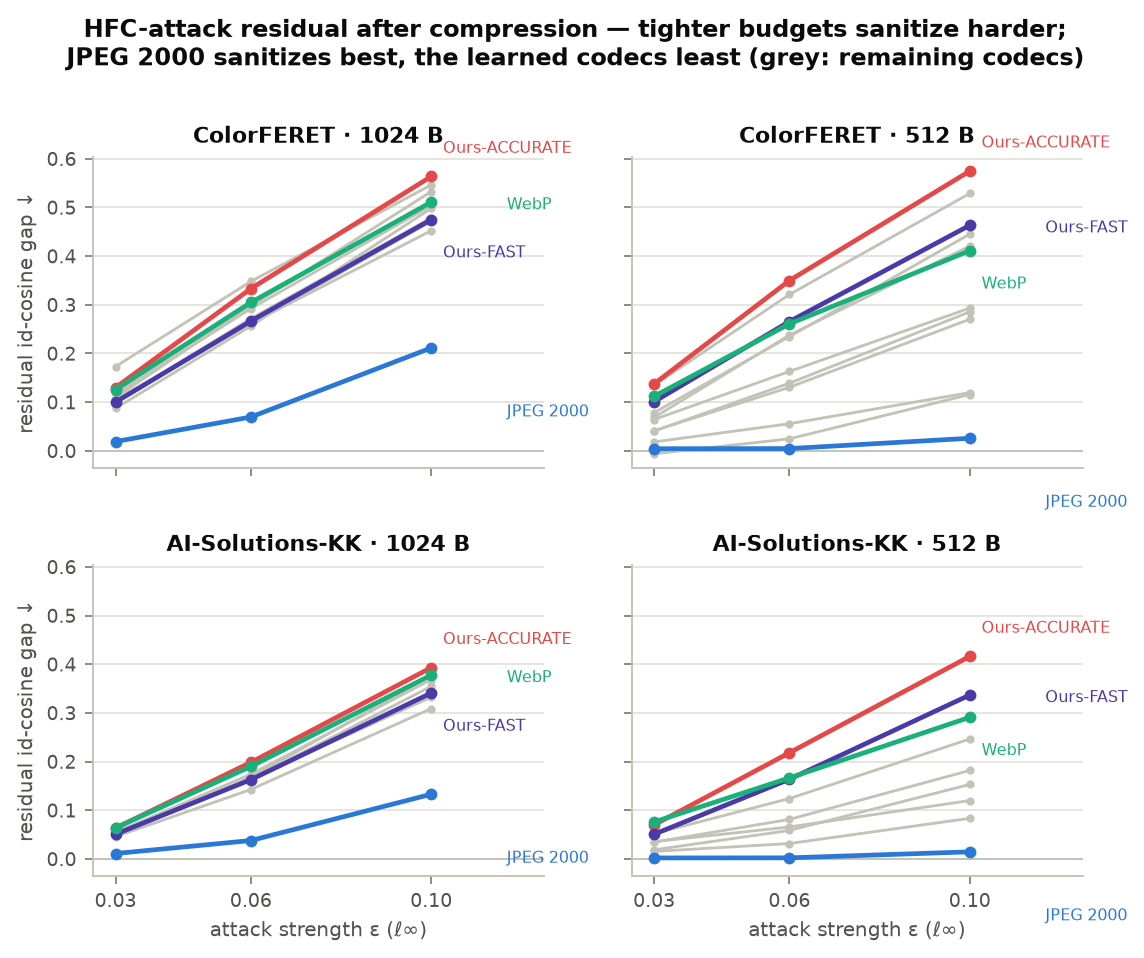}
  \caption{HFC-attack sanitization: residual identity-cosine gap (four-anchor mean;
  lower means the perturbation was removed) versus attack strength $\varepsilon$, per
  codec, for both datasets (rows) and byte budgets (columns) --- the graphical form of
  Tables~\ref{tab:sanitization-hfc} and~\ref{tab:sanitization-hfc-kk}. The three
  findings of this section are visible: for the classical codecs the 512\,B panel sits
  below its 1024\,B counterpart at every $\varepsilon$ on Color~FERET and at
  $\varepsilon \ge 0.06$ on AI-Solutions-KK (the budget dominates; at $\varepsilon=0.03$
  on KK the two budgets nearly coincide and WebP inverts), JPEG~2000 is the lowest curve
  wherever the residual is non-trivial, and Ours-ACCURATE is the highest full-support
  curve at $\varepsilon \ge 0.06$ in every panel --- with the two trained curves' budgets
  lying essentially on top of each other, unlike any classical codec. Ours-FAST and
  Ours-ACCURATE are drawn on all four panels. Grey curves are the remaining codecs; on the
  Color~FERET panels these include the four reduced-coverage rows of
  Table~\ref{tab:sanitization-hfc} (JPEG-FzT, JPEG-AI, bmshj2018, mbt2018, measured on
  $300$ crops), one of which --- JPEG-AI --- runs above Ours-ACCURATE at
  $\varepsilon \le 0.06$ in the 1024\,B panel.}
  \label{fig:sanitization}
\end{figure}

The sanitization hypothesis is that passing the adversarial crop through the codec restores
the matcher's behavior toward the clean case. We quantify this with the \emph{residual}: the
identity-cosine gap that survives compression,
$\text{residual} = \cos(\text{clean-comp}, \text{ref}) -
\cos(\text{adv-comp}, \text{ref})$, averaged over the four anchor matchers (two, where a
codec was attacked under only two --- marked $^\ddagger$ in the tables). Both cosines are
taken over the same crops, so a codec attacked on a subset is compared with itself rather
than against a different population.
Averaging over four matchers --- three of them (\texttt{arcface\_antelopev2},
\texttt{lvface\_l}, \texttt{topofr\_r100}) architecturally independent of the frozen
EdgeFace side-stream to which our trained codecs are anchored --- makes this residual far
less susceptible to evaluator circularity than a single-anchor score. A
small residual means the adversarial and clean crops compress to almost the same identity ---
the perturbation has been sanitized; a large residual means the attack survives the codec.
This continuous residual is the headline metric we report throughout; it substitutes for
an attack-success-rate-versus-bytes curve (which would need a fixed accept/reject
threshold per matcher) and captures the same quantity --- how much of the attack survives
compression --- without committing to one operating threshold. The threshold-based
attack-success-rate-versus-bytes curve is therefore not reported here.
Table~\ref{tab:sanitization-hfc} reports the residual per codec across the
$\varepsilon \times$ budget grid, and Figure~\ref{fig:sanitization} visualizes the same
sweep.

\begin{table}[t]
  \centering
  \caption{HFC-attack sanitization on Color~FERET: residual identity-cosine gap surviving
  compression (mean over the four anchor matchers; \emph{lower} is better, i.e.\ stronger
  sanitization), across attack strength $\varepsilon$ and byte budget. Rows are ordered
  strongest sanitizer first. Tighter budgets sanitize far more aggressively for the
  classical codecs; the two trained codecs are almost budget-independent. Generated
  directly from the stored residuals, so every printed number is reproducible from
  \texttt{outputs/adversarial/sanitization.csv}. $^\ddagger$ marks reduced coverage:
  JPEG-AI, bmshj2018 and mbt2018 are measured on two anchors
  (\texttt{arcface\_antelopev2}, \texttt{lvface\_l}) rather than four, and those three
  plus JPEG-FzT were attacked on a $300$-crop subset against $11{,}335$ crops for every
  other row. Those four rows are printed for completeness but are excluded from the
  best/worst comparison, because a residual over $300$ crops is not comparable with one
  over the full set. \shadelegend}
  \label{tab:sanitization-hfc}
  \small
  \adjustbox{max width=\textwidth}{\input{tables/sanitization_hfc.tex}}
\end{table}

Three findings stand out. First, the \textbf{byte budget dominates for the classical
codecs}: across every classical codec and attack strength, the 512\,B operating point
leaves a smaller residual than 1024\,B, usually by a factor of two or more (e.g.\ JPEG at
$\varepsilon=0.06$ drops the residual from $0.261$ at 1024\,B to $0.056$ at 512\,B; the
narrowest gap is WebP's $0.112$ vs.\ $0.126$ at $\varepsilon=0.03$). The two trained codecs
are the exception: their budgets are essentially tied --- Ours-ACCURATE's 512\,B cell sits
$0.008$--$0.017$ \emph{above} its 1024\,B cell and Ours-FAST's two budgets agree to within
$0.012$ with no consistent sign --- because their rate control degrades the reconstruction
gracefully instead of quantising the high-frequency band away. This is consistent with the hypothesis --- a tighter budget forces heavier
quantization, which discards more of the high-frequency band the HFC attack lives in.
Second, \textbf{JPEG~2000 is the strongest sanitizer} at every grid point where the
residual is non-trivial, holding it at or below $0.07$ except in the hardest
$\varepsilon=0.10$/1024\,B corner (at $\varepsilon=0.03$/512\,B HEIF edges it out, on a
residual indistinguishable from zero). Third,
the \textbf{learned codecs are the weakest sanitizers at the tight budget}: Ours-ACCURATE
preserves the largest
residual of the full-support rows at every grid point ($0.130$--$0.575$; the reduced-coverage
JPEG-AI row is higher still at $\varepsilon \le 0.06$/1024\,B, but on $300$ crops), and
Ours-FAST is the second-weakest row at
512\,B for $\varepsilon \ge 0.06$ --- though at 1024\,B, where its residual barely moves
while the classical codecs' residuals blow up, Ours-FAST is only mid-pack (e.g.\ $0.267$ at
$\varepsilon=0.06$, between JPEG's $0.261$ and JPEG~XL's $0.292$). The likely cause is that
their nonlinear synthesis transforms reconstruct high-frequency structure that the
classical transform codecs simply throw away. The effect is not confined to the learned
family --- WebP is the weakest classical codec in five of the six cells (HEIF is marginally
weaker in the $\varepsilon=0.10$/1024\,B corner, $0.533$ vs.\ $0.511$), sitting above
Ours-FAST at every 1024\,B point and below it at 512\,B for $\varepsilon \ge 0.06$, and for
the same reason: its reconstructions are the sharpest.
The caveat is that this sanitization is bought with image quality --- the same heavy
quantization that removes the perturbation also removes the clean identity signal, which is
why the 512\,B clean-compressed cosines themselves are low; the right way to read the table
is jointly with the clean-budget verification results elsewhere in the report, not as a
free defense. The trade-off is stark when the two axes are paired explicitly: JPEG~2000,
the strongest sanitizer of the field (residual $0.069$ at $\varepsilon{=}0.06$/1024\,B),
is simultaneously the worst standard codec on clean identity (EER $3.23\%$ at
112\,px/1024\,B under ArcFace, Section~\ref{sec:benchmark}), whereas Ours-ACCURATE --- the
best codec on clean identity at the tight budget (EER $0.22\%$ at 112\,px/512\,B under
ArcFace, against JPEG~2000's $19.5\%$) --- leaves the largest residual in the table
($0.333$ at the same grid point, and $0.350$ at 512\,B), with Ours-FAST mid-pack at
$0.267$. Sanitization strength and clean
identity preservation are, on this evidence, two ends of the same quantization dial
rather than independently optimizable properties.

\paragraph{CLIP-surrogate attack.} The same sanitization picture holds for the
transfer-based CLIP-surrogate I-FGSM attack, and it is even more favourable to the defender.
Table~\ref{tab:sanitization-clip} reports the residual per codec across the
$\varepsilon \times$ budget grid, averaged over the four anchor matchers
(\texttt{arcface\_antelopev2}, \texttt{lvface\_l}, \texttt{topofr\_r100},
\texttt{edgeface\_xs}). At matched $\varepsilon$ and budget the CLIP residuals are smaller
than the HFC ones in \emph{every} cell (for JPEG-FzT the HFC row compared against is its
reduced $300$-crop measurement) --- for
JPEG~2000 the residual never exceeds $0.06$ across the whole grid, and at 512\,B every
classical codec's residual is at or below $\approx 0.17$ (the trained codecs reach
$\approx 0.26$) --- because
the CLIP-surrogate perturbation is both a weaker attack to begin with and is largely
quantized away once it passes through the codec. The budget ordering also reappears for the
classical codecs, without exception (the 512\,B column is below the 1024\,B column for
every classical codec at every $\varepsilon$), although for the two trained codecs the two
budgets are essentially tied; and JPEG~2000 is again the strongest sanitizer, except at
512\,B and $\varepsilon \le 0.06$ where HEIF's slightly negative residual edges it out,
while WebP is the weakest of the classical codecs at every $\varepsilon$. The conclusion is that compression sanitizes a
transfer-based attack at least as effectively as the frequency-domain one --- the codec does
not need the perturbation to live in the very highest frequencies in order to remove it.

\begin{table}[t]
  \centering
  \caption{CLIP-surrogate I-FGSM sanitization on Color~FERET: residual identity-cosine gap
  surviving compression (mean over the four anchor matchers
  \texttt{arcface\_antelopev2}, \texttt{lvface\_l}, \texttt{topofr\_r100},
  \texttt{edgeface\_xs}; lower is better), across attack strength $\varepsilon$ and byte
  budget. Residuals are smaller than under the HFC attack
  (Table~\ref{tab:sanitization-hfc}) in \emph{every} cell: the CLIP surrogate transfers
  less of its perturbation through compression than the high-frequency attack does. (For
  JPEG-FzT that comparison is against its reduced $300$-crop HFC row; JPEG-AI and the
  CompressAI baselines are absent here because they were never attacked with this
  surrogate.)
  Every row, including both trained codecs, is the four-anchor mean, and each is taken
  over the same $2{,}000$ attacked crops under both cosines. \shadelegend}
  \label{tab:sanitization-clip}
  \small
  \adjustbox{max width=\textwidth}{\input{tables/sanitization_clip.tex}}
\end{table}

\paragraph{Li-AE prototypical-autoencoder attack.} Sanitization of the trained-surrogate
Li-AE attack completes the picture and confirms that the codec-level ordering is a property
of the \emph{defense}, not of any one attack.
Table~\ref{tab:sanitization-liae} reports the Li-AE residuals per codec across the
$\varepsilon \times$ budget grid, averaged over the same four anchor matchers. All three
findings reappear: the byte budget dominates for every \emph{classical} codec (the 512\,B
column is far below the 1024\,B column; the two trained codecs, covered at
$\varepsilon=0.06$ only, are again the exception, with $0.251$ vs.\ $0.257$ for Ours-FAST
and $0.310$ vs.\ $0.299$ for Ours-ACCURATE), JPEG~2000 is the strongest sanitizer at every
grid point but one (residual $\le 0.12$ except in the hardest $\varepsilon=0.10$/1024\,B
corner; at $\varepsilon=0.03$/512\,B HEIF's $-0.006$ edges out its $0.002$, as under HFC),
and WebP is the weakest classical sanitizer --- but not JPEG-FzT, which on this
full-support pass is mid-pack, below AVIF at every $\varepsilon$ and below JPEG~XL at five
of the six ($\varepsilon{=}0.03$/1024\,B is a near-tie the other way, $0.113$ vs.\ $0.112$). At
$\varepsilon=0.06$/512\,B, the one cell where the trained codecs are covered at that
budget, they are the two weakest rows in the table. The one attack-specific effect is that
at matched $\varepsilon$ and budget the Li-AE residuals sit \emph{slightly above} the HFC
ones in most 1024\,B cells --- clearly so for JPEG~2000, JPEG and WebP, whereas JPEG~XL and
the trained codecs sit slightly below, and at 512\,B the two attacks are level on average
(e.g.\ JPEG~2000 at $\varepsilon{=}0.06$/1024\,B leaves $0.116$ under Li-AE versus $0.069$
under HFC and $0.021$ under CLIP): the trained proxy-AE perturbation, refined by an
intermediate-level attack, lives partly in the mid-level feature band rather than purely in
the top frequencies a codec quantizes hardest, so a little more of it survives compression
at the looser budget.
It is nonetheless heavily sanitized --- the $0.278$ worst-corner JPEG~2000 residual still
falls to $0.052$ at 512\,B --- so the practical conclusion is unchanged: compression
sanitizes all three no-box attacks, the effect is bought with byte budget, and the strength
ordering of the \emph{attacks} (Li-AE $\approx$ HFC $>$ CLIP, Table~\ref{tab:attack-strength})
carries over to how much of each \emph{survives} the codec (Li-AE $\approx$ HFC $>$ CLIP
residuals, with Li-AE the larger of the near-tied pair in most 1024\,B cells).

\begin{table}[t]
  \centering
  \caption{Li-AE prototypical-autoencoder sanitization on Color~FERET: residual
  identity-cosine gap surviving compression (mean over the four anchor matchers
  \texttt{arcface\_antelopev2}, \texttt{lvface\_l}, \texttt{topofr\_r100},
  \texttt{edgeface\_xs}; lower is better), across attack strength $\varepsilon$ and byte
  budget. The codec ordering closely matches the HFC and CLIP tables
  (Tables~\ref{tab:sanitization-hfc},~\ref{tab:sanitization-clip}) --- except that
  JPEG-FzT, measured here on the full $11{,}335$ crops rather than the $300$-crop HFC
  subset, ranks mid-pack; residuals sit slightly
  above HFC in most $1024$\,B cells, as the ILA-refined perturbation is not purely
  high-frequency. The two trained codecs were run against this attack at
  $\varepsilon=0.06$ only, so their other columns read ``---''; they are shown rather
  than dropped, because at the one $\varepsilon$ they cover they are the two weakest
  sanitizers at 512\,B ($0.310$ and $0.251$, against WebP's $0.243$), which is the finding
  this section reports against our own codec --- at 1024\,B WebP's $0.327$ is weaker still.
  \shadelegend}
  \label{tab:sanitization-liae}
  \small
  \adjustbox{max width=\textwidth}{\input{tables/sanitization_liae.tex}}
\end{table}

\subsection{Ours-as-defense}
\tldr{Run as a fixed input transform, our trained codecs sanitize all three attacks --- their
residuals sit in the same range as the \emph{weaker} classical codecs --- but they do
\emph{not} break the sanitization-vs-quality trade-off: at 1024\,B Ours-FAST sanitizes like
a mid-pack classical codec and Ours-ACCURATE like the weakest one, neither like JPEG~2000,
and at 512\,B with $\varepsilon \ge 0.06$ the two of them are the weakest rows in the
table. Ours-ACCURATE, the higher-quality model, is consistently the weaker sanitizer of
the two.}

The most interesting case for this project is whether our trained codecs --- which are
optimized to preserve the identity signal under a hard byte budget --- can sanitize the
adversarial perturbation \emph{while} retaining clean-image identity, breaking the
sanitization-vs-quality trade-off that the generic codecs exhibit. Ours-FAST (the tiny
$\sim$1.35\,M-parameter model) and Ours-ACCURATE (the $\sim$18.7\,M-parameter model with the
identity side-stream and refine head) are plausible candidates either way: the identity
side-stream could carry a compact, attack-resistant identity descriptor, or it could equally
re-inject perturbed high-frequency detail through the refine head.

Table~\ref{tab:ours-defense} answers the question by placing both trained codecs beside the
strongest (JPEG~2000) and weakest (WebP) classical sanitizers, for all three attacks, averaged
over the same anchor matchers. The verdict is that \textbf{Ours-as-defense sanitizes, but
does not win}. Three observations follow. First, the trained codecs are genuine sanitizers:
their residuals are finite and ordered as the attack hypothesis predicts (CLIP
below HFC throughout), though --- unlike the classical codecs --- the two byte budgets
barely differ for them: across both datasets and all three attacks Ours-ACCURATE's 512\,B
cell sits $0.002$--$0.023$ \emph{above} its 1024\,B cell in every case, while Ours-FAST's
two budgets agree to within $0.014$ with no consistent sign.
Second, they do \emph{not}
match JPEG~2000 --- under HFC at $\varepsilon=0.06$/1024\,B, Ours-FAST leaves a residual of
$0.267$ and Ours-ACCURATE $0.333$, versus JPEG~2000's $0.069$; at that grid point Ours-FAST
is level with JPEG ($0.261$) and Ours-ACCURATE sits above WebP ($0.305$), the weakest
classical sanitizer, and at 512\,B with $\varepsilon \ge 0.06$ the two trained codecs are
the weakest rows in the table --- nowhere do they behave like the transform codecs
that aggressively discard high frequencies. Third, and consistently, \textbf{the
higher-quality model is the weaker defense}: Ours-ACCURATE leaves a larger residual than
Ours-FAST at every comparable cell, which is exactly the signature of the refine head
re-injecting fine detail --- including adversarial high-frequency content --- that
Ours-FAST's coarser reconstruction throws away. This is the sanitization-vs-quality
trade-off reappearing \emph{within} our own model family rather than being broken by it: the
side-stream preserves clean identity but does not selectively reject the perturbation. The
practical reading is that our codecs are competitive identity-preserving compressors but are
not, on this evidence, a free adversarial purifier; a defender who wants maximal
sanitization should still reach for a tighter byte budget or a transform codec like
JPEG~2000.

\begin{table}[t]
  \centering
  \caption{Ours-as-defense residual comparison on Color~FERET: residual identity-cosine gap
  (mean over the anchor matchers; lower is better) for the two trained codecs against the
  strongest (JPEG~2000) and weakest classical (WebP) sanitizers, under all three attacks at
  the two budgets and $\varepsilon=0.06$. Every row, including Ours-ACCURATE, is the
  four-anchor mean over \texttt{arcface\_antelopev2}, \texttt{lvface\_l},
  \texttt{topofr\_r100}, and \texttt{edgeface\_xs}: the TopoFR-R100 $\times$ Ours-ACCURATE
  pairing is populated here. Lower is stronger sanitization. Every cell is the four-anchor
  mean of the \texttt{residual} column of
  \texttt{outputs/adversarial/sanitization.csv}, so the rows reproduce the corresponding
  cells of Tables~\ref{tab:sanitization-hfc}--\ref{tab:sanitization-liae}. Li-AE for Ours is
  likewise evaluated on Color~FERET (third column group), completing the three-attack picture:
  for the learned codecs Li-AE $\approx$ HFC $\gg$ CLIP, the same coarse ordering as for the
  classical codecs (Table~\ref{tab:attack-strength}), although for the learned pair Li-AE
  falls just \emph{below} HFC rather than just above it.}
  \label{tab:ours-defense}
  \small
  \begin{tabular}{lcccccc}
    \toprule
    & \multicolumn{2}{c}{HFC, $\varepsilon=0.06$} & \multicolumn{2}{c}{CLIP, $\varepsilon=0.06$} & \multicolumn{2}{c}{Li-AE, $\varepsilon=0.06$} \\
    \cmidrule(lr){2-3}\cmidrule(lr){4-5}\cmidrule(lr){6-7}
    Codec & 512\,B & 1024\,B & 512\,B & 1024\,B & 512\,B & 1024\,B \\
    \midrule
    JPEG~2000     & \textbf{0.005} & \textbf{0.069} & \textbf{$-0.001$} & \textbf{0.021} & \textbf{0.014} & \textbf{0.116} \\
    WebP          & 0.261 & 0.305 & 0.085 & 0.151 & 0.243 & 0.327 \\
    Ours-FAST     & 0.266 & 0.267 & 0.111 & 0.106 & 0.251 & 0.257 \\
    Ours-ACCURATE & 0.350 & 0.333 & 0.144 & 0.137 & 0.310 & 0.299 \\
    \bottomrule
  \end{tabular}
\end{table}

\subsection{Generalization to AI-Solutions-KK}
\label{subsec:adv-kk}
\tldr{Repeating the full sweep on in-the-wild AI-Solutions-KK reproduces every Color~FERET
finding: the byte budget dominates at $\varepsilon \ge 0.06$, JPEG~2000 is the strongest
sanitizer, and Ours-ACCURATE is the weakest --- background clutter and non-frontal pose do
not change the ordering.}

To check that the sanitization picture is not an artefact of controlled studio portraits, we repeat
the sweep on the in-the-wild AI-Solutions-KK set, where background clutter and
non-frontal pose give the attacker more high-frequency cover. All three attacks (HFC,
CLIP-surrogate, and Li-AE) now run for all nine codecs --- the seven classical codecs and both
trained codecs as defenses --- so the full three-attack sanitization picture is complete on
\emph{both} datasets. All four anchor
matchers completed on this run, so every row is a clean four-matcher mean.
Table~\ref{tab:sanitization-hfc-kk} reports the HFC residuals, and the three Color~FERET
findings reappear, with the same qualifications: the 512\,B budget sanitizes far harder
than 1024\,B at $\varepsilon \ge 0.06$ (at $\varepsilon=0.03$ the two budgets are close and
WebP even inverts, $0.076$ vs.\ $0.064$), JPEG~2000
is the strongest sanitizer at every grid point (residual $\le 0.14$ even in the hardest
$\varepsilon=0.10$/1024\,B corner), and Ours-ACCURATE is the weakest at
$\varepsilon \ge 0.06$, peaking at $0.416$ --- though at $\varepsilon=0.03$ WebP is
marginally weaker still, and Ours-FAST, as on Color~FERET, is near-worst at 512\,B but only
mid-pack at 1024\,B. Under the CLIP-surrogate attack the KK residuals are again uniformly
smaller --- JPEG~2000 stays $\le 0.023$ across the whole grid and Ours-ACCURATE peaks at
$0.171$ --- so the in-the-wild setting, if anything, sanitizes slightly harder than
Color~FERET at matched budget. The KK Li-AE pass reproduces the same ordering: JPEG~2000 is
again the strongest sanitizer (four-anchor mean residual $0.07$ at $\varepsilon{=}0.06$/$1024$\,B,
$0.012$ at $512$\,B), WebP the weakest \emph{classical} sanitizer ($0.19$/$0.13$, quoting
$1024$\,B/$512$\,B), and --- as on
Color~FERET --- \textbf{the learned codecs are the weakest defenders at the tighter budget}: Ours-ACCURATE leaves
the largest $512$\,B residual ($0.19$/$0.21$, essentially tied with WebP at $1024$\,B) and Ours-FAST a little less ($0.16$/$0.16$), the same
``higher-quality model sanitizes less'' signature seen under HFC. So the Ours-as-defence result
holds identically on KK: our codecs sanitize no-box attacks about as well as the weakest classical
codecs, not better.
The practical conclusion is unchanged and holds on both
datasets: compression sanitizes no-box attacks, the effect is bought with image quality, and
the learned codecs do not break that trade-off.

\begin{table}[t]
  \centering
  \caption{HFC-attack sanitization on AI-Solutions-KK: residual identity-cosine gap
  surviving compression (mean over the four anchor matchers \texttt{arcface\_antelopev2},
  \texttt{lvface\_l}, \texttt{topofr\_r100}, \texttt{edgeface\_xs}; lower is better), across
  attack strength $\varepsilon$ and byte budget. The ordering matches Color~FERET
  (Table~\ref{tab:sanitization-hfc}): JPEG~2000 sanitizes hardest and Ours-ACCURATE least,
  and tighter budgets sanitize harder at $\varepsilon \ge 0.06$ (at $\varepsilon=0.03$ the
  two budgets nearly coincide, and WebP inverts). Every cell here rests on the same
  $2{,}000$-crop attacked subset --- $1{,}999$ for JPEG~XL at 512\,B, one crop the codec
  could not produce --- so no row needs a coverage caveat; JPEG-AI and the CompressAI
  baselines are absent because they were never attacked on this dataset. \shadelegend}
  \label{tab:sanitization-hfc-kk}
  \small
  \adjustbox{max width=\textwidth}{\input{tables/sanitization_hfc_kk.tex}}
\end{table}

\paragraph{Scope and future work: codec-aware attacks.} This study evaluates the
\emph{no-box} attacker exclusively; a codec-aware adversary is not yet evaluated. Because the
credential codec is a public standard once deployed, such an adversary could attack
\emph{through} the Ours decoder --- using backward-pass differentiable approximation (BPDA)
or straight-through gradient estimation to bypass the non-differentiable quantization --- so
whether compression still sanitizes against that stronger attacker is scheduled as future
work.

%% file: sections/12_significance.tex
\section{Statistical Significance}
\label{sec:significance}
\tldr{We test whether the accuracy gaps between codecs are real or could be noise: paired McNemar tests on verification decisions and DeLong tests on ROC areas, with Benjamini--Hochberg correction for the many comparisons. The headline result is that at 112\,px/1024\,B the strongest modern codecs (AVIF, HEIF, JPEG~XL, WebP, JPEG-AI, and our trained \texttt{ours\_accurate}) are \emph{operationally interchangeable}: most of their pairwise differences do reach significance after Benjamini--Hochberg correction, but the effect sizes are practically negligible (ROC-AUC gaps at the fourth decimal). On Color~FERET under EdgeFace-XS four pairs fail significance outright (AVIF$\approx$WebP, HEIF$\approx$JPEG~XL, JPEG~XL$\approx$Ours-FAST, HEIF$\approx$Ours-FAST); across all four Color~FERET anchors nine of $220$ tests fail, and on AI-Solutions-KK only seven of $440$ --- including JPEG-AI$\approx$Ours-ACCURATE on the independent ArcFace anchor. Yet every one of these codecs beats JPEG and JPEG~2000 decisively.}

A ranking of codecs by equal-error rate (EER) is only useful if the gaps between
them are larger than what sampling noise alone would produce. This section attaches
formal significance to the head-to-head differences reported in the benchmark
(Section~\ref{sec:benchmark}). The full testing protocol --- pairing scheme,
trial construction, and operating points --- is given in
Section~\ref{sec:protocol}; here we recap it only briefly and then read the
results off the per-pair table.

\subsection{Method recap}
\tldr{Two complementary paired tests --- McNemar on the accept/reject decisions and DeLong on the ROC areas --- run on the \emph{same} verification trials for both codecs, then corrected for multiplicity with Benjamini--Hochberg FDR.}

Because every codec is evaluated on the \emph{identical} set of verification
trials, the comparisons are paired and we can test the difference between two
codecs directly rather than comparing two independent error estimates. We use two
tests per codec pair:

\begin{itemize}
  \item \textbf{McNemar's test}~\cite{mcnemar1947} on the binary
  accept/reject decisions. For a pair of codecs $(a,b)$ it counts the discordant
  trials --- those where $a$ is correct and $b$ is wrong ($b$) versus the reverse
  ($c$) --- and forms the $\chi^2$ statistic from $b$ and $c$. It is sensitive to
  differences in the thresholded decision, i.e.\ accuracy at a fixed operating
  point.
  \item \textbf{DeLong's test}~\cite{delong1988} on the area under the ROC
  curve (AUC). It compares the two codecs' AUCs while accounting for the
  covariance induced by sharing the same trials, and is sensitive to
  threshold-independent separability.
\end{itemize}

Reporting both guards against a pair that looks tied at one operating point but
differs across the whole curve, or vice versa. Each test produces a raw
$p$-value; we treat a pair as significant only after multiple-comparison
correction (below). Unless stated otherwise the numbers in this section are for
the EdgeFace-XS matcher~\cite{george2024edgeface} on Color~FERET at the 112\,px
working resolution and the 1024-byte budget, the same anchor configuration used
elsewhere in the report; all values are read directly from the per-pair
significance table on disk.

\subsection{Pairwise codec comparisons}
\tldr{Aligned (uncompressed) reference beats every codec significantly. JPEG~2000 is the worst by a wide margin and loses to everyone. The interesting outcome is a cluster of modern codecs so close that their pairwise differences are either statistically non-significant or significant-but-negligible.}

Table~\ref{tab:sig-cf-edgeface} lists representative pairwise outcomes. Two
patterns are unambiguous. First, the \texttt{aligned} (uncompressed) reference is
significantly better than every codec, and JPEG~2000 is significantly worse than
every other method --- both at the floor of the McNemar $p$-values
($p_\mathrm{adj}=0$ to machine precision against all comparators, with AUC
dropping from $0.9998$ for aligned to $0.9868$ for JPEG~2000). At 1024 bytes and
112\,px, JPEG~2000 is simply not competitive for this task.

\begin{figure}[t]
  \centering
  \includegraphics[width=0.92\linewidth]{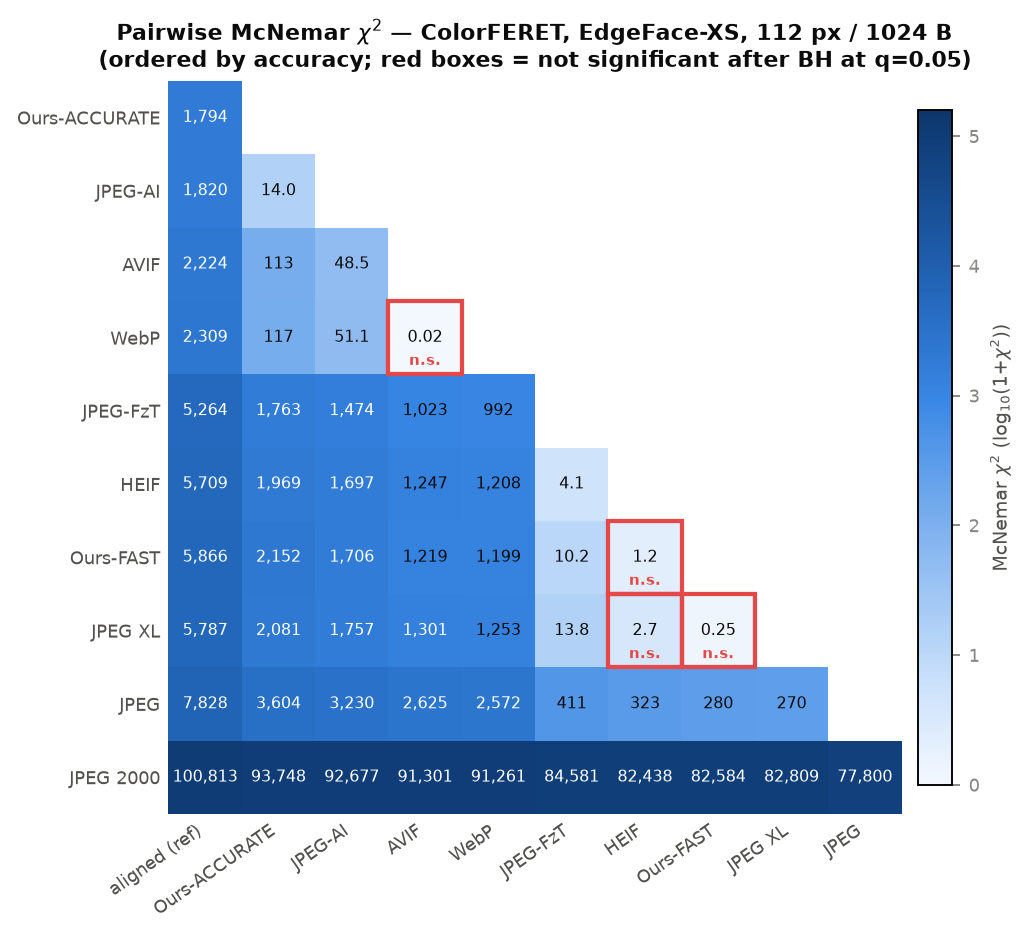}
  \caption{The complete pairwise McNemar matrix behind
  Table~\ref{tab:sig-cf-edgeface} (Color~FERET, EdgeFace-XS, 112\,px/1024\,B; rows and
  columns ordered by accuracy, colour is $\log_{10}(1+\chi^2)$). Only two cells
  survive Benjamini--Hochberg correction as non-significant (red boxes): AVIF/WebP
  and HEIF/JPEG~XL. Everything else is significant, but the $\chi^2$ magnitudes span
  four orders --- from the modern-codec near-ties ($\chi^2<20$) to the JPEG~2000 row
  ($\chi^2\approx10^5$) --- which is why significance alone does not rank the modern
  cluster.}
  \label{fig:significance-matrix}
\end{figure}

Second, and more informative, the strongest modern codecs are \emph{operationally}
interchangeable. With millions of paired trials per cell the McNemar test has enormous
power, so most pairwise differences do cross the significance threshold --- but the
effect sizes are negligible. The per-cell trial count is every mated pair plus a fixed
random sample of $2.0\times10^{6}$ non-mated pairs: $2\,094\,299$ on Color~FERET
($95\,839$ mated) and $3\,515\,807$ on AI-Solutions-KK ($1\,515\,807$ mated). These are
therefore not the $\sim$$64$\,M (Color~FERET) / $\sim$$153.7$\,M (KK) all-pairs totals
quoted in the dataset summary (Table~\ref{tab:dataset-summary}) --- the non-mated side is
sub-sampled, because the full impostor set adds cost without adding resolution at these
counts --- so the two figures are not in conflict. Note also that this stage draws a
smaller impostor sample than the accuracy grid of \S\ref{sec:protocol}, which caps at
$5.0\times10^{6}$: the paired tests here need the \emph{same} trials under both codecs
rather than the tightest possible ROC, so the counts in this section and in the rate--EER
tables are not expected to match.
A further caveat is statistical dependence: the all-pairs $\binom{N}{2}$ protocol makes
these millions of trials heavily dependent (each subject recurs in many pairs), so the
pair-level McNemar/DeLong $p$-values --- which bottom out at machine zero --- are
power-inflated and must not be read as effect magnitudes. We therefore lead the ranking
narrative with the subject-level bootstrap confidence intervals and Cliff's-$\delta$ effect
sizes of Section~\ref{subsec:rank-based}, and de-emphasise the pair-level $p$-values
accordingly. The headline rate--EER tables (Tables~\ref{tab:rate-eer-cf-arcface}
onward) carry \textbf{subject-level 95\% bootstrap CIs} (identity-cluster
resampling), so the reader can see the honest uncertainty directly --- on
AI-Solutions-KK, with only 105 identities, these CIs are an order of magnitude wider
than a pair-level interval would suggest. Only four modern-codec pairs fail to reach
significance even at this trial count: AVIF vs.\ WebP ($p_\mathrm{adj}=0.89$; discordant-decision
counts $b=6562$ vs.\ $c=6545$, essentially balanced), HEIF vs.\ JPEG~XL
($p_\mathrm{adj}=0.11$), and --- under the promoted $3$M-step checkpoint --- JPEG~XL vs.\
Ours-FAST ($p_\mathrm{adj}=0.63$) and HEIF vs.\ Ours-FAST ($p_\mathrm{adj}=0.29$). The pairs that do clear the threshold clear it narrowly ---
AVIF vs.\ JPEG-AI and JPEG-AI vs.\ WebP have McNemar $\chi^2$ below $20$ (against
$\chi^2$ in the thousands for the aligned and JPEG~2000 comparisons) and AUC gaps at
the fourth decimal place, so the rank order among them carries no practical weight. In
practical terms: choosing among AVIF, WebP, HEIF, JPEG~XL, and JPEG-AI at this
operating point is a matter of encoder availability and licensing, not measurable
identity preservation --- yet all of them clear JPEG and JPEG~2000 by a comfortable,
significant margin. Figure~\ref{fig:significance-matrix} renders the full pairwise
matrix: the four-orders-of-magnitude spread in $\chi^2$ between the near-ties
($\chi^2<3$) and the JPEG~2000 comparisons ($\chi^2\approx10^5$) is the quantitative
form of the ``significant but negligible'' distinction this section is built on.

Third, our two trained codecs (\texttt{ours\_accurate}, \texttt{ours\_fast}) are in
the run, and they land squarely in this modern-codec cluster. Both are decisively
better than JPEG~2000 --- \texttt{ours\_accurate} vs.\ JPEG~2000 gives a McNemar
$\chi^2$ of $85000$ and an AUC gap of $0.0127$ ($p_\mathrm{adj}\approx 0$) --- and both
are, in turn, significantly worse than the \texttt{aligned} reference, exactly as every
lossy codec is. Against the modern codecs the picture matches the AVIF/WebP story: the
McNemar tests reach significance (large $\chi^2$, $p_\mathrm{adj}$ far below any
threshold) because the paired trial counts are enormous, but the underlying AUC
differences are at the fourth decimal or smaller --- \texttt{ours\_accurate} vs.\
JPEG-AI differs in AUC by only $\sim$$1.8\times10^{-5}$, and vs.\ AVIF by
$\sim$$1.7\times10^{-4}$. The effect is statistically real but practically negligible:
Ours is competitive with the strongest general-purpose codecs at this operating point.
These three Ours-vs-modern-codec comparisons carry one further caveat: the flagship table is
anchored on \texttt{edgeface\_xs}, and \texttt{ours\_accurate}'s identity side-stream is
trained against a frozen EdgeFace-S embedding, so an EdgeFace-family evaluator is not
independent of Ours. The \texttt{ours\_accurate}-vs-\{JPEG-AI, AVIF, WebP\} gaps on this
anchor are therefore confounded and should be read together with the held-out CVLFace-IR101
and ArcFace results of Section~\ref{sec:benchmark}, which are architecturally independent of
the side-stream. \textbf{We have run that independent-matcher check directly}: re-testing
the same $112$\,px$/1024$\,B pairs under \texttt{arcface\_antelopev2} and
\texttt{cvlface\_ir101} (both architecturally independent of the EdgeFace side-stream)
reproduces the edgeface conclusion --- \texttt{ours\_accurate} vs.\ \{JPEG-AI, AVIF, WebP\}
differ only at the fourth AUC decimal ($|\Delta\mathrm{AUC}|\le 2\times10^{-4}$; under
CVLFace-IR101 the Ours-vs-AVIF pair is not even significant, $p_\mathrm{adj}=0.34$), matching
the pattern seen on \texttt{edgeface\_xs}. The ``operationally interchangeable'' verdict is
therefore \emph{not} an artefact of the circular evaluator. (The rank \emph{concordance}
recomputed with the EdgeFace family excluded is essentially unchanged --- see
\S\ref{subsec:rank-based}.)
Finally, the two variants are separable from each other --- \texttt{ours\_accurate}
beats \texttt{ours\_fast} ($0.9923$ vs.\ $0.9897$ accuracy, $\chi^2=1381$,
$p_\mathrm{adj}\approx 4\times10^{-302}$), confirming that the accuracy-oriented variant
buys a small but statistically solid gain over the fast one.

\begin{table}[t]
  \centering
  \caption{Representative pairwise significance on Color~FERET, EdgeFace-XS, 112\,px, 1024\,B. $\chi^2$ and $p$ are McNemar on the accept/reject decisions; AUC$_a$/AUC$_b$ are the DeLong ROC areas whose difference is tested separately. $p_\mathrm{adj}$ is the Benjamini--Hochberg-corrected McNemar $p$; \textbf{sig.} marks pairs that remain significant after correction at a false-discovery rate of $q=0.05$. All values from the on-disk significance table.}
  \label{tab:sig-cf-edgeface}
  \small
  \adjustbox{max width=\textwidth}{\input{tables/sig_cf_edgeface.tex}}
\end{table}

One nuance worth flagging: the two tests do not always agree at the margin. For
several modern-codec pairs the DeLong AUC test reaches significance (its raw $p$
is tiny) even though McNemar does not, because the ROC area is estimated from many
more comparisons than the handful of discordant thresholded decisions. We take the
McNemar decision test as primary for the operating-point ranking and report the
AUC test alongside it; where they diverge, the codecs differ in curve shape but
not in accept/reject behaviour at the chosen threshold.

\subsection{Multiple-comparison correction}
\tldr{With ten codecs plus the aligned reference there are $55$ pairwise tests in every cell, so we control the false-discovery rate with Benjamini--Hochberg at $q=0.05$; only the near-tied modern-codec pairs survive correction as non-significant.}

A full pairwise comparison over the eleven conditions that clear the coverage filter
(the \texttt{aligned} reference plus ten codecs) yields $\binom{11}{2}=55$ tests
\emph{per} matcher, resolution, and budget cell --- $220$ on Color~FERET and $440$ on
AI-Solutions-KK over the cells reported here. Testing that many hypotheses inflates the chance of a spurious
``significant'' result, so we adjust the raw McNemar $p$-values with the
Benjamini--Hochberg false-discovery-rate procedure~\cite{benjamini1995fdr},
reported as $p_\mathrm{adj}$ in Table~\ref{tab:sig-cf-edgeface} and in the
\texttt{p\_adj} / \texttt{significant} columns of the on-disk table. Because the
genuine effects here are enormous (raw $p$-values down to machine zero), the
correction changes almost nothing: every pair that was significant before
correction remains so, and the pairs flagged non-significant (at the
$q=0.05$ false-discovery threshold) are only the near-ties: nine of the $220$
Color~FERET tests and seven of the $440$ KK tests, every one of them a pair drawn from
the modern-codec cluster or from the two legacy codecs (JPEG$\approx$JPEG-FzT,
JPEG$\approx$JPEG~XL). The conclusion is therefore robust to the
multiplicity adjustment rather than balanced on its edge.

\subsection{Significance on AI-Solutions-KK}
\tldr{The in-the-wild KK trials confirm and sharpen the Color~FERET picture: aligned
beats every codec, JPEG~2000 is the worst, and --- crucially --- the modern-codec
near-ties that Color~FERET left undecided are \emph{all} resolved as significant on KK,
because its $\sim$2\,million trials per cell give the tests far more power. The KK grid
is complete: all four anchors at both resolutions, $55$ codec pairs per cell, $440$ tests.}

The AI-Solutions-KK significance grid is complete on disk
(Section~\ref{sec:artifacts}): all four anchor matchers
(\texttt{arcface\_antelopev2}, \texttt{lvface\_l}, \texttt{topofr\_r100},
\texttt{edgeface\_xs}) at both the $112$\,px working resolution and the $224$\,px source,
$55$ codec pairs per cell, $440$ tests in all. Nothing is read off a partial cell and no
resolution substitution is needed.

Table~\ref{tab:sig-kk} lists, for every cell, the pairs that fail significance together
with the largest contrast in that cell. The two qualitative patterns carry over intact:
the \texttt{aligned} reference is significantly better than every codec, and JPEG~2000 is
significantly worse than every other method --- its AUC falls to $0.936$--$0.965$ at
$112$\,px and $0.955$--$0.978$ at $224$\,px, against $\approx 0.999$ for aligned on every
anchor --- all at the floor of the McNemar $p$-values. The sharper result is what happens to the
modern-codec pairs. On Color~FERET under EdgeFace-XS four pairs are statistically
indistinguishable after correction (AVIF$\approx$WebP, HEIF$\approx$JPEG~XL,
HEIF$\approx$Ours-FAST, JPEG~XL$\approx$Ours-FAST); the nominated tie-pair,
JPEG-AI$\approx$WebP, is in fact \emph{significant} there, so the observed ties are
different pairs than predicted. On AI-Solutions-KK the grid is $55$ pairs per cell across
four anchors and two resolutions --- $440$ tests --- and only \textbf{seven} fail. Six of
those are at the $112$\,px working resolution and one at $224$\,px, so the in-the-wild
picture is close to all-significant without being it.

That is roughly what an in-the-wild benchmark should give: each KK cell aggregates
$3{,}515{,}807$ verification trials against $2{,}094{,}299$ on Color~FERET, so even
fourth-decimal AUC gaps clear the threshold. AVIF vs.\ WebP, a tie on Color~FERET, is
significant in all eight KK cells --- on the three side-stream-independent anchors it
clears the threshold on an AUC difference of only $0.8$--$4.0\times10^{-4}$. The
practical reading is unchanged --- the modern codecs are
\emph{operationally} interchangeable, with AUC gaps far below any deployment-relevant
tolerance --- but the statistical verdict on KK is that they are nonetheless distinct.

\begin{table}[t]
  \centering
  \caption{Pairwise significance on AI-Solutions-KK at 1024\,B, all four anchors at both
  the $112$\,px working resolution and the $224$\,px source. $\chi^2$ is McNemar on the
  accept/reject decisions; AUC$_a$/AUC$_b$ are DeLong ROC areas; $p_\mathrm{adj}$ is the
  Benjamini--Hochberg corrected McNemar $p$. The grid is $55$ codec pairs per cell,
  $440$ in all; the table prints every pair that \emph{fails} significance --- that is
  the finding --- together with the largest contrast in each cell to set the scale.
  Seven of the $440$ fail, six of them at $112$\,px on the ArcFace and LVFace-L anchors,
  one at $224$\,px. All values from the on-disk significance table.}
  \label{tab:sig-kk}
  \small
  \adjustbox{max width=\textwidth}{%
  \adjustbox{max width=\textwidth}{\input{tables/sig_kk.tex}}}
\end{table}

The trained codecs are in the KK run at full coverage, on every anchor and at both
resolutions. The Color~FERET verdict repeats: \texttt{aligned} beats both trained
variants decisively (on \texttt{edgeface\_xs} at 224\,px, $\chi^2$ of $6539$ and $6917$,
$p_\mathrm{adj}\approx 0$), and the two variants remain separable from each other
everywhere --- but by a margin that shrinks sharply with resolution. At $112$\,px the
Ours-ACCURATE-vs-Ours-FAST contrast is enormous ($\chi^2$ between $10{,}440$ and
$18{,}244$ across the four anchors); at $224$\,px it falls to between $4$ and $134$, and
on \texttt{edgeface\_xs} it only just clears the threshold ($\chi^2=4.2$,
$p_\mathrm{adj}=0.040$, an AUC gap of $5\times10^{-5}$). The accurate variant's advantage
over the fast one is therefore a working-resolution effect: at the $224$\,px source, where
both variants have more pixels and a harder budget, they converge.

Where the two datasets differ is in \emph{which} pairs tie. On Color~FERET the four ties
in the EdgeFace-XS cell are all among the modern block codecs and Ours-FAST (the other
five, spread over the remaining three anchors, add AVIF$\approx$JPEG-AI,
AVIF$\approx$WebP, JPEG-AI$\approx$WebP, AVIF$\approx$Ours-ACCURATE and the legacy
JPEG$\approx$JPEG-FzT); on AI-Solutions-KK the ties involve
JPEG-FzT and JPEG~XL --- and, on ArcFace at $112$\,px, \textbf{JPEG-AI and Ours-ACCURATE}
($\chi^2=3.2$, $p_\mathrm{adj}=0.074$). That is the one pairing the report has most reason
to scrutinise, since JPEG-AI is the strongest competitor: on the in-the-wild set at the
working resolution, under an anchor entirely independent of our identity side-stream, the
two are statistically indistinguishable.

\subsection{Rank-based analysis across the matcher roster}
\label{subsec:rank-based}
\tldr{The McNemar/DeLong tests above are pairwise and per-matcher; a Friedman test plus
Kendall's coefficient of concordance ask the complementary question --- do the matchers
\emph{agree} on the codec ranking? They do, strongly ($W=0.85$), which is the statistical
backing for the backbone-invariance claim of Section~\ref{sec:benchmark}.}

\begin{figure}[t]
  \centering
  \includegraphics[width=0.9\linewidth]{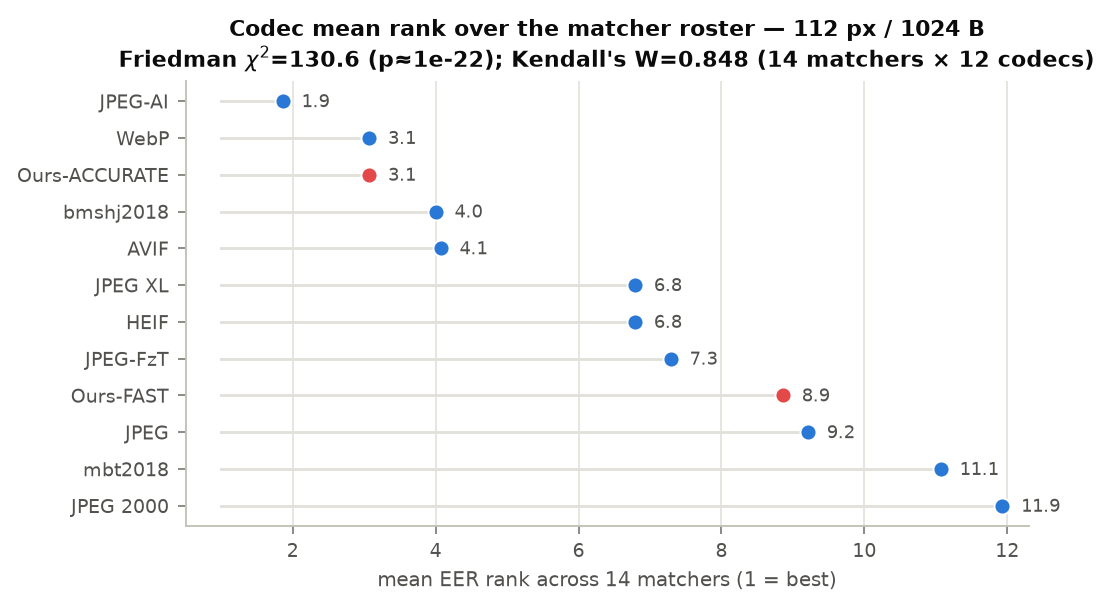}
  \caption{Mean EER rank of each codec across the fourteen fully-covered matchers at
  $112$\,px$/1024$\,B (1 = lowest EER). The tight concordance ($W=0.85$) means these
  mean ranks are representative of essentially every individual matcher's ranking:
  JPEG-AI leads, the trained Ours-ACCURATE ties WebP for second, and mbt2018 and
  JPEG~2000 trail under every backbone. All twelve codecs are ranked --- ten
  off-the-shelf plus both custom variants --- since every one has a stored EER for every
  roster matcher at this cell, which is what a complete matcher $\times$ codec matrix
  requires. Two cautions. The CompressAI baselines (\texttt{bmshj2018} $4.00$,
  \texttt{mbt2018} $11.07$) are ranked on the $300$-crop subset, so their positions are
  not comparable with the other ten. And a mean rank is not an effect size: the gap
  between rank $1.86$ and rank $3.07$ is a fraction of a percentage point of EER
  (Table~\ref{tab:posthoc} gives the effect sizes), whereas the gap to JPEG~2000 is two
  orders of magnitude.}
  \label{fig:codec-mean-rank}
\end{figure}

The pairwise tests establish that individual codec gaps are real; they do not directly
quantify whether the \emph{ranking} is stable across matchers. To do that we take the
per-matcher EERs at the $112$\,px$/1024$\,B operating point for the fourteen matchers with
complete codec coverage and run rank-based tests over the $14\times12$ (matcher $\times$
codec) grid. A Friedman test decisively rejects the null that the codecs are
interchangeable ($\chi^2=130.6$, $p\approx 1.3\times10^{-22}$), and \textbf{Kendall's
coefficient of concordance is $W=0.85$} --- a strong agreement, meaning the matchers rank
the codecs almost the same way regardless of backbone. The mean ranks
(Figure~\ref{fig:codec-mean-rank}) put JPEG-AI first at $1.86$, with the trained
Ours-ACCURATE and WebP tied behind it at $3.07$ and AVIF at $4.07$, and JPEG~2000
($11.93$) and \texttt{mbt2018} ($11.07$) at the bottom; this is the same ordering the
per-matcher grids show, now with a single concordance statistic behind it. The apparent
fourth place of \texttt{bmshj2018} ($4.00$) should not be read as a result: its cells come
from the $300$-crop Color~FERET subset, so it is ranked against codecs measured on
$11{,}335$.
Because this concordance is computed over a matcher roster that includes the EdgeFace family
--- to which Ours-ACCURATE's identity side-stream is anchored --- we recomputed it with all
four EdgeFace-family matchers removed. Agreement does not weaken; it \emph{strengthens},
from $W=0.85$ over all fourteen matchers to $W=0.89$ over the remaining ten. The
backbone-invariance of the codec ranking is therefore not an artefact of the EdgeFace
matchers --- if anything those four are the least concordant members of the roster. The
held-out CVLFace-IR101 and ArcFace rows supply the same independent cross-check on the
Ours cells specifically.

Table~\ref{tab:posthoc} reports Holm-corrected Wilcoxon signed-rank tests and Cliff's
$\delta$ effect sizes for the headline pairs across the roster: Ours-ACCURATE beats
JPEG~2000 with the maximal effect ($\delta=-1.00$: it wins on every matcher) and beats both
Ours-FAST ($\delta=-0.49$) and JPEG-FzT ($\delta=-0.36$), while it is statistically tied
with WebP ($\delta=+0.00$, $p_{\text{Holm}}=0.36$) and with AVIF ($\delta=-0.03$,
$p_{\text{Holm}}=0.24$) --- so
the cross-matcher rank evidence places Ours-ACCURATE level with the two best general-purpose
codecs, not behind them. These subject-level effect sizes, together with the bootstrap
$95\%$ confidence intervals on EER and FNMR carried in the master accuracy grid
(\texttt{eer\_lo}/\texttt{eer\_hi} and the \texttt{fnmr\_*\_lo}/\texttt{hi} columns), are the
primary basis for the ranking claims --- we treat them, rather than the power-inflated
pair-level $p$-values, as the effect-size ground truth.

\begin{table}[t]
  \centering \small
  \caption{Rank-based cross-matcher comparison at $112$\,px$/1024$\,B (fourteen matchers with
  complete codec coverage). Wilcoxon signed-rank $p$ is Holm-corrected over the family;
  Cliff's $\delta$ is the effect size ($-1$ = codec $a$ lower EER on every matcher). Global
  agreement: Friedman $\chi^2=130.6$ ($p\approx 1.3\times10^{-22}$), Kendall's $W=0.85$.}
  \label{tab:posthoc}
  \adjustbox{max width=\textwidth}{\input{tables/posthoc_stats.tex}}
\end{table}

%% file: sections/13_discussion.tex
\section{Discussion, Limitations, and Roadmap}
\label{sec:discussion}
\tldr{This is the honest current-state section: a uniform sub-1\,kB benchmark across ten off-the-shelf codecs, two datasets, and four matchers is in place and supports clear identity-preservation conclusions; the custom learned codec is competitive at 1024\,B (runner-up to JPEG-AI on the 224\,px identity-cosine headline, but leading it at the 112\,px verification working resolution --- at both budgets on both datasets --- and beating it on the FNMR@FMR$=10^{-4}$ operating point at 512\,B on Color~FERET); the four anchors span the full resolution grid on both datasets. One extension remains in progress: the Ours cells of the fourteen-model roster at the non-112\,px AI-Solutions-KK resolutions (Color~FERET is complete for every codec).}

This section separates what the report already establishes from what is still in
flight. We state the current pipeline state explicitly so that a reader can
distinguish a settled finding from a placeholder, and so that the remaining work is
scoped rather than implied.

\paragraph{Findings register.} Every headline number quoted in this report is
re-derived from the committed pipeline outputs into a machine-readable findings
register (Section~\ref{sec:artifacts}, with a JSON twin and a
schema README). Each register row records the claim, the exact value, the dataset,
matcher, resolution, and budget it holds at, and the source file it is computed
from, so any quantitative statement in the text can be traced --- and, after a
pipeline re-run, re-checked --- mechanically. The interpretation figures added
throughout the report (the distortion-vs-identity scatter, the fairness-disparity
panel, the recompression heatmaps, the sanitization-residual sweep, the significance
matrix, and the cross-matcher rank plot) are generated from the same committed
outputs by a single script and are archived alongside the register.

\subsection{What we have established}
\tldr{A single, uniform protocol now compares the codecs at a hard sub-1\,kB budget on two datasets and four matchers, and it already separates the codecs that keep identity from the ones that destroy it.}

The core contribution that is fully in place is a \emph{uniform} evaluation
protocol applied identically to every codec. All methods compress the same
pixel-aligned crops to the same hard byte budgets, the reconstructions are embedded
by the same set of face matchers, and verification is scored by the same routine.
This removes the usual confound where codec comparisons differ in alignment, crop,
or operating point rather than in the codec itself.

Concretely, the benchmark spans ten off-the-shelf codecs --- JPEG, JPEG\,2000, WebP,
JPEG\,XL, AVIF, HEIF, JPEG-FzT, JPEG-AI, and the learned baselines
\texttt{bmshj2018}~\cite{balle2018hyperprior} and \texttt{mbt2018}~\cite{minnen2018joint}
from CompressAI~\cite{begaint2020compressai} --- plus our two custom variants. It is run
on both datasets (the controlled Color~FERET studio portraits~\cite{phillips2000feret} and
the in-the-wild AI-Solutions-KK set) and embedded by all four anchor matchers
(\texttt{arcface\_antelopev2}~\cite{deng2019arcface}, \texttt{lvface\_l},
\texttt{topofr\_r100}, and \texttt{edgeface\_xs}~\cite{george2024edgeface}). For
each cell we report EER and FNMR at standard operating points, and on Color~FERET we
add paired significance tests (McNemar~\cite{mcnemar1947},
DeLong~\cite{delong1988}, with Benjamini--Hochberg FDR
control~\cite{benjamini1995fdr}; see Section~\ref{sec:significance}). Image-quality
measurements (PSNR, SSIM, and perceptual metrics) accompany the identity scores so
that pixel fidelity and identity fidelity can be read side by side.

The headline qualitative finding is robust across this grid: at the 1024\,B budget
the modern codecs --- WebP, AVIF, JPEG\,XL, HEIF, and JPEG-AI --- preserve the
identity signal well enough that EER stays close to the uncompressed baseline
(Color~FERET, 112\,px, ArcFace: $0.09$--$0.19\%$ against $0.04\%$ aligned),
whereas JPEG\,2000 collapses outright ($3.23\%$) and legacy JPEG, which still
survives this particular cell at $0.25\%$, collapses as soon as the budget tightens to
512\,B ($33.35\%$) or the resolution rises to 224\,px ($2.75\%$). For example, at
112\,px and 1024\,B on
Color~FERET the identity cosine is $0.851$
for JPEG against $0.921$ for WebP, a gap
that is visible in both the rate--EER tables and the per-codec montages
(Figure~\ref{fig:cmp-1kb}). The collapse of JPEG and JPEG\,2000 is driven by
their behaviour at extreme compression rather than by any artefact of our protocol,
which the quality study corroborates.

Our custom learned codec is the second established result. The accurate variant
(\emph{Ours-ACCURATE}, $\approx$18.7\,M parameters) is competitive with JPEG-AI on
identity at 1024\,B and
degrades gracefully toward the harder 512\,B operating point. On the in-the-wild KK
set at 112\,px the codec-comparison panel places \emph{Ours-ACCURATE} at an
identity cosine of $0.938$ at 1024\,B and $0.874$ at 512\,B, ahead of
JPEG-AI at both budgets ($0.921$ and $0.850$); the lead is mirrored on Color~FERET at
112\,px ($0.948$ vs.\ $0.930$ for JPEG-AI at 1024\,B), so at the verification working
resolution \emph{Ours-ACCURATE} is the strongest codec on identity while running
natively and enforcing the byte budget exactly. These 112\,px figures are at or above the
same variant's 224\,px scores; at that higher source resolution the ranking flips, with
JPEG-AI leading the identity-cosine headline ($0.958$/$0.947$ on Color~FERET/KK at
1024\,B) against \emph{Ours-ACCURATE}'s $0.947$/$0.934$, so \emph{Ours-ACCURATE} is the
runner-up there --- the 112\,px and 224\,px figures are different operating points, not a
discrepancy. Because
the identity side-stream's frozen anchor is an EdgeFace model, Ours-ACCURATE's headline
ranking is read off the anchors that are architecturally independent of it (ArcFace,
LVFace-L) rather than the in-grid EdgeFace evaluator; see the limitation below. The codec
reaches an exact budget by a binary search over a frozen 64-entry gain table and
packs a self-describing container, so every reported byte count is a true on-disk
size rather than a target.

Beyond the main benchmark, four focused studies are wired into the same pipeline:
the effect of working resolution (64/96/112/168/224\,px, with 112\,px as the
verification working resolution), demographic fairness using the KK Monk skin-tone,
age, and gender attributes~\cite{monk2023mst}, compressed-on-compressed
recompression, and no-box adversarial robustness with the codec acting as a
sanitiser. The Color~FERET instances of these studies have produced results;
their completion status on KK is detailed below.


\subsection{Limitations and threats to validity}
\tldr{JPEG-AI is now fully scored on the AI-Solutions-KK grid --- both the \emph{verification} grid (level with Ours-ACCURATE at the top of the in-the-wild 1024\,B field) and the \emph{quality} matrix --- at both 112 and 224\,px, and the §6 montages render the real JPEG-AI reconstruction at both resolutions; no HEIF stand-in remains. A few extension studies remain deferred.}

We list the known gaps so that no figure or table is over-read.

\paragraph{JPEG-AI on KK: verification and quality grids now complete.} JPEG-AI was
originally Color-FERET-only, but we have since compressed the full AI-Solutions-KK set
with the reference codec at \emph{both} 112 and 224\,px (both budgets) and scored it: the
\emph{verification} (EER/FNMR) grid now carries JPEG-AI rows on KK under all four anchors
plus the held-out CVLFace-IR101 (Table~\ref{tab:heldout-cvlface}), and JPEG-AI
is level with Ours-ACCURATE at the top of the in-the-wild $1024$\,B field (mean
FNMR$@10^{-4}$ $2.92\%$ against $2.93\%$, a $0.01$\,pp gap on overlapping confidence
intervals). The KK
JPEG-AI \emph{quality} scores are measured at both resolutions, now at full support
($n{=}17\,534$, up from the earlier $2000$/$400$-crop decode pilots): at \textbf{112\,px} PSNR
$30.5$/$28.3$\,dB, SSIM $0.930$/$0.893$, LPIPS $0.056$/$0.103$; at
\textbf{224\,px} PSNR $31.2$/$28.5$\,dB, SSIM $0.910$/$0.856$, LPIPS $0.138$/$0.233$,
quoted at $1024$/$512$\,B. The §\ref{sec:quality} \emph{visual montage} grids
render the \emph{real} JPEG-AI reconstruction at \emph{both} $112$ and $224$\,px on KK. No
HEIF stand-in remains, and every verification and quality number is measured on genuine
JPEG-AI output.

\paragraph{The learned codec spans the anchor grid; the wider roster now spans resolution.}
Both custom variants are now scored across the full resolution sweep
($64/96/112/168/224$\,px) and both byte budgets on all four anchor matchers and both
datasets, so the competitive 1024\,B claim and the 512\,B degradation are grounded in
completed cells. The \emph{extra ten} FR models of the fourteen-model roster are now also
embedded across $64$--$224$\,px for the classical and transform codecs: the
full-resolution model-effect heatmaps (Figures~\ref{fig:model-effect-224-cf}
and~\ref{fig:model-effect-224-kk}) confirm the codec ranking is stable across backbone
\emph{and} resolution. On Color~FERET the roster is now closed: all twelve
configurations, including JPEG-AI, both CompressAI baselines and both Ours variants,
carry an EER under all fourteen matchers at all five resolutions and both budgets. What
remains is on AI-Solutions-KK, where the two Ours variants are still missing two to five
of the ten non-anchor matchers at $64$/$96$/$168$\,px (JPEG-AI was compressed there at
$112$/$224$\,px only, and the CompressAI baselines were never run on KK); these do
not bear on the headline codec ranking.

\paragraph{Evaluator independence for the custom codec.} Ours-ACCURATE's identity
side-stream is anchored to a frozen EdgeFace-S embedding
(Section~\ref{subsec:codec-side}). Because EdgeFace is also one of the in-grid
evaluators, the EdgeFace verification numbers for Ours-ACCURATE are not fully
architecturally independent of the codec, so we read its headline ranking off the
anchors that are disjoint from the side-stream (ArcFace and LVFace-L). To test this
directly we also re-scored every reconstruction on a fully held-out backbone
(\texttt{cvlface\_ir101}, IR-101/WebFace12M, from a family used nowhere else in the
codec): Section~\ref{subsec:codec-heldout} shows the codec ranking, the
ACCURATE\,$>$\,FAST ordering, and the graceful $512$\,B degradation all survive there ---
on KK at $512$\,B Ours-ACCURATE is in fact the best codec on this independent matcher ---
so the circularity does not drive the result. The originally
specified separately-sourced ArcFace-iresnet100 side-stream anchor was not available on
disk; retraining with it (or any backbone outside the evaluation roster) would remove the
circularity at the source and is the intended fix.

\paragraph{Encode/decode speed is now measured.} The codec-properties table
(Table~\ref{tab:codec-properties}, in Section~\ref{sec:related}) reports the
provenance and deployment characteristics --- ownership, standard year, licensing,
hardware-decoder availability, and hard-budget fit rate --- for every benchmarked codec,
and encode/decode \emph{latency} is now measured under controlled single-core-CPU
(\texttt{taskset}) and single-GPU conditions in Section~\ref{subsec:speed}
(Table~\ref{tab:speed}, Figure~\ref{fig:speed-trend}). The headline latency findings ---
classical encode spanning three orders of magnitude, classical asymmetry (cheap decode),
and the learned codecs' $4$--$6\times$ GPU speedup --- are folded into that section.

\paragraph{The significance grid and the stress studies have landed.} The
recompression (Section~\ref{sec:recompression}) and adversarial-sanitisation
(Section~\ref{sec:adversarial}) studies are now complete on \emph{both} datasets, and
reproduce their Color~FERET orderings on KK. The paired significance tests are likewise
complete on both: $55$ codec pairs per cell across all four anchors, at $112$\,px on
Color~FERET ($220$ tests) and at both $112$ and $224$\,px on AI-Solutions-KK ($440$).
Cross-dataset significance claims are therefore fully supported on every anchor. What
Color~FERET does \emph{not} carry is a $224$\,px significance grid --- its resolution
sweep is reported through EER rather than through paired tests --- so any
$224$\,px-specific significance statement rests on KK alone.

\paragraph{Generalisation.} Two datasets, one controlled and one in-the-wild, give
useful coverage but do not span every capture condition (for example heavy pose or
occlusion beyond what KK contains), and all conclusions are conditioned on the four
chosen matchers. The findings should be read as evidence within this protocol, not
as a claim about every face matcher or capture setting.

\subsection{Current pipeline state}
\tldr{Both custom variants are fully in the grid across all four anchors, both budgets, and the full resolution sweep on both datasets; the recompression and adversarial studies have landed on both datasets; the extra ten FR models are now embedded across $64$--$224$\,px for the classical/transform codecs (full-resolution model-effect), with only their learned-codec cells at the non-112\,px resolutions still to fill.}

A snapshot of what is done and what is still computing, as of this draft:

\begin{itemize}[leftmargin=1.4em,itemsep=2pt,topsep=2pt]
  \item \textbf{Custom codec --- in the grid.} Both variants (Ours-FAST,
        $\approx$1.35\,M parameters; Ours-ACCURATE, $\approx$18.7\,M with the identity
        side-stream and gated refine head) are embedded across all four anchor
        matchers, both budgets, and the full resolution sweep on both datasets --- all
        $11{,}335$ Color~FERET crops embedded per anchor ($17{,}534$ on
        AI-Solutions-KK) --- so their verification numbers are final on the anchors.
  \item \textbf{Stress studies complete on both datasets.} The recompression matrices
        (Section~\ref{sec:recompression}) and the adversarial sanitisation study
        (Section~\ref{sec:adversarial}) are complete and reproduce on KK. All three no-box
        attacks (HFC, CLIP-surrogate, and Li-AE) are scored on both datasets for all nine
        codecs, including both Ours-as-defence variants --- at all three perturbation
        budgets, except Li-AE\,$\times$\,Ours on Color~FERET, which is measured at
        $\varepsilon=0.06$ only.
  \item \textbf{Fourteen-model roster --- complete on Color~FERET, near-complete on KK.}
        The extra ten FR models beyond the four anchors are now embedded at
        $64/96/168/224$\,px for every codec on Color~FERET, extending the
        backbone-invariance heatmaps to the source resolution
        (Figures~\ref{fig:model-effect-224-cf}--\ref{fig:model-effect-224-kk}). The
        only remaining gaps are on AI-Solutions-KK: the two Ours variants lack two to
        five of the ten non-anchor matchers at $64/96/168$\,px, JPEG-AI was compressed on
        KK at $112$/$224$\,px only, and the CompressAI baselines were never run there.
        All four anchors are populated at every KK resolution for every codec compressed
        there, so the headline codec ranking is unaffected.
\end{itemize}

\subsection{Remaining work}
\tldr{The to-do list is bounded: fill the Ours cells for the extra ten matchers at the non-112\,px AI-Solutions-KK resolutions (Color~FERET is complete for all twelve configurations across resolutions). Reviewer-requested follow-ups --- a print-scan/2D-barcode channel simulation, a codec-aware adversarial attack, a full-corpus train/test dedup scan, a two-seed retrain of Ours-FAST, and a KK budget sweep --- are noted as future work.}

The outstanding items, in roughly the order they will land, are:

\begin{enumerate}[leftmargin=1.6em,itemsep=2pt,topsep=2pt]
  \item \textbf{The last wider-roster cells on AI-Solutions-KK.} Color~FERET is complete
        for all fourteen matchers at $64/96/168/224$\,px for every codec and the
        $224$\,px model-effect heatmaps are in Section~\ref{sec:benchmark}; what remains
        is the Ours columns for the extra ten matchers at the non-112\,px KK resolutions,
        plus the KK JPEG-AI compressions outside $112$/$224$\,px (the per-image reference
        decoders are the bottleneck).
  \item \textbf{Reviewer-requested follow-ups.} Beyond the roster gap above, the following
        extensions would further stress the protocol: a print-scan / 2D-barcode channel
        simulation, a codec-aware adversarial attack, a full-corpus train/test
        deduplication scan, a two-seed retrain of Ours-FAST, and a KK budget sweep.
\end{enumerate}

Table~\ref{tab:status} consolidates the per-study completion status across the two
datasets, so a reader can see at a glance which conclusions are settled and which are
still in flight.

\begin{table}[t]
  \centering
  \caption{Consolidated study status by dataset. \emph{done} = complete and reported;
  \emph{partial} = available for a subset of matchers/conditions; \emph{pending} = still
  computing. Reconstruction quality is done for the classical, neural, JPEG-AI, and
  trained codecs. The custom codec
  spans the full resolution sweep ($64/96/112/168/224$\,px) on both datasets for the four
  anchors; the extra ten FR models are now embedded across $64$--$224$\,px for the
  classical/transform codecs (only their learned-codec cells at non-112\,px remain).
  Fairness uses all four anchor matchers. Significance covers all four anchors on both
  datasets ($220$ tests on Color~FERET at $112$\,px, $440$ on KK at $112$ and $224$\,px).
  The recompression study is complete on both datasets. All three no-box attacks (HFC, CLIP-surrogate, and Li-AE) are
  scored on both datasets for all nine codecs, including both Ours-as-defence variants, at
  all three perturbation budgets --- the one exception being Li-AE\,$\times$\,Ours on
  Color~FERET, measured at $\varepsilon=0.06$ only.}
  \label{tab:status}
  \small
  \begin{tabular}{lll}
    \toprule
    Study & Color~FERET & AI-Solutions-KK \\
    \midrule
    Codec benchmark            & done    & done    \\
    Reconstruction quality     & done    & done    \\
    Custom codec (Ours)        & done    & done    \\
    Resolution/preprocessing   & done    & done    \\
    Demographic fairness       & done    & done    \\
    Recompression              & done    & done    \\
    Adversarial sanitization   & done    & done    \\
    Statistical significance   & done    & done    \\
    \bottomrule
  \end{tabular}
\end{table}

%% file: sections/14_conclusion.tex
\section{Conclusion}
\label{sec:conclusion}
\tldr{Sub-1\,kB identity-preserving face compression is achievable: modern transform codecs---above all JPEG-AI---retain the matcher's identity signal at 1024 bytes where legacy JPEG collapses, and a compact custom learned codec tracks them closely on identity (runner-up to JPEG-AI at the 224\,px headline cell, but ahead of it at the 112\,px working point, at both budgets) while degrading more gracefully toward the harder 512-byte budget. The section closes with number-backed deployment guidance.}

This report set out to determine whether a pixel-aligned face crop can be stored in at
most 1024 bytes (and, more aggressively, 512 bytes) without destroying the identity signal
a face matcher relies on. The answer is yes. Across two datasets---the controlled
Color~FERET studio portraits~\cite{phillips2000feret} and the in-the-wild AI-Solutions-KK
set---and four anchor matchers spanning ArcFace~\cite{deng2019arcface}, LVFace, TopoFR and
EdgeFace~\cite{george2024edgeface}, the benchmark shows a clear split between codecs.
Legacy JPEG~\cite{wallace1992jpeg} \emph{does} fit the 1\,kB budget on 112\,px crops (100\%
compliance, EER $0.25\%$ under ArcFace) but collapses at 512 bytes ($33.4\%$ EER) or at
224\,px, where block artifacts wreck verification. JPEG~2000~\cite{taubman2002jpeg2000},
though it always fits the budget, is the \emph{worst} codec at $38$ of $40$ $1024$\,B
anchor operating points ($3.23\%$ EER at 112\,px/1024\,B versus $0.04\%$ aligned; the two
exceptions, Color~FERET at 224\,px under ArcFace and EdgeFace-XS, are where legacy JPEG is
worse still), worst on average, and the dominant fairness amplifier---so it must be avoided
outright (at the tighter $512$\,B/$112$\,px point legacy JPEG is worse still, $33.4\%$ EER
versus JPEG~2000's $19.5\%$). The codecs that actually preserve identity at 1024 bytes are
the modern transform/learned ones---WebP~\cite{google2010webp},
JPEG~XL~\cite{alakuijala2019jpegxl}, AVIF~\cite{aomavif}, HEIF, and especially
JPEG-AI~\cite{ascenso2023jpegai}---keeping equal-error rates within $0.15$\,pp of aligned
(Color~FERET, 112\,px, ArcFace: JPEG-AI/WebP/AVIF $0.09\%$, HEIF $0.18\%$, JPEG~XL $0.19\%$
vs $0.04\%$). The CompressAI baselines~\cite{balle2018hyperprior,minnen2018joint}
(bmshj2018, mbt2018) join this group only on full-coverage evidence: their $1024$/$512$\,B
cells were scored on a $300$-crop subset ($n_{\mathrm{neg}}=3\,375$) and are not
comparable---mbt2018 reads $2.50\%$ there, an artefact the body disowns---whereas at the
full-coverage $960$/$768$\,B budgets they are low and monotone ($0.13$--$0.21\%$ for
mbt2018, $0.17$--$0.28\%$ for bmshj2018). Our custom codec---a tiny FAST variant and a
larger ACCURATE variant with an identity side-stream and refine head---tracks these strong
codecs closely on identity at 1\,kB. At the 224\,px/1024\,B headline cell it is runner-up
to JPEG-AI, but at 112\,px the ranking inverts: Ours-ACCURATE leads on identity-cosine
($0.948$ vs $0.930$) and beats JPEG-AI on FNMR@$10^{-4}$ at 512 bytes on Color~FERET. Its
headline ranking uses anchors architecturally independent of its EdgeFace side-stream,
and---enforcing the byte budget via a binary search over a frozen gain table in a
self-describing container---it degrades more gracefully as the budget tightens to 512
bytes.

Beyond the headline result, the surrounding studies turn ``feasible'' into
``deployable''---each contributing a quantitative rule, not just direction. The resolution
analysis fixes 112\,px as the practical verification point: downsampling from 224\,px
retains $99.97\%$ of the AC spectral energy and, for strong matchers, over $0.99$
embedding cosine, while freeing budget; below 112\,px, clean EER roughly doubles by
64\,px. The budget sweep locates the sub-1\,kB error floor: classical codecs are flat at
$\geq 960$\,B but spike below ${\sim}800$\,B (AVIF $0.11\%\!\rightarrow\!2.71\%$, JPEG
$0.28\%\!\rightarrow\!33.4\%$, 960 to 512\,B), while byte-budgeted learned codecs cross it
unharmed (Ours-ACCURATE $0.14$--$0.22\%$ across the range). The fairness study (Monk
skin-tone~\cite{monk2023mst}, age, gender) finds compression mildly widens the
between-group EER gap while preserving group ordering, overwhelmingly a codec-choice
effect: on a uniform basis (MST5/6/7/9/10, MST8 excluded) JPEG~2000 amplifies the
skin-tone disparity $11.2$--$20.0\times$ over baseline on the three
side-stream-independent anchors---ArcFace $13.7\times$ ($0.37\!\rightarrow\!5.08$\,pp),
LVFace-L $11.2\times$ ($0.33\!\rightarrow\!3.74$\,pp), TopoFR-R100 $20.0\times$
($0.26\!\rightarrow\!5.22$\,pp), i.e.\ $3.4$--$5.0$\,pp added---while every other codec,
ours included, adds at most $1.7$\,pp (up to ${\sim}5\times$ for the mid codecs and
Ours-FAST, but from a far smaller base). The multiplier also depends on the baseline, not
just the codec: on EdgeFace-XS, whose uncompressed disparity is already $1.53$\,pp, the
same absolute JPEG~2000 damage ($+4.96$\,pp, to $6.49$\,pp) reads as only $4.2\times$. The
recompression study yields two rules: never route a sub-1\,kB face through JPEG~2000, and
never transcode JPEG~XL into an HEVC/AV1 still format --- the worst chain,
JPEG~XL$\to$HEIF, reaches $17.14\%$ EER (Color~FERET) and $24.17\%$ (AI-Solutions-KK) at
1024\,B, a $+16$--$21$\,pp jump over the benign JPEG~XL$\to$JPEG~XL diagonal
($1.01$/$3.28\%$); JPEG~XL$\to$AVIF fails the same way, milder ($8.00$/$15.34\%$,
$+7.4$/$+12.7$\,pp). Every other modern chain costs under about 1\,pp. The adversarial
study shows compression itself sanitizes no-box perturbations~\cite{goodfellow2015fgsm},
but sanitization strength and clean-identity preservation are two ends of the same
quantization dial---JPEG~2000 sanitizes best because it destroys the most signal---so
compression is an incidental defense, not a designed one. Finally, the significance
analysis confirms which gaps are real: the modern-codec cluster's gaps are statistically
significant but practically negligible on Color~FERET, remaining operationally
interchangeable; every collapse above is significant at machine-precision $p$-values, and
the ranking itself is backbone-invariant (Kendall's $W=0.85$ across fourteen matchers).

\paragraph{Deployment guidance.} Table~\ref{tab:deployment-guidance} condenses the
benchmark, properties, speed, fairness, and recompression findings into the concrete
recommendations a face-on-document integrator needs; every entry is backed by the
quoted section. The fourteen-model roster is now complete on Color~FERET: every codec
---classical, transform, JPEG-AI, the two CompressAI baselines and both Ours
variants---carries an EER under all fourteen matchers at all five resolutions and both
budgets. The gaps that remain are confined to AI-Solutions-KK, where JPEG-AI was
compressed at the two headline resolutions (112 and 224\,px) only, the Ours variants
still lack two to five of the ten non-anchor matchers at 64/96/168\,px, and the
CompressAI baselines were never run; all four anchor matchers, however, are populated at
every KK resolution for every codec compressed there. These remaining cells are documented in
Section~\ref{sec:discussion}; we expect them to sharpen, rather than overturn, the
conclusions above. All three no-box
attacks (HFC, CLIP-surrogate, and Li-AE) are now measured on both datasets for all nine
codecs, including both Ours-as-defence variants---complete at all three perturbation
budgets except Li-AE\,$\times$\,Ours on Color~FERET, which is measured at
$\varepsilon=0.06$ only---as are the KK JPEG-AI scoring pass and
encode/decode speed, all reported in the body.


%% file: sections/15_artifacts.tex
\section{Artifacts}
\label{sec:artifacts}
\tldr{Every number, table and figure in this report is derived from the machine-readable
files listed here. They are grouped by pipeline stage; the body text refers to results, not
to files, so this section is the single place where the two are connected.}

The report is generated, not transcribed: each table and figure is emitted by a script that
reads one or more of the files below, so a claim in the text can always be traced back to
the measurement that produced it. This section is the index. Paths are given relative to the
project's output root; the datasets themselves are described in Section~\ref{sec:datasets}
and the AI-Solutions-KK $112$ and $224$\,px aligned crops are published separately
(Section~\ref{sec:datasets-alignment}). Table~\ref{tab:artifacts} indexes the files by
pipeline stage.

Three files carry most of the report. \textbf{\texttt{accuracy/metrics.csv}} is the master
verification table: one row per (dataset, matcher, source) cell --- $4{,}632$ rows, one
per unique (dataset, matcher, source-tag) --- with EER, FNMR at
FMR~$\in\{10^{-2},10^{-3},10^{-4}\}$ and the FMR actually realised at each, the
mated/non-mated pair counts, and the aligned reference EER with its delta.
Percentile-bootstrap $95\%$ confidence intervals are carried for the four anchor matchers
and \texttt{cvlface\_ir101}, the cells for which the report quotes intervals, not for the
whole fourteen-model roster. A NaN FNMR is a deliberate value, not a gap: it marks an
operating point with too few impostor scores above it to be estimable in that cell
(\S\ref{sec:protocol}). Achieved byte sizes are \emph{not} in this file --- they live in
\texttt{codec\_comparison/file\_size\_summary.csv}.
\textbf{\texttt{quality/quality\_summary.csv}} is its
full-reference counterpart: median PSNR, SSIM, MS-SSIM, LPIPS and DISTS per
(dataset, codec, resolution, budget) cell together with the per-cell sample count $n$.
\textbf{\texttt{report\_summary/key\_findings.csv}} is the derived register that the
headline claims are read from, with a JSON twin and a schema documented in
\texttt{report\_summary/README.md}.

\begin{table}[t]
  \centering
  \footnotesize
  \caption{Report artifacts by pipeline stage. Every table and figure in this report is
  generated from one of these files, or --- for the two preprocessing/crop-tightness
  ablations of Section~\ref{sec:ablations} --- recomputed directly from the committed
  embedding arrays by its generator.}
  \label{tab:artifacts}
  \setlength{\tabcolsep}{4pt}
  \begin{tabular}{@{}>{\raggedright\arraybackslash}p{0.45\textwidth}>{\raggedright\arraybackslash}p{0.50\textwidth}@{}}
    \toprule
    File & Contents \\
    \midrule
    \multicolumn{2}{@{}l}{\emph{Verification accuracy (Sections~\ref{sec:benchmark},
    \ref{sec:codec})}}\\
    \texttt{accuracy/metrics.csv} & Master grid: EER, FNMR@FMR (bootstrap CIs on the
      anchor matchers), realised FMR, pair counts, $\Delta$EER vs.\ the aligned
      reference \\
    \texttt{accuracy/preproc\_through\_codec\_kk.csv} & Pre-processing operators pushed
      through each codec (Section~\ref{subsec:preproc-through}) \\
    \texttt{embeddings/manifest.csv} (one per dataset, on the dataset store rather than
      under the output root) & Provenance of every embedding array: how
      many crops were embedded per (matcher, source) and how many were missing, so the
      support behind any cell in the grid above can be checked directly \\
    \midrule
    \multicolumn{2}{@{}l}{\emph{Image quality (Section~\ref{sec:quality})}}\\
    \texttt{quality/quality\_summary.csv} & Median PSNR/SSIM/MS-SSIM/LPIPS/DISTS and $n$
      per (dataset, codec, res, budget) \\
    \texttt{quality/quality\_\{colorferet,kk\}.csv} & The same metrics per dataset before
      aggregation \\
    \texttt{quality/fiq\_\{colorferet,kk\}.csv} & Face-image-quality scores per codec cell
      (Section~\ref{subsec:fiq}) \\
    \texttt{quality/codec\_verify.csv} & Per-crop reconstruction check for the custom
      codec, including per-tertile luma error \\
    \texttt{quality/idcos\_tail\_512.csv} & Worst-case identity tail at $512$\,B: p5 and
      median per-image reconstruction-vs-original cosine (Section~\ref{sec:codec}) \\
    \texttt{quality/speed\_*.csv} & Encode/decode latency: classical, learned and JPEG-AI
      codecs on CPU; learned codecs and JPEG-AI on GPU; plus the merged
      \texttt{speed\_benchmark.csv} \\
    \midrule
    \multicolumn{2}{@{}l}{\emph{Fairness (Section~\ref{sec:fairness})}}\\
    \texttt{accuracy/fairness\_\{colorferet,kk\}.csv} & Per-subgroup EER by skin tone, age
      and gender \\
    \texttt{accuracy/fairness\_disparity\_\-\{colorferet,kk\}.csv} & Max$-$min between-group
      disparity per codec and matcher \\
    \texttt{accuracy/disparity\_ci\_kk.csv} & Subject-level cluster-bootstrap CIs on the KK
      disparities \\
    \texttt{accuracy/fmr\_fairness\_\{colorferet,kk\}.csv} & Disparity measured at a fixed
      FMR rather than at EER \\
    \midrule
    \multicolumn{2}{@{}l}{\emph{Statistical testing (Section~\ref{sec:significance})}}\\
    \texttt{accuracy/significance\_\{colorferet,kk\}.csv} & Pairwise McNemar and DeLong
      tests with BH-FDR-corrected $p$-values \\
    \texttt{accuracy/significance\_*\_r\{res\}\_b\{budget\}.csv} & The same tests per
      matcher $\times$ resolution $\times$ budget cell \\
    \midrule
    \multicolumn{2}{@{}l}{\emph{Studies}}\\
    \texttt{recompression/recompression\_*.csv} & Compressed-on-compressed chain EERs
      (Section~\ref{sec:recompression}) \\
    \texttt{adversarial/sanitization.csv} & Attack strength and post-codec sanitisation
      per (dataset, matcher, attack, $\varepsilon$, codec, budget), with the clean and
      attacked post-codec cosines, the residual, and the crop count $n$ behind each cell;
      three attacks $\times$ three $\varepsilon$ over all twelve codecs
      (Section~\ref{sec:adversarial}) \\
    \texttt{resolution\_information/*.csv} & Spectral retention, embedding decomposition
      and EER by resolution (Section~\ref{sec:ablations}) \\
    \texttt{ablation/s7\_ablation.csv} & Custom-codec component ablation
      (Section~\ref{sec:codec}) \\
    \texttt{difficulty/*.csv} & Per-sample difficulty and its predictors
      (Section~\ref{sec:difficulty}) \\
    \texttt{contamination/*.csv} & Train/test overlap and duplicate-image scans \\
    \midrule
    \multicolumn{2}{@{}l}{\emph{Codec metadata and derived summaries}}\\
    \texttt{codec\_properties.csv} & Provenance, licensing, implementation and footprint
      per codec (Table~\ref{tab:codec-properties}) \\
    \texttt{codec\_comparison/file\_size\_summary.csv} & Achieved file-size distribution
      and budget-compliance rate \\
    \texttt{codec\_comparison/jpegai\_*.csv} & JPEG-AI decoder trade-off and its KK grid
      (Section~\ref{subsec:jpegai-speed}) \\
    \texttt{codec\_comparison/res\{112,224\}/\-comparison.json} & Per-(dataset, codec,
      budget) identity cosine, PSNR/SSIM/LPIPS and achieved bytes; source of the
      codec-comparison tables \\
    \texttt{report\_summary/key\_findings.\{csv,json\}} & Derived register the headline
      claims are read from; \texttt{report\_summary/README.md} documents its schema \\
    \texttt{report\_summary/derived\_stats.json} & Scalars quoted in the text: the
      quality-vs-identity Spearman $\rho$ and the fairness-amplification factors \\
    \texttt{posthoc\_scalars.txt}, \texttt{posthoc\_stats.tex} (written under
      \texttt{report/tables/} rather than the output root: they are generated report
      inputs, derived from \texttt{accuracy/metrics.csv}, not pipeline measurements) &
      Friedman $\chi^2$, Kendall's $W$ and the codec mean ranks over the
      $14\times12$ matcher $\times$ codec grid; Holm-corrected Wilcoxon $p$ and Cliff's
      $\delta$ per codec pair (Section~\ref{sec:significance}) \\
    \bottomrule
  \end{tabular}
\end{table}